\documentclass{article}

    \PassOptionsToPackage{numbers, compress}{natbib}

    \usepackage[final]{neurips_2025}

\usepackage[utf8]{inputenc} 
\usepackage[T1]{fontenc}    
\usepackage{hyperref}       
\usepackage{url}            
\usepackage{booktabs}       
\usepackage{amsfonts}       
\usepackage{subcaption}    
\usepackage{nicefrac}       
\usepackage{microtype}      
\usepackage{xcolor}         
\usepackage{amsmath}
\usepackage{amssymb}
\usepackage{caption}
\usepackage{graphicx}
\usepackage{dsfont}
\usepackage{wrapfig}
\usepackage{xspace}
\usepackage{titlecaps}
\usepackage{cleveref}
\usepackage{amsthm}
\usepackage{dsfont}
\usepackage{wrapfig}
\usepackage{makecell}
\usepackage{booktabs}
\usepackage{tablefootnote}

\usepackage{listings}

\usepackage[ruled, noend]{algorithm2e}

\newtheorem{theorem}{Theorem}

\newtheorem{lemma}{Lemma}
\Crefname{equation}{Eq.}{Eqs.}
\Crefname{algocf}{Algorithm}{Algorithms}

\newlength{\itemizelength} 
\title{Document Summarization with\\ Conformal Importance Guarantees}

\author{%
  Bruce Kuwahara\thanks{Equal contribution} \\
  Signal 1 AI \\
  Toronto, Canada \\
  \And
  Chen-Yuan Lin$^*$ \\ 
  Signal 1 AI \\
  Toronto, Canada \\
  \And
  Xiao Shi Huang$^*$ \\ 
  Signal 1 AI \\
  Toronto, Canada \\
  \And
  Kin Kwan Leung \\ 
  Layer 6 AI \\
  Toronto, Canada \\
  \And
  Jullian Arta Yapeter \\ 
  Signal 1 AI \\
  Toronto, Canada \\
  \And
  Ilya Stanevich \\ 
  Signal 1 AI \\
  Toronto, Canada \\
  \And
  Felipe Perez \\ 
  Signal 1 AI \\
  Toronto, Canada \\
  \And
  Jesse C. Cresswell \\ 
  Layer 6 AI \\
  Toronto, Canada \\
}

\begin{document}

\maketitle

\begin{abstract}

Automatic summarization systems have advanced rapidly with large language models (LLMs), yet they still lack reliable guarantees on inclusion of critical content in high-stakes domains like healthcare, law, and finance. In this work, we introduce Conformal Importance Summarization, the first framework for importance-preserving summary generation which uses conformal prediction to provide rigorous, distribution-free coverage guarantees. By calibrating thresholds on sentence-level importance scores, we enable extractive document summarization with user-specified coverage and recall rates over critical content. Our method is model-agnostic, requires only a small calibration set, and seamlessly integrates with existing black-box LLMs. Experiments on established summarization benchmarks demonstrate that Conformal Importance Summarization achieves the theoretically assured information coverage rate. Our work suggests that Conformal Importance Summarization can be combined with existing techniques to achieve reliable, controllable automatic summarization, paving the way for safer deployment of AI summarization tools in critical applications. Code is available at \href{https://github.com/layer6ai-labs/conformal-importance-summarization}{github.com/layer6ai-labs/conformal-importance-summarization}.
\vspace{-10pt}
\end{abstract}

\vspace{-4pt}
\section{Introduction}
\label{sec:introduction}
\vspace{-4pt}

Summarization is a widely performed task in many domains, from media \citep{zhang2024benchmarking} and legal documents \citep{preti2024automatic, kanapala2019text} to scientific articles \citep{cachola-etal-2020-tldr} and clinical reports \citep{croxford2025development}. Recent advances in large language models (LLMs) have significantly improved the quality of summary generation \citep{10.5555/3648699.3648939, Achiam2023GPT4TR, grattafiori2024llama, yang2024qwen2}, with methods such as prompt-based generation \citep{chang2024booookscore, zhang2024systematic} and fine-tuned transformer models \cite{Vaswani2017Jun} exhibiting superior generalization and adaptability over classical natural language processing (NLP) methods \citep{liu2019textsummarizationpretrainedencoders, 10.5555/3524938.3525989}. However, in critical domains any error in an AI-generated summary can have serious consequences \citep{asgari2024framework}. For example, even with evidence that AI summarizers can reduce physician workloads and alleviate burnout \citep{ghatnekar2021digital}, lack of consistency and the need to verify the AI's work remains a concern for physicians in practice \citep{shah2025physician}. Despite the improvements mentioned above, no existing method \emph{guarantees} retention of important content which could, for example, ensure safety in a high-stakes application like healthcare \citep{croxford2025current}.

Conformal prediction \cite{vovk2005algorithmic, shafer2008tutorial} has recently risen in popularity as it provides distribution-free, finite-sample coverage guarantees \cite{angelopoulos2022gentle}, and has shown promise in classification \cite{romano2020aps, angelopoulos2021raps, huang2024saps}, regression \cite{burnaev2014efficiency,MESSOUDI2021108101, romano2019conformalized}, and language tasks such as factual question answering \cite{kumar2023conformal, quach2024conformal, mohri2024factuality, cherian2024large, feng2025}. In this work, we introduce \textbf{Conformal Importance Summarization}, the first application of conformal prediction to document summarization which provides statistical guarantees on the inclusion of important content. 

Our contributions are as follows:
\begin{itemize}
\setlength{\leftskip}{\itemizelength}
    \item We formalize the problem of importance-preserving document summarization with statistical guarantees through the conformal prediction framework.
    \item We introduce a method that calibrates sentence-level importance scores, allowing summary generation with user-specified error ($\alpha$) and recall ($\beta$) rates.
    \item We evaluate our method across multiple summarization benchmarks, demonstrating empirical importance coverage as expected from our theory, and quantifying the utility of our approach.
\end{itemize}

\vspace{-4pt}
\section{Background and Terminology}
\label{sec:background}
\vspace{-4pt}
\subsection{Document Summarization}\label{subsec:document_summarization}
\vspace{-4pt}
Summarization is the task of producing a concise version of a source document that preserves its most important content. There are two major categories of approaches \cite{koh2022survey}: \textbf{extractive summarization}, which selects spans of text (typically sentences) taken verbatim from the source text, and \textbf{abstractive summarization} which generates new sentences that paraphrase or synthesize information from the source text. While abstractive summarization is extremely accessible due to the advent of instruction-tuned LLMs \cite{wei2022finetuned, sanh2022multitask}, extractive summarization can be more suitable to high-stakes domains because it limits the possibility of hallucinations and remains more faithful to the source's meaning. While our main focus is on extractive summarization, we also show how the two approaches can be combined to benefit from the improved fluency and conciseness that abstractive summarization offers.

Classical extractive methods such as TextRank \citep{mihalcea-tarau-2004-textrank} rely on heuristics or graph-based techniques. Modern extractive models, such as BERTSum \citep{liu-lapata-2019-text}, leverage pretrained language models like BERT \cite{devlin2019bertpretrainingdeepbidirectional} to encode sentence-level representations and classify sentence importance. Recent trends in summarization also include reinforcement learning for optimizing summary-level objectives directly (e.g., ROUGE \citep{DBLP:conf/iclr/PaulusXS18, lin-2004-rouge}), fact-consistency tuning using entailment models \citep{kryscinski-etal-2020-evaluating}, and hybrid extractive-abstractive pipelines \citep{chen-bansal-2018-fast}. 

In this work, we show that extractive summarization, which scores and ranks text spans, is naturally compatible with the calibration step of conformal prediction and can achieve statistical guarantees on the retention of important sentences.

\vspace{-4pt}
\subsection{Conformal Prediction}\label{subsec:conformal_prediction}
\vspace{-4pt}
Consider a classification or regression problem with inputs $x\in \mathcal{X}$ associated with ground-truth values $y^*\in \mathcal{Y}$ drawn jointly from a distribution $(x, y^*)\sim \mathbb{P}$. Conformal prediction (CP) \citep{vovk2005algorithmic, shafer2008tutorial} first calibrates a threshold $\hat q$ based on labeled data, then predicts a set of output values $C_{\hat q}(x_\text{test})\subseteq \mathcal{Y}$ for any new datapoint $x_\text{test}$, while guaranteeing \emph{coverage} with a user-defined error rate $\alpha$,
\begin{equation}\label{eq:coverage-guarantee}
     \mathbb{P}[y_\text{test}^* \in C_{\hat q}(x_\text{test})] \geq 1 - \alpha.
\end{equation}
Remarkably, the coverage guarantee is distribution-free and valid in finite samples, making CP a versatile tool to provide robust guarantees on correctness in a wide variety of scenarios \cite{angelopoulos2022gentle}. 

Given a model $f_\theta:\mathcal{X} \to \mathcal{Y}$, CP defines a \emph{conformal score} function $S:\mathcal{X} \times \mathcal{Y}\to \mathbb{R}$, where larger values indicate worse agreement between $f_\theta(x)$ and $y^*$. Upon computing $S$ over $n$ calibration datapoints, the threshold ${\hat q}$ is set as the $\tfrac{\lceil{(n+1)(1-\alpha)}\rceil}{n}$ quantile of the conformal scores. For any new input, prediction sets are generated by including all output values for which the conformal score is below the threshold $\hat q$,
\begin{equation}\label{eq:prediction-set}
    C_{\hat q}(x_\text{test})  = \{y\in \mathcal{Y} \mid s(x_\text{test}, y) < \hat q \}.
\end{equation}
As long as $x_\text{test}$ is exchangeable with the calibration data drawn from $\mathbb{P}$, \Cref{eq:coverage-guarantee} will hold. For equal coverage levels $1-\alpha$, the usefulness of prediction sets can be judged by their size \cite{romano2020aps, angelopoulos2021raps, huang2024saps}, with smaller average set sizes $\mathbb{E}\vert C_{\hat q}\vert $ indicating prediction sets that are more useful in downstream tasks \cite{cresswell2024, cresswell2025conformal}. The quality of $ C_{\hat q}$ is largely driven by the accuracy and calibration of $f_\theta$, and the design of $S$ for expressing the model's confidence.

\vspace{-4pt}
\subsection{Conformal Factuality for Question-Answering}\label{subsec:conformal_factuality}
\vspace{-4pt}

Extending CP to language tasks requires rethinking the problem setup and design of the conformal score $S$. Whereas classification tasks have a finite label set $\mathcal{Y}$, open-ended question-answering has an effectively infinite output space with many semantically equivalent responses, making it incompatible with the standard CP framework described in \Cref{subsec:conformal_prediction}.

\looseness=-1 Conformal factuality \cite{mohri2024factuality} overcomes these challenges by replacing prediction sets with entailment sets, aiming to give responses that are entailed by the ground-truth $y^*$ with high probability \cite{bowman2015entailment}. Given a natural language question $x$ and response generated by an LLM, conformal factuality first decomposes the generated text into subclaims $\hat y=\{c_1, \dots, c_p\}$. We use $T(x, y^*)\subseteq \hat y$ to denote the subset of generated claims which are entailed by the ground-truth $y^*$ for question $x$. Then, individual claims are filtered out based on a calibrated threshold $\hat q$. The remaining claims constitute a response $y\subseteq \hat{y}$ which is factual with high probability,
\begin{equation}\label{eq:conformal_factuality}
    \mathbb{P}[y \subseteq T(x, y^*)] \geq 1 - \alpha.
\end{equation}

To set $\hat q$, each claim's factuality is first evaluated based on heuristics such as model confidence or self-evaluation. Then, the conformal score is assigned as the greatest factuality level out of all claims not entailed by $y^*$. After computing the conformal score for each $(x, y^*)$ in the calibration set, the overall conformal threshold $\hat q$ is set as the $\tfrac{\lceil{(n+1)(1-\alpha)}\rceil}{n}$ quantile of scores. 

On test data, claims with assessed factuality level less than $\hat q$ are filtered out. The sets of retained claims satisfy \Cref{eq:conformal_factuality} by Theorem 3.1 of \cite{mohri2024factuality}. Since longer, more detailed answers are more helpful, the quality of the final response is judged by how many of the generated claims can be retained, subject to meeting the coverage guarantee in \Cref{eq:conformal_factuality}. Quality is dictated by how well the factuality of individual claims is assessed, and the prevalence of non-factual claims in the generated response, which can be improved with a stronger LLM, adjustments to the score function, better grounding through retrieved context, or reasoning \cite{cherian2024large, feng2025, rubin2025conformal}.

\vspace{-4pt}
\section{Extractive Summarization via Conformal Importance}
\label{sec:method}

\vspace{-4pt}
We take inspiration from conformal factuality to go beyond question-answering and provide statistical guarantees on extractive summarization. By calibrating a threshold on an importance score, we ensure that the final summary preserves the most salient information while maintaining a bounded risk of omitting critical content. 

Formally, our goal is to produce a shorter version $y$ of the long-text $x$ which, with high probability, retains all important information. We take the long-text $x\in \mathcal{X}$ to consist of multiple sentences, $x=\{c_1, \dots, c_{p}\}$, and let $y^*\subseteq x$ be the ground-truth set of important sentences from $x$.\footnote{For simplicity we break $x$ by sentences, but any span could be used. We discuss cases where $y^*\not\subseteq x$ below.} We then filter out sentences from $x$ based on a calibrated threshold $\hat q$ leaving a subset $y\subseteq x$ which retains all important information with high probability,
\begin{equation}\label{eq:conformal_importance}
    \mathbb{P}[y^* \subseteq y] \geq 1 - \alpha.
\end{equation}
We note a key difference between \Cref{eq:conformal_factuality} and \Cref{eq:conformal_importance}: conformal factuality aims for high precision that retained claims are factual, while conformal importance aims for high recall that important sentences are retained. As long as high recall is ensured, shorter summaries are preferable which allows us to measure the quality of $y$ as the proportion of sentences removed. In the remainder of this section we develop a more general framework for error control than used in conformal factuality, describe our method to extract summaries, and theoretically prove that it obeys a coverage guarantee.

\textbf{Generalizing the Coverage Guarantee.} \quad One limitation of the conformal factuality framework is that \Cref{eq:conformal_factuality} rigidly considers a response with any number of non-factual claims to be a failure case. A more general framework would allow the user to set their own tolerance on how many non-factual claims could appear in an acceptable response \cite{angelopoulos2024conformal}. Hence, for conformal importance we relax the coverage guarantee \Cref{eq:conformal_importance} so that summaries are acceptable if they retain at least a fraction $\beta$ of the important sentences. Let
\begin{equation}\label{eq:fraction_retained}
B(y; y^*) = \frac{|y\cap y^*|}{|y^*|}
\end{equation}
be the \emph{recall}, i.e. the fraction of important sentences retained by $y$. Then the relaxed coverage guarantee is
\begin{equation}\label{eq:generalized_conformal_importance}
\mathbb{P}[B(y; y^*) \geq\beta] \geq 1 - \alpha.
\end{equation}
Of course, this recovers \Cref{eq:conformal_importance} when $\beta=1$. 

\begin{figure}[t]
    \centering
        \begin{subfigure}{0.68\textwidth}
        \includegraphics[width=0.97\linewidth]{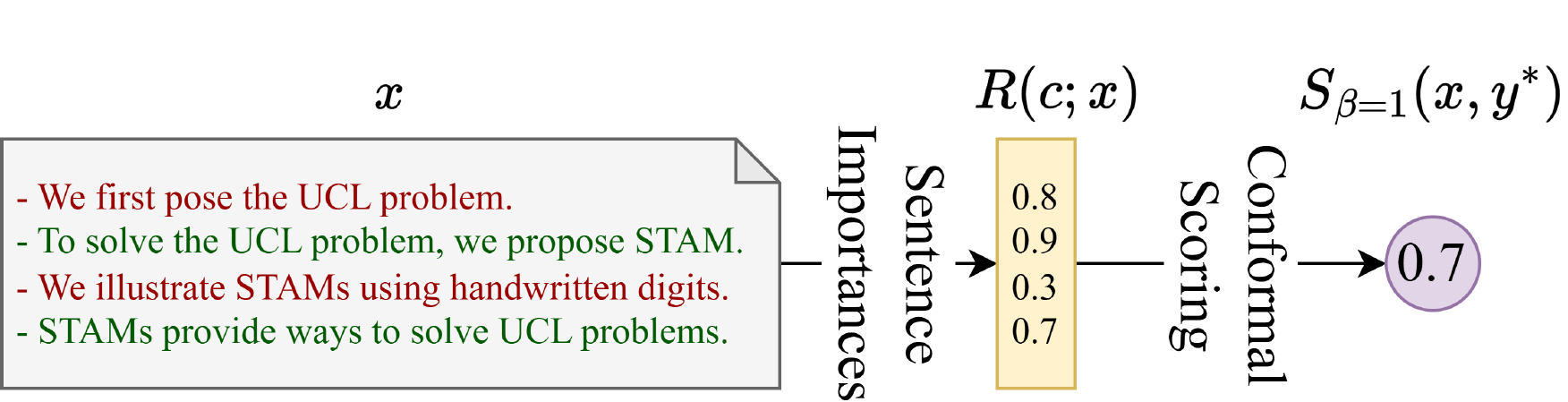}
        \caption{}
        \label{fig:fig1a}
        \end{subfigure}
        \begin{subfigure}{0.3\textwidth}
            \includegraphics[width=0.97\linewidth]{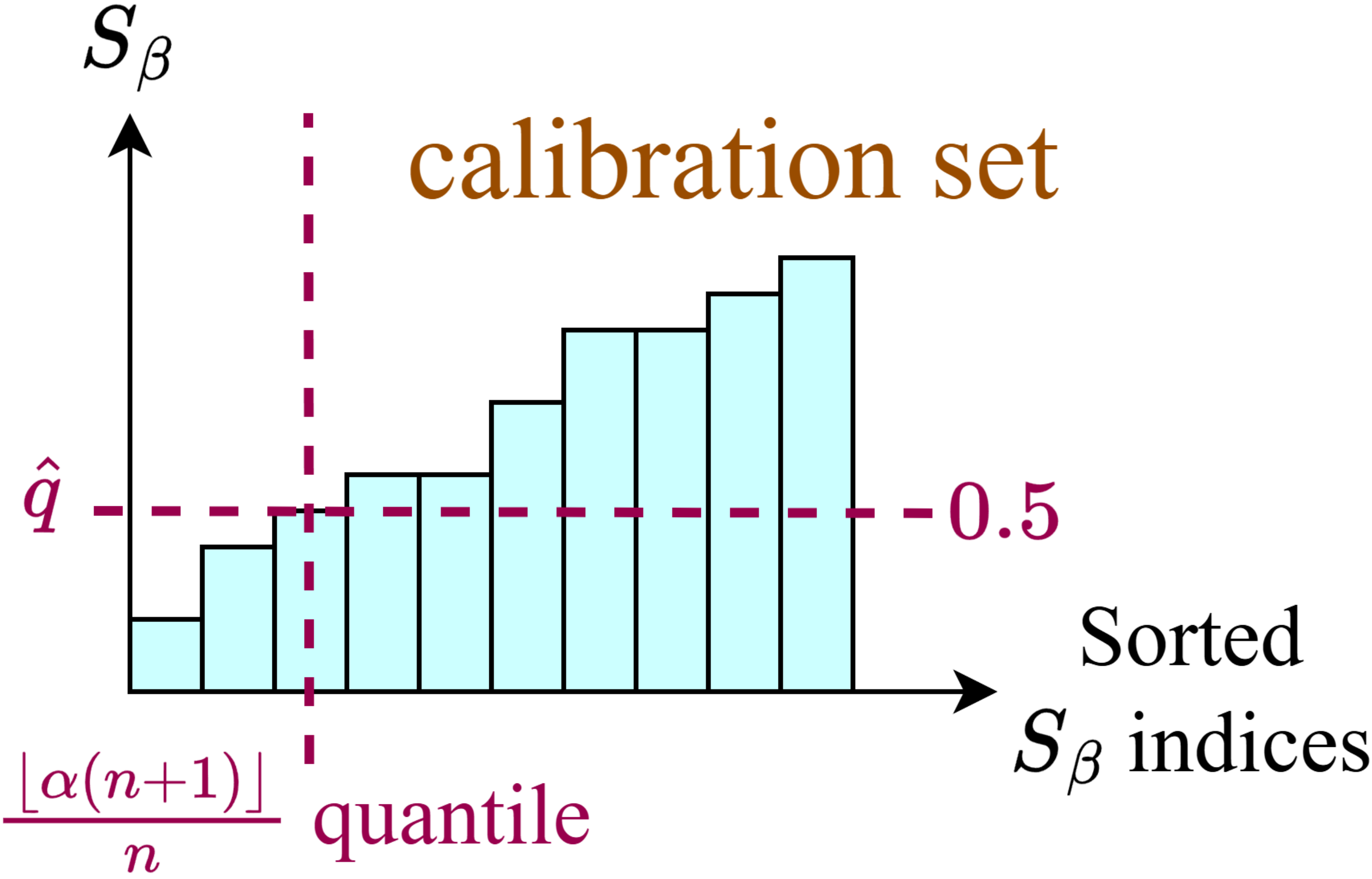}
        \caption{}
        \label{fig:fig1b}
        \end{subfigure}

        \begin{subfigure}{0.95\textwidth}
        \includegraphics[width=0.97\linewidth]{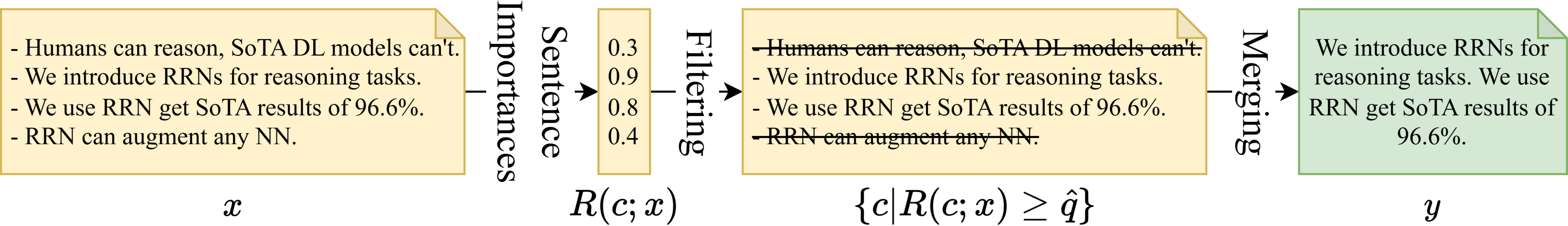}
        \caption{}
        \label{fig:fig1c}
        \end{subfigure}
        \caption{\looseness=-1 (a) Steps to compute the conformal score for a labeled datapoint. Ground-truth important sentences are coloured in green. $S_{\beta=1}$ is the smallest value of $R(c;x)$ for any important sentence. (b) The conformal threshold $\hat{q}$ is computed as a quantile of the sorted conformal scores over the calibration set. (c) At inference, only sentences with importance score $R(c;x)\geq \hat q$ are retained in the summary.}
    \label{fig:fig1}
    \vspace{-10pt}
\end{figure}

\textbf{Conformal Importance Summarization.} \quad  Given the long-text $x$ we assign an importance score to each sentence $c$ as $R(c;x)$ such that $R(c;x)\geq 0$ with larger scores indicating higher estimated importance. In practice, the design of the importance score is key to the performance of our method, and we discuss various options below. Similar to \cite{mohri2024factuality}, we define a filtering function based on importance scores
\begin{equation}\label{eq:filter_importance}
    F_q(x) = \{c\in x \mid R(c;x) \geq q\}.
\end{equation}
This function satisfies both $F_0(x) = x$ and $F_\infty(x) = \varnothing$ so that $F_q$ filters out more sentences as $q$ increases, and satisfies a nesting property: $F_q(x) \subseteq F_{q'}(x)$ for $q\geq q'$ \citep{gupta2022nested}.

To determine the appropriate threshold $\hat q$ we use CP calibration with conformal scores over a labeled calibration dataset. For each pair of long-text $x_i$ and ground-truth summary $y_i^*$ we compute
\begin{equation}\label{eq:conformal_importance_score}
S_\beta(x_i, y_i^*) := \max\{ q \in \mathbb{R}^+ \mid  B(F_q(x_i); y_i^*) \geq \beta\}.
\end{equation} 
We note that the maximum always exists by the definition of $F_q(x)$. This score computes the largest threshold $q$ such that at least a fraction $\beta$ of the important sentences are retained after filtration. A larger $q$ (and hence larger $S_\beta$) is preferable as it will produce a more concise summary $y_i=F_q(x_i)$. 

Noting that $B(F_q(x); y^*)$ is a non-increasing function of $q$, we have the following observation.
\begin{lemma}\label{lemma}
For a fixed $\hat q$, we have
\begin{equation}
S_\beta (x, y^*) \geq \hat q\Longleftrightarrow B(F_{\hat q}(x); y^*) \geq \beta.
\end{equation}
\end{lemma}
\vspace{-12pt}
\begin{proof}
First, assume that $S_\beta (x, y^*) \geq \hat{q}$. By the definition in \Cref{eq:conformal_importance_score}, there exists $q' \geq \hat{q}$ such that $B(F_{q'}(x); y^*) \geq \beta$. Since $B(F_q(x); y^*)$ is non-increasing in $q$, and $q'\geq \hat{q}$, it follows that $B(F_{\hat{q}}(x); y^*) \geq \beta$. On the other hand, now assume $B(F_{\hat{q}}(x); y^*)\geq \beta$. By the definition of maximum, we have $S_\beta(x, y^*) \geq \hat{q}$ directly from \Cref{eq:conformal_importance_score}.
\end{proof}
\vspace{-8pt}

With all conformal scores $S_\beta(x_i, y_i^*)$ computed on the calibration set, we choose the overall conformal threshold $\hat q$ as the $\frac{\lfloor \alpha (n+1)\rfloor}{n}$ quantile.\footnote{The different quantile compared to conformal factuality is necessary, as we have shifted to focus on recall instead of precision in the coverage guarantee and conformal score function.} Given a new long-text $x_\text{test}$ we score each sentence's importance as $R(c;x_\text{test})$, and filter out any sentence with importance less than $\hat q$, returning $y_\text{test} = F_{\hat q}( x_\text{test})$, as shown in \Cref{fig:fig1}. This procedure satisfies our generalized coverage guarantee in \Cref{eq:generalized_conformal_importance}.
\begin{theorem}\label{theorem}
Let $\{x_i, y_i^*\}_{i=1}^{n+1}$ be exchangeable and $0<\beta\leq 1$. Let $\hat{q}$ be the $\frac{\lfloor\alpha (n+1) \rfloor}{n}$-th quantile of the scores $\{S_\beta(x_i, y_i^*)\}_{i=1}^n$, which we assume to be distinct without loss of generality. Then for $\alpha \in[\frac{1}{n+1}, 1]$, we have 
\vspace{-4pt}
\begin{equation}
1-\alpha \leq \mathbb{P}[B(F_{\hat{q}}(x_{n+1}); y_{n+1}^*)\geq \beta] < 1 - \alpha + \frac{1}{n+1}.
\end{equation}
\end{theorem}
\vspace{-12pt}
\begin{proof} \label{proof:coverage}
Let $s_i = S_\beta (x_i, y_i^*)$ for $i=1,\ldots, n$, and $s_\text{test} = S_\beta(x_{n+1}, y_{n+1}^*)$. Without loss of generality, we assume the scores are sorted $s_1 < s_2\ldots <s_n$. By exchangeability of the $x_i$'s, and for any $k=1,\ldots,n$, we have
\vspace{-4pt}
\begin{equation}
\mathbb{P}[s_\text{test} \geq s_k] = 1 - \frac{k}{n+1}.
\end{equation}
Noting that $\hat q= s_{\lfloor \alpha (n+1)\rfloor}$ by definition of the quantile, and that $\alpha \geq \frac{1}{n+1}$, we obtain
\vspace{-4pt}
\begin{equation}\label{eq:proof_intermediate}
\mathbb{P}[s_\text{test} \geq \hat{q}] = 1 - \frac{\lfloor \alpha (n+1)\rfloor}{n+1} \geq 1 - \alpha.
\end{equation}
By \Cref{lemma} we then have 
\vspace{-4pt}
\begin{equation}
\mathbb{P}[B(F_{\hat{q}}(x_{n+1}); y_{n+1}^*)\geq \beta] \geq 1 - \alpha.
\end{equation}

On the other hand, from \Cref{eq:proof_intermediate} we see that
\begin{equation}
\mathbb{P}[s_\text{test} \geq\hat{q}] = 1 - \frac{\lfloor \alpha (n+1)\rfloor}{n+1} < 1 - \frac{\alpha (n+1) - 1}{n+1} = 1 - \alpha + \frac{1}{n+1},
\end{equation}
which shows that
\vspace{-4pt}
\begin{equation}
1-\alpha \leq \mathbb{P}[B(F_{\hat{q}}(x_{n+1}); y_{n+1}^*)\geq \beta] < 1 - \alpha + \frac{1}{n+1}.
\end{equation}
\end{proof}

\vspace{-10pt}
\textbf{Design Choices.}\quad  Beyond the free parameters $\alpha$ and $\beta$ which can be set according to the user's error tolerances, our method involves design choices around the ground-truth $y^*$ and the importance score function $R(c;x)$. We defined important sentences as those sentences appearing in $y^*$, but the source of ground truth remains flexible. On benchmark datasets for extractive summarization, $y^*$ may be provided at the sentence level as a subset of $x$. For other datasets, a summary may be given, but not as verbatim sentences selected from $x$, in which case we assume sentences in $y^*$ can be matched to a unique source from $x$. When $y^*$ is unavailable, and cannot be collected from domain experts, automated generation techniques using prompted LLMs can select important sentences. However, we note the strong possibility of bias if the same LLM is used for $R(c;x)$ \cite{NEURIPS2024_7f1f0218, deutsch-etal-2022-limitations}. We describe concrete methods for sentence matching and ground truth generation in \Cref{sec:datasets}.

Even for the same long-texts $x$, different users may consider different parts important. For example, when summarizing clinical notes produced over the course of a patient's hospital stay, a doctor and an administrator may have different views on which details are important. Conformal Importance Summarization can accommodate these different viewpoints; for the same data $x$, each user can indicate their preferences $y^*$ on the calibration set, and receive a threshold $\hat q$ tailored to their needs. It may also be beneficial to guide the meaning of importance via $R(c;x)$, for instance by describing within an LLM prompt the type of information which should be scored highly. Indeed, designing $R(c;x)$ is the most direct way to influence the ultimate performance of our method, so we offer and experimentally compare several possibilities that can be grouped into two classes. 

\textbf{LLM Scoring} - An LLM is prompted to return a score between 0 and 1 of how important $c$ is, given $x$ as context.  We test five different LLMs, \textbf{GPT-4o mini}~\cite{openai2024gpt4omini}, \textbf{Gemini 2.0 Flash-Lite}~\cite{mallick2025gemini2}, \textbf{Gemini 2.5 Flash}~\cite{google2025gemini25flash}, \textbf{Llama3-8B}~\cite{grattafiori2024llama}, and  \textbf{Qwen3-8B}~\cite{yang2025qwen3technicalreport}, all with the same prompt given in \Cref{app:GPTprompt}. The first three used public APIs, while the latter two were hosted locally on a 48 GB A6000 GPU.

\textbf{Embedding Similarity Scoring} - Sentence-level embeddings are commonly used in extractive summarization, and we demonstrate how they can be leveraged to give coverage guarantees. Using a sentence-level embedding model (e.g. SBERT \cite{reimers-gurevych-2019-sentence}), distances between all embeddings are computed to form a graph. Then, one of four aggregation methods is used to produce importance scores: \textbf{Cosine Similarity Centrality}~\citep{ramirez-orta-milios-2021-unsupervised} builds a fully connected similarity graph and assigns importance as the weighted centrality of a node; \textbf{Sentence Centrality}~\citep{gong2022improving} creates a directed graph where each sentence’s importance is computed based on similarity to later sentences only; \textbf{GUSUM}~\citep{gokhan2022gusum} creates an undirected graph where edges are cosine similarities, and the importance score is augmented with sentence-level features like position and length; \textbf{LexRank}~\citep{erkan2004lexrank} uses a similar undirected graph construction, but filters out weak connections, and then applies a PageRank-like algorithm \cite{page1999pagerank} over node centrality to rank the importance of sentences.

\textbf{Abstractive Post-processing.}\quad Extractive summarization limits the possibility of hallucination, but may produce text which does not flow smoothly. On the other hand, using LLMs for direct abstractive summarization does not give precise control over coverage and recall. Abstractive summarization essentially contains two subtasks: identifying important information, and synthesizing it into a concise and grammatically correct summary \cite{chen-bansal-2018-fast, gehrmann-etal-2018-bottom}. Direct prompting forces the LLM to perform both tasks at once. We argue that disentangling these subtasks can lead to better information retention, similar to how retrieval-augmented generation \cite{lewis2020rag} splits question-answering into the two subtasks of information retrieval and answer generation.

We propose a hybrid extractive-abstractive pipeline which first identifies important information, then synthesizes it. First, Conformal Importance Summarization extracts important information from $x$ as $y=F_{\hat{q}}(x)$ with guarantees on coverage and recall. Then, extracted sentences $y$ are passed to an LLM which is prompted to preserve \emph{all} information, and only improve conciseness and flow. While there is no guarantee the LLM will maintain coverage, below we test how successful it can be in practice, and compare it to pure abstractive summarization. The prompts used are given in \Cref{app:GPTprompt}.

\vspace{-4pt}
\section{Experimental Setup}
\label{sec:setup}
\vspace{-4pt}

Our experiments are designed to validate the conformal guarantee given by \Cref{theorem}, and compare Conformal Importance Summarization to existing summarization methods that do not provide guarantees. We run ablations to understand (\emph{i}) the influence of both $\alpha$ and $\beta$; (\emph{ii}) the design of the importance score function $R(c;x)$; and, (\emph{iii}) for datasets without explicit ground-truth labels, the label creation method. Finally, we compare pure abstractive summarization with an LLM to our hybrid extractive-abstractive approach.

\vspace{-8pt}
\subsection{Datasets}\label{sec:datasets}
\vspace{-4pt}
We use 5 datasets to evaluate the performance of our framework: \textbf{ECTSum}~\citep{mukherjee-etal-2022-ectsum} contains complete transcripts from corporate earnings calls, as well as expert-curated extractive summaries at the sentence level; \textbf{CSDS}~\citep{lin-etal-2021-csds} is a dataset of Chinese language customer-client conversations. Although the summaries are abstractive, each conversation has sentence-level labels for use as an extractive benchmark; \textbf{CNN/DM}~\citep{10.5555/2969239.2969428} covers news sourced from CNN and The Daily Mail with human-written summary sentences; SciTLDR~\citep{cachola-etal-2020-tldr} consists of summaries of scientific papers sourced from both authors and peer-reviewers, and we use two versions where the input is either the full text (\textbf{TLDR-Full}), or just the abstract, introduction, and conclusion (\textbf{TLDR-AIC}); \textbf{MTS-Dialog}~\citep{mts-dialog} is a collection of doctor-patient conversations and corresponding summaries intended to cover dialogue material.

\begin{table}[t]
    \centering
        \caption{Dataset Details}
    \label{tab:dataset}
    \setlength{\tabcolsep}{3pt}
    \small
    \begin{tabular}{lllrrrr}
        \toprule
        Dataset & Subset Filtering & Labeling Method & $|\mathcal{D}_{\text{cal}}|$& $|\mathcal{D}_{\text{test}}|$ & Average Length $p$ & Label Positive Rate\\
        \midrule
        ECT & All & Provided & 100 & 2322 & 45.6& 0.10 \\
        CSDS & Valid + Test & Provided & 100 & 1500 & 25.5& 0.27 \\
        CNN/DM & 1000 samples & SBERT Opt. & 100 & 900 & 31.0& 0.10 \\
        TLDR-AIC & >1 summary & SBERT Opt. & 100 & 1043 & 40.7& 0.06 \\
        TLDR-Full & >1 summary & SBERT Opt. & 100 & 1043 & 216 & 0.0014 \\
        MTS & >1 input sentence & GPT-4o mini & 100 & 1029 & 6.4& 0.81 \\
        \bottomrule
    \end{tabular}
    \vspace{-12pt}
\end{table}

For each dataset, a random subset ($n=100$) of all datapoints is sampled to form the calibration dataset. All remaining samples form the test dataset, except for CSDS where we use only the original validation and test sets, and CNN/DailyMail where we use 900 samples to reduce resource requirements, as shown in \Cref{tab:dataset}. Other details on dataset preparation are given in \Cref{app:details}.

While ECTSum and CSDS contain sentence-level labels, the other datasets require them to be generated by comparing the summary text to the original sentences. For CNN/DM and SciTLDR, a greedy labeling process based on SBERT~\citep{reimers-gurevych-2019-sentence} embedding cosine similarity is used, which we describe fully in \Cref{app:details} with \Cref{alg:greedy-extractive}. Due to the heavy context-dependency of MTS, we instead queried GPT-4o mini to generate the labels, which tended to classify sentences as important at a higher rate. We share the prompt used in \Cref{app:GPTprompt}. While neither GPT-based nor SBERT-based labels would completely match a domain expert's preferences, it is sufficient to evaluate the theoretical guarantees and performance of Conformal Importance Summarization.

\vspace{-4pt}
\subsection{Metrics}\label{sec:metrics}
\vspace{-4pt}
To evaluate the quality of Conformal Importance Summarization, we use several complementary metrics. First, independent of the conformal framework, we assess the ranking quality of importance scores $R(c;x)$ using the area under the precision-recall curve (AUPRC), which evaluates how well each method distinguishes between important and unimportant sentences. AUPRC captures the trade-off between conciseness (precision) and completeness (recall).

Second, we evaluate the conformal framework by fixing values for the target error rate $\alpha$ and recall $\beta$, and measuring the proportion of sentences removed, akin to set size in traditional conformal prediction, which reflects how effectively content is compressed while preserving key information.

Finally, the empirical values of coverage and recall actually acheived are relevant. The recall of a given summary is $B(y; y^*)$ defined in \Cref{eq:fraction_retained}, while the empirical coverage is binary, computed as $B(y; y^*)\geq\beta$. Both these measures are then averaged over the test set.

\subsection{Baselines}\label{sec:baselines}
No existing methods provide coverage guarantees for extractive summarization. In lieu of such existing baselines, we implement the several importance scoring methods described in \Cref{sec:method}, and also compare to the empirical coverage attained by LLMs without our conformal framework. In particular, we use GPT-4o mini to directly label each sentence as important or not, given the long-text as context but without access to the ground-truth summary. The prompt we use is provided in \Cref{app:GPTprompt}. Whereas Conformal Importance Summarization can provide summaries for any choice of $\alpha$, this baseline only provides a single value empirically, with no user control.

For the targeted evaluations of importance functions via AUPRC, we also include the ground-truth importance rate as a reference point, as it represents the performance of a random scoring function.

\vspace{-4pt}
\section{Results}
\label{sec:results}
\vspace{-4pt}

\begin{figure}[t]
    \centering
    \includegraphics[width=0.85\linewidth, trim={10 10 10 10}, clip]{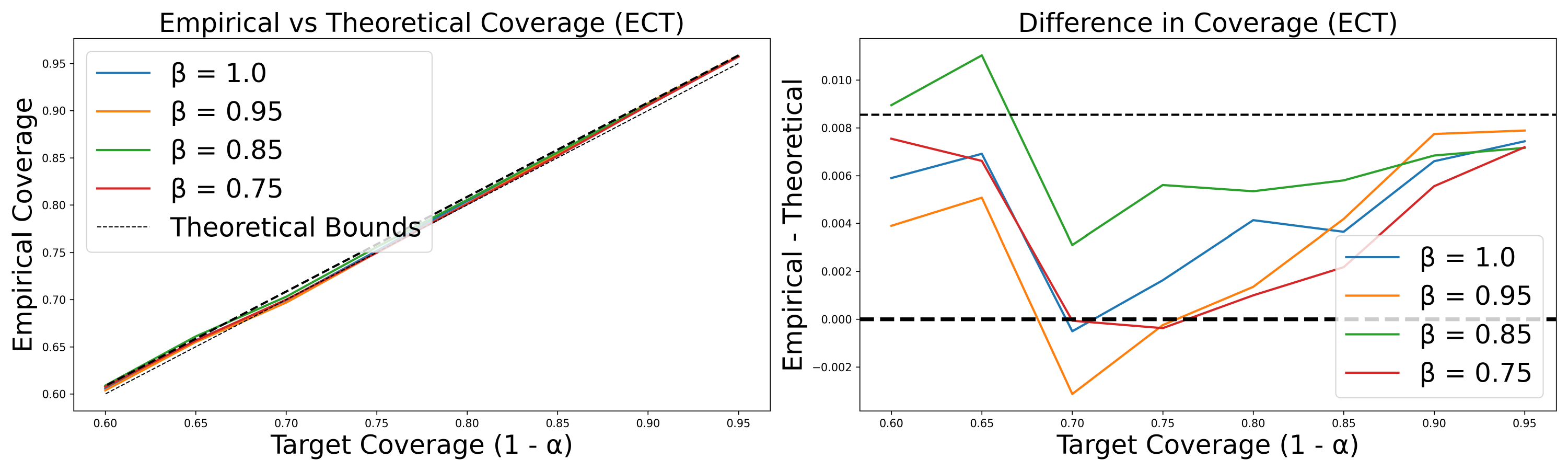}
    \caption{User-specified target coverage 1-$\alpha$ versus average empirical coverage, on ECTSum using Gemini 2.5 Flash scores. Dashed lines show theoretical bounds given in \Cref{theorem}. Results are averaged over 400 random splits of calibration and test data.}
    \label{fig:conformal_calibration}
    \vspace{-10pt}
\end{figure}

\subsection{Testing the Coverage Guarantee}\label{sec:empirical}

First, we empirically verify the theoretical guarantees proved in \Cref{sec:method}. We run Conformal Importance Summarization on the calibration set for a specified $\alpha$ and $\beta$, and measure the empirical coverage on the test set. The coverage rate should lie between $1-\alpha $ and $1 - \alpha + \frac{1}{n+1}$ in expectation, per \Cref{theorem}. Since the coverage is a random variable depending on the calibration data, we repeat the process a total of 400 times with randomized data splits, and average the coverage. \Cref{fig:conformal_calibration} shows results for the ECTSum dataset with Gemini 2.5 Flash scores under several values of $\alpha$ and $\beta$. We see that the average coverage consistently lies between the theoretical bounds, as expected. Deviations from the bounds are minor, and result from the variance of the coverage random variable. The same plots for other datasets and scoring functions are shown in Appendix \ref{app:coverage_plots}.

\subsection{Effect of $\alpha$ and $\beta$}\label{sec:alphabeta}

\begin{wrapfigure}[13]{r}{0.45\textwidth}
  \begin{center}
  \vspace{-23pt}
    \includegraphics[width=1.0\linewidth, trim={28 0 70 40}, clip]{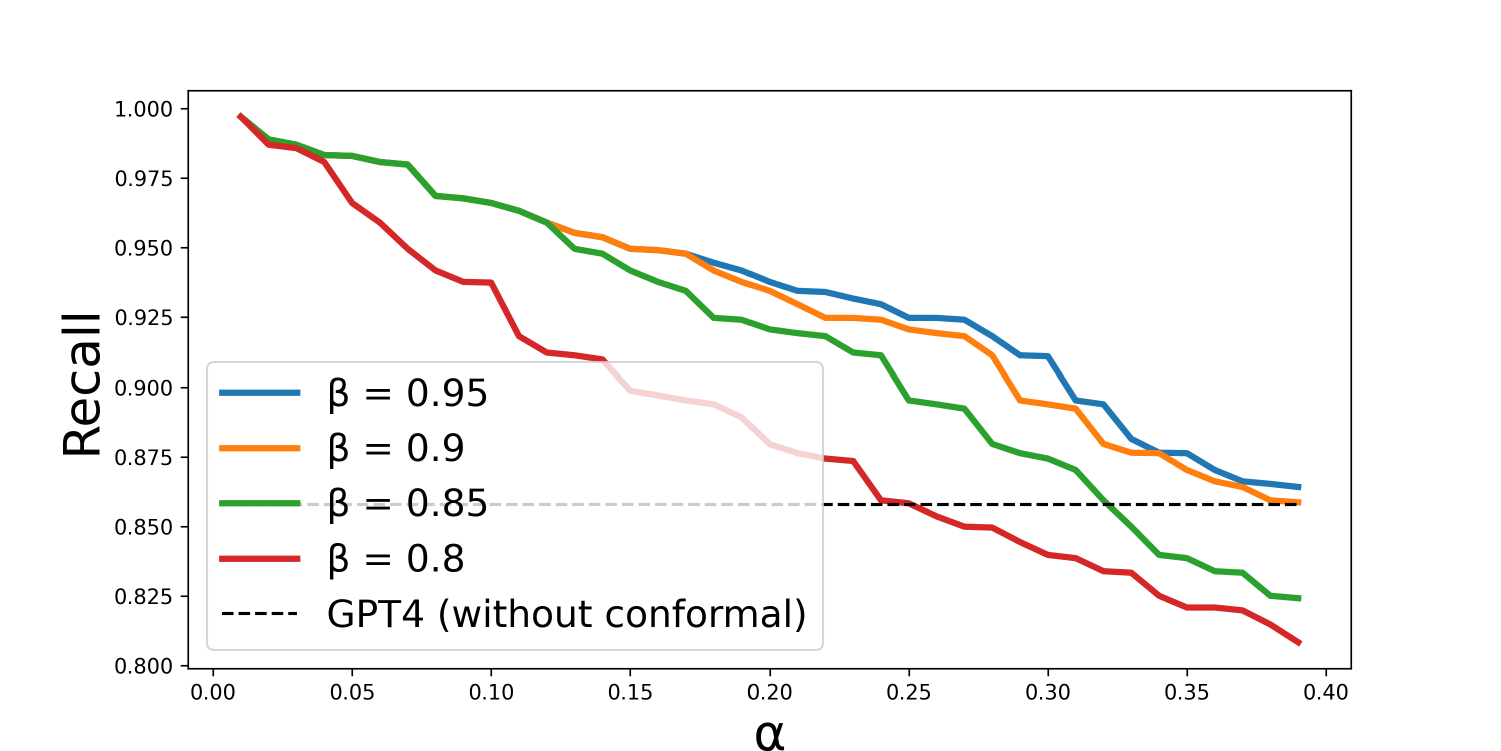}
  \end{center}
  \vspace{-8pt}
    \caption{Target error rate $\alpha$ versus empirical recall $B(y; y^*)$ of important sentences in summaries, averaged over the CSDS test set. The dashed line shows GPT-4o mini performance without using conformal prediction.}
    \label{fig:recall_vs_alpha}
\end{wrapfigure}
Next, we demonstrate that by tuning $\alpha$ and $\beta$, one can control the tradeoff between conciseness and completeness of the resultant summary. \Cref{fig:recall_vs_alpha} using GPT-4o mini on CSDS shows that a higher allowable error rate $\alpha$ results in lower empirical recall, as does a lower target recall $\beta$. Note that $\beta$ is the \emph{minimum} target recall for a $1-\alpha$ portion of the generated summaries, therefore the empirical recall is usually larger than $\beta$ in practice for small $\alpha$. 

\Cref{fig:recall_vs_alpha} (dashed line) also shows the baseline empirical recall when using GPT-4o mini as a standalone extractive summarization tool, as described in \Cref{sec:baselines}. This demonstrates how our conformal method allows more fine-grained control over recall and coverage than relying on LLMs directly. The same plots for other datasets and scoring functions are shown in Appendix \ref{app:error_recall}.

\begin{figure}[t]
    \centering
        \begin{subfigure}{0.48\textwidth}
        \includegraphics[width=0.95\linewidth, trim={40 7 60 50}, clip]{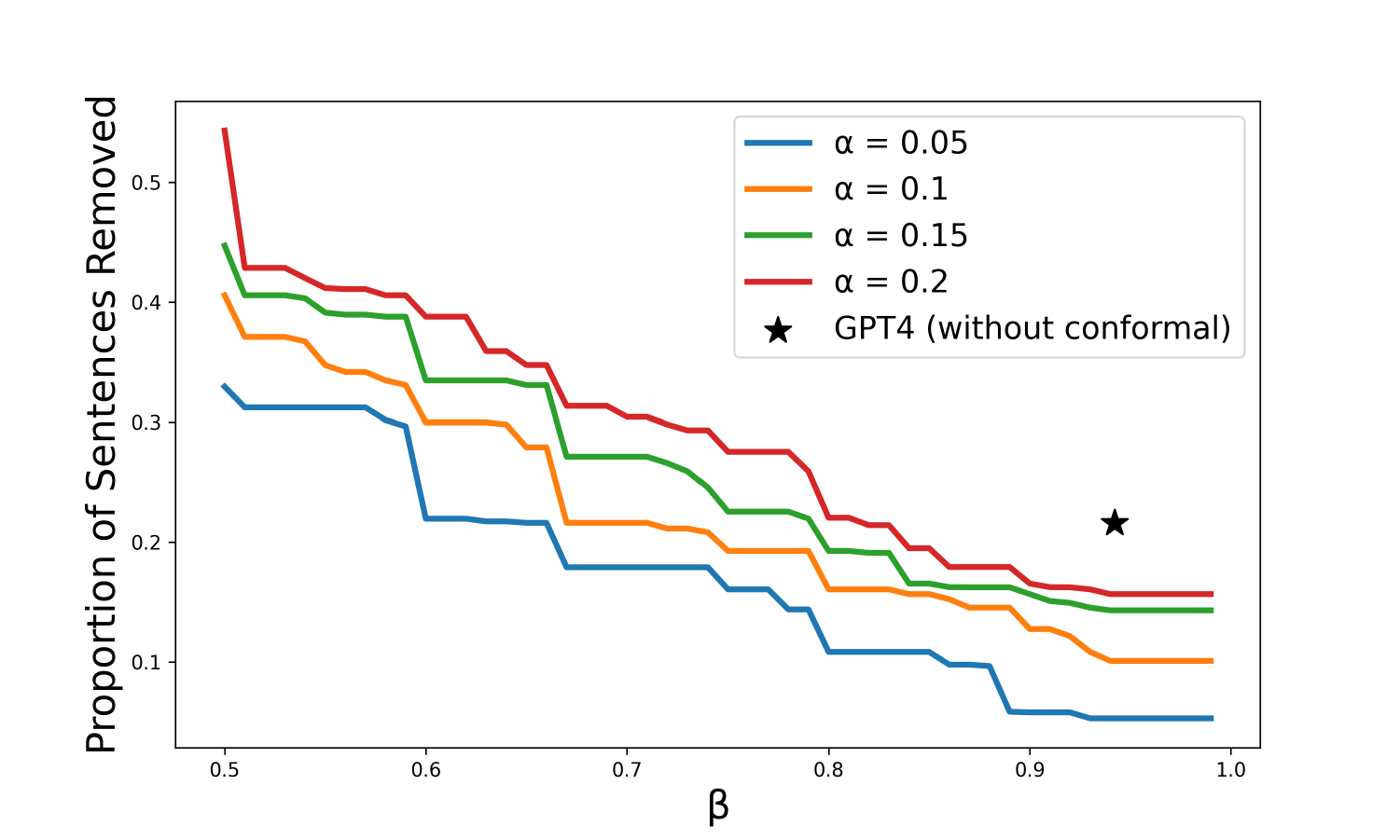}
        \end{subfigure}
        \begin{subfigure}{0.48\textwidth}
            \includegraphics[width=0.95\linewidth, trim={40 7 60 50}, clip]{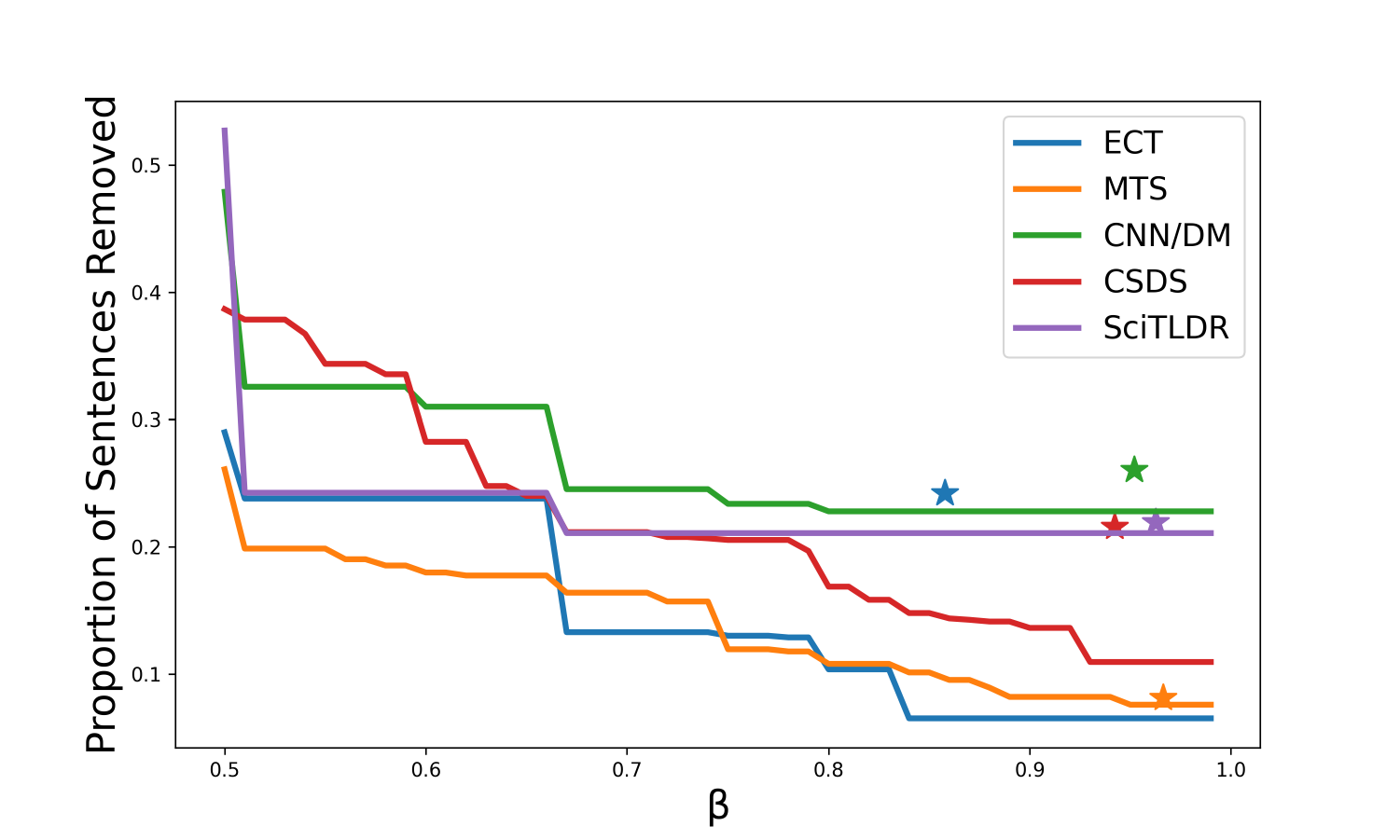}
        \end{subfigure}
        \caption{Target recall $\beta$ vs. proportion of sentences removed (conciseness). \textbf{Left:} Lines indicate different values for the target error rate $\alpha$ on CSDS. \textbf{Right:} Lines indicate different datasets ($\alpha=0.1$). Stars indicate the empirical recall and conciseness achieved by GPT-4o mini without conformal prediction.}
    \label{fig:reduction_versus_beta}
    \vspace{-10pt}
\end{figure}

High recall alone is not the objective, since summaries also must be concise. \Cref{fig:reduction_versus_beta} shows the proportion of sentences removed as the target recall $\beta$ is varied. Higher $\beta$ values are more conservative and retain more sentences, while lower values lead to shorter summaries but may miss more important information. We see that for a given value of $\beta$, higher values of $\alpha$ also result in shorter summaries, since greater risk tolerance enables more sentences to be removed. These trends are consistent across datasets and scoring functions, as shown in Appendix \ref{app:recall_conciseness}.

Again, we contrast this with the non-conformal baseline shown as stars in \Cref{fig:reduction_versus_beta}. Direct extraction with GPT-4o mini produces more concise summaries than our method for a given recall level, but offers no control over what that recall level is. In cases where concise summaries are the top priority, our method allows the user to tune $\beta$ down (or $\alpha$ up) to produce shorter summaries than the baseline.
\vspace{-10pt}

\subsection{Importance Score Function Performance}
The efficacy of Conformal Importance Summarization is highly dependent on the design of the importance score $R(c;x)$. Here we compare the performance of the alternatives described in \Cref{sec:method} across datasets. \Cref{tab:performance_precesion_reduction} (left columns) shows the sample-averaged AUPRC between the scores and ground-truth importance labels, as well as the dataset positive rate. Gemini 2 and 2.5's scores perform the best across all datasets, but GPT-4o mini also demonstrates superior performance compared to classical NLP methods. Smaller, open-source models still perform admirably compared to NLP approaches. Constraining GPT-4o mini to output a binary score rather than a float negatively affects its performance. 

\begin{figure}[t]
    \centering
        \begin{subfigure}{0.6\textwidth}
    \includegraphics[width=1.0\linewidth]{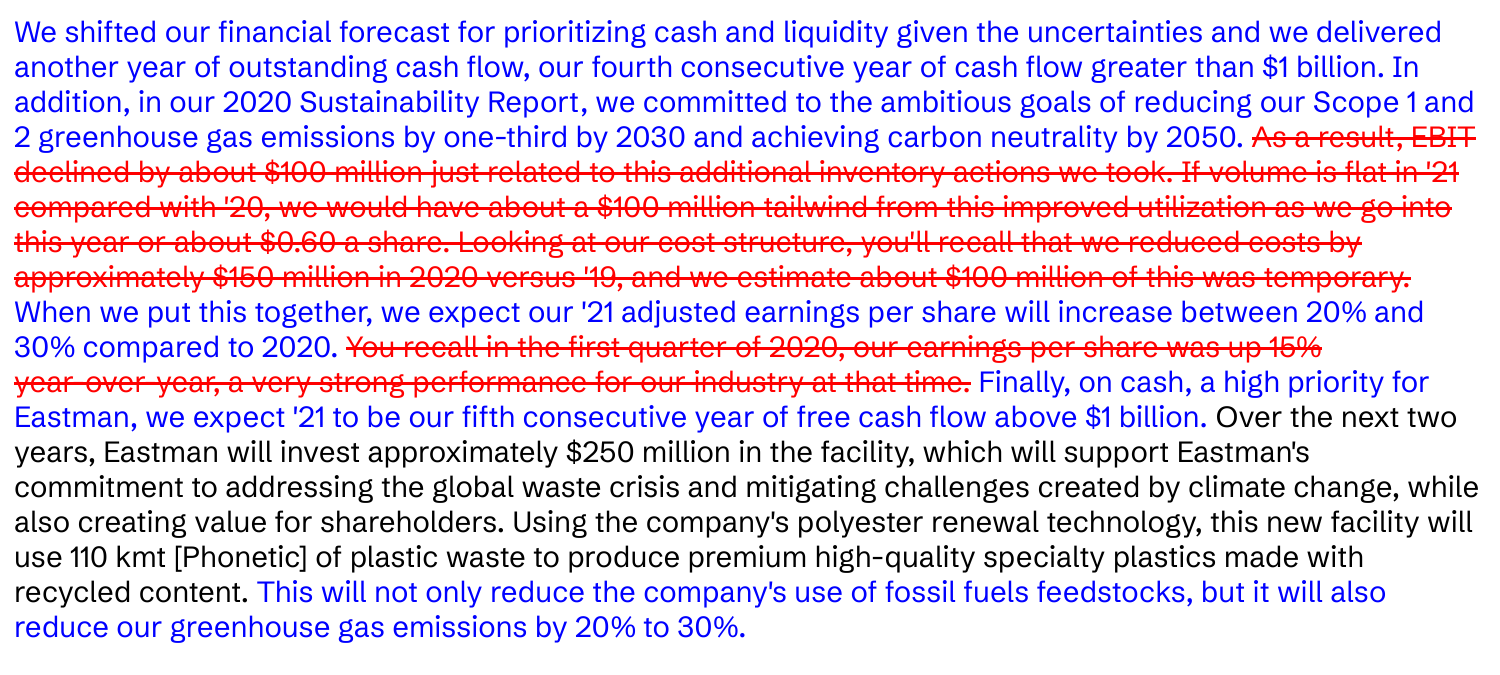}   
        \end{subfigure}
        \begin{subfigure}{0.39\textwidth}
            \includegraphics[width=0.95\linewidth, trim={10 35 10 10}, clip]{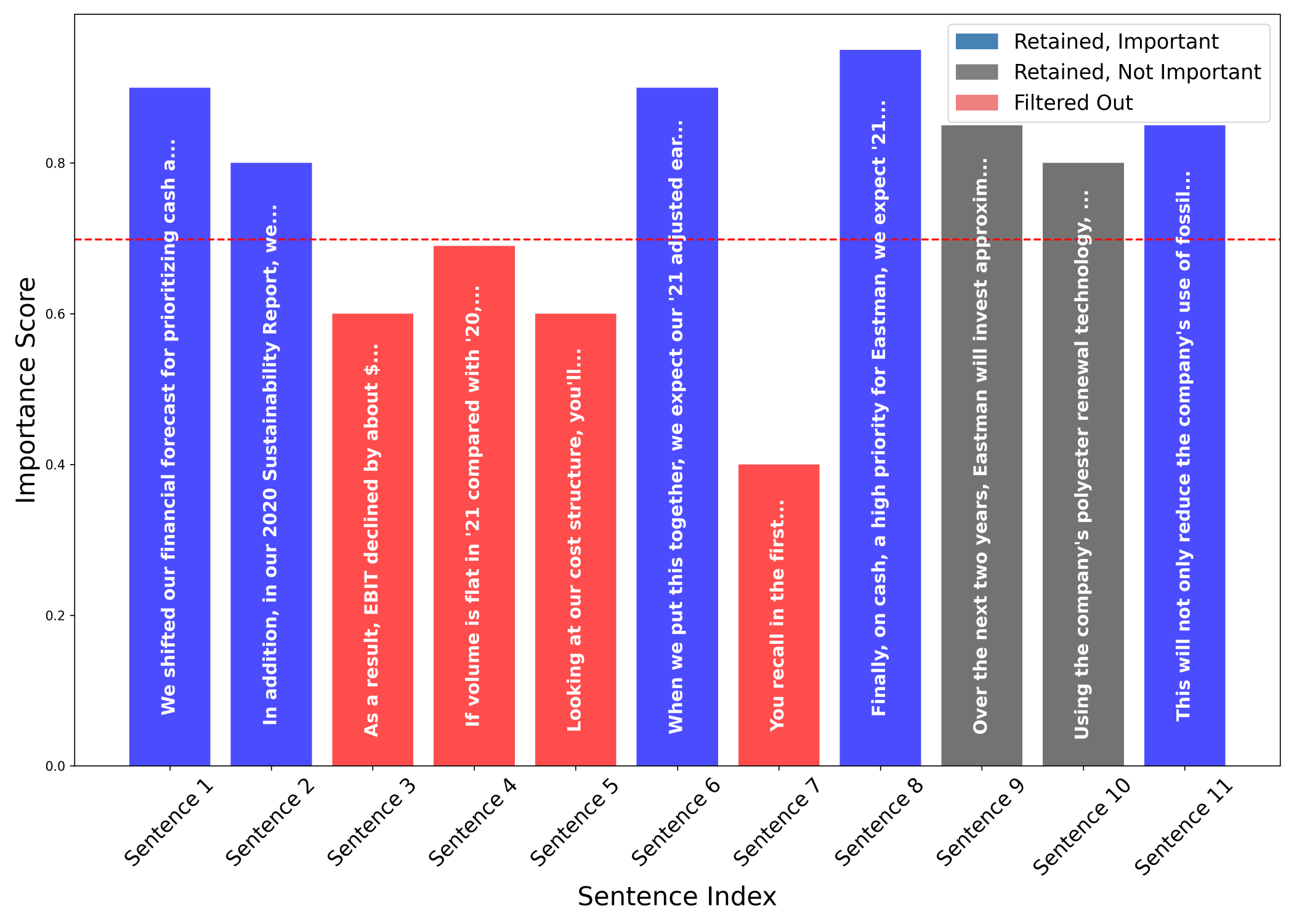}
        \end{subfigure}
    \caption{Example of Conformal Importance Summarization using Gemini 2.5 Flash scores. \textbf{Left:} Ground-truth summary sentences in blue; text with a strikethrough indicates sentences filtered out by Conformal Importance Summarization ($\alpha=0.2$, $\beta=0.8$). On this example, all important sentences were retained ($B(y;y^*)=1$) while $36\%$ of all sentences were filtered out. \textbf{Right:} Sentence-wise importance scores $R(c;x)$ compared to the conformal threshold $\hat q$ (dashed line).}
    \label{fig:example}
    \vspace{-10pt}
\end{figure}

\Cref{tab:performance_precesion_reduction} (right columns) shows the performance of different score functions as measured by the proportion of sentences removed with $\alpha=0.2$ and $\beta=0.8$. Once again, the Gemini models and GPT-4o mini generally perform the best by having the greatest reduction in length for this level of coverage. One example summary  with filtered and retained sentences is shown in \Cref{fig:example}.

\begin{table}[t]
    \centering
        \caption{Performance comparison of importance scoring methods. \textbf{Left:} AUPRC of importance rankings compared to ground truth. AUPRC of the original article indicates the proportion of all sentences labeled as important. \textbf{Right:} Conciseness of summaries (proportion of sentences removed) under Conformal Importance Summarization with $\alpha=0.2$ and $\beta=0.8$. Higher is better.
}
    \label{tab:performance_precesion_reduction}
    \small{
    \centering
    \resizebox{\textwidth}{!}{
        \setlength{\tabcolsep}{3pt} 
    \begin{tabular}{l|cccclc|cccclc}
        \toprule
        Importance Score & \multicolumn{6}{c}{AUPRC $\uparrow$}& \multicolumn{6}{c}{Proportion of Sentences Removed $\uparrow$}\\
        & ECT & CSDS & CNN/DM & TLDR-AIC  &TLDR-Full& MTS & ECT & CSDS & CNN/DM & TLDR-AIC  &TLDR-Full& MTS \\
        \midrule
        Original Article & 0.10 & 0.27 & 0.10 & 0.06  &0.014& 0.81 & 0.00 & 0.00 &0.00 &0.00  &0.00&0.00\\
        \midrule
        Cos. Sim. Centrality & 0.22 & 0.34 & 0.34 & 0.35  &0.14& 0.86 & 0.22 & 0.11 & 0.18 & 0.29  &0.5& 0.18 \\
        Sentence Centrality & 0.14 & 0.34 & 0.29 & 0.28  &0.09& 0.86 & 0.17 & 0.08 & 0.22 & 0.30  &0.50& 0.10 \\
        GUSUM & 0.21 & 0.44 & 0.33 & 0.21  &0.09& 0.90 & 0.11 & 0.24 & 0.27 & 0.15  &0.20& 0.13 \\
        LexRank & 0.22 & 0.43 & 0.32 & 0.32  &0.14& *\tablefootnote{LexRank's bag of words embedding step fails on this dataset, hence no scores are available.}& 0.16 & 0.12 & 0.20 & 0.37  &0.35& *$^4$\\
        \midrule
        GPT-4o mini (binary) & 0.12 & 0.34 & 0.13 & 0.08  &0.02& 0.83 & 0.24 & 0.22 & 0.26 & 0.22  &0.28& 0.08 \\
        GPT-4o mini & 0.30 & 0.49 & 0.34 & 0.33  &0.20& 0.93 & 0.24 & 0.25 & \textbf{0.30} & 0.40  &0.26& 0.16 \\
        Llama3-8B & 0.18 & 0.39 & 0.22 & 0.15  &0.05& 0.92 & 0.13 & 0.11 & 0.14 & 0.11  &0.17& 0.14 \\
        Qwen3-8B & 0.17 & 0.38 & 0.22 & 0.16  &0.04& 0.91 & 0.13 & 0.11 & 0.09 & 0.14  &0.12& \textbf{0.22} \\
        Gemini 2.0 Flash-Lite & 0.35 & 0.68 & \textbf{0.42} & \textbf{0.39}  &\textbf{0.25}& \textbf{0.95} & 0.28 & 0.46 & 0.25 & 0.40  &0.45& 0.13 \\
        Gemini 2.5 Flash & \textbf{0.43} & \textbf{0.69} & 0.36 & 0.34  &0.24& 0.94 & \textbf{0.37} & \textbf{0.49} & 0.26 & \textbf{0.41}  &\textbf{0.47}& 0.14 \\
        \bottomrule
 
    \end{tabular}
    }
    }
    \vspace{-8pt}
\end{table}

\subsection{Ablations}\label{sec:ablations}
    \vspace{-4pt}
First, we perform ablations over the labeling method for datasets lacking explicit ground truth. Specifically, in \Cref{app:rouge_score} we test ROUGE-1, -2, and -L scores \cite{lin-2004-rouge} instead of SBERT cosine similarity, in conjunction with \Cref{alg:greedy-extractive} to create importance labels, then re-evaluate AUPRC and conciseness performance for all importance scoring methods. Aside from small changes in the scores, the performance trend is similar, and no fundamental difference in our conclusions would have been made with another labeling method. 

Our method relies on a labeled calibration set which could be expensive to curate. In \Cref{app:cal_set_size} we perform an ablation over the calibration set size $n$. Compared to $n=100$ used throughout this work, having as few as 50 labeled examples can still produce comparable results.

    \vspace{-4pt}
\subsection{Comparison to Direct Abstractive and Hybrid Extractive-Abstractive Summarization}\label{subsec:extractive_abstractive_comparison}
    \vspace{-4pt}

Next, we evaluate existing LLMs prompted to directly summarize text while retaining at least a fraction $\beta$ of important information. To give these models guidance on \emph{what} information is considered important, we add 10 examples from the calibration set to the prompt, enabling in-context learning \cite{NEURIPS2020_1457c0d6}. We also test our proposed hybrid pipeline from \Cref{sec:method} where an LLM starts with our extractive summary and is prompted to retain \emph{all} information. If successful, this would preserve the empirical coverage level, while improving paragraph flow and potentially shortening the text further. We use Gemini 2.5 Flash for the extractive component, as it showed the strongest results in \Cref{tab:performance_precesion_reduction}, and test GPT-4o mini as the abstractive model (similar plots for Gemini 2.5 Flash are given in \Cref{app:extractive_abstractive_comparison}).

\begin{figure}[t]
    \centering
        \begin{subfigure}{0.32\textwidth}
        \includegraphics[width=1.0\linewidth, trim={8 10 10 10}, clip]{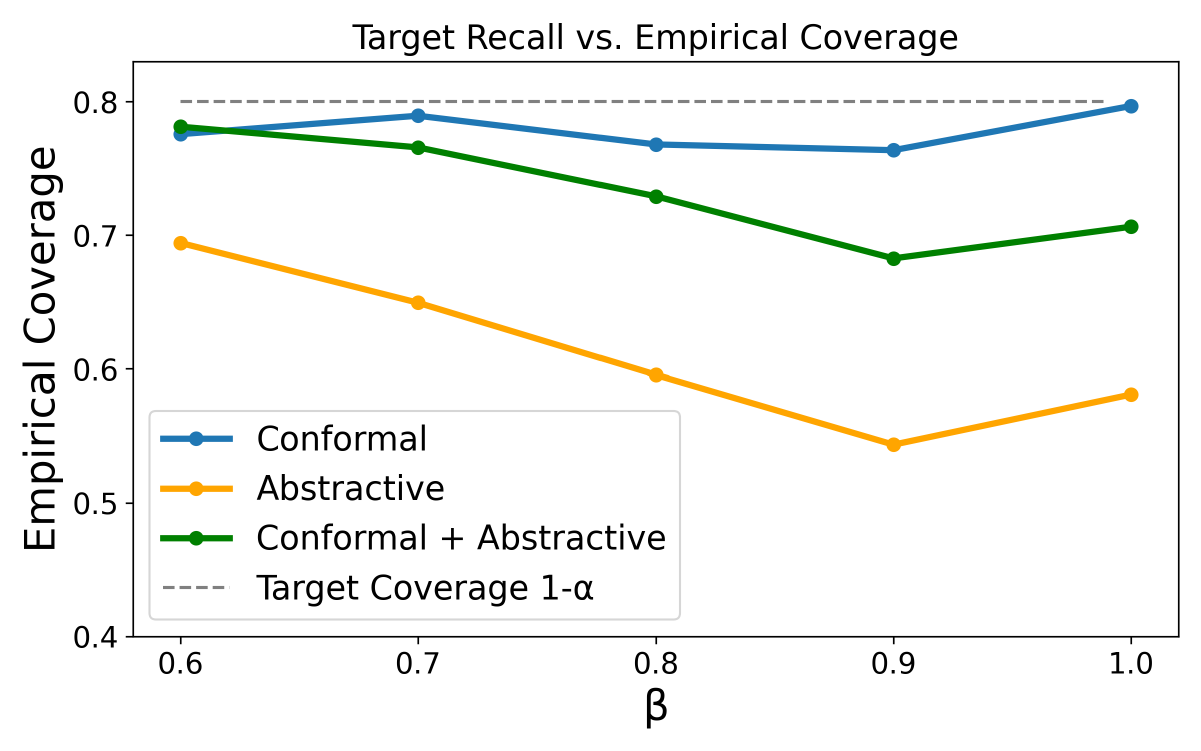}
        \end{subfigure} \
        \begin{subfigure}{0.32\textwidth}
            \includegraphics[width=1.0\linewidth, trim={8 10 10 10}, clip]{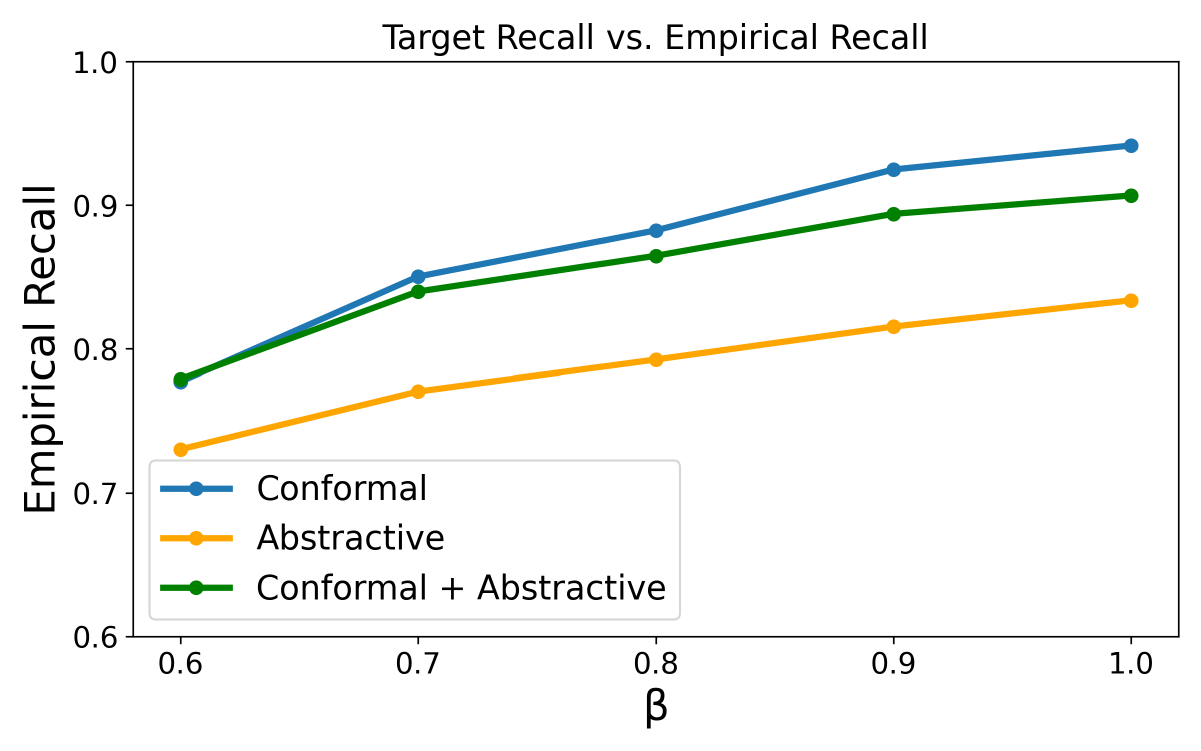}
        \end{subfigure} \
                \begin{subfigure}{0.32\textwidth}
            \includegraphics[width=1.0\linewidth, trim={8 10 10 10}, clip]{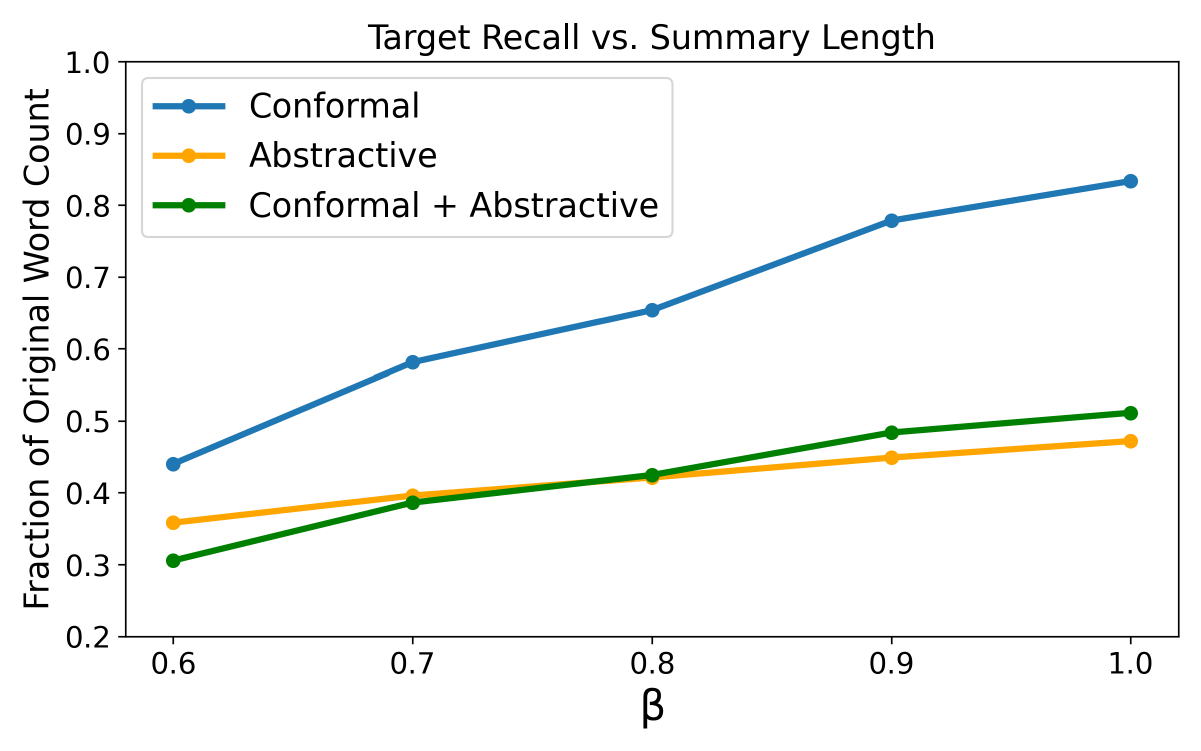}
        \end{subfigure}
        \caption{Comparison between extractive summarization with our method, abstractive summarization with an LLM, and our hybrid proposal on ECTSum. Here the target coverage is $1-\alpha=0.8$, the conformal approach uses Gemini 2.5 Flash scoring, and the abstractive model is GPT-4o mini.}
\label{fig:extractive_abstractive_comparison_gpt}
\vspace{-10pt}
\end{figure}

Unlike extractive summarization, in abstractive summarization determining whether the important aspects of a ground-truth sentence have been preserved is subjective. Hence, we use a proxy for recall $B(y; y^*)$ based on semantic entailment: each ground-truth important sentence in $y^*$ is compared to the  generated summary $y$ using an LLM-based evaluator to check if the important aspects of the sentence have been retained. In practice we use GPT-4o mini with prompts given in Appendix \ref{app:GPTprompt}.

In \Cref{fig:extractive_abstractive_comparison_gpt} using ECTSum we first observe that direct prompting to retain a fraction $\beta$ of important information is not effective, achieving coverage on only 55-70\% of summaries. While the hybrid approach has lower coverage than conformal alone, it greatly outperforms direct prompting. Notably, its improved recall and coverage are achieved with the same level of conciseness as direct prompting. We attribute this to how the hybrid approach separates out the two subtasks of identifying important information, and synthesizing it into a concise summary. Perhaps surprisingly, we note that direct prompting for different levels of $\beta$ does give some level of control over recall.

\vspace{-4pt}
\section{Conclusions and Limitations}
\label{sec:conclusion}
\vspace{-4pt}

Our results show that Conformal Importance Summarization provides distribution-free coverage guarantees for extractive summarization. It allows integration with both LLMs and classical NLP techniques to achieve flexible and reliable summary generation with control over the tradeoff between completeness and conciseness. Furthermore, we show that these results hold with a variety of importance scores and labeling methods, which could reflect different user preferences for importance.

Our experiments suggest that recent LLMs are well-suited to judge sentence-level importance. LLMs prompted to directly judge importance can achieve higher sentence removal rates than our conformal method for a single recall level, but do not provide any control over the desired recall or conciseness. An LLM prompted to directly summarize inputs while retaining all important information tends to produce very concise summaries, but with much lower coverage and recall than requested. Separating the subtasks of judging importance and synthesizing information greatly improves recall without sacrificing conciseness. Since we did not perform extensive prompt tuning, we suspect that greater performance could be achieved with our method through tuning, or through reasoning models such as Deepseek-R1 \citep{guo2025deepseek}.

\looseness=-1 While our benchmark suite spanned the domains of finance, customer service, news, science, and medicine with both English and Chinese examples, we note that it only contained two datasets with explicit ground-truth labels for extractive summarization, while the other three datasets required label refinement. Our datasets are somewhat limited in maximum length - with the exception being SciTLDR-Full at an average length of 216 sentences. Meanwhile, electronic health records may exceed 200,000 words \citep{croxford2025development}. Implementing Conformal Importance Summarization for such a problem would be a valuable step towards validating its practical utility and establishing standards for $\alpha$ and $\beta$.

Towards the goal of summarizing longer documents, it is possible to extend our framework to spans of text other than sentences, which we focused on for convenience. Simply break the long-text into spans, score their importance, and filter out spans with scores lower than the calibrated threshold. For example, paragraph-sized spans would reduce the number of spans that need to be scored for long documents. Due to the lack of existing datasets with long documents and labeled extractive summaries at the paragraph level, we have not performed these experiments.

Future work could extend our framework to more general forms of labels. One possibility is to perform extractive summarization with non-binary labels, such as an annotator's Likert scale rating of importance, and provide guarantees of including sentences with a specified fraction of the total importance weight. A more ambitious goal is to extend the framework fully to abstractive summarization. This would enable us to better leverage LLMs' natural abstractive capabilities, allow better evaluations on a wide range of datasets, and provide more natural sounding summaries.

\clearpage 

\bibliographystyle{plainnat}
\bibliography{bib}

\begin{thebibliography}{68}
\providecommand{\natexlab}[1]{#1}
\providecommand{\url}[1]{\texttt{#1}}
\expandafter\ifx\csname urlstyle\endcsname\relax
  \providecommand{\doi}[1]{doi: #1}\else
  \providecommand{\doi}{doi: \begingroup \urlstyle{rm}\Url}\fi

\bibitem[Achiam et~al.(2023)]{Achiam2023GPT4TR}
Josh Achiam et~al.
\newblock {GPT-4 Technical Report}.
\newblock \emph{arXiv:2303.08774}, 2023.

\bibitem[Angelopoulos and Bates(2021)]{angelopoulos2022gentle}
Anastasios~N. Angelopoulos and Stephen Bates.
\newblock A gentle introduction to conformal prediction and distribution-free uncertainty quantification.
\newblock \emph{arXiv:2107.07511}, 2021.

\bibitem[Angelopoulos et~al.(2021)Angelopoulos, Bates, Jordan, and Malik]{angelopoulos2021raps}
Anastasios~N. Angelopoulos, Stephen Bates, Michael Jordan, and Jitendra Malik.
\newblock Uncertainty sets for image classifiers using conformal prediction.
\newblock In \emph{International Conference on Learning Representations}, 2021.

\bibitem[Angelopoulos et~al.(2024)Angelopoulos, Bates, Fisch, Lei, and Schuster]{angelopoulos2024conformal}
Anastasios~Nikolas Angelopoulos, Stephen Bates, Adam Fisch, Lihua Lei, and Tal Schuster.
\newblock Conformal risk control.
\newblock In \emph{The Twelfth International Conference on Learning Representations}, 2024.

\bibitem[Asgari et~al.(2024)Asgari, Monta{\~n}a-Brown, Dubois, Khalil, Balloch, and Pimenta]{asgari2024framework}
Elham Asgari, Nina Monta{\~n}a-Brown, Magda Dubois, Saleh Khalil, Jasmine Balloch, and Dominic Pimenta.
\newblock {A framework to assess clinical safety and hallucination rates of LLMs for medical text summarisation}.
\newblock \emph{npj Digital Medicine}, 8\penalty0 (1), 2024.
\newblock \doi{10.1038/s41746-025-01670-7}.

\bibitem[Basu~Mallick and Kilpatrick(2025)]{mallick2025gemini2}
Shrestha Basu~Mallick and Logan Kilpatrick.
\newblock {Gemini 2.0: Flash, Flash-Lite and Pro}.
\newblock \url{https://developers.googleblog.com/en/gemini-2-family-expands/}, February 2025.
\newblock Accessed May 15, 2025.

\bibitem[Ben~Abacha et~al.(2023)Ben~Abacha, Yim, Fan, and Lin]{mts-dialog}
Asma Ben~Abacha, Wen-wai Yim, Yadan Fan, and Thomas Lin.
\newblock An empirical study of clinical note generation from doctor-patient encounters.
\newblock In \emph{Proceedings of the 17th Conference of the European Chapter of the Association for Computational Linguistics}, pages 2291--2302. Association for Computational Linguistics, 2023.
\newblock \doi{10.18653/v1/2023.eacl-main.168}.

\bibitem[Bowman et~al.(2015)Bowman, Angeli, Potts, and Manning]{bowman2015entailment}
Samuel~R. Bowman, Gabor Angeli, Christopher Potts, and Christopher~D. Manning.
\newblock A large annotated corpus for learning natural language inference.
\newblock In \emph{Proceedings of the 2015 Conference on Empirical Methods in Natural Language Processing}, pages 632--642. Association for Computational Linguistics, 2015.
\newblock \doi{10.18653/v1/D15-1075}.

\bibitem[Brown et~al.(2020)Brown, Mann, Ryder, Subbiah, Kaplan, Dhariwal, Neelakantan, Shyam, Sastry, Askell, Agarwal, Herbert-Voss, Krueger, Henighan, Child, Ramesh, Ziegler, Wu, Winter, Hesse, Chen, Sigler, Litwin, Gray, Chess, Clark, Berner, McCandlish, Radford, Sutskever, and Amodei]{NEURIPS2020_1457c0d6}
Tom Brown, Benjamin Mann, Nick Ryder, Melanie Subbiah, Jared~D Kaplan, Prafulla Dhariwal, Arvind Neelakantan, Pranav Shyam, Girish Sastry, Amanda Askell, Sandhini Agarwal, Ariel Herbert-Voss, Gretchen Krueger, Tom Henighan, Rewon Child, Aditya Ramesh, Daniel Ziegler, Jeffrey Wu, Clemens Winter, Chris Hesse, Mark Chen, Eric Sigler, Mateusz Litwin, Scott Gray, Benjamin Chess, Jack Clark, Christopher Berner, Sam McCandlish, Alec Radford, Ilya Sutskever, and Dario Amodei.
\newblock Language models are few-shot learners.
\newblock In \emph{Advances in Neural Information Processing Systems}, volume~33, pages 1877--1901, 2020.

\bibitem[Burnaev and Vovk(2014)]{burnaev2014efficiency}
Evgeny Burnaev and Vladimir Vovk.
\newblock Efficiency of conformalized ridge regression.
\newblock In \emph{Proceedings of The 27th Conference on Learning Theory}, volume~35, pages 605--622, 2014.

\bibitem[Cachola et~al.(2020)Cachola, Lo, Cohan, and Weld]{cachola-etal-2020-tldr}
Isabel Cachola, Kyle Lo, Arman Cohan, and Daniel Weld.
\newblock {TLDR}: Extreme summarization of scientific documents.
\newblock In \emph{Findings of the Association for Computational Linguistics: EMNLP 2020}, pages 4766--4777. Association for Computational Linguistics, 2020.
\newblock \doi{10.18653/v1/2020.findings-emnlp.428}.

\bibitem[Chang et~al.(2024)Chang, Lo, Goyal, and Iyyer]{chang2024booookscore}
Yapei Chang, Kyle Lo, Tanya Goyal, and Mohit Iyyer.
\newblock Booookscore: A systematic exploration of book-length summarization in the era of {LLM}s.
\newblock In \emph{The Twelfth International Conference on Learning Representations}, 2024.

\bibitem[Chen and Bansal(2018)]{chen-bansal-2018-fast}
Yen-Chun Chen and Mohit Bansal.
\newblock Fast abstractive summarization with reinforce-selected sentence rewriting.
\newblock In \emph{Proceedings of the 56th Annual Meeting of the Association for Computational Linguistics (Volume 1: Long Papers)}, pages 675--686, 2018.
\newblock \doi{10.18653/v1/P18-1063}.

\bibitem[Cherian et~al.(2024)Cherian, Gibbs, and Candès]{cherian2024large}
John~J. Cherian, Isaac Gibbs, and Emmanuel~J. Candès.
\newblock Large language model validity via enhanced conformal prediction methods.
\newblock In \emph{Advances in Neural Information Processing Systems}, volume~37, pages 114812--114842, 2024.

\bibitem[Chowdhery et~al.(2023)Chowdhery, Narang, Devlin, Bosma, Mishra, Roberts, Barham, Chung, Sutton, Gehrmann, Schuh, Shi, Tsvyashchenko, Maynez, Rao, Barnes, Tay, Shazeer, Prabhakaran, Reif, Du, Hutchinson, Pope, Bradbury, Austin, Isard, Gur-Ari, Yin, Duke, Levskaya, Ghemawat, Dev, Michalewski, Garcia, Misra, Robinson, Fedus, Zhou, Ippolito, Luan, Lim, Zoph, Spiridonov, Sepassi, Dohan, Agrawal, Omernick, Dai, Pillai, Pellat, Lewkowycz, Moreira, Child, Polozov, Lee, Zhou, Wang, Saeta, Diaz, Firat, Catasta, Wei, Meier-Hellstern, Eck, Dean, Petrov, and Fiedel]{10.5555/3648699.3648939}
Aakanksha Chowdhery, Sharan Narang, Jacob Devlin, Maarten Bosma, Gaurav Mishra, Adam Roberts, Paul Barham, Hyung~Won Chung, Charles Sutton, Sebastian Gehrmann, Parker Schuh, Kensen Shi, Sasha Tsvyashchenko, Joshua Maynez, Abhishek Rao, Parker Barnes, Yi~Tay, Noam Shazeer, Vinodkumar Prabhakaran, Emily Reif, Nan Du, Ben Hutchinson, Reiner Pope, James Bradbury, Jacob Austin, Michael Isard, Guy Gur-Ari, Pengcheng Yin, Toju Duke, Anselm Levskaya, Sanjay Ghemawat, Sunipa Dev, Henryk Michalewski, Xavier Garcia, Vedant Misra, Kevin Robinson, Liam Fedus, Denny Zhou, Daphne Ippolito, David Luan, Hyeontaek Lim, Barret Zoph, Alexander Spiridonov, Ryan Sepassi, David Dohan, Shivani Agrawal, Mark Omernick, Andrew~M. Dai, Thanumalayan~Sankaranarayana Pillai, Marie Pellat, Aitor Lewkowycz, Erica Moreira, Rewon Child, Oleksandr Polozov, Katherine Lee, Zongwei Zhou, Xuezhi Wang, Brennan Saeta, Mark Diaz, Orhan Firat, Michele Catasta, Jason Wei, Kathy Meier-Hellstern, Douglas Eck, Jeff Dean, Slav Petrov, and Noah Fiedel.
\newblock {PaLM: Scaling Language Modeling with Pathways}.
\newblock \emph{Journal of Machine Learning Research}, 24\penalty0 (240):\penalty0 1--113, 2023.

\bibitem[Cresswell et~al.(2024)Cresswell, Sui, Kumar, and Vouitsis]{cresswell2024}
Jesse~C. Cresswell, Yi~Sui, Bhargava Kumar, and No\"{e}l Vouitsis.
\newblock {Conformal Prediction Sets Improve Human Decision Making}.
\newblock In \emph{Proceedings of the 41st International Conference on Machine Learning}, volume 235, pages 9439--9457, 2024.

\bibitem[Cresswell et~al.(2025)Cresswell, Kumar, Sui, and Belbahri]{cresswell2025conformal}
Jesse~C. Cresswell, Bhargava Kumar, Yi~Sui, and Mouloud Belbahri.
\newblock {Conformal Prediction Sets Can Cause Disparate Impact}.
\newblock In \emph{The Thirteenth International Conference on Learning Representations}, 2025.

\bibitem[Croxford et~al.(2025{\natexlab{a}})Croxford, Gao, Pellegrino, Wong, Wills, First, Liao, Goswami, Patterson, and Afshar]{croxford2025current}
Emma Croxford, Yanjun Gao, Nicholas Pellegrino, Karen Wong, Graham Wills, Elliot First, Frank Liao, Cherodeep Goswami, Brian Patterson, and Majid Afshar.
\newblock Current and future state of evaluation of large language models for medical summarization tasks.
\newblock \emph{npj Health Systems}, 2\penalty0 (1):\penalty0 6, 2025{\natexlab{a}}.

\bibitem[Croxford et~al.(2025{\natexlab{b}})Croxford, Gao, Pellegrino, Wong, Wills, First, Schnier, Burton, Ebby, Gorskic, Kalscheur, Khalil, Pisani, Rubeor, Stetson, Liao, Goswami, Patterson, and Afshar]{croxford2025development}
Emma Croxford, Yanjun Gao, Nicholas Pellegrino, Karen~K. Wong, Graham Wills, Elliot First, Miranda Schnier, Kyle Burton, Cris~G. Ebby, Jillian Gorskic, Matthew Kalscheur, Samy Khalil, Marie Pisani, Tyler Rubeor, Peter Stetson, Frank Liao, Cherodeep Goswami, Brian Patterson, and Majid Afshar.
\newblock Development and validation of the provider documentation summarization quality instrument for large language models.
\newblock \emph{Journal of the American Medical Informatics Association}, pages 1050--1060, 2025{\natexlab{b}}.

\bibitem[Deutsch et~al.(2022)Deutsch, Dror, and Roth]{deutsch-etal-2022-limitations}
Daniel Deutsch, Rotem Dror, and Dan Roth.
\newblock On the limitations of reference-free evaluations of generated text.
\newblock In \emph{Proceedings of the 2022 Conference on Empirical Methods in Natural Language Processing}, pages 10960--10977. Association for Computational Linguistics, 2022.
\newblock \doi{10.18653/v1/2022.emnlp-main.753}.

\bibitem[Devlin et~al.(2019)Devlin, Chang, Lee, and Toutanova]{devlin2019bertpretrainingdeepbidirectional}
Jacob Devlin, Ming-Wei Chang, Kenton Lee, and Kristina Toutanova.
\newblock {BERT: Pre-training of Deep Bidirectional Transformers for Language Understanding}.
\newblock In \emph{Proceedings of the 2019 Conference of the North {A}merican Chapter of the Association for Computational Linguistics: Human Language Technologies, Volume 1 (Long and Short Papers)}, 2019.
\newblock \doi{10.18653/v1/N19-1423}.

\bibitem[Erkan and Radev(2004)]{erkan2004lexrank}
G{\"u}nes Erkan and Dragomir~R Radev.
\newblock Lexrank: Graph-based lexical centrality as salience in text summarization.
\newblock \emph{Journal of Artificial Intelligence Research}, 22:\penalty0 457--479, 2004.

\bibitem[Feng et~al.(2025)Feng, Sui, Hou, Cresswell, and Wu]{feng2025}
Naihe Feng, Yi~Sui, Shiyi Hou, Jesse~C. Cresswell, and Ga~Wu.
\newblock Response quality assessment for retrieval-augmented generation via conditional conformal factuality.
\newblock In \emph{Proceedings of the 48th International ACM SIGIR Conference on Research and Development in Information Retrieval}, page 2832–2836, 2025.
\newblock ISBN 9798400715921.
\newblock \doi{10.1145/3726302.3730244}.
\newblock URL \url{https://doi.org/10.1145/3726302.3730244}.

\bibitem[Gehrmann et~al.(2018)Gehrmann, Deng, and Rush]{gehrmann-etal-2018-bottom}
Sebastian Gehrmann, Yuntian Deng, and Alexander Rush.
\newblock {Bottom-Up Abstractive Summarization}.
\newblock In \emph{Proceedings of the 2018 Conference on Empirical Methods in Natural Language Processing}, 2018.
\newblock \doi{10.18653/v1/D18-1443}.

\bibitem[Ghatnekar et~al.(2021)Ghatnekar, Faletsky, and Nambudiri]{ghatnekar2021digital}
Shilpa Ghatnekar, Adam Faletsky, and Vinod~E Nambudiri.
\newblock Digital scribe utility and barriers to implementation in clinical practice: A scoping review.
\newblock \emph{Health and Technology}, 11\penalty0 (4):\penalty0 803--809, 2021.

\bibitem[Gokhan et~al.(2022)Gokhan, Smith, and Lee]{gokhan2022gusum}
Tuba Gokhan, Phillip Smith, and Mark Lee.
\newblock {GUSUM: Graph-based Unsupervised Summarization Using Sentence Features Scoring and Sentence-BERT}.
\newblock In \emph{Proceedings of TextGraphs-16: Graph-based Methods for Natural Language Processing}, pages 44--53. Association for Computational Linguistics, 2022.

\bibitem[Gong et~al.(2022)Gong, Zhu, Qi, Tong, Lu, and Wu]{gong2022improving}
Shuai Gong, Zhenfang Zhu, Jiangtao Qi, Chunling Tong, Qiang Lu, and Wenqing Wu.
\newblock {Improving extractive document summarization with sentence centrality}.
\newblock \emph{PLOS ONE}, 17\penalty0 (7):\penalty0 1--16, 07 2022.
\newblock \doi{10.1371/journal.pone.0268278}.

\bibitem[{Google Developers}(2025)]{google2025gemini25flash}
{Google Developers}.
\newblock {Developers can now start building with Gemini 2.5 Flash}.
\newblock \url{https://blog.google/products/gemini/gemini-2-5-flash-preview/}, April 2025.
\newblock Accessed May 15, 2025.

\bibitem[Grattafiori et~al.(2024)]{grattafiori2024llama}
Aaron Grattafiori et~al.
\newblock {The Llama 3 Herd of Models}.
\newblock \emph{arXiv:2407.21783}, 2024.

\bibitem[Guo et~al.(2025)Guo, Yang, Zhang, Song, Zhang, Xu, Zhu, Ma, Wang, Bi, et~al.]{guo2025deepseek}
Daya Guo, Dejian Yang, Haowei Zhang, Junxiao Song, Ruoyu Zhang, Runxin Xu, Qihao Zhu, Shirong Ma, Peiyi Wang, Xiao Bi, et~al.
\newblock Deepseek-r1: Incentivizing reasoning capability in llms via reinforcement learning.
\newblock \emph{arXiv:2501.12948}, 2025.

\bibitem[Gupta et~al.(2022)Gupta, Kuchibhotla, and Ramdas]{gupta2022nested}
Chirag Gupta, Arun~K. Kuchibhotla, and Aaditya Ramdas.
\newblock Nested conformal prediction and quantile out-of-bag ensemble methods.
\newblock \emph{Pattern Recognition}, 127:\penalty0 108496, 2022.
\newblock ISSN 0031-3203.
\newblock \doi{10.1016/j.patcog.2021.108496}.

\bibitem[Hermann et~al.(2015)Hermann, Kocisky, Grefenstette, Espeholt, Kay, Suleyman, and Blunsom]{10.5555/2969239.2969428}
Karl~Moritz Hermann, Tomas Kocisky, Edward Grefenstette, Lasse Espeholt, Will Kay, Mustafa Suleyman, and Phil Blunsom.
\newblock Teaching machines to read and comprehend.
\newblock In \emph{Advances in Neural Information Processing Systems}, volume~28, 2015.

\bibitem[Huang et~al.(2024)Huang, Xi, Zhang, Yao, Qiu, and Wei]{huang2024saps}
Jianguo Huang, Huajun Xi, Linjun Zhang, Huaxiu Yao, Yue Qiu, and Hongxin Wei.
\newblock Conformal prediction for deep classifier via label ranking.
\newblock In \emph{Proceedings of the 41st International Conference on Machine Learning}, 2024.

\bibitem[Kanapala et~al.(2019)Kanapala, Pal, and Pamula]{kanapala2019text}
Ambedkar Kanapala, Sukomal Pal, and Rajendra Pamula.
\newblock Text summarization from legal documents: A survey.
\newblock \emph{Artificial Intelligence Review}, 51:\penalty0 371--402, 2019.

\bibitem[Koh et~al.(2022)Koh, Ju, Liu, and Pan]{koh2022survey}
Huan~Yee Koh, Jiaxin Ju, Ming Liu, and Shirui Pan.
\newblock An empirical survey on long document summarization: Datasets, models, and metrics.
\newblock \emph{ACM Comput. Surv.}, 55\penalty0 (8), December 2022.
\newblock ISSN 0360-0300.
\newblock \doi{10.1145/3545176}.

\bibitem[Kryscinski et~al.(2020)Kryscinski, McCann, Xiong, and Socher]{kryscinski-etal-2020-evaluating}
Wojciech Kryscinski, Bryan McCann, Caiming Xiong, and Richard Socher.
\newblock Evaluating the factual consistency of abstractive text summarization.
\newblock In \emph{Proceedings of the 2020 Conference on Empirical Methods in Natural Language Processing (EMNLP)}, pages 9332--9346, 2020.
\newblock \doi{10.18653/v1/2020.emnlp-main.750}.

\bibitem[Kumar et~al.(2023)Kumar, Lu, Gupta, Palepu, Bellamy, Raskar, and Beam]{kumar2023conformal}
Bhawesh Kumar, Charlie Lu, Gauri Gupta, Anil Palepu, David Bellamy, Ramesh Raskar, and Andrew Beam.
\newblock Conformal prediction with large language models for multi-choice question answering.
\newblock \emph{arXiv:2305.18404}, 2023.

\bibitem[Lewis et~al.(2020)Lewis, Perez, Piktus, Petroni, Karpukhin, Goyal, K\"{u}ttler, Lewis, Yih, Rockt\"{a}schel, Riedel, and Kiela]{lewis2020rag}
Patrick Lewis, Ethan Perez, Aleksandra Piktus, Fabio Petroni, Vladimir Karpukhin, Naman Goyal, Heinrich K\"{u}ttler, Mike Lewis, Wen-tau Yih, Tim Rockt\"{a}schel, Sebastian Riedel, and Douwe Kiela.
\newblock {Retrieval-Augmented Generation for Knowledge-Intensive NLP Tasks}.
\newblock In \emph{Advances in Neural Information Processing Systems}, volume~33, pages 9459--9474, 2020.

\bibitem[Lin(2004)]{lin-2004-rouge}
Chin-Yew Lin.
\newblock {ROUGE}: A package for automatic evaluation of summaries.
\newblock In \emph{Text Summarization Branches Out}, pages 74--81, Barcelona, Spain, July 2004. Association for Computational Linguistics.

\bibitem[Lin et~al.(2021)Lin, Ma, Zhu, Xiang, Zhou, Zhang, and Zong]{lin-etal-2021-csds}
Haitao Lin, Liqun Ma, Junnan Zhu, Lu~Xiang, Yu~Zhou, Jiajun Zhang, and Chengqing Zong.
\newblock {CSDS: A Fine-Grained Chinese Dataset for Customer Service Dialogue Summarization}.
\newblock In \emph{Proceedings of the 2021 Conference on Empirical Methods in Natural Language Processing}, pages 4436--4451, 2021.
\newblock \doi{10.18653/v1/2021.emnlp-main.365}.

\bibitem[Liu and Lapata(2019{\natexlab{a}})]{liu-lapata-2019-text}
Yang Liu and Mirella Lapata.
\newblock Text summarization with pretrained encoders.
\newblock In \emph{Proceedings of the 2019 Conference on Empirical Methods in Natural Language Processing and the 9th International Joint Conference on Natural Language Processing (EMNLP-IJCNLP)}, pages 3730--3740, 2019{\natexlab{a}}.
\newblock \doi{10.18653/v1/D19-1387}.

\bibitem[Liu and Lapata(2019{\natexlab{b}})]{liu2019textsummarizationpretrainedencoders}
Yang Liu and Mirella Lapata.
\newblock Text summarization with pretrained encoders.
\newblock In \emph{Proceedings of the 2019 Conference on Empirical Methods in Natural Language Processing and the 9th International Joint Conference on Natural Language Processing (EMNLP-IJCNLP)}, pages 3730--3740, 2019{\natexlab{b}}.
\newblock \doi{10.18653/v1/D19-1387}.

\bibitem[Messoudi et~al.(2021)Messoudi, Destercke, and Rousseau]{MESSOUDI2021108101}
Soundouss Messoudi, Sébastien Destercke, and Sylvain Rousseau.
\newblock Copula-based conformal prediction for multi-target regression.
\newblock \emph{Pattern Recognition}, 120:\penalty0 108101, 2021.
\newblock ISSN 0031-3203.
\newblock \doi{https://doi.org/10.1016/j.patcog.2021.108101}.

\bibitem[Mihalcea and Tarau(2004)]{mihalcea-tarau-2004-textrank}
Rada Mihalcea and Paul Tarau.
\newblock {T}ext{R}ank: Bringing order into text.
\newblock In \emph{Proceedings of the 2004 Conference on Empirical Methods in Natural Language Processing}, pages 404--411. Association for Computational Linguistics, July 2004.

\bibitem[Mohri and Hashimoto(2024)]{mohri2024factuality}
Christopher Mohri and Tatsunori Hashimoto.
\newblock Language models with conformal factuality guarantees.
\newblock In \emph{Proceedings of the 41st International Conference on Machine Learning}, 2024.

\bibitem[Mukherjee et~al.(2022)Mukherjee, Bohra, Banerjee, Sharma, Hegde, Shaikh, Shrivastava, Dasgupta, Ganguly, Ghosh, and Goyal]{mukherjee-etal-2022-ectsum}
Rajdeep Mukherjee, Abhinav Bohra, Akash Banerjee, Soumya Sharma, Manjunath Hegde, Afreen Shaikh, Shivani Shrivastava, Koustuv Dasgupta, Niloy Ganguly, Saptarshi Ghosh, and Pawan Goyal.
\newblock {{ECTSum: A New Benchmark Dataset For Bullet Point Summarization of Long Earnings Call Transcripts}}.
\newblock In \emph{Proceedings of the 2022 Conference on Empirical Methods in Natural Language Processing}, pages 10893--10906, 2022.
\newblock \doi{10.18653/v1/2022.emnlp-main.748}.

\bibitem[{OpenAI}(2025)]{openai2024gpt4omini}
{OpenAI}.
\newblock {GPT-4o mini: Advancing cost-efficient intelligence}.
\newblock \url{https://openai.com/index/gpt-4o-mini-advancing-cost-efficient-intelligence/}, 2025.
\newblock Accessed May 15, 2025.

\bibitem[Page et~al.(1999)Page, Brin, Motwani, and Winograd]{page1999pagerank}
Lawrence Page, Sergey Brin, Rajeev Motwani, and Terry Winograd.
\newblock {The PageRank citation ranking: Bringing order to the web}.
\newblock Technical report, Stanford Infolab, 1999.

\bibitem[Panickssery et~al.(2024)Panickssery, Bowman, and Feng]{NEURIPS2024_7f1f0218}
Arjun Panickssery, Samuel~R. Bowman, and Shi Feng.
\newblock {LLM Evaluators Recognize and Favor Their Own Generations}.
\newblock In \emph{Advances in Neural Information Processing Systems}, volume~37, pages 68772--68802, 2024.

\bibitem[Paulus et~al.(2018)Paulus, Xiong, and Socher]{DBLP:conf/iclr/PaulusXS18}
Romain Paulus, Caiming Xiong, and Richard Socher.
\newblock A deep reinforced model for abstractive summarization.
\newblock In \emph{International Conference on Learning Representations}, 2018.

\bibitem[Preti et~al.(2024)Preti, Giannone, Favalli, and Romagnoli]{preti2024automatic}
David Preti, Cristina Giannone, Andrea Favalli, and Raniero Romagnoli.
\newblock {Automatic Summarization of Legal Texts, Extractive Summarization using LLMs}.
\newblock \emph{Ital-IA 2024: 4th National Conference on Artificial Intelligence}, 2024.

\bibitem[Quach et~al.(2024)Quach, Fisch, Schuster, Yala, Sohn, Jaakkola, and Barzilay]{quach2024conformal}
Victor Quach, Adam Fisch, Tal Schuster, Adam Yala, Jae~Ho Sohn, Tommi~S. Jaakkola, and Regina Barzilay.
\newblock Conformal language modeling.
\newblock In \emph{The Twelfth International Conference on Learning Representations}, 2024.

\bibitem[Ramirez-Orta and Milios(2021)]{ramirez-orta-milios-2021-unsupervised}
Juan Ramirez-Orta and Evangelos Milios.
\newblock Unsupervised document summarization using pre-trained sentence embeddings and graph centrality.
\newblock In \emph{Proceedings of the Second Workshop on Scholarly Document Processing}, pages 110--115. Association for Computational Linguistics, 2021.
\newblock \doi{10.18653/v1/2021.sdp-1.14}.

\bibitem[Reimers and Gurevych(2019)]{reimers-gurevych-2019-sentence}
Nils Reimers and Iryna Gurevych.
\newblock {Sentence-{BERT}: Sentence Embeddings using {S}iamese {BERT}-Networks}.
\newblock In \emph{Proceedings of the 2019 Conference on Empirical Methods in Natural Language Processing and the 9th International Joint Conference on Natural Language Processing (EMNLP-IJCNLP)}, pages 3982--3992. Association for Computational Linguistics, 2019.
\newblock \doi{10.18653/v1/D19-1410}.

\bibitem[Romano et~al.(2019)Romano, Patterson, and Candès]{romano2019conformalized}
Yaniv Romano, Evan Patterson, and Emmanuel Candès.
\newblock Conformalized quantile regression.
\newblock In \emph{Advances in Neural Information Processing Systems}, volume~32, 2019.

\bibitem[Romano et~al.(2020)Romano, Sesia, and Candès]{romano2020aps}
Yaniv Romano, Matteo Sesia, and Emmanuel Candès.
\newblock Classification with valid and adaptive coverage.
\newblock In \emph{Advances in Neural Information Processing Systems}, volume~33, 2020.

\bibitem[Rubin-Toles et~al.(2025)Rubin-Toles, Gambhir, Ramji, Roth, and Goel]{rubin2025conformal}
Maxon Rubin-Toles, Maya Gambhir, Keshav Ramji, Aaron Roth, and Surbhi Goel.
\newblock Conformal language model reasoning with coherent factuality.
\newblock In \emph{The Thirteenth International Conference on Learning Representations}, 2025.

\bibitem[Sanh et~al.(2022)Sanh, Webson, Raffel, Bach, Sutawika, Alyafeai, Chaffin, Stiegler, Raja, Dey, Bari, Xu, Thakker, Sharma, Szczechla, Kim, Chhablani, Nayak, Datta, Chang, Jiang, Wang, Manica, Shen, Yong, Pandey, Bawden, Wang, Neeraj, Rozen, Sharma, Santilli, Fevry, Fries, Teehan, Scao, Biderman, Gao, Wolf, and Rush]{sanh2022multitask}
Victor Sanh, Albert Webson, Colin Raffel, Stephen Bach, Lintang Sutawika, Zaid Alyafeai, Antoine Chaffin, Arnaud Stiegler, Arun Raja, Manan Dey, M~Saiful Bari, Canwen Xu, Urmish Thakker, Shanya~Sharma Sharma, Eliza Szczechla, Taewoon Kim, Gunjan Chhablani, Nihal Nayak, Debajyoti Datta, Jonathan Chang, Mike Tian-Jian Jiang, Han Wang, Matteo Manica, Sheng Shen, Zheng~Xin Yong, Harshit Pandey, Rachel Bawden, Thomas Wang, Trishala Neeraj, Jos Rozen, Abheesht Sharma, Andrea Santilli, Thibault Fevry, Jason~Alan Fries, Ryan Teehan, Teven~Le Scao, Stella Biderman, Leo Gao, Thomas Wolf, and Alexander~M Rush.
\newblock Multitask prompted training enables zero-shot task generalization.
\newblock In \emph{International Conference on Learning Representations}, 2022.

\bibitem[Shafer and Vovk(2008)]{shafer2008tutorial}
Glenn Shafer and Vladimir Vovk.
\newblock A tutorial on conformal prediction.
\newblock \emph{Journal of Machine Learning Research}, 9\penalty0 (3), 2008.

\bibitem[Shah et~al.(2025)Shah, Crowell, Jeong, Devon-Sand, Smith, Yang, Ma, Liang, Delahaie, Hsia, Shanafelt, Pfeffer, Sharp, Lin, and Garcia]{shah2025physician}
Shreya~J. Shah, Trevor Crowell, Yejin Jeong, Anna Devon-Sand, Margaret Smith, Betsy Yang, Stephen~P. Ma, April~S. Liang, Clarissa Delahaie, Caroline Hsia, Tait Shanafelt, Michael~A. Pfeffer, Christopher Sharp, Steven Lin, and Patricia Garcia.
\newblock {Physician Perspectives on Ambient AI Scribes}.
\newblock \emph{JAMA Network Open}, 8\penalty0 (3):\penalty0 e251904--e251904, 2025.

\bibitem[Vaswani et~al.(2017)Vaswani, Shazeer, Parmar, Uszkoreit, Jones, Gomez, Kaiser, and Polosukhin]{Vaswani2017Jun}
Ashish Vaswani, Noam Shazeer, Niki Parmar, Jakob Uszkoreit, Llion Jones, Aidan~N Gomez, {\L}ukasz Kaiser, and Illia Polosukhin.
\newblock Attention is all you need.
\newblock In \emph{Advances in Neural Information Processing Systems}, volume~30, 2017.

\bibitem[Vovk et~al.(2005)Vovk, Gammerman, and Shafer]{vovk2005algorithmic}
Vladimir Vovk, Alexander Gammerman, and Glenn Shafer.
\newblock \emph{Algorithmic Learning in a Random World}.
\newblock Springer, 2005.

\bibitem[Wei et~al.(2022)Wei, Bosma, Zhao, Guu, Yu, Lester, Du, Dai, and Le]{wei2022finetuned}
Jason Wei, Maarten Bosma, Vincent Zhao, Kelvin Guu, Adams~Wei Yu, Brian Lester, Nan Du, Andrew~M. Dai, and Quoc~V Le.
\newblock Finetuned language models are zero-shot learners.
\newblock In \emph{International Conference on Learning Representations}, 2022.

\bibitem[Yang et~al.(2024)Yang, Yang, Zhang, Hui, Zheng, Yu, Li, Liu, Huang, Wei, Lin, Yang, Tu, Zhang, Yang, Yang, Zhou, Lin, Dang, Lu, Bao, Yang, Yu, Li, Xue, Zhang, Zhu, Men, Lin, Li, Tang, Xia, Ren, Ren, Fan, Su, Zhang, Wan, Liu, Cui, Zhang, and Qiu]{yang2024qwen2}
An~Yang, Baosong Yang, Beichen Zhang, Binyuan Hui, Bo~Zheng, Bowen Yu, Chengyuan Li, Dayiheng Liu, Fei Huang, Haoran Wei, Huan Lin, Jian Yang, Jianhong Tu, Jianwei Zhang, Jianxin Yang, Jiaxi Yang, Jingren Zhou, Junyang Lin, Kai Dang, Keming Lu, Keqin Bao, Kexin Yang, Le~Yu, Mei Li, Mingfeng Xue, Pei Zhang, Qin Zhu, Rui Men, Runji Lin, Tianhao Li, Tianyi Tang, Tingyu Xia, Xingzhang Ren, Xuancheng Ren, Yang Fan, Yang Su, Yichang Zhang, Yu~Wan, Yuqiong Liu, Zeyu Cui, Zhenru Zhang, and Zihan Qiu.
\newblock {Qwen2.5 Technical Report}.
\newblock \emph{arXiv:2412.15115}, 2024.

\bibitem[Yang et~al.(2025)Yang, Li, Yang, Zhang, Hui, Zheng, Yu, Gao, Huang, Lv, Zheng, Liu, Zhou, Huang, Hu, Ge, Wei, Lin, Tang, Yang, Tu, Zhang, Yang, Yang, Zhou, Zhou, Lin, Dang, Bao, Yang, Yu, Deng, Li, Xue, Li, Zhang, Wang, Zhu, Men, Gao, Liu, Luo, Li, Tang, Yin, Ren, Wang, Zhang, Ren, Fan, Su, Zhang, Zhang, Wan, Liu, Wang, Cui, Zhang, Zhou, and Qiu]{yang2025qwen3technicalreport}
An~Yang, Anfeng Li, Baosong Yang, Beichen Zhang, Binyuan Hui, Bo~Zheng, Bowen Yu, Chang Gao, Chengen Huang, Chenxu Lv, Chujie Zheng, Dayiheng Liu, Fan Zhou, Fei Huang, Feng Hu, Hao Ge, Haoran Wei, Huan Lin, Jialong Tang, Jian Yang, Jianhong Tu, Jianwei Zhang, Jianxin Yang, Jiaxi Yang, Jing Zhou, Jingren Zhou, Junyang Lin, Kai Dang, Keqin Bao, Kexin Yang, Le~Yu, Lianghao Deng, Mei Li, Mingfeng Xue, Mingze Li, Pei Zhang, Peng Wang, Qin Zhu, Rui Men, Ruize Gao, Shixuan Liu, Shuang Luo, Tianhao Li, Tianyi Tang, Wenbiao Yin, Xingzhang Ren, Xinyu Wang, Xinyu Zhang, Xuancheng Ren, Yang Fan, Yang Su, Yichang Zhang, Yinger Zhang, Yu~Wan, Yuqiong Liu, Zekun Wang, Zeyu Cui, Zhenru Zhang, Zhipeng Zhou, and Zihan Qiu.
\newblock Qwen3 technical report, 2025.

\bibitem[Zhang et~al.(2024{\natexlab{a}})Zhang, Yu, and Zhang]{zhang2024systematic}
Haopeng Zhang, Philip~S Yu, and Jiawei Zhang.
\newblock A systematic survey of text summarization: From statistical methods to large language models.
\newblock \emph{ACM Computing Surveys}, 2024{\natexlab{a}}.

\bibitem[Zhang et~al.(2020)Zhang, Zhao, Saleh, and Liu]{10.5555/3524938.3525989}
Jingqing Zhang, Yao Zhao, Mohammad Saleh, and Peter Liu.
\newblock {PEGASUS}: Pre-training with extracted gap-sentences for abstractive summarization.
\newblock In \emph{Proceedings of the 37th International Conference on Machine Learning}, volume 119, pages 11328--11339, 2020.

\bibitem[Zhang et~al.(2024{\natexlab{b}})Zhang, Ladhak, Durmus, Liang, McKeown, and Hashimoto]{zhang2024benchmarking}
Tianyi Zhang, Faisal Ladhak, Esin Durmus, Percy Liang, Kathleen McKeown, and Tatsunori~B Hashimoto.
\newblock Benchmarking large language models for news summarization.
\newblock \emph{Transactions of the Association for Computational Linguistics}, 12:\penalty0 39--57, 2024{\natexlab{b}}.

\end{thebibliography}


\newpage
\appendix

\section{Additional Experimental Details}
\label{app:details}

\subsection{Dataset Processing Details}
Some modifications were required to make datasets amenable to our extractive summarization setting.
\begin{itemize}
\setlength{\leftskip}{\itemizelength}
    \item \textbf{TLDR-AIC}: We selected the Abstract-Introduction-Conclusion subset to provide an appropriate amount of input context for summarization. A large portion of this dataset consists of single-sentence summaries. We focus on samples where multiple versions of single-sentence summaries have been written from different perspectives (e.g. author, reviewer), a total of 1143 samples, and pool those sentences into a single summary as the ground truth.
    \item \textbf{TLDR-Full}: This version of SciTLDR uses the full text of a scientific paper as the input, making inputs much larger and summaries a much smaller fraction of the long-text. Otherwise, processing is the same as TLDR-AIC.
    \item \textbf{MTS-Dialog}: This dataset contains question and answer-style conversations between doctors and patients. Hence, certain sentences require context for correct interpretation. To accommodate this, each question from the doctor and all subsequent patient responses were merged into a single sentence unit, following the pattern "Doctor: \textit{questions}. Patient: \textit{answers}". If a sample input contained two or fewer sentences after this merge, the entire sample is removed from the dataset; 1129 samples remained after this filtering.
    
\end{itemize}

\paragraph{Dataset Licenses}

ECT is released under a GNU GPL license. CSDS is provided for use without a specific license. CNN/DM and SciTLDR are both released under an Apache-2.0 License. MTS-Dialog is released under a Creative Commons Attribution 4.0 International license. Hence, our usages of these five datasets are permissible under their respective licenses.

\subsection{Greedy Optimization for Extractive Summarization Labeling}

The following greedy oracle labeling method is used to provide sentence‐level importance labels for datasets that lack sentence‐level ground‐truth annotations. All we require is that a reference summary $r$ be available, which could be abstractive rather than extractive. The output of the algorithm is an extractive summary that can be used as ground truth $y^*$ in Conformal Importance Summarization. First, for a long-text $x$ we compute a similarity score $V$ between each sentence $c_i$ and the reference summary, then sort sentences by score in descending order. We then iterate through the ranked sentences in a greedy manner and add them to the extractive summary if the overall similarity of the combined extractive summary to $r$ increases by at least $\delta$. Mathematically, if the current extractive summary is $y^*_\text{curr}$ with similarity to $r$ of $V(y^*_\text{curr}; r)$, the sentence $c_i$ is added to $y^*_\text{curr}$ if
\begin{equation}
V(y^*_\text{curr}\cup\{c_i\}; r) - V(y^*_\text{curr}; r) > \delta.
\end{equation}
After one iteration through the sentences in $x$, we return $y^*_\text{curr}$ to be used as the ground truth extractive summary $y^*$. This process is depicted in \Cref{alg:greedy-extractive}.

\SetKwInput{KwIn}{Input}      
\SetKwInput{KwOut}{Output}    
\SetKwProg{Fn}{Function}{}{end} 

\begin{algorithm}[ht]
\caption{Greedy Optimization for Extractive‐Summarization Labeling}
\label{alg:greedy-extractive}

\KwIn{Sentences $x = [c_1,\dots,c_p]$, reference summary $r$, scoring function $V(\cdot;\cdot)$, threshold $\delta$}
\KwOut{Extractive summary $y^*$}

Compute $v_i = V(c_i; r)$ for all $i$\;
Sort indices by descending $v$ giving the permutation $\pi = [\pi_1,\dots,\pi_n]$\;

$y^*_\text{curr} \leftarrow \emptyset$

$V_\text{curr} \leftarrow 0$

\For{$j = 1$ \KwTo $p$}{
  $i \leftarrow \pi_j$\;
  $\Delta \leftarrow V(y^*_\text{curr}\cup\{c_i\}; r) - V_\text{curr}$\;
  \If{$\Delta > \delta$}{
    $y^*_\text{curr} \leftarrow y^*_\text{curr} \cup \{c_i\}$\;
    $V_\text{curr} \leftarrow V_\text{curr} + \Delta$\;
  }
}

\Return $y^*_\text{curr}$
\end{algorithm}

\section{LLM Prompts}
\label{app:GPTprompt}

\textbf{Ground Truth Labelling.}\quad To generate LLM-based ground-truth labels for important sentences in extractive summarization where sentence-level labels were not available, we provided input sentences as separate strings to GPT-4o mini, along with the existing summary text from the dataset. The prompt in Listing~\ref{lst:system-prompt-importance} was used to produce a list of ground-truth labels corresponding to each input sentence.

\begin{lstlisting}[caption=System Prompt for Creating Sentence-Level Ground-Truth Labels,label=lst:system-prompt-importance]
"""
Evaluate whether each input claim is included in the summary text. The output labels, corresponding to each input claim, should be either 0 or 1, indicating whether the corresponding claim, or the information it carries, is indeed included in the actual summary. For example, if claim_1's information is contained in the summary, then label_1 should be 1; if information carried in claim_3 cannot be found in the summary text, then label_3 should be 0.

Summary text:
{summary_text}

List of claims:
{[claim_text]}
"""
\end{lstlisting}

\textbf{Importance Scoring.}\quad To generate LLM-based scores $R(c_i; y^*)$ for Conformal Importance Summarization, the source text was provided to an LLM along with the individual sentences from that text as separate strings. The prompt in Listing~\ref{lst:system-prompt-float-scores} was used to produce a list of importance scores between 0 and 1 corresponding to each input sentence.

\begin{lstlisting}[caption=System Prompt for Creating Importance Scores,label=lst:system-prompt-float-scores]
"""
Please evaluate the importance of each input claim in the original text, based on how the information carried in the claim is aligned with the overall message. Please provide a importance score for EACH input claim. Each output score should be a two decimal float number ranged between 0 and 1, indicating how important the corresponding input claim is in the context of the text document. For example, if claim 1's information is highly aligned with that of the input text, and very likely to be included in the summary, then score 1 should be close to 1, say greater than 0.8; if information carried in claim 3 is trivial or only remotely related to the central message of the text, and is not worthy of inclusion in the summary, then score 3 should be close to 0, say less than 0.2.

Original text:
{original_text}

List of claims:
{[claim_text]}
"""
\end{lstlisting}

We also experimented with prompting the LLM to give binary scores $R(c_i; y^*)$, rather than floats. The prompt in  Listing~\ref{lst:system-prompt-binary-scores} was used to produce a list of importance scores corresponding to each input sentence as above.

\begin{lstlisting}[caption=System Prompt for Creating Importance Scores with Binary Restriction,label=lst:system-prompt-binary-scores]
"""
Evaluate the importance of each input claim in the original text, based on how the information carried in the claim is aligned with the overall message. Please provide a binary importance score for EACH input claim. Each output score should be either 0 or 1, indicating whether the corresponding input claim is important enough in the context of the text document to be included in the summary. For example, if claim 1's information is highly aligned with that of the input text, and very likely to be included in the summary, then score 1 should be 1; if information carried in claim 3 is trivial or only remotely related to the central message of the text, and is not worthy of inclusion in the summary, then score 3 should be 0.

Original text:
{original_text}

List of claims:
{[claim_text]}
"""
\end{lstlisting}

\textbf{Direct Abstractive Summarization.}\quad We tested the ability of instruction-tuned LLMs to directly create abstractive summaries that retain a specified fraction $\beta$ of important information. To guide the LLM as to what type of information was important for a given dataset, ten examples from the calibration dataset were provided to enable in-context learning \cite{NEURIPS2020_1457c0d6}.

\begin{lstlisting}[caption=System Prompt for Direct Abstractive Summarization,label=lst:system-prompt-abstractive]
"""
Here are examples of what constitutes important information to include in the summary:

{examples_text}

Create an abstractive summary of the following text.

Requirements:
- Aim to retain at least {beta}% of the important information
- Use your own words and phrasing (abstractive, not extractive)

Input text to summarize:
{input_text}
"""
\end{lstlisting}

\textbf{Hybrid Extractive-Abstractive Summarization.}\quad To disentangle the subtasks of importance assessment and summary synthesis, we applied abstractive summarization with an LLM as a post-processing step after Conformal Importance Summarization. The abstractive step used the following prompt that specifies \emph{all} information should be retained from the extractive summary input.

\begin{lstlisting}[caption=System Prompt for Abstractive Summarization as a Post-Processing Step,label=lst:system-prompt-hybrid]
"""
Requirements:
- Use more concise language to make the text shorter
- Retain all of the information from the input text
- Improve flow, coherence, and readability
"""
\end{lstlisting}

\textbf{Sentence-level Recall Estimation.}\quad Unlike for extractive summaries, determining if an abstractive summary has retained a specific piece of important information requires a judgement call. Hence, we estimated recall of sentence-level information using GPT4o-mini prompted to determine if the content of a sentence is retained in an abstractive summary.

\begin{lstlisting}[caption=System Prompt for Determining if a Ground-truth Important Sentence is Retained in an Abstractive Summary,label=lst:system-prompt-recall]
"""
You will be given an important sentence from the original text and a generated summary. Your goal is to determine whether the important sentence given is retained in the generated summary.

Important sentence:
{important_sentence}

Generated summary:
{summary}

Output True if the important sentence is retained in the generated summary. Output False otherwise.
"""
\end{lstlisting}

\section{Additional Experimental Results}
\label{app:rougeexperiment}

\subsection{Coverage Plots for all Datasets and Methods}\label{app:coverage_plots}
\begin{figure}[t]
    \centering
        \begin{subfigure}{0.48\textwidth}
        \includegraphics[width=0.95\linewidth]{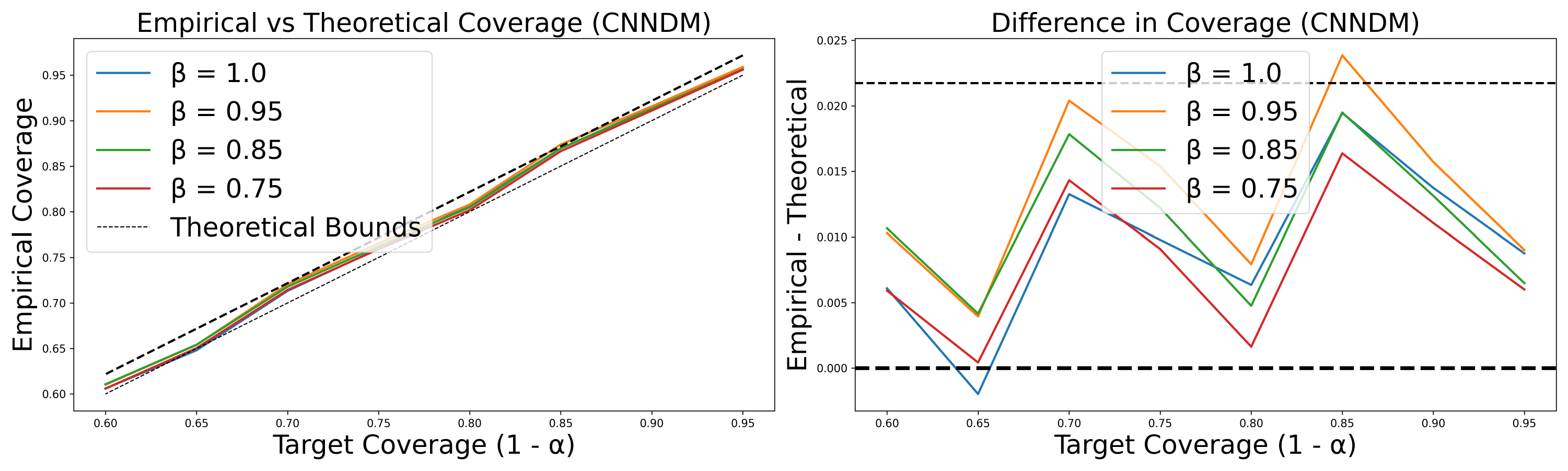}
        \caption{Cosine Similarity Centrality}
        \end{subfigure}
                \begin{subfigure}{0.48\textwidth}
        \includegraphics[width=0.95\linewidth]{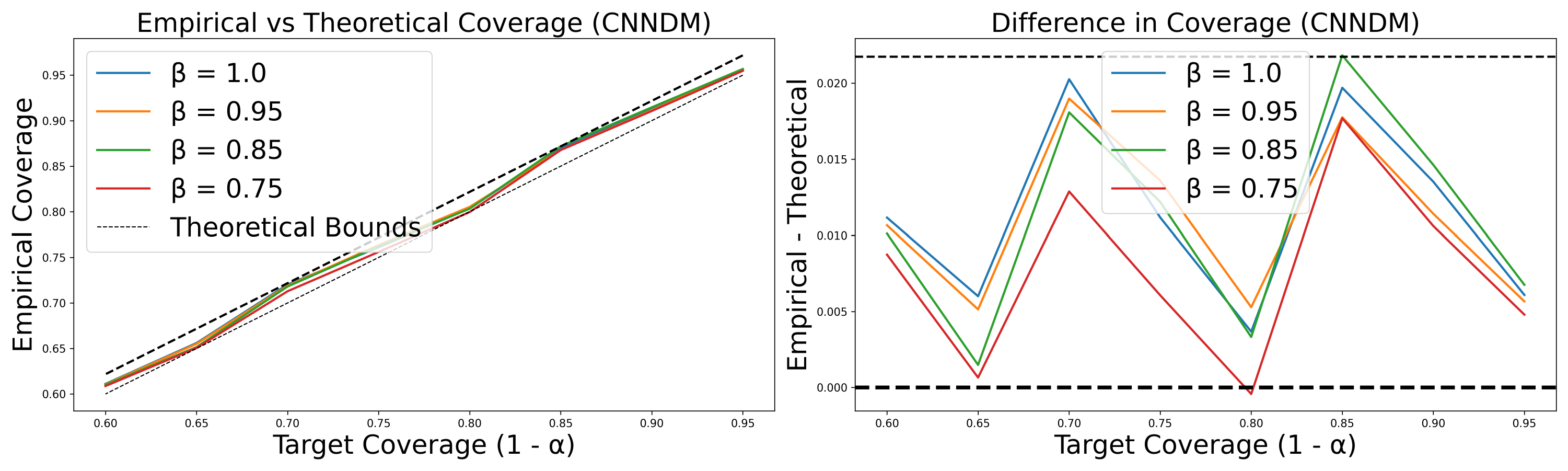}
        \caption{Sentence Centrality}
        \end{subfigure}
        \begin{subfigure}{0.48\textwidth}
        \includegraphics[width=0.95\linewidth]{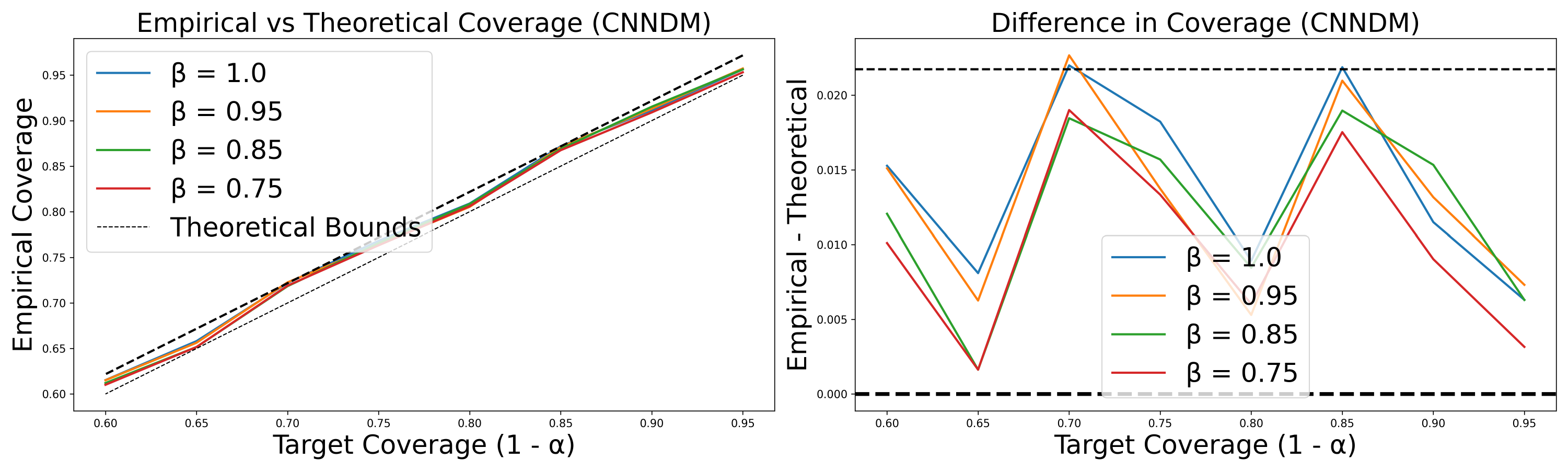}
        \caption{GUSUM}
        \end{subfigure}
        \begin{subfigure}{0.48\textwidth}
        \includegraphics[width=0.95\linewidth]{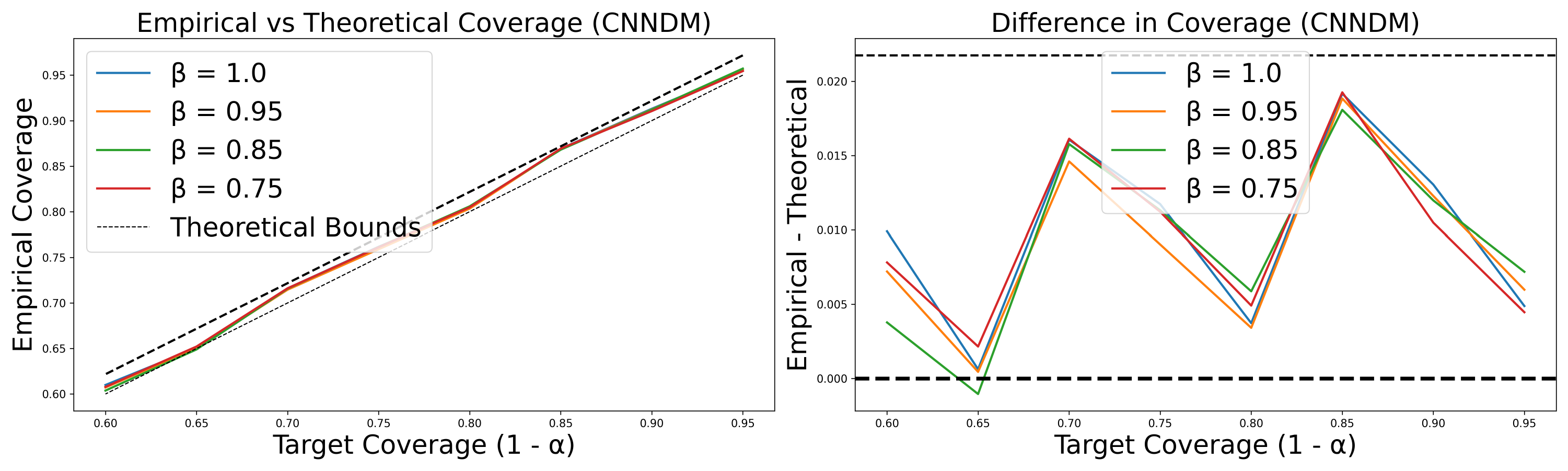}
        \caption{GPT-4o mini}
        \end{subfigure}
                \begin{subfigure}{0.48\textwidth}
        \includegraphics[width=0.95\linewidth]{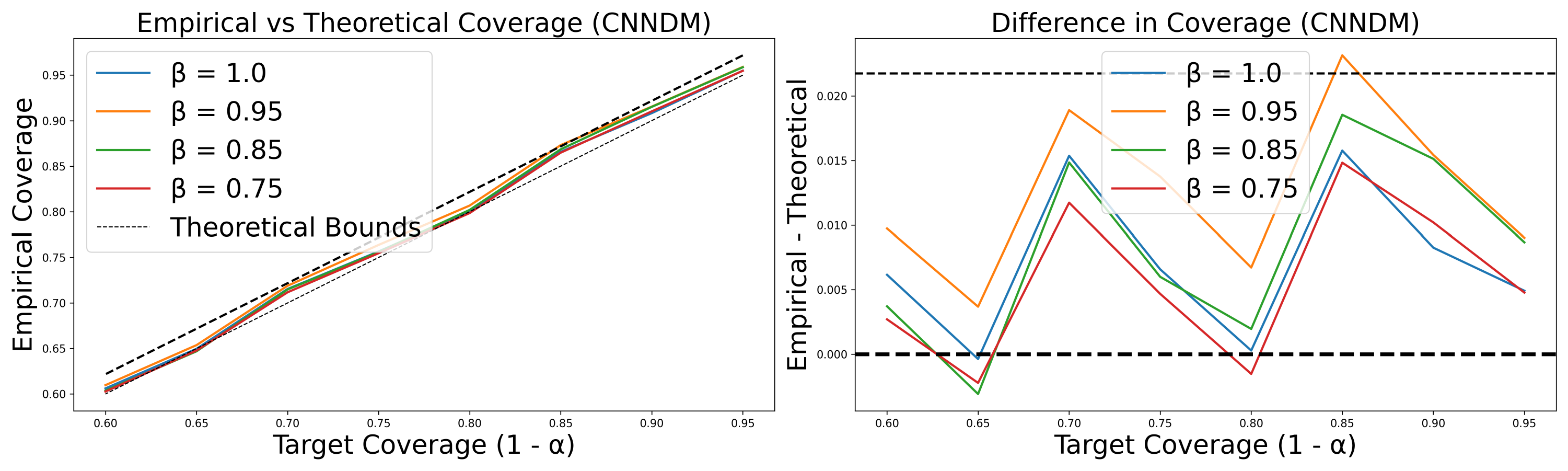}
                    \caption{Llama 3}
        \end{subfigure}
        \begin{subfigure}{0.48\textwidth}
        \includegraphics[width=0.95\linewidth]{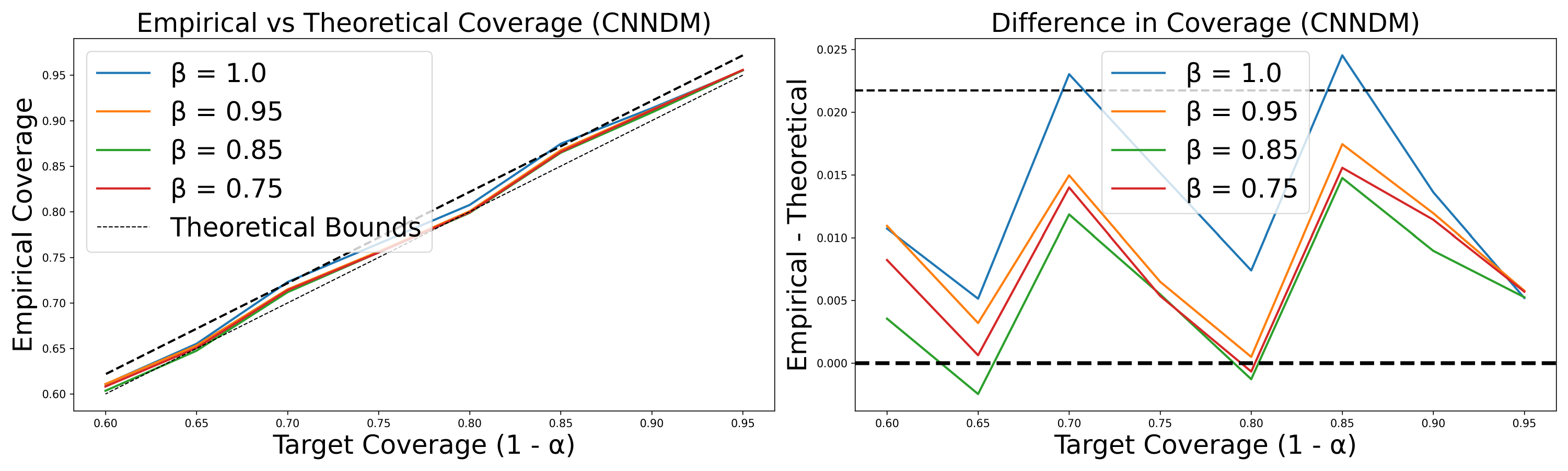}
        \caption{Qwen 3}
        \end{subfigure}
        \begin{subfigure}{0.48\textwidth}
    \includegraphics[width=0.95\linewidth]{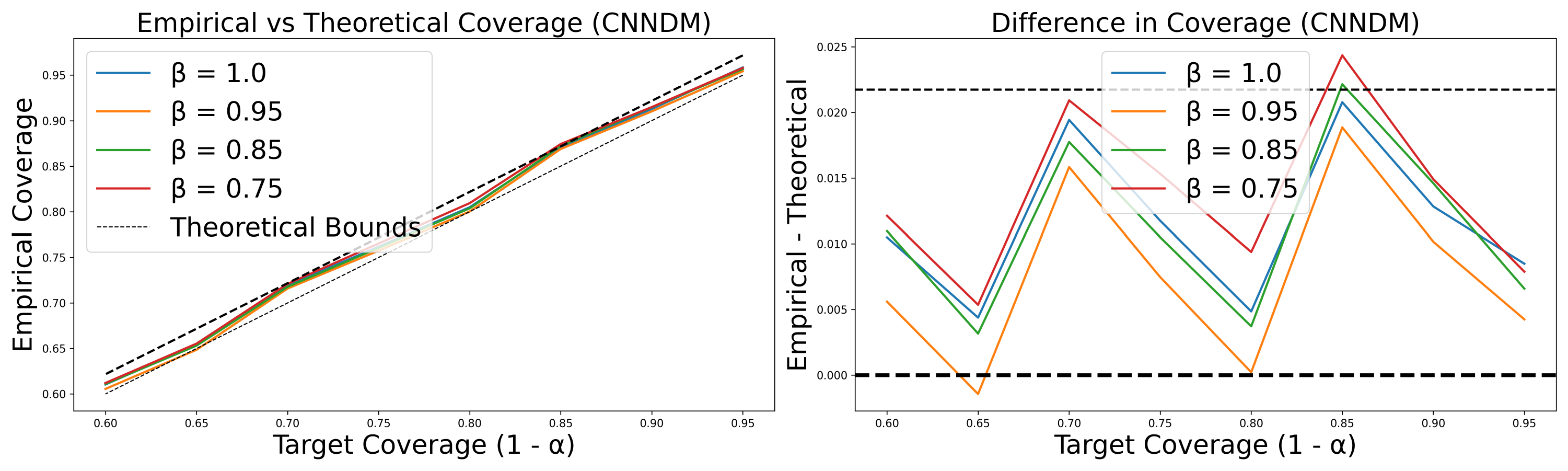}
        \caption{Gemini 2.0 Flash-Lite}
        \end{subfigure}
        \begin{subfigure}{0.48\textwidth}
        \includegraphics[width=0.95\linewidth]{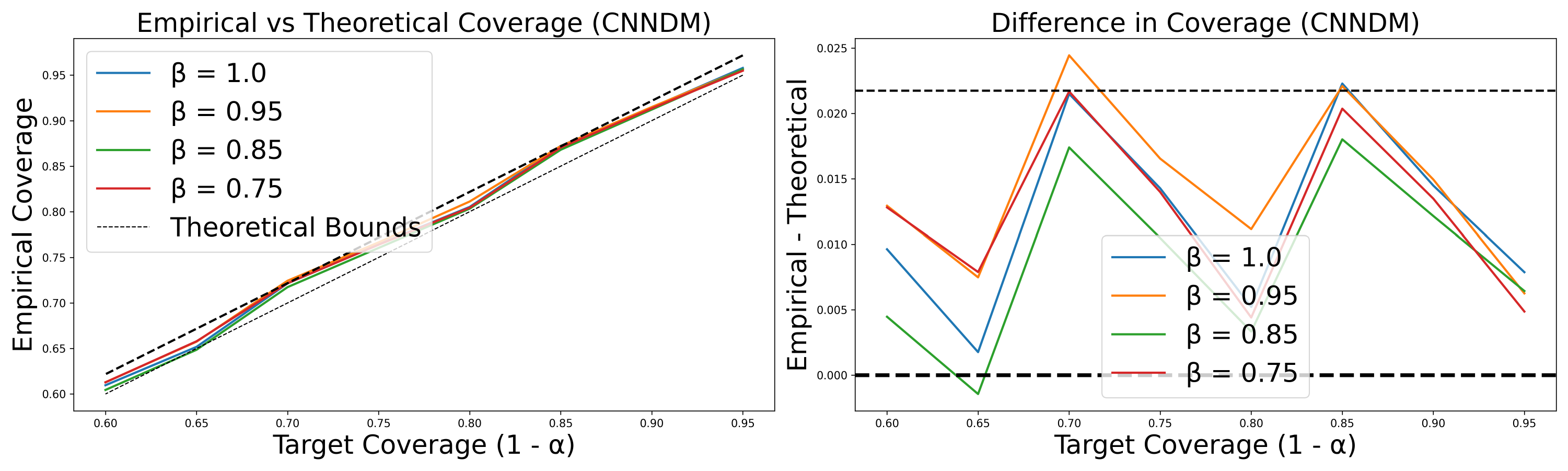}
        \caption{Gemini 2.5 Flash}
        \end{subfigure}
        \caption{User-specified probability of achieving coverage (1-$\alpha$) vs. empirical probability of achieving coverage, on the CNN/DM dataset. Dashed lines show theoretical bounds given in \Cref{theorem}. Results are averaged over 400 random splits of calibration and test data.}
    \label{fig:calibration_CNNDM}
\end{figure}

\begin{figure}[t]
    \centering
        \begin{subfigure}{0.48\textwidth}
        \includegraphics[width=0.95\linewidth]{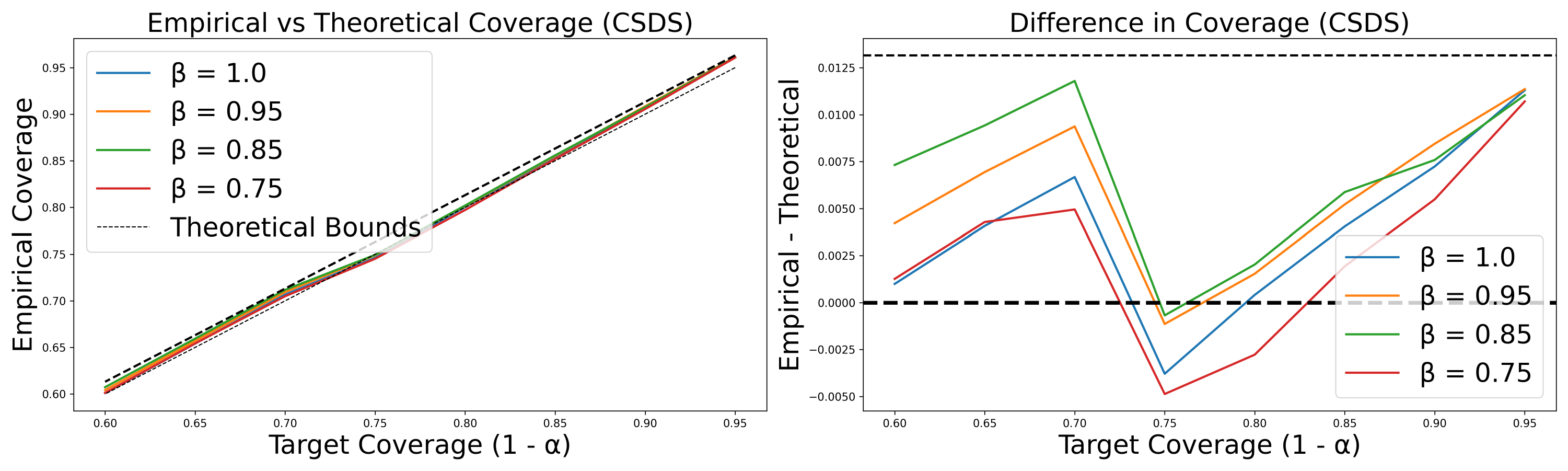}
        \caption{Cosine Similarity Centrality}
        \end{subfigure}
        \begin{subfigure}{0.48\textwidth}
        \includegraphics[width=0.95\linewidth]{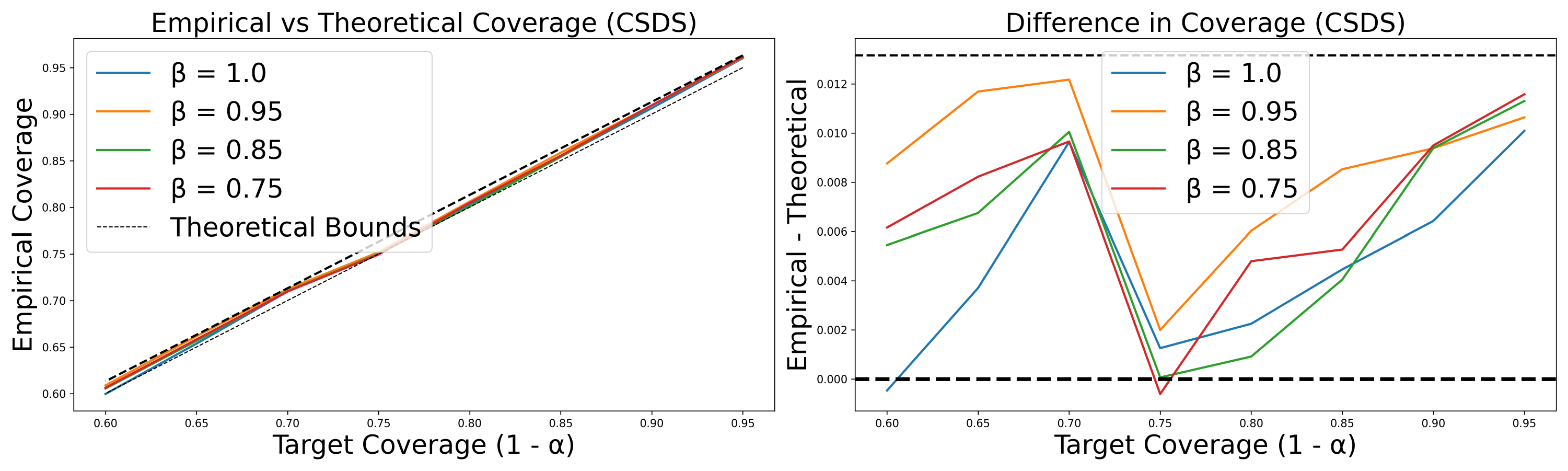}
        \caption{Sentence Centrality}
        \end{subfigure}
        \begin{subfigure}{0.48\textwidth}
    \includegraphics[width=0.95\linewidth]{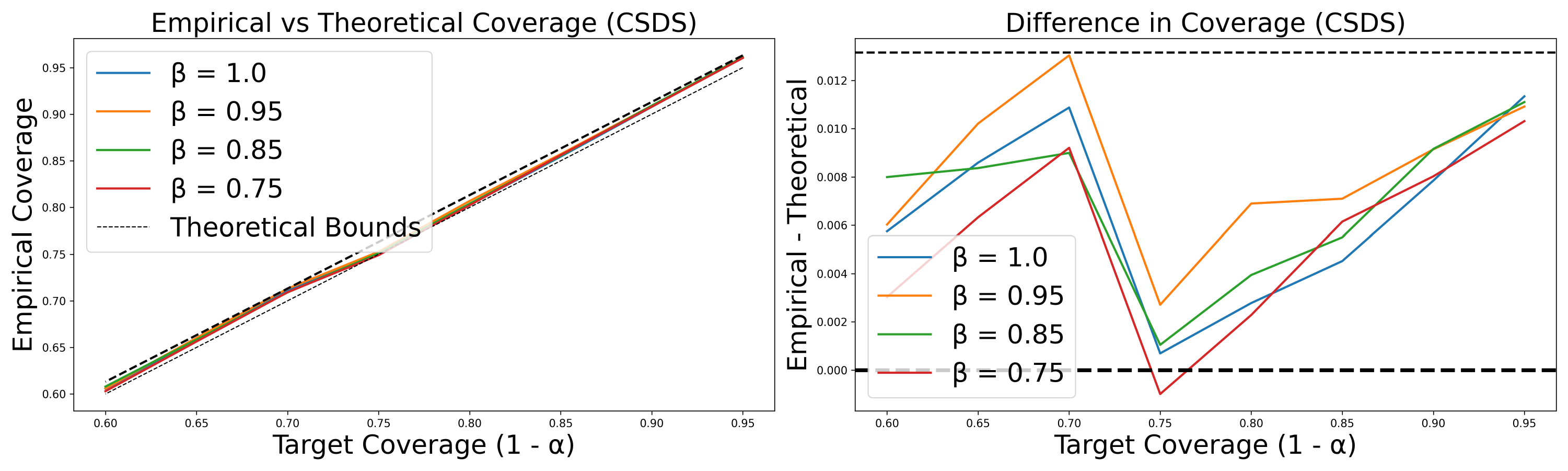}
        \caption{GUSUM}
        \end{subfigure}
        \begin{subfigure}{0.48\textwidth}
        \includegraphics[width=0.95\linewidth]{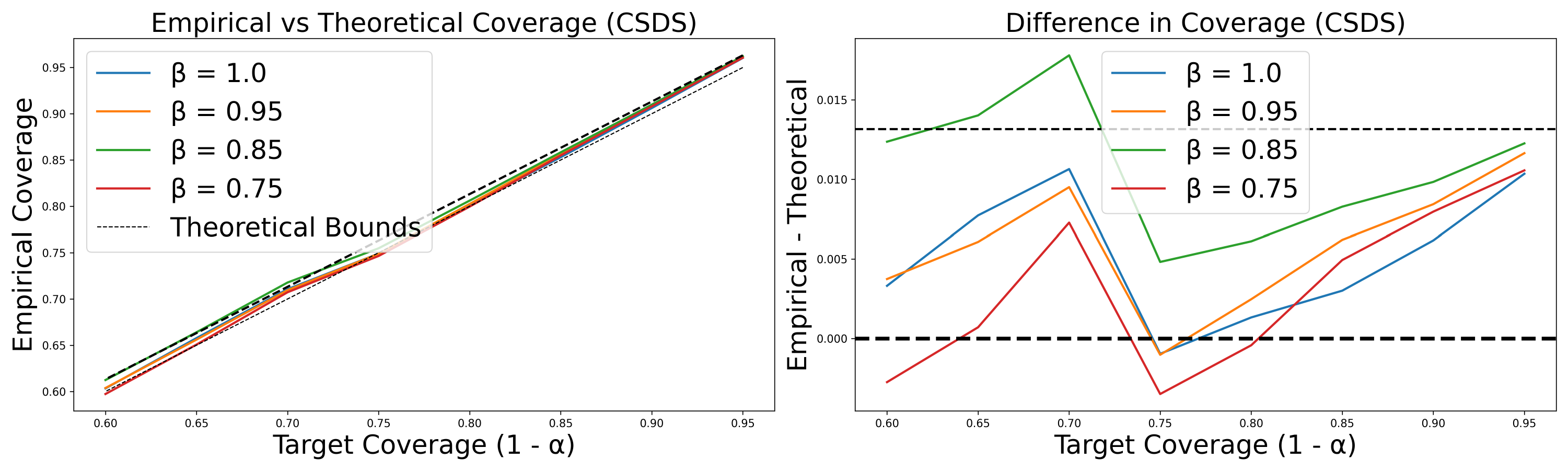}
        \caption{LexRank}
        \end{subfigure}
        \begin{subfigure}{0.48\textwidth}
        \includegraphics[width=0.95\linewidth]{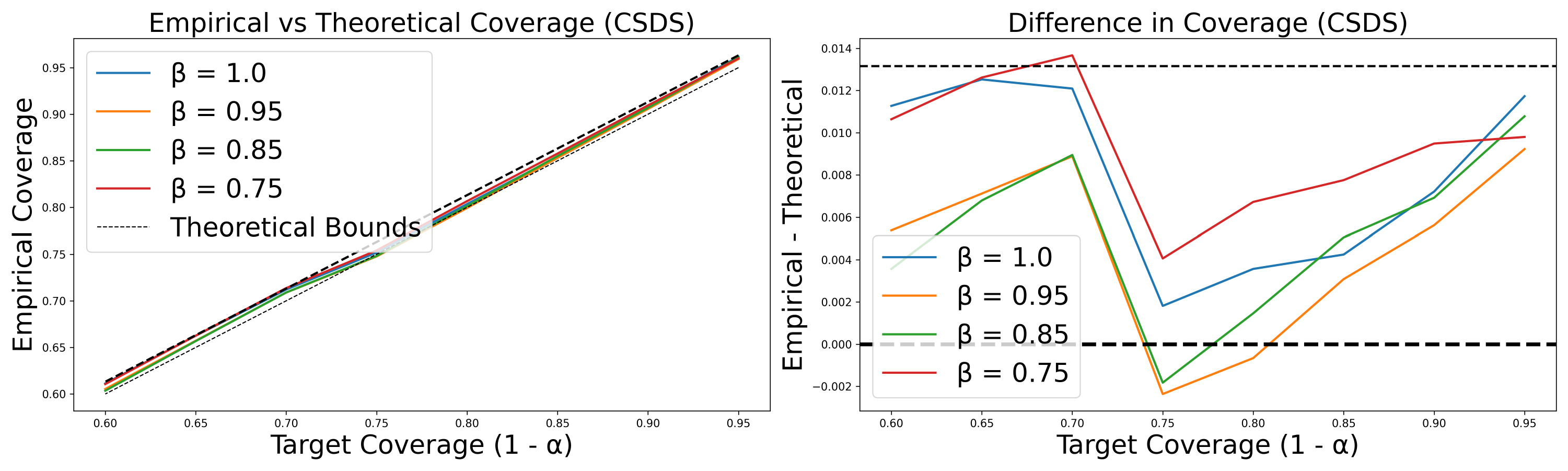}
        \caption{Llama 3}
        \end{subfigure}
        \begin{subfigure}{0.48\textwidth}
        \includegraphics[width=0.95\linewidth]{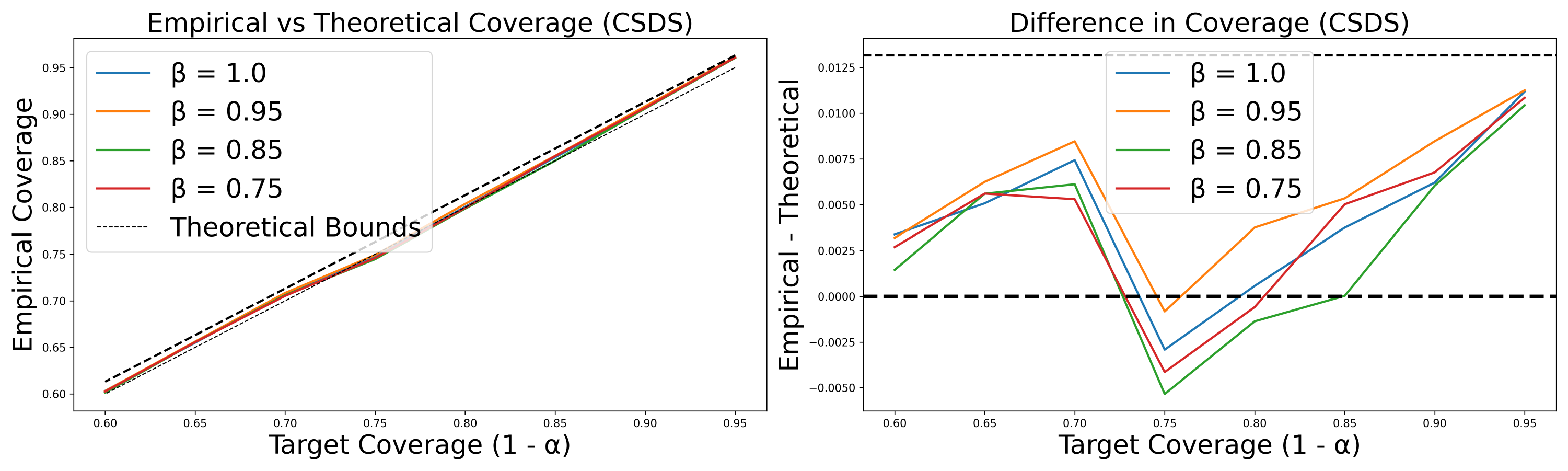}
        \caption{Qwen 3}
        \end{subfigure}
        \begin{subfigure}{0.48\textwidth}
        \includegraphics[width=0.95\linewidth]{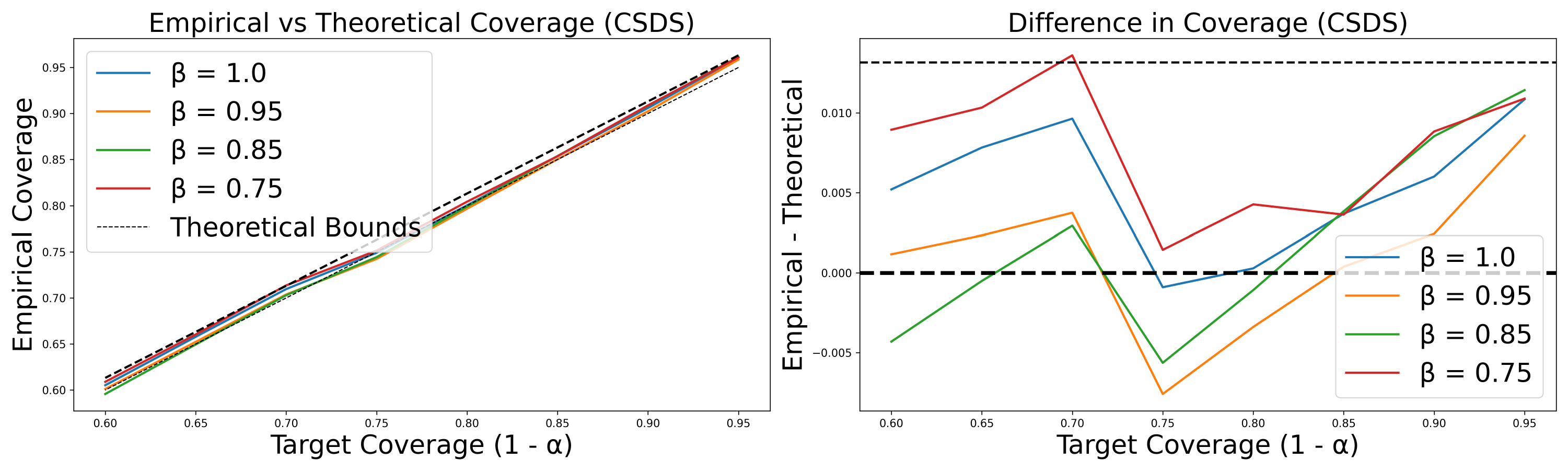}
        \caption{Gemini 2.0 Flash-Lite}
        \end{subfigure}
        \begin{subfigure}{0.48\textwidth}
        \includegraphics[width=0.95\linewidth]{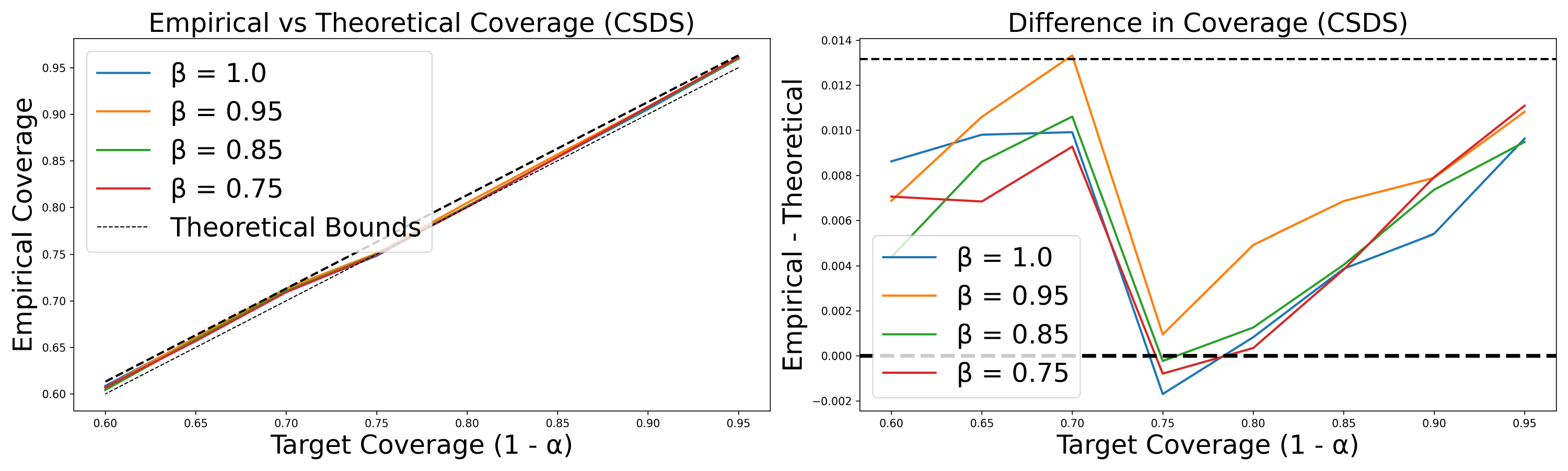}
        \caption{Gemini 2.5 Flash}
        \end{subfigure}
        \caption{User-specified probability of achieving coverage (1-$\alpha$) vs. empirical probability of achieving coverage, on the CSDS dataset. Dashed lines show theoretical bounds given in \Cref{theorem}. Results are averaged over 400 random splits of calibration and test data.}
    \label{fig:calibration_CSDS}
\end{figure}

\begin{figure}[t]
    \centering
        \begin{subfigure}{0.48\textwidth}
        \includegraphics[width=0.95\linewidth]{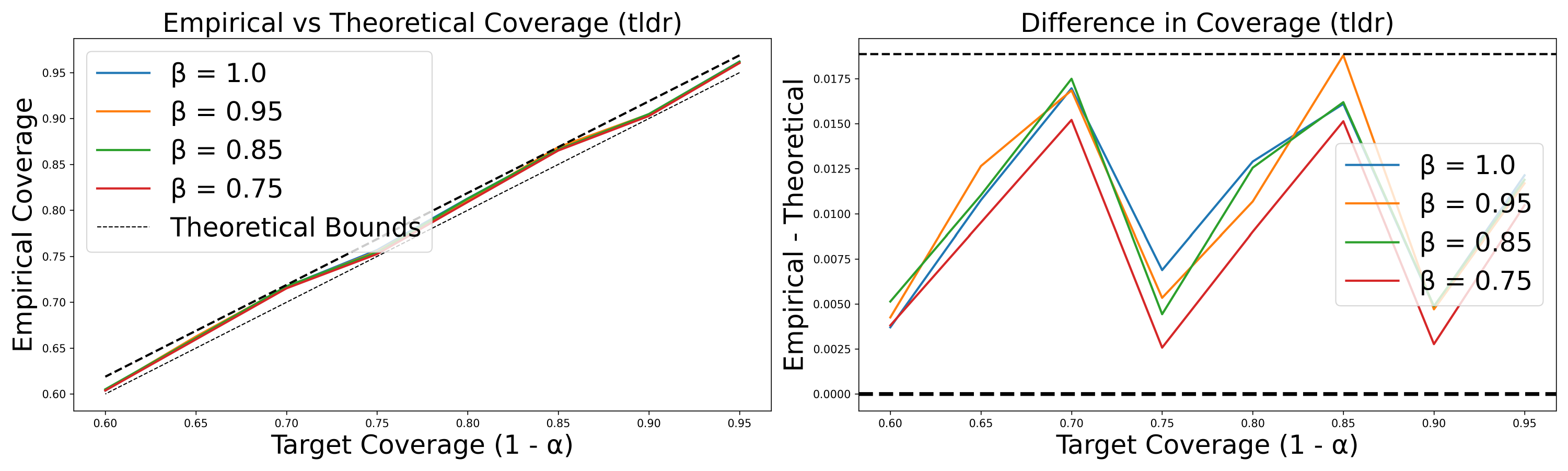}
        \caption{Cosine Similarity Centrality}
        \end{subfigure}
        \begin{subfigure}{0.48\textwidth}
        \includegraphics[width=0.95\linewidth]{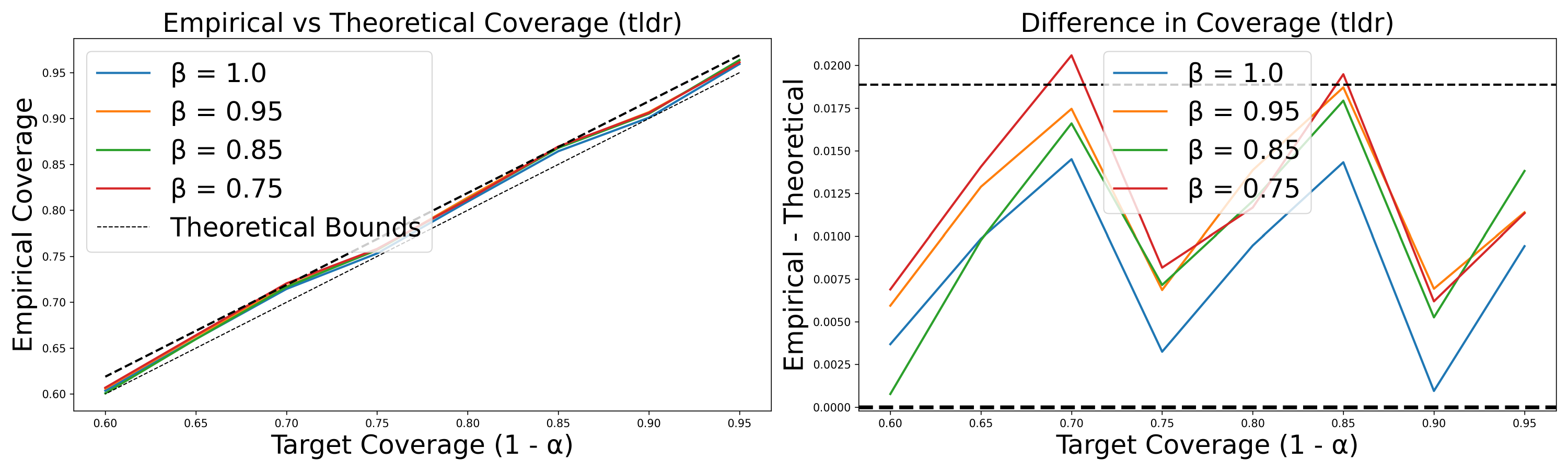}
        \caption{Sentence Centrality}
        \end{subfigure}
        \begin{subfigure}{0.48\textwidth}
        \includegraphics[width=0.95\linewidth]{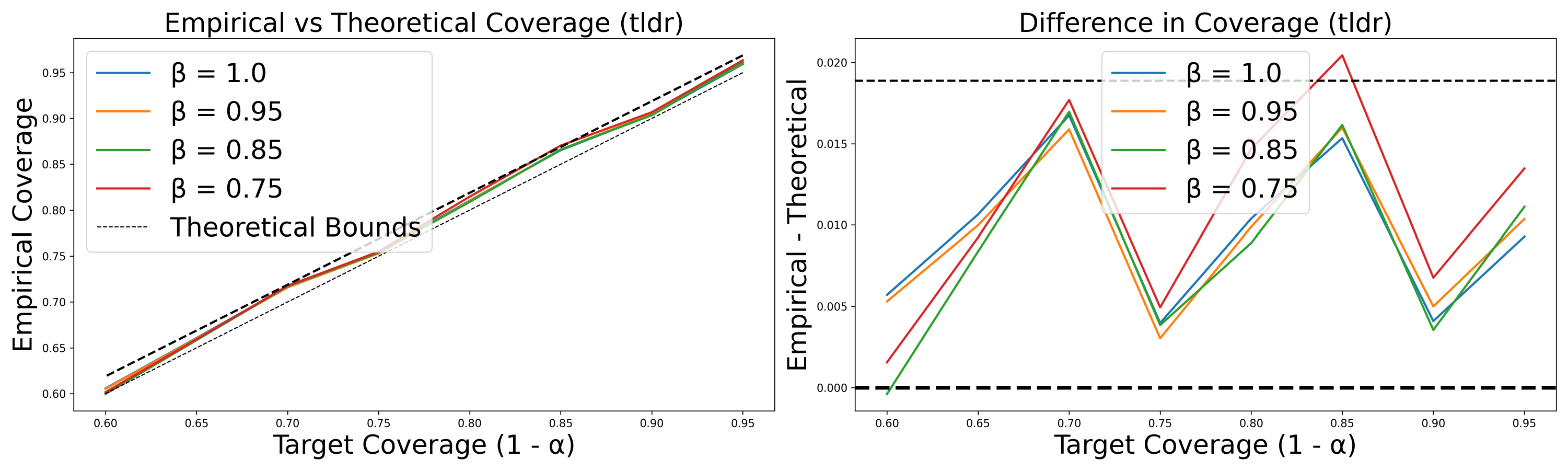}
        \caption{GUSUM}
        \end{subfigure}
        \begin{subfigure}{0.48\textwidth}
        \includegraphics[width=0.95\linewidth]{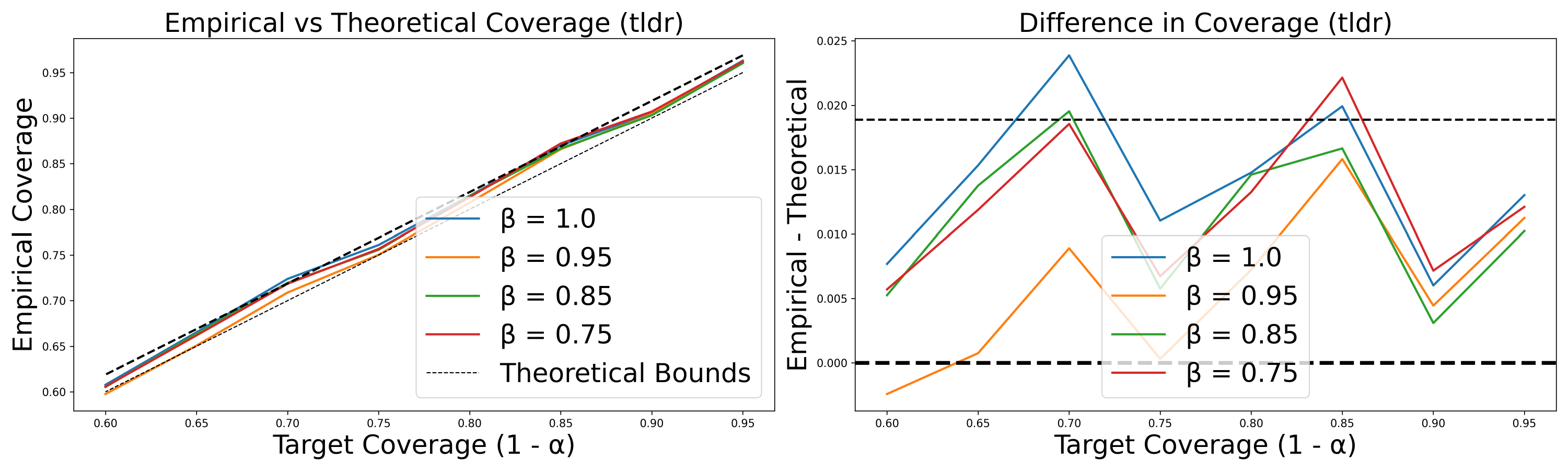}
        \caption{LexRank}
        \end{subfigure}
        \begin{subfigure}{0.48\textwidth}
        \includegraphics[width=0.95\linewidth]{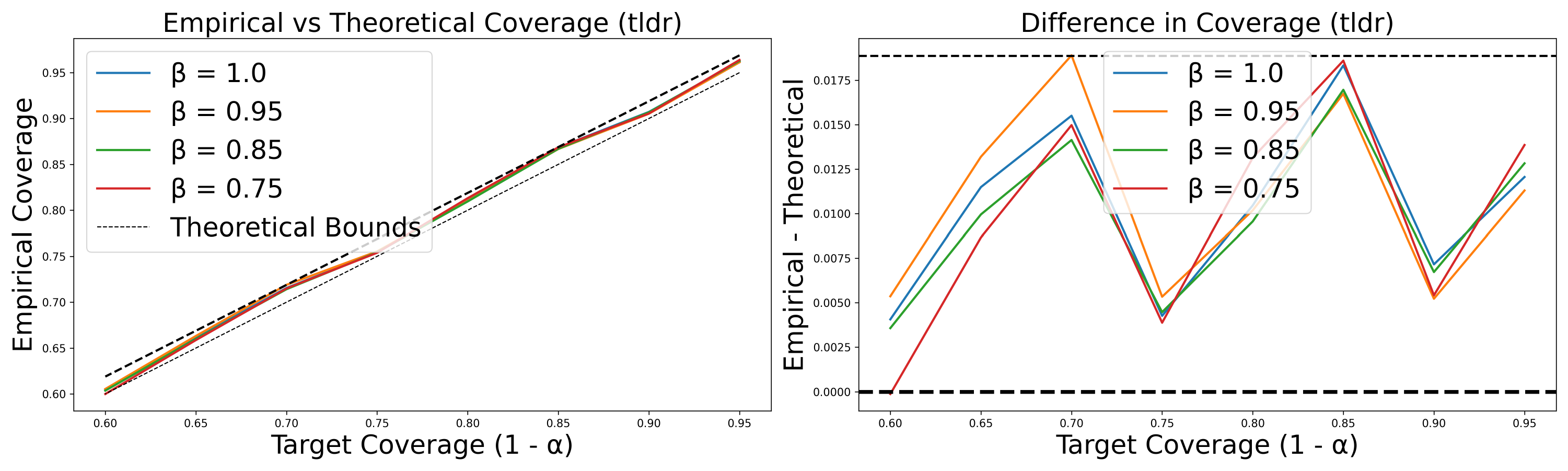}
        \caption{GPT-4o mini}
        \end{subfigure}
        \begin{subfigure}{0.48\textwidth}
        \includegraphics[width=0.95\linewidth]{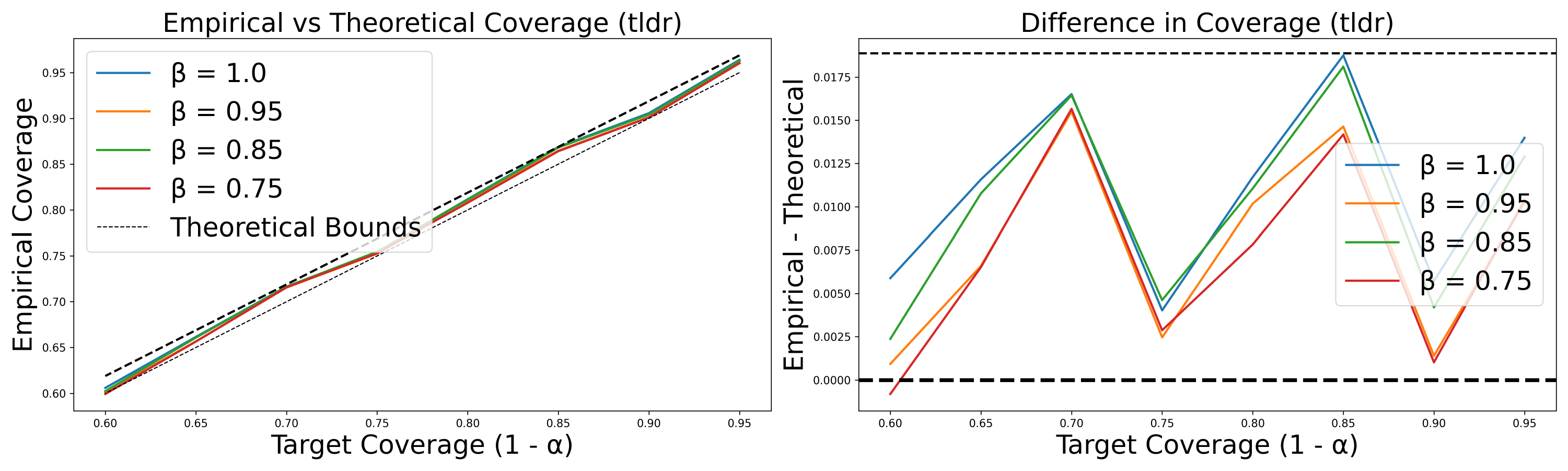}
                \caption{Llama 3}
        \end{subfigure}
        \begin{subfigure}{0.48\textwidth}
        \includegraphics[width=0.95\linewidth]{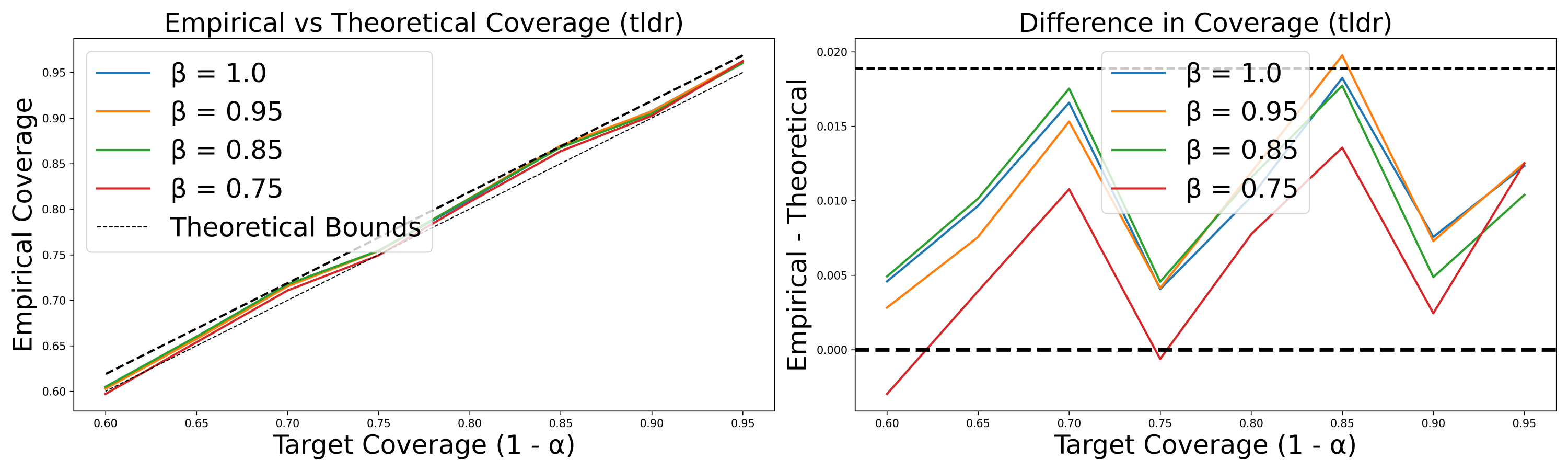}
        \caption{Qwen 3}
        \end{subfigure}
        \begin{subfigure}{0.48\textwidth}
        \includegraphics[width=0.95\linewidth]{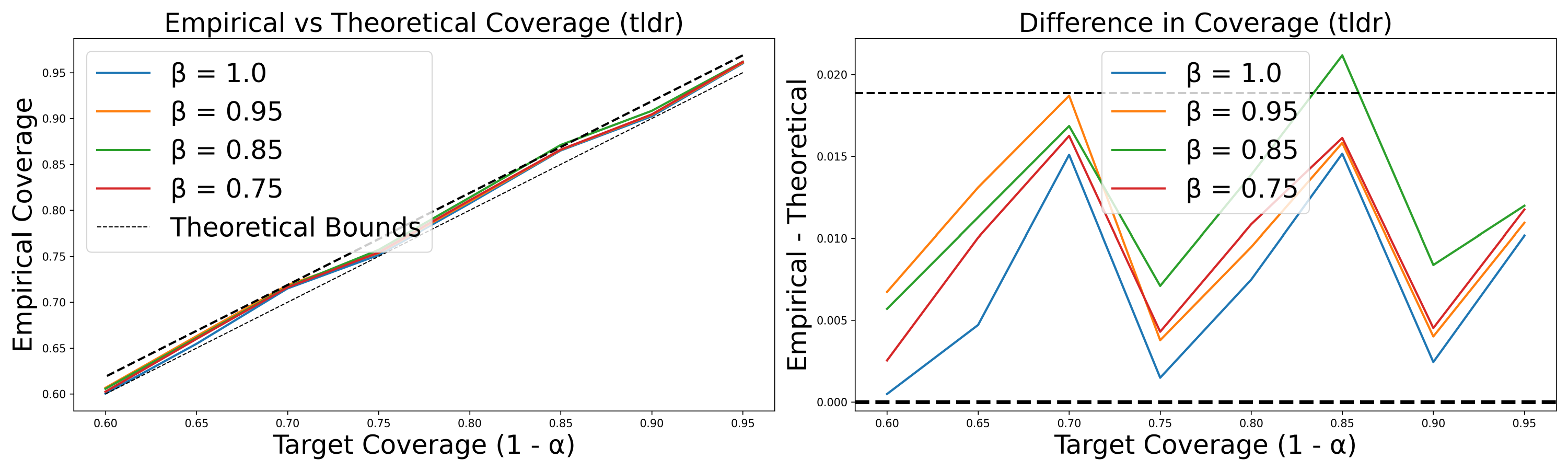}
        \caption{Gemini 2.0 Flash-Lite}
        \end{subfigure}
        \begin{subfigure}{0.48\textwidth}
        \includegraphics[width=0.95\linewidth]{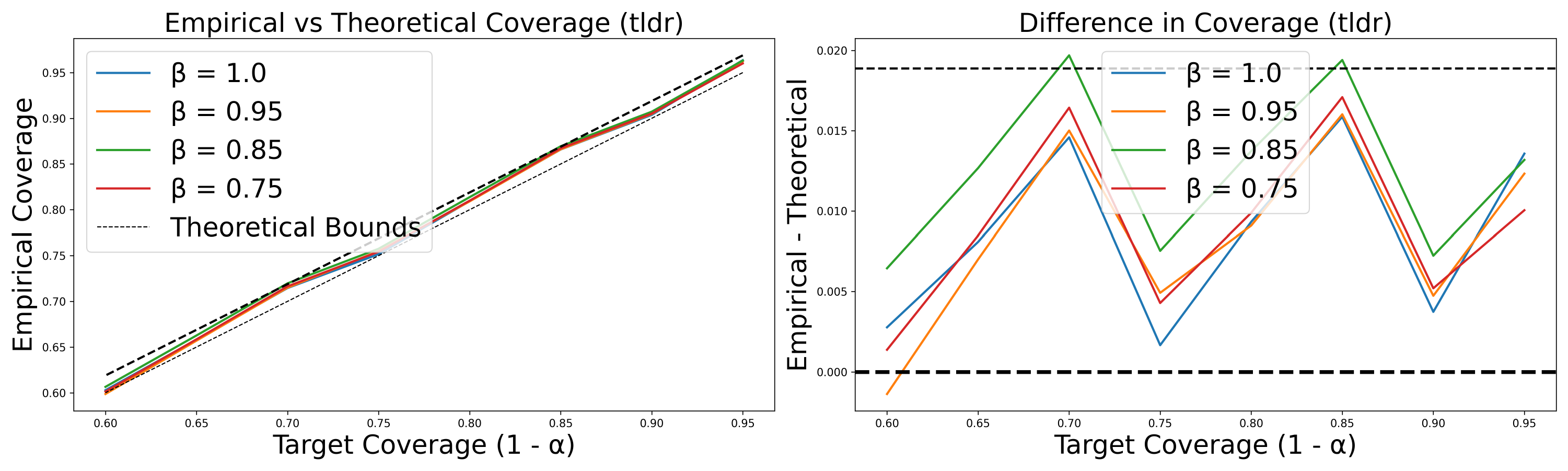}
        \caption{Gemini 2.5 Flash}
        \end{subfigure}
        \caption{User-specified probability of achieving coverage (1-$\alpha$) vs. empirical probability of achieving coverage, on the TLDR-AIC dataset. Dashed lines show theoretical bounds given in \Cref{theorem}. Results are averaged over 400 random splits of calibration and test data.}
    \label{fig:calibration_tldr}
\end{figure}

\begin{figure}[t]
    \centering
        \begin{subfigure}{0.48\textwidth}
        \includegraphics[width=0.95\linewidth]{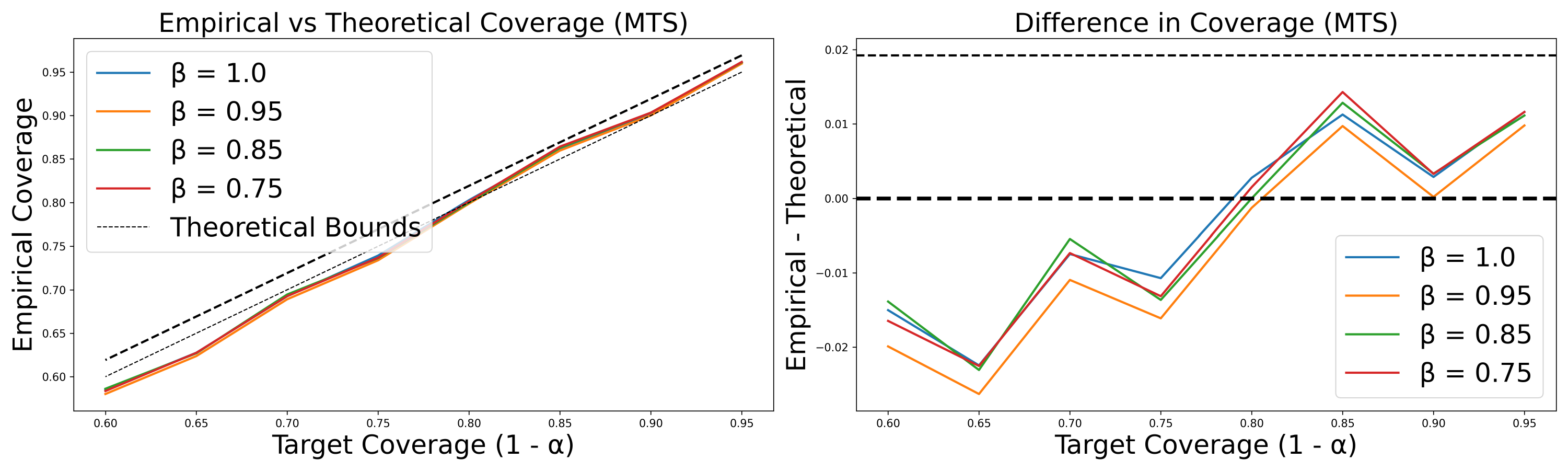}
        \caption{Cosine Similarity Centrality}
        \end{subfigure}
        \begin{subfigure}{0.48\textwidth}
        \includegraphics[width=0.95\linewidth]{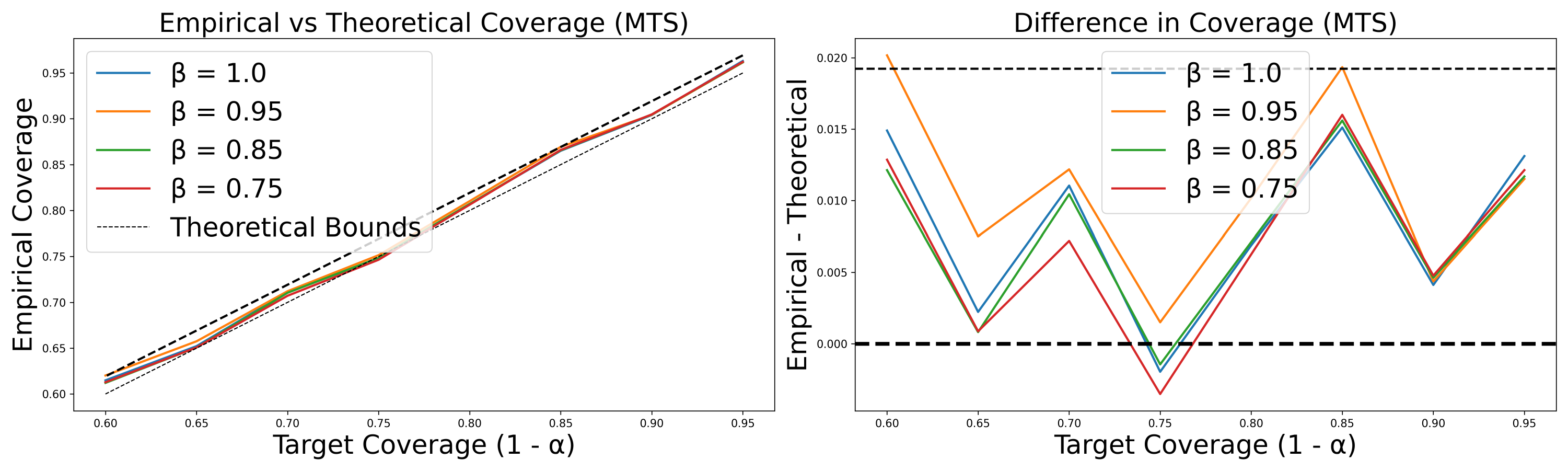}
        \caption{Sentence Centrality}
        \end{subfigure}
        \begin{subfigure}{0.48\textwidth}
        \includegraphics[width=0.95\linewidth]{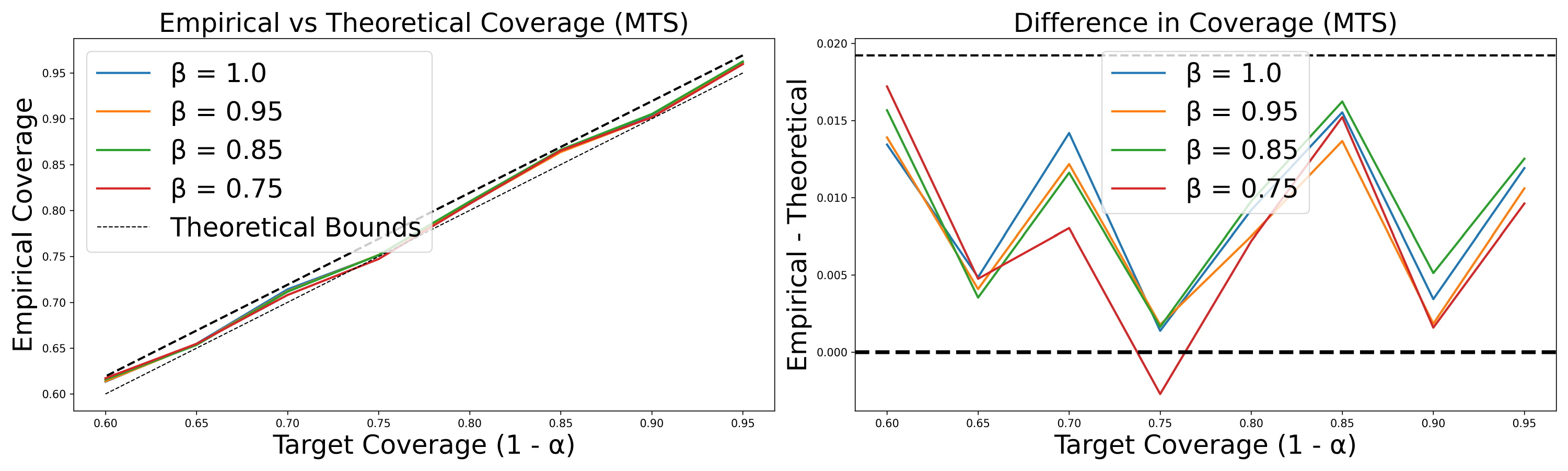}
        \caption{GUSUM}
        \end{subfigure}
        \begin{subfigure}{0.48\textwidth}
        \includegraphics[width=0.95\linewidth]{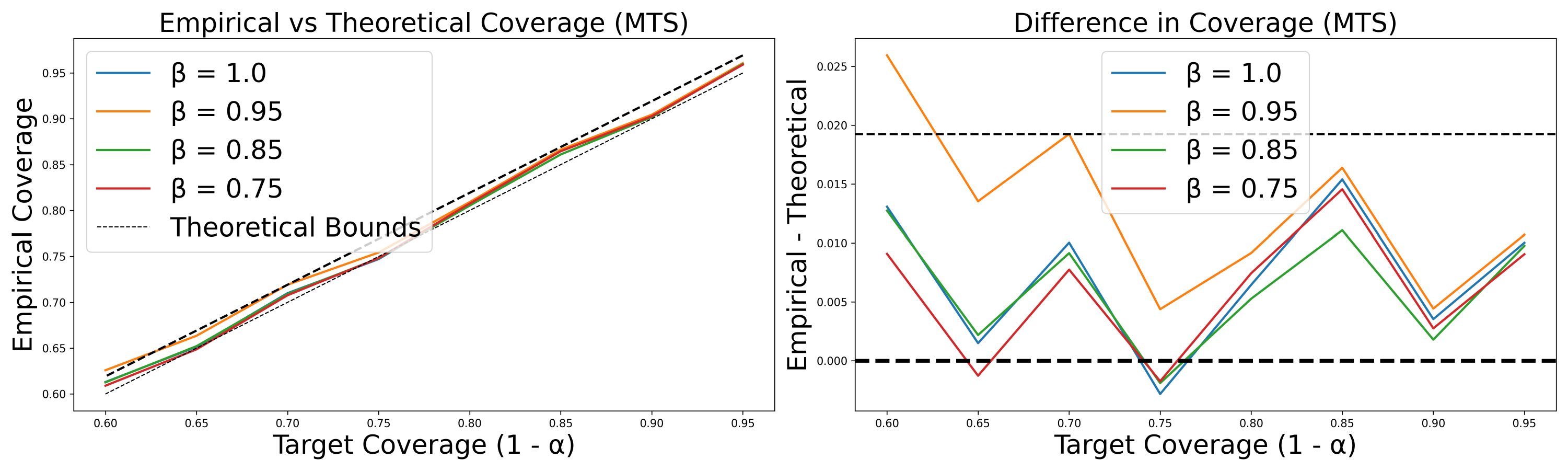}
        \caption{GPT-4o mini}
        \end{subfigure}
        \begin{subfigure}{0.48\textwidth}
        \includegraphics[width=0.95\linewidth]{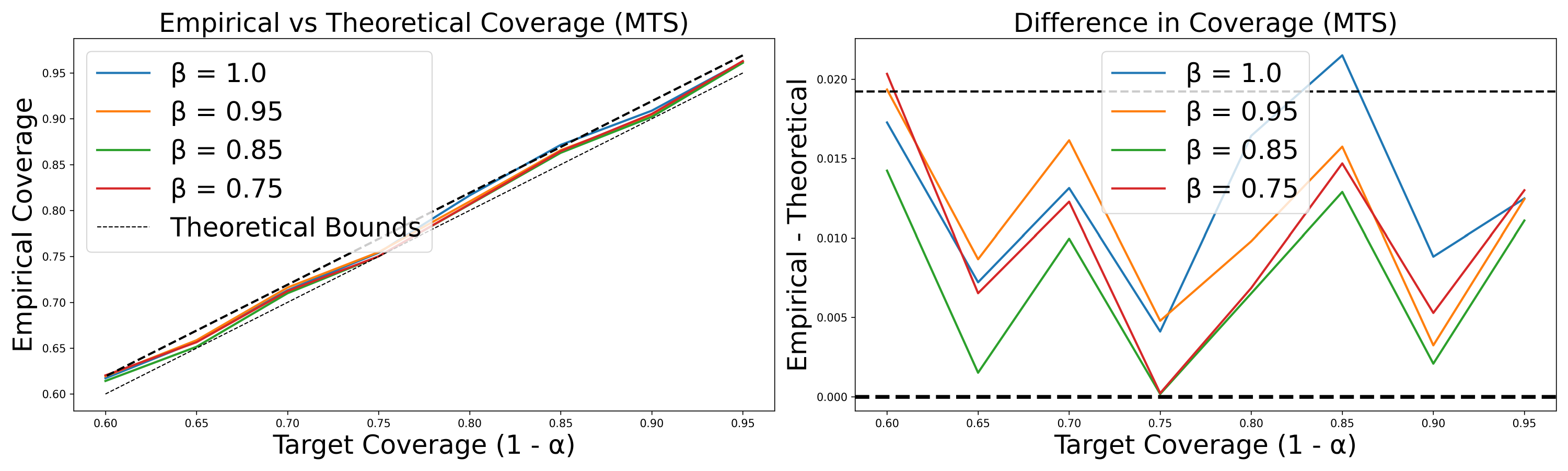}
        \caption{Llama 3}
        \end{subfigure}
        \begin{subfigure}{0.48\textwidth}
        \includegraphics[width=0.95\linewidth]{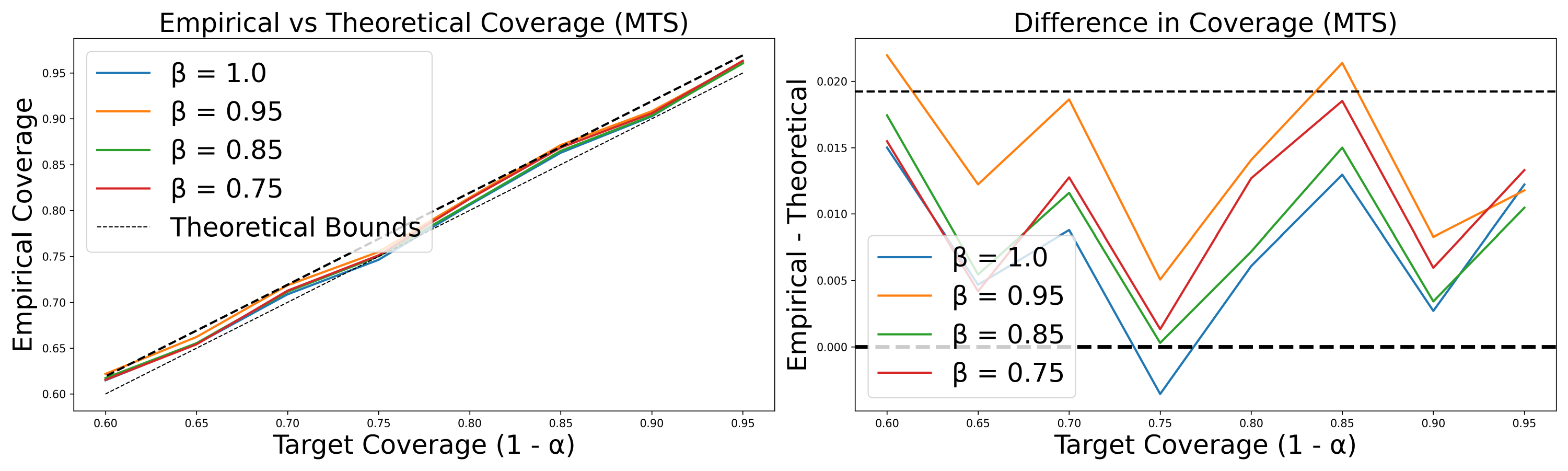}
        \caption{Qwen 3}
        \end{subfigure}
        \begin{subfigure}{0.48\textwidth}
        \includegraphics[width=0.95\linewidth]{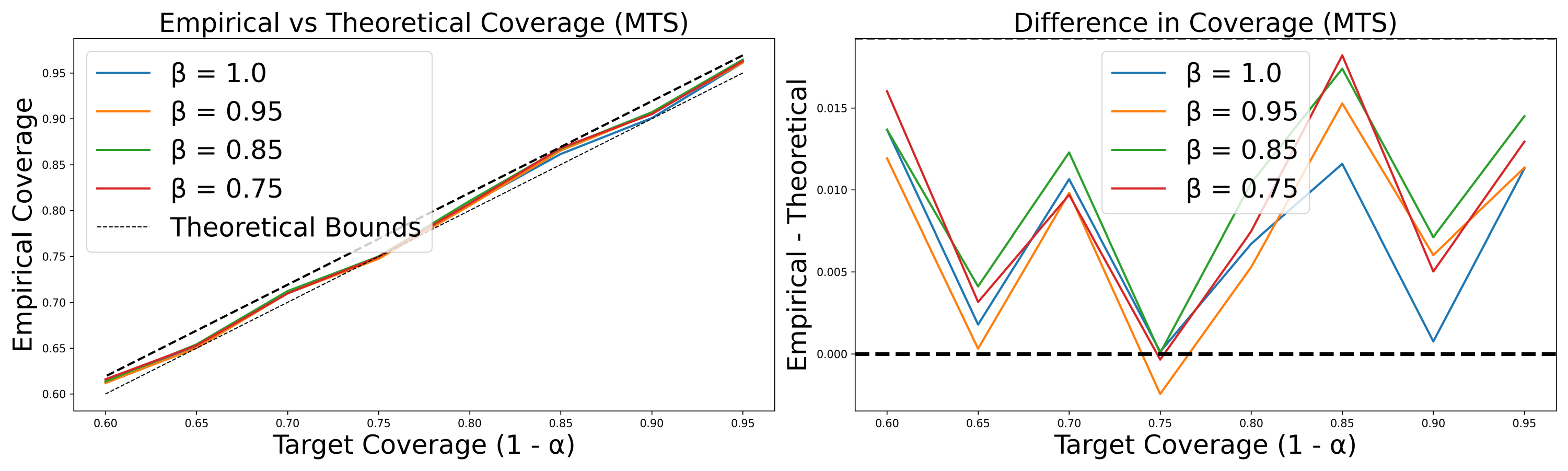}
        \caption{Gemini 2.0 Flash-Lite}
        \end{subfigure}
        \begin{subfigure}{0.48\textwidth}
        \includegraphics[width=0.95\linewidth]{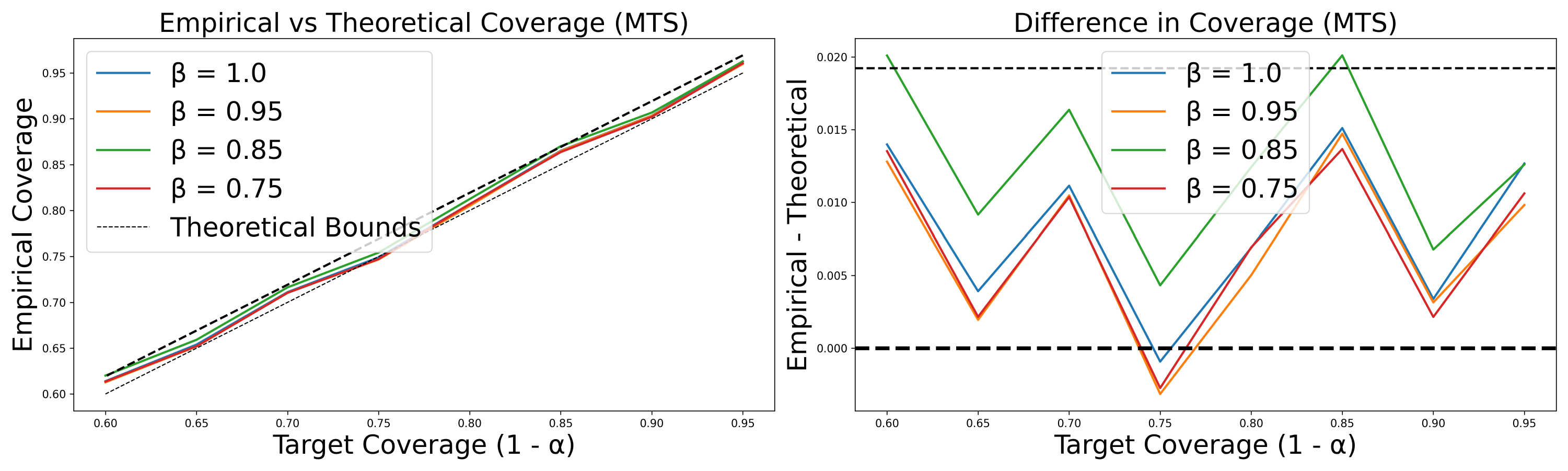}
        \caption{Gemini 2.5 Flash}
        \end{subfigure}
        \caption{User-specified probability of achieving coverage (1-$\alpha$) versus empirical probability of achieving coverage, on MTS dataset. Dashed lines show theoretical bounds given in \Cref{theorem}. Results are averaged over 400 random splits of calibration and test data.}
    \label{fig:calibration_MTS}
\end{figure}

\begin{figure}[t]
    \centering
        \begin{subfigure}{0.48\textwidth}
        \includegraphics[width=0.95\linewidth]{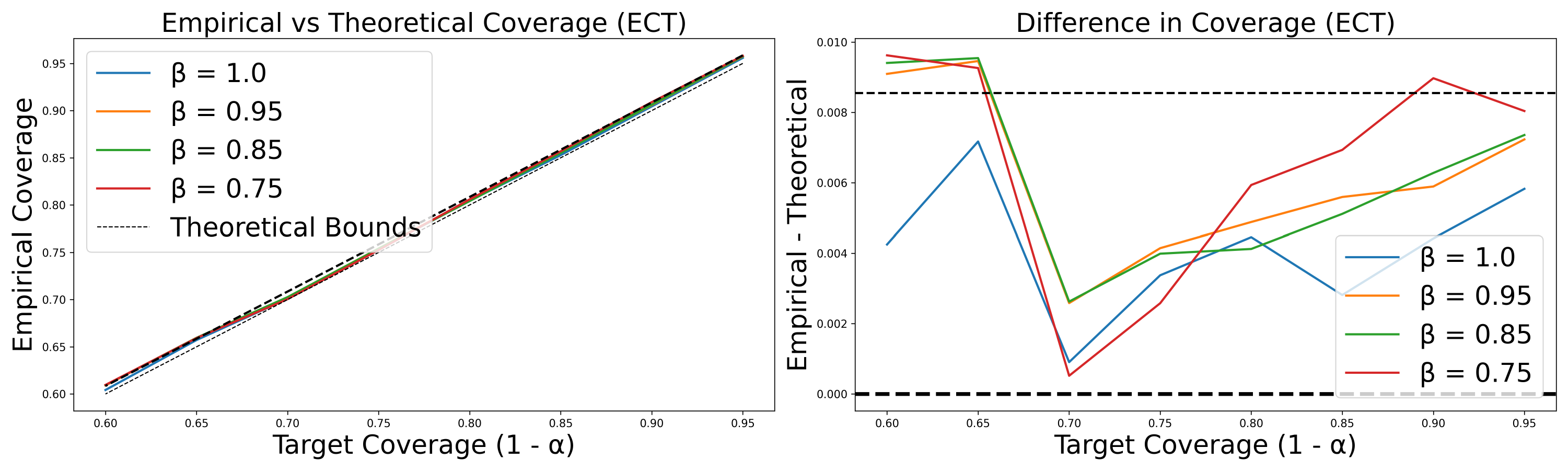}
        \caption{Cosine Similarity Centrality}
        \end{subfigure}
        \begin{subfigure}{0.48\textwidth}
        \includegraphics[width=0.95\linewidth]{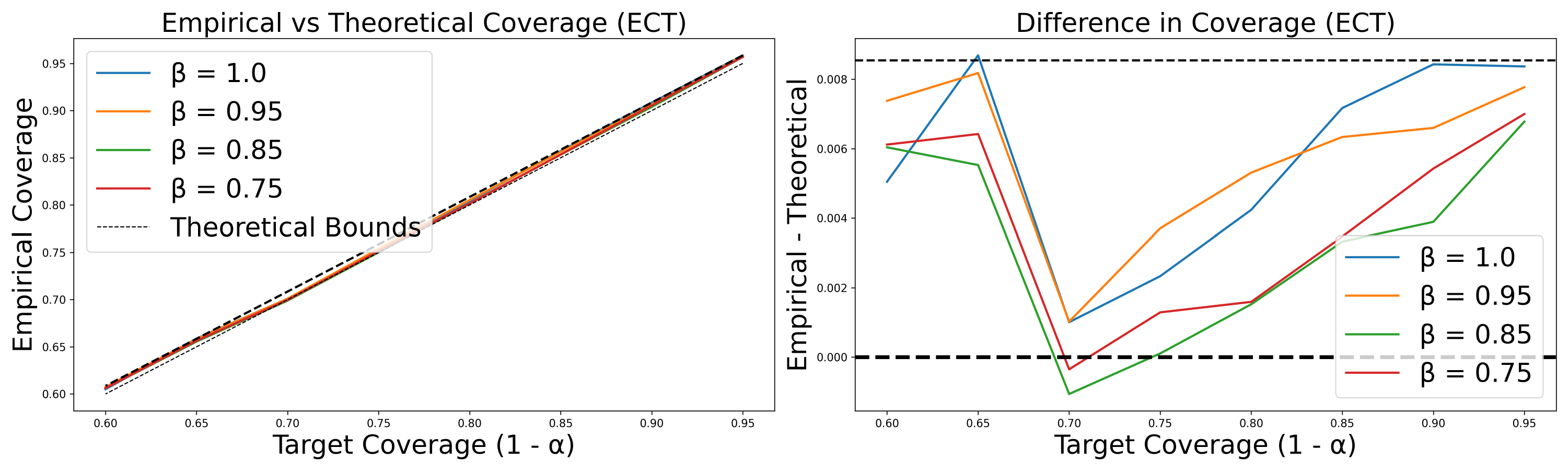}
        \caption{Sentence Centrality}
        \end{subfigure}
        \begin{subfigure}{0.48\textwidth}
        \includegraphics[width=0.95\linewidth]{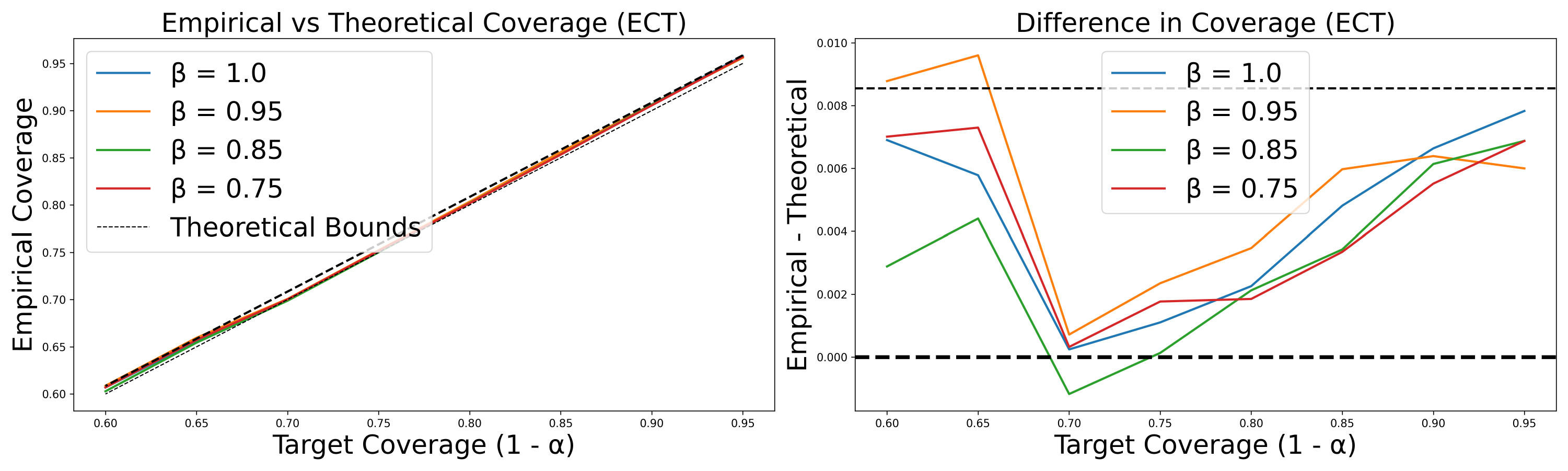}
        \caption{GUSUM}
        \end{subfigure}
        \begin{subfigure}{0.48\textwidth}
        \includegraphics[width=0.95\linewidth]{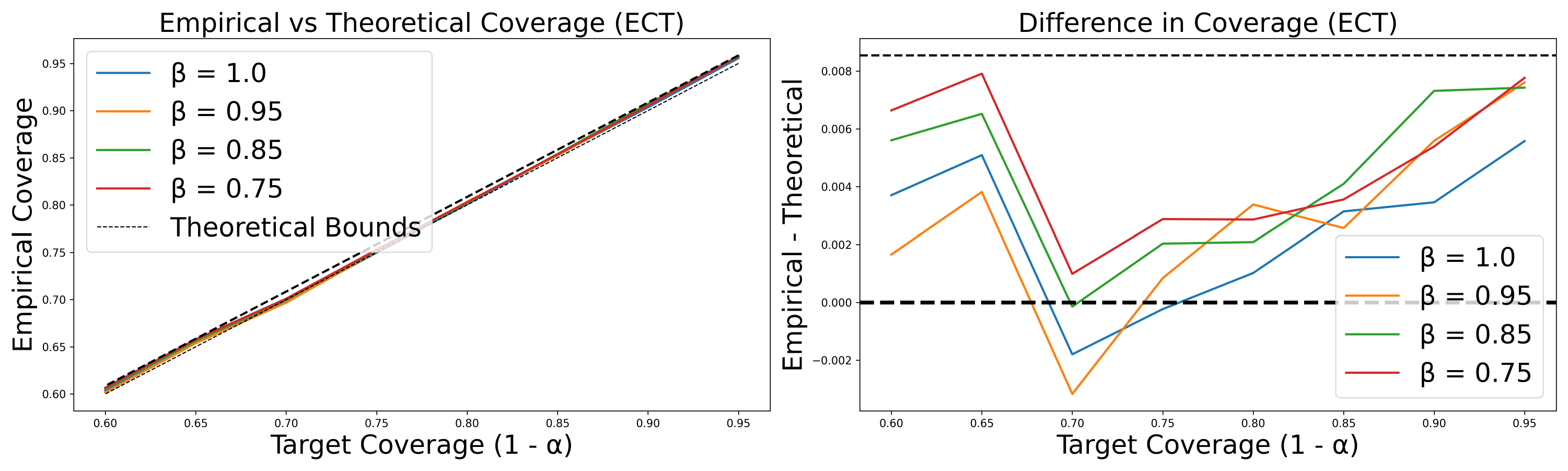}
        \caption{LexRank}
        \end{subfigure}
        \begin{subfigure}{0.48\textwidth}
        \includegraphics[width=0.95\linewidth]{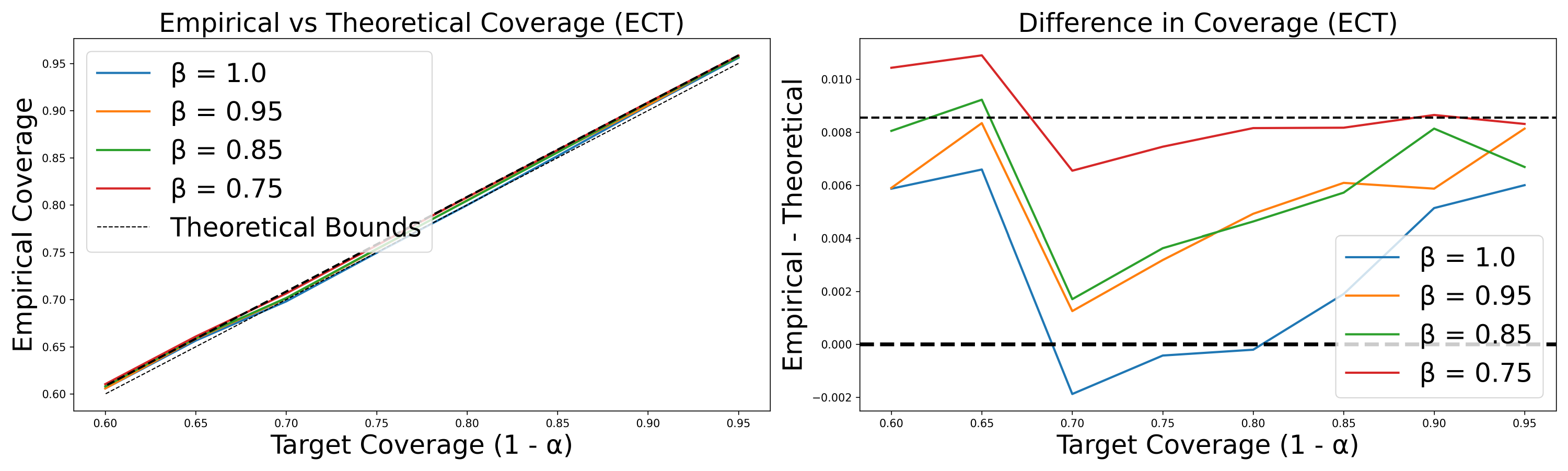}
        \caption{GPT-4o mini}
        \end{subfigure}
        \begin{subfigure}{0.48\textwidth}
        \includegraphics[width=0.95\linewidth]{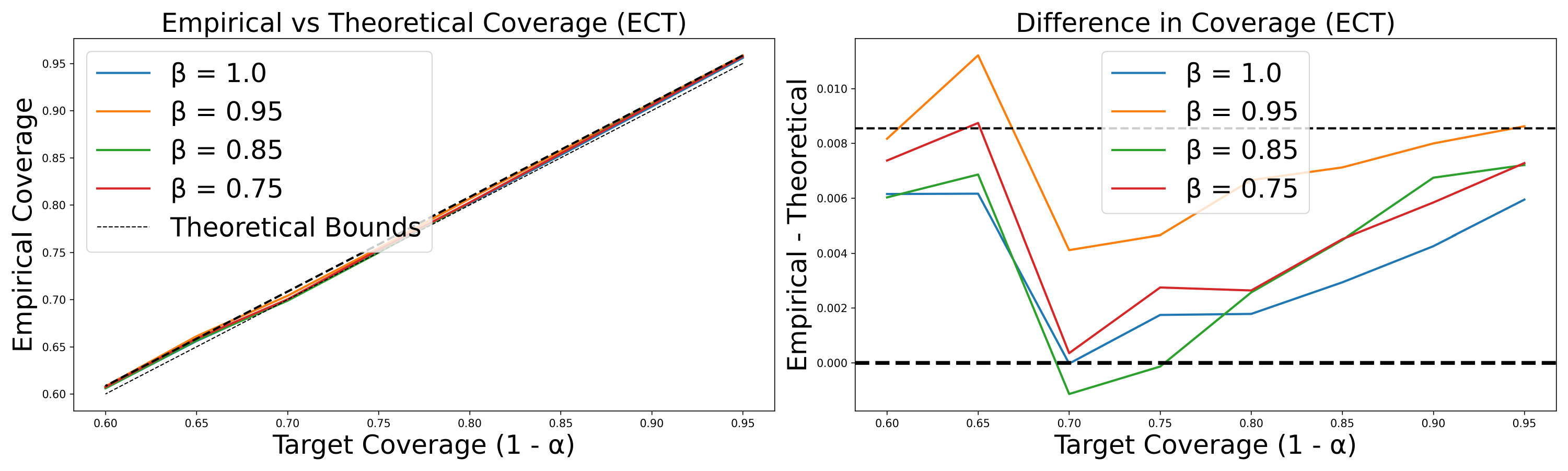}
        \caption{Llama 3}
        \end{subfigure}
        \begin{subfigure}{0.48\textwidth}
        \includegraphics[width=0.95\linewidth]{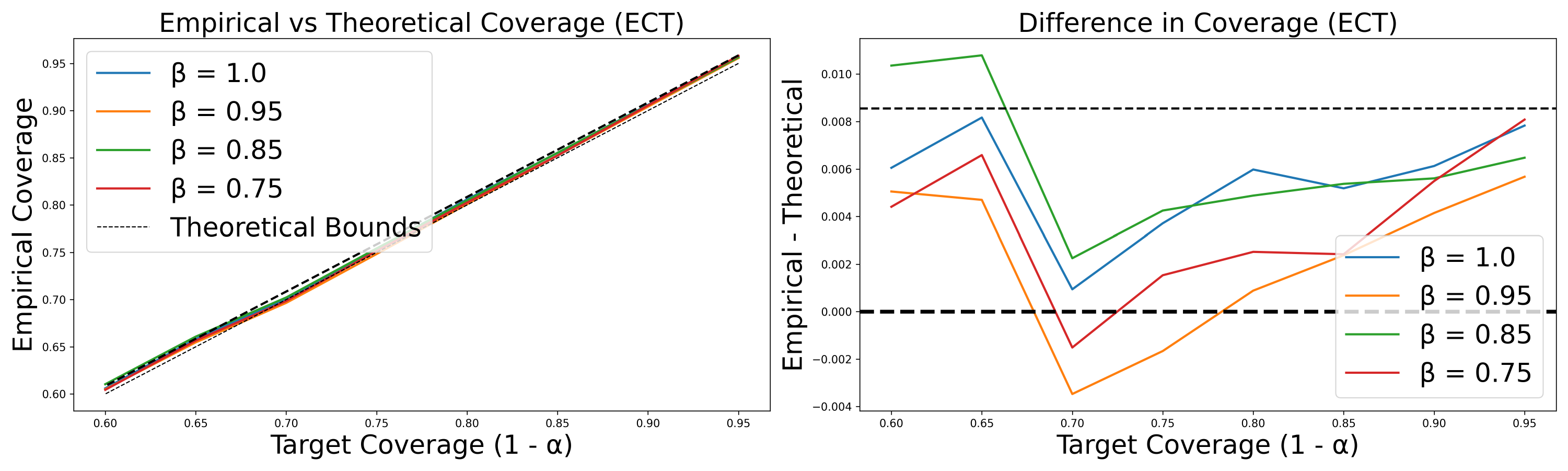}
        \caption{Qwen 3}
        \end{subfigure}
        \begin{subfigure}{0.48\textwidth}
        \includegraphics[width=0.95\linewidth]{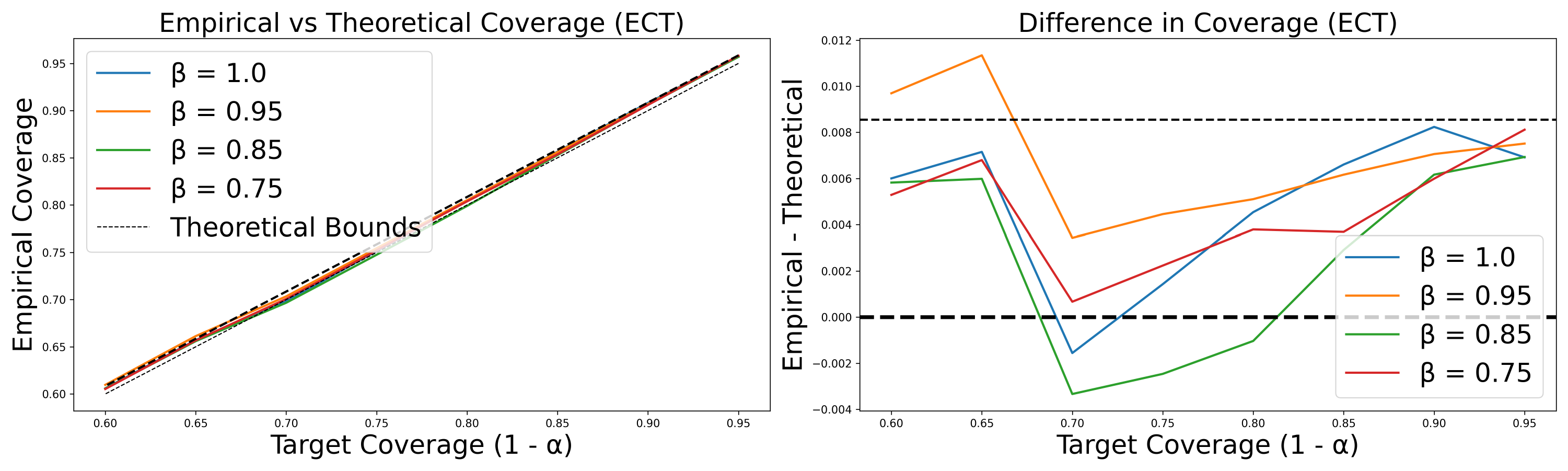}
        \caption{Gemini 2.0 Flash-Lite}
        \end{subfigure}
        \begin{subfigure}{0.48\textwidth}
        \includegraphics[width=0.95\linewidth]{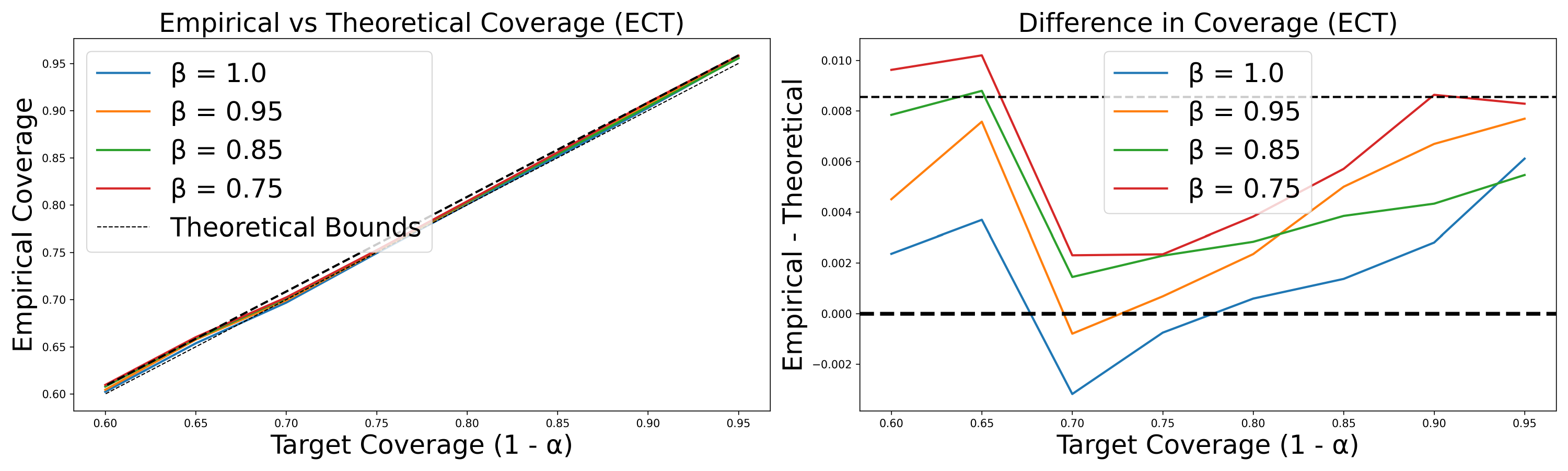}
        \caption{Gemini 2.5 Flash}
        \end{subfigure}
        \caption{User-specified probability of achieving coverage (1-$\alpha$) vs. empirical probability of achieving coverage, on the ECT dataset. Dashed lines show theoretical bounds given in \Cref{theorem}. Results are averaged over 400 random splits of calibration and test data.}
    \label{fig:calibration_ECT}
\end{figure}

To supplement the results in \Cref{sec:empirical}, here we show the empirical coverage obtained by all methods across all datasets. This confirms that all numerical comparisons in our work are fair in that they all achieve the expected coverage level. In Figures~\ref{fig:calibration_CNNDM} - \ref{fig:calibration_ECT} we shows plots analogous to \Cref{fig:conformal_calibration} in the main text. All parameters are identical to that in the main text. We see a similar trend where nearly all datapoints fall within the bounds for all plots, with occasional small deviations due to the inherent randomness involved and finite sample sizes.

\clearpage

\clearpage
\subsection{Error Rate vs. Recall Plots for all Datasets and Methods}\label{app:error_recall}
\begin{figure}[t]
    \centering
        \begin{subfigure}{0.48\textwidth}
        \includegraphics[width=0.95\linewidth, trim={30 0 60 40}, clip]{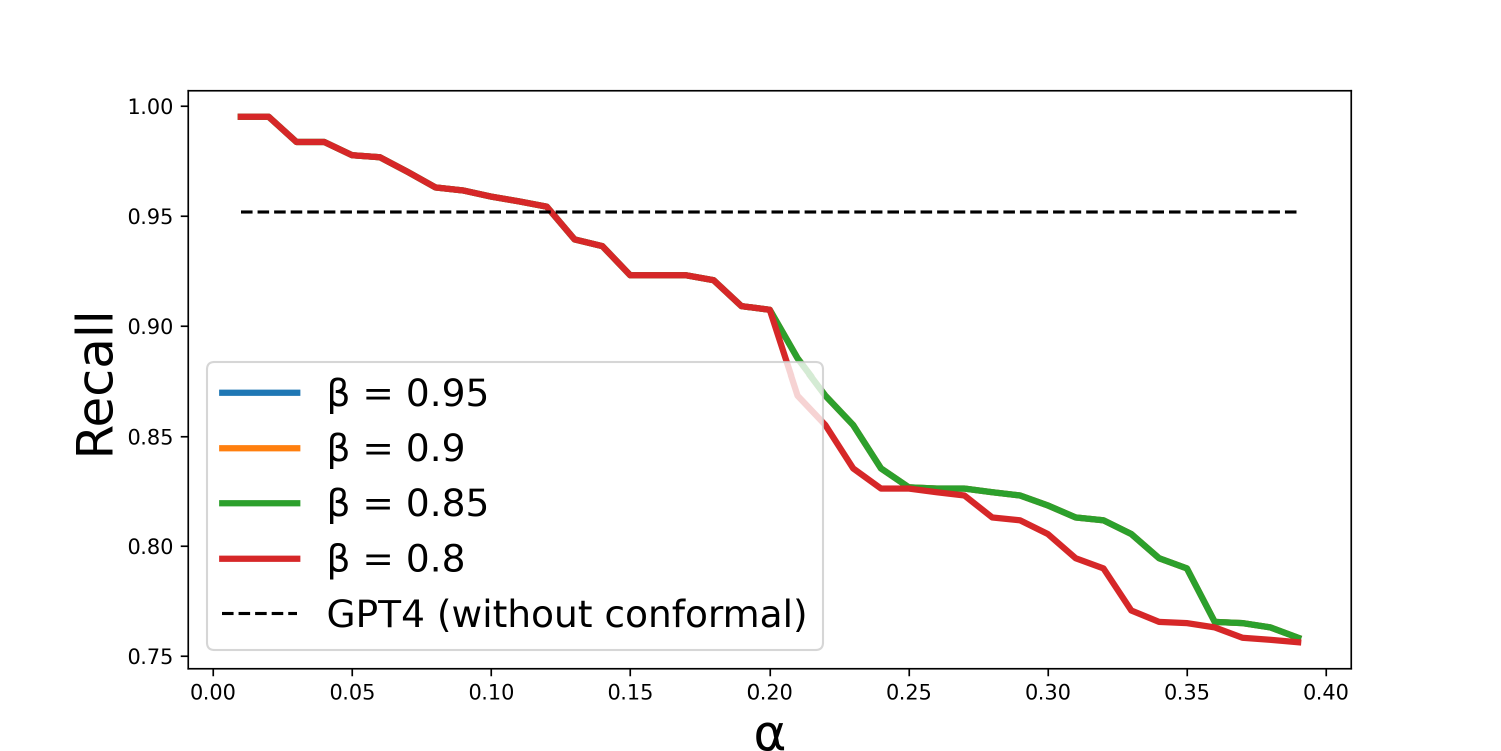}
        \caption{Cosine Similarity Centrality}
        \end{subfigure}
        \begin{subfigure}{0.48\textwidth}
        \includegraphics[width=0.95\linewidth, trim={30 0 60 40}, clip]{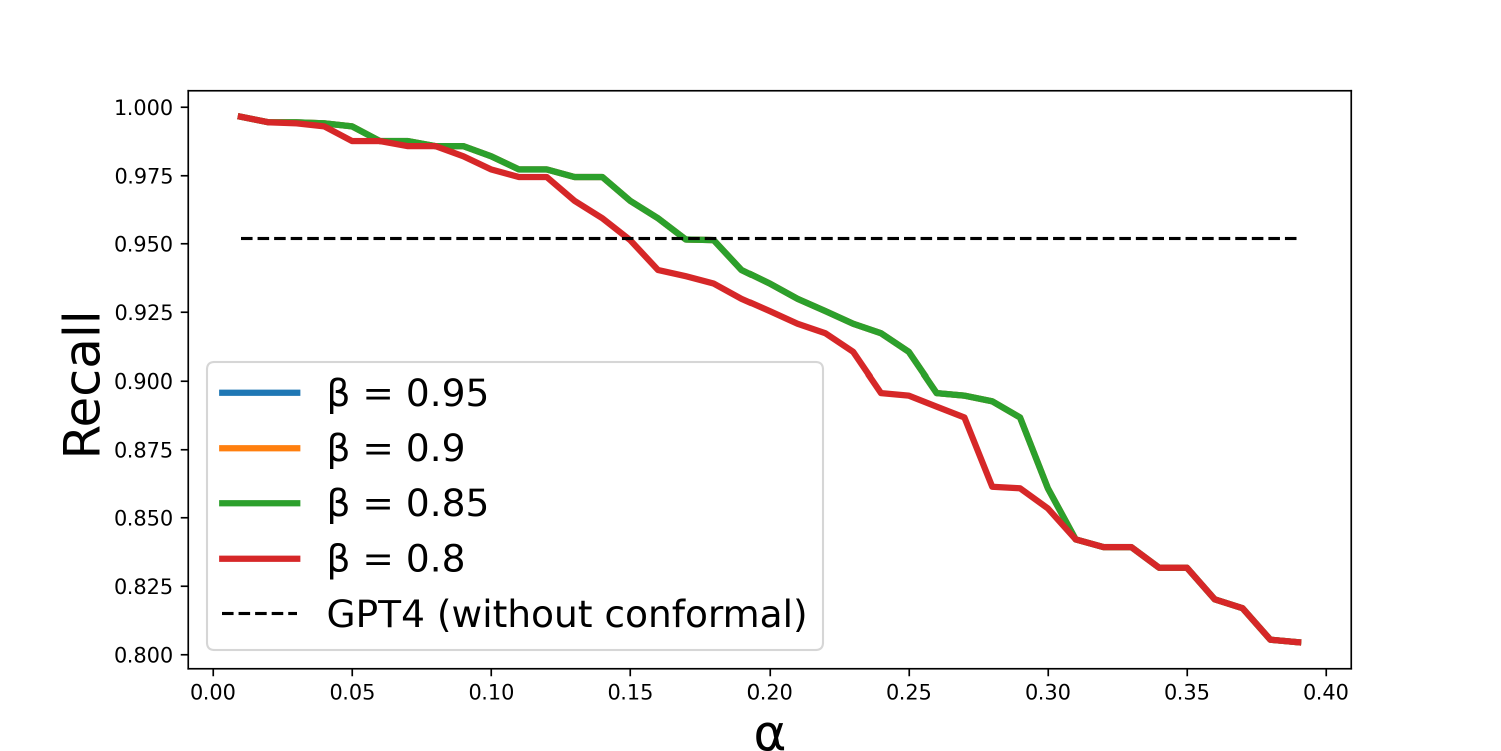}
        \caption{Sentence Centrality}
        \end{subfigure}
        \begin{subfigure}{0.48\textwidth}
        \includegraphics[width=0.95\linewidth, trim={30 0 60 40}, clip]{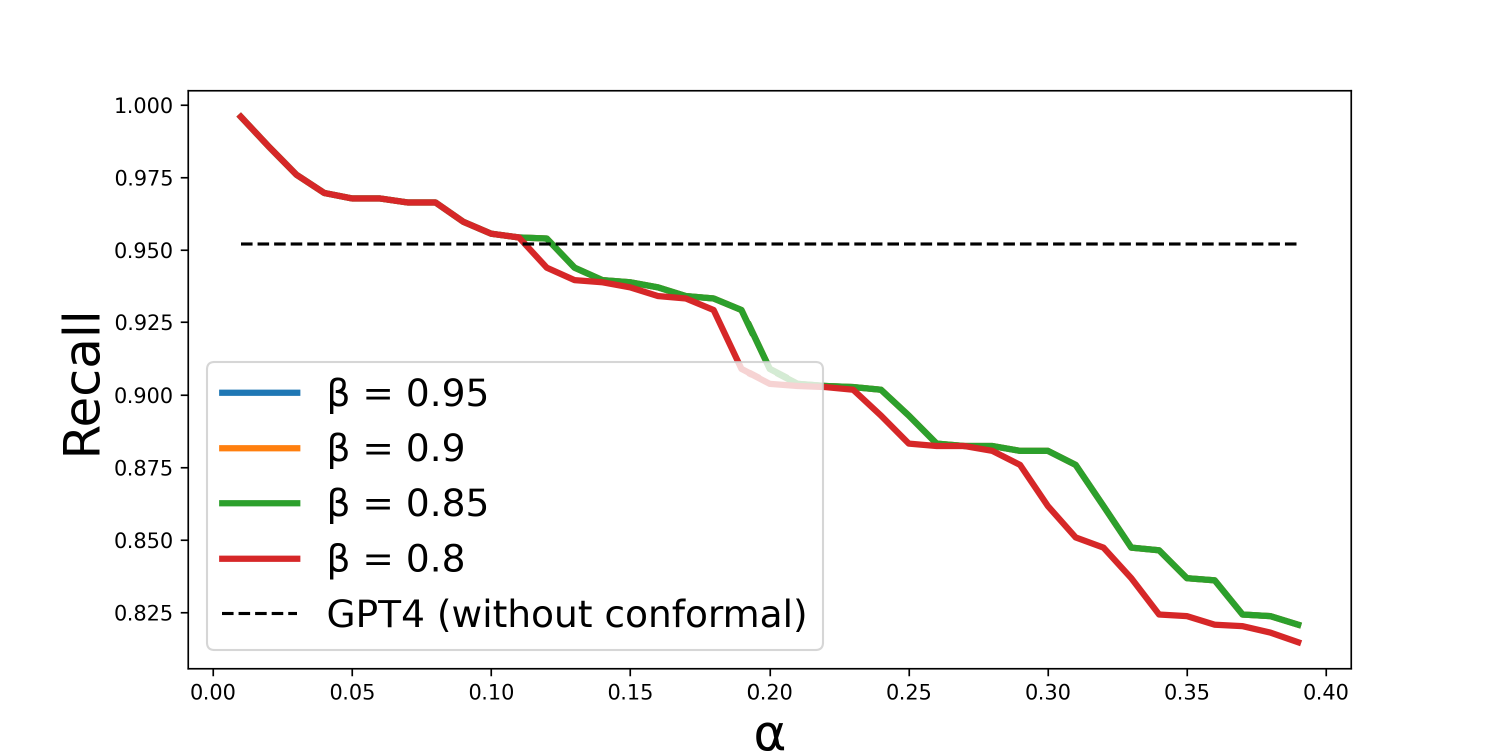}
        \caption{GUSUM}
        \end{subfigure}
        \begin{subfigure}{0.48\textwidth}
        \includegraphics[width=0.95\linewidth, trim={30 0 60 40}, clip]{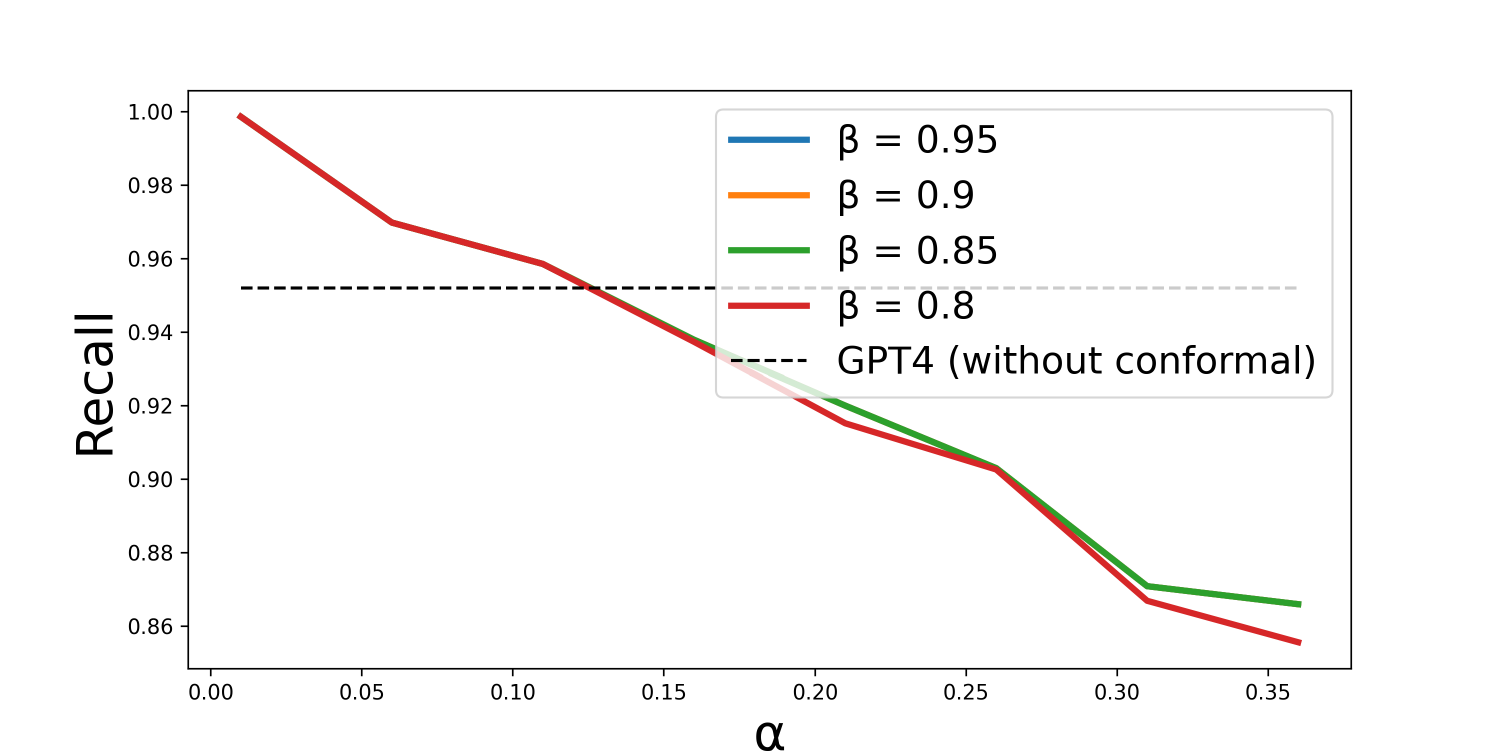}
        \caption{LexRank}
        \end{subfigure}
        \begin{subfigure}{0.48\textwidth}
        \includegraphics[width=0.95\linewidth, trim={30 0 60 40}, clip]{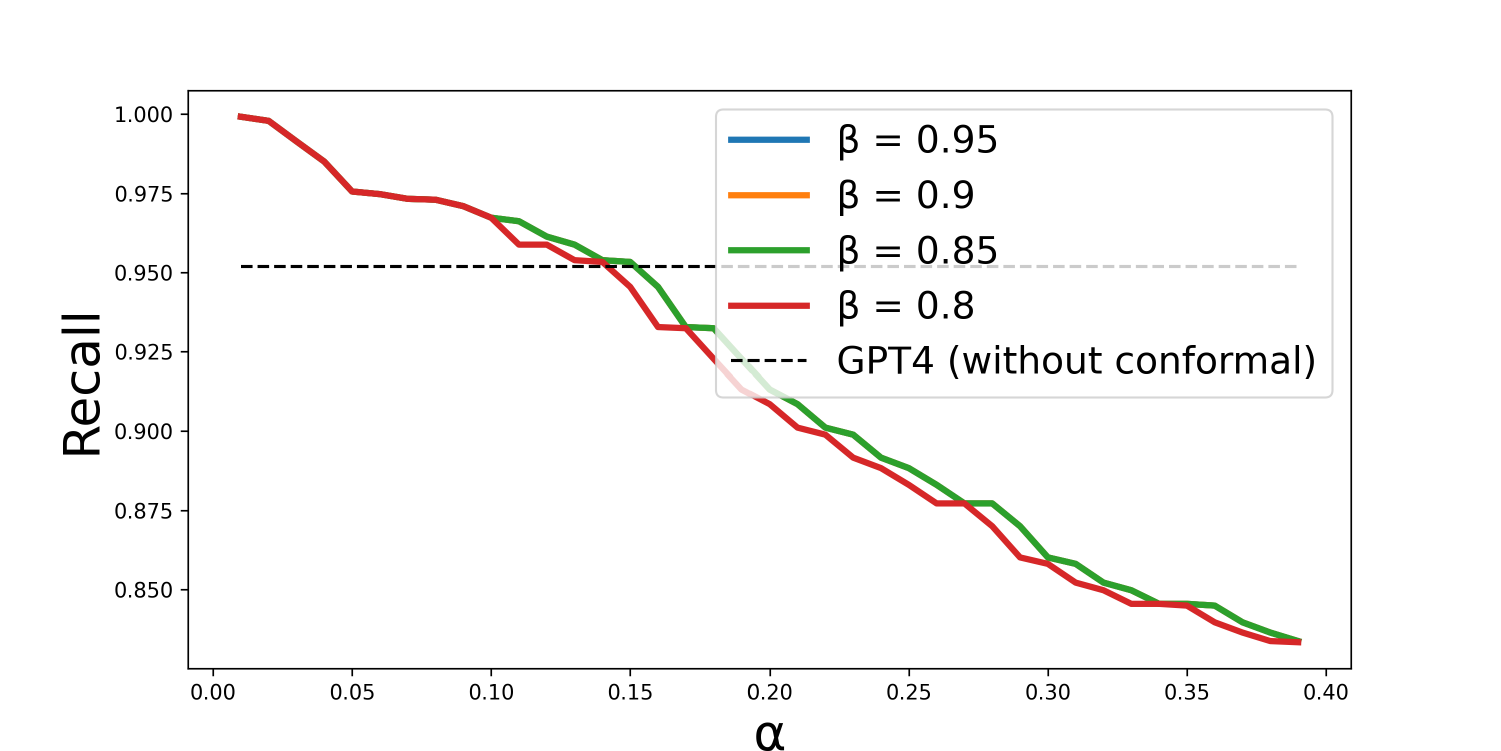}
        \caption{GPT-4o mini}
        \end{subfigure}
        \begin{subfigure}{0.48\textwidth}
        \includegraphics[width=0.95\linewidth, trim={30 0 60 40}, clip]{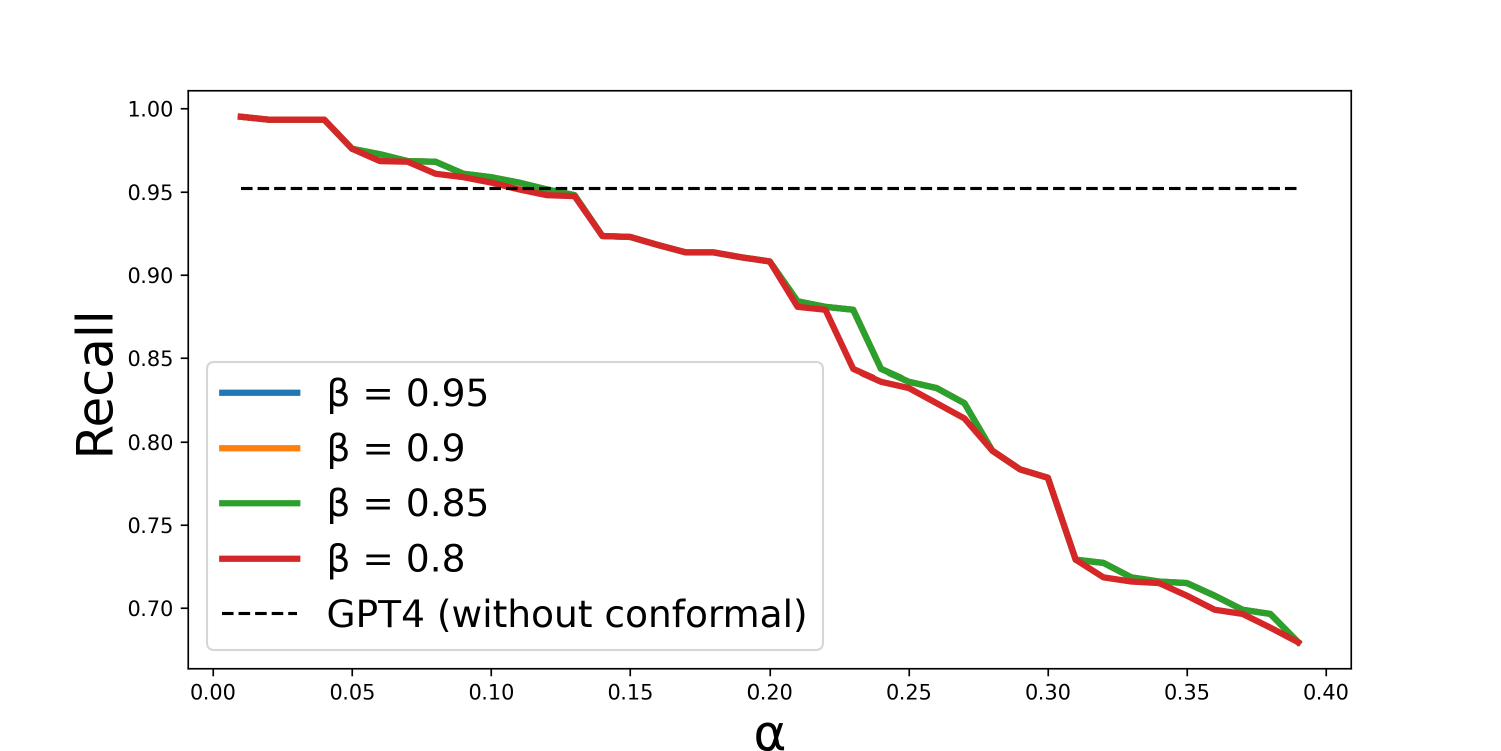}
        \caption{Llama 3}
        \end{subfigure}
        \begin{subfigure}{0.48\textwidth}
        \includegraphics[width=0.95\linewidth, trim={30 0 60 40}, clip]{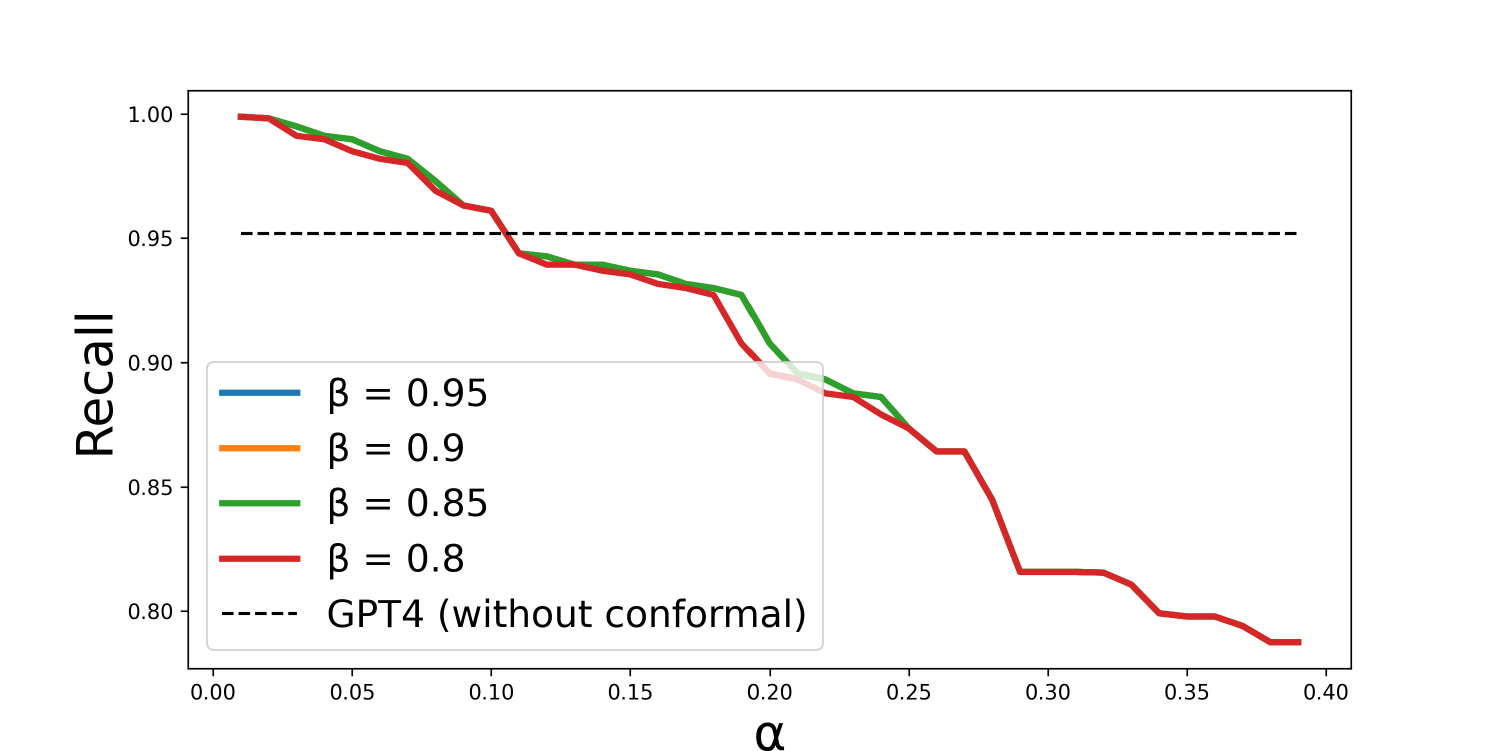}
        \caption{Qwen 3}
        \end{subfigure}
        \begin{subfigure}{0.48\textwidth}
        \includegraphics[width=0.95\linewidth, trim={30 0 60 40}, clip]{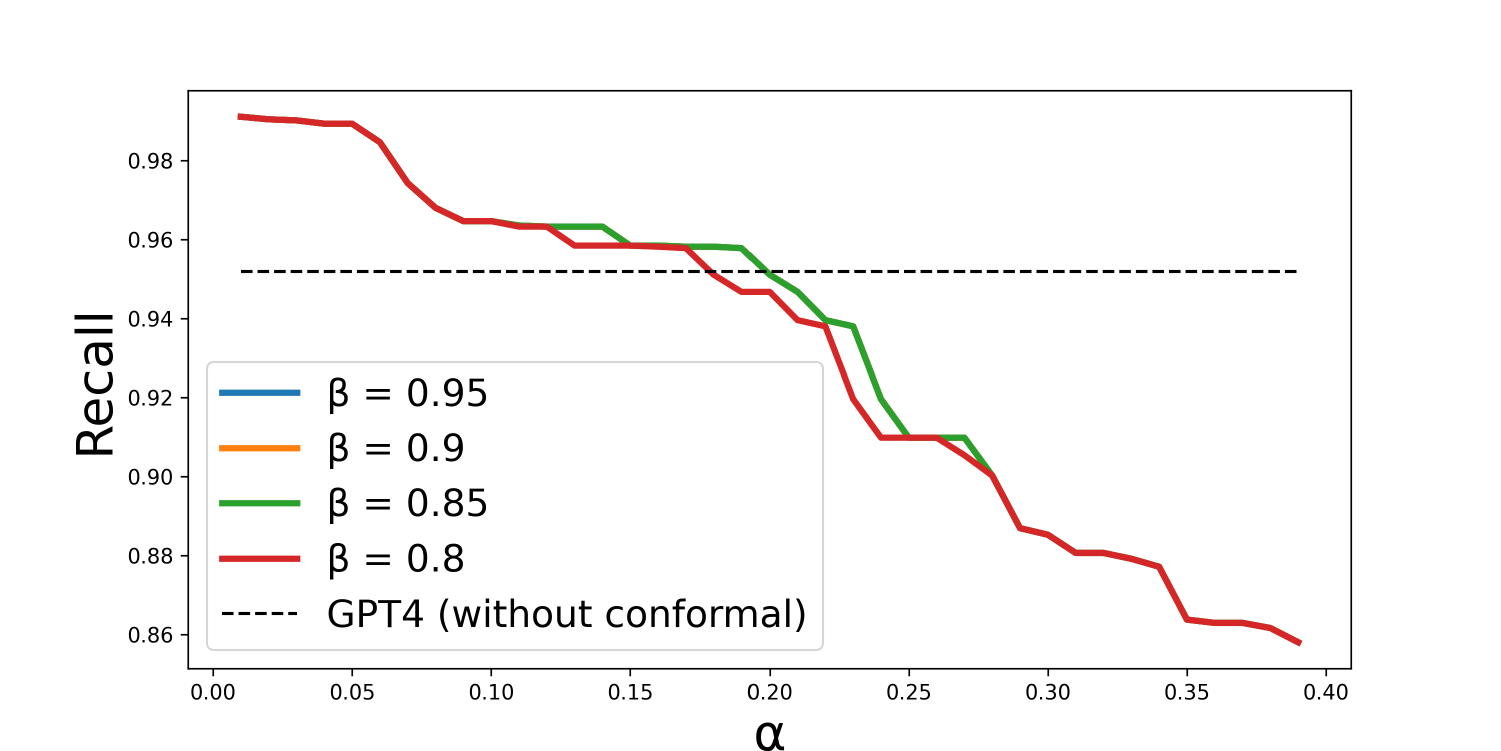}
        \caption{Gemini 2.0 FlashLite}
        \end{subfigure}
        \begin{subfigure}{0.48\textwidth}
        \includegraphics[width=0.95\linewidth, trim={30 0 60 40}, clip]{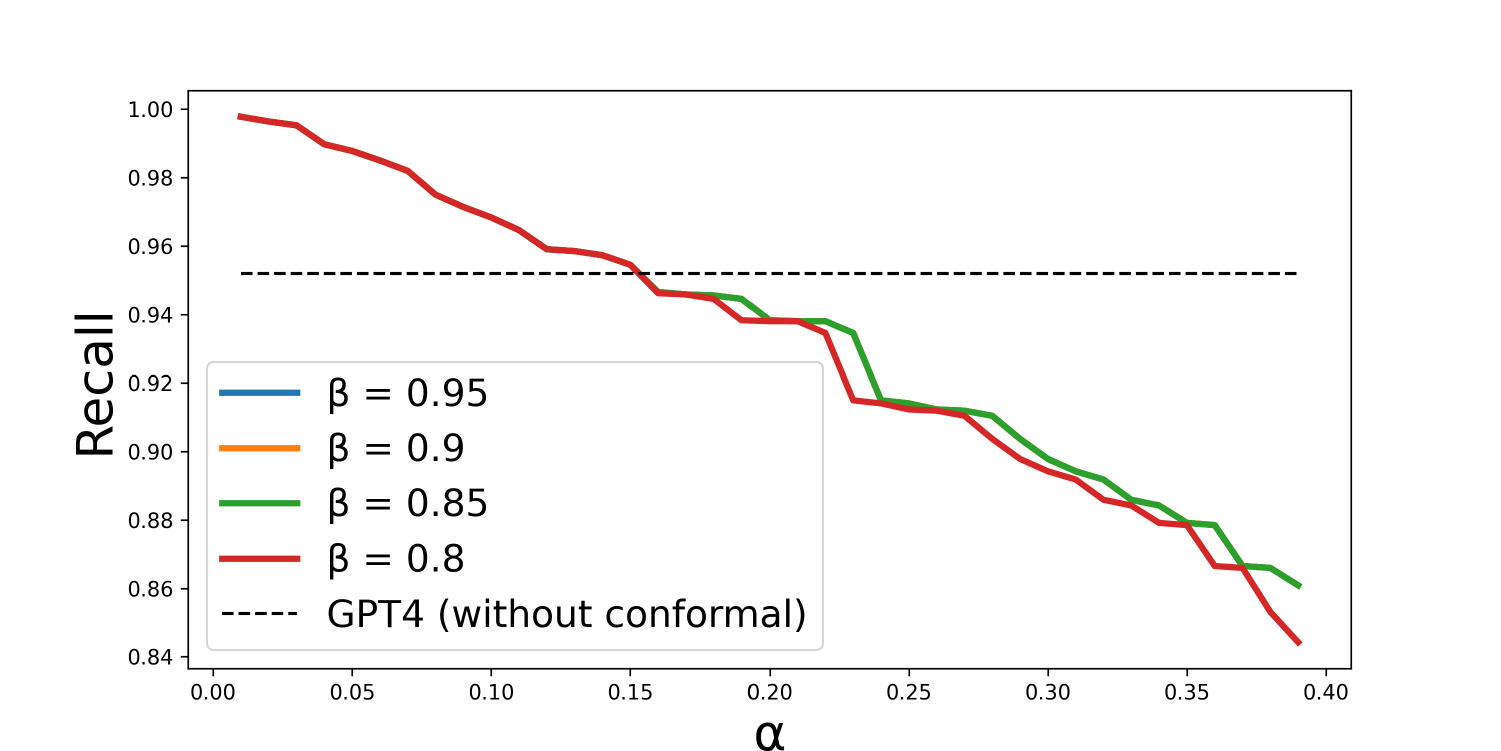}
        \caption{Gemini 2.5 Flash}
        \end{subfigure}
        \caption{Target error rate $\alpha$ versus empirical recall $B(y; y^*)$ of important sentences in summaries, averaged over the CNN/DM test set. The dashed line shows GPT-4o mini performance without using conformal prediction. Several curves may overlap when there are only a few discrete levels of empirical recall possible, making some values of $\beta$ equivalent.}
    \label{fig:recall_versus_alpha_CNNDM}
\end{figure}

\begin{figure}[t]
    \centering
        \begin{subfigure}{0.48\textwidth}
        \includegraphics[width=0.95\linewidth, trim={30 0 60 40}, clip]{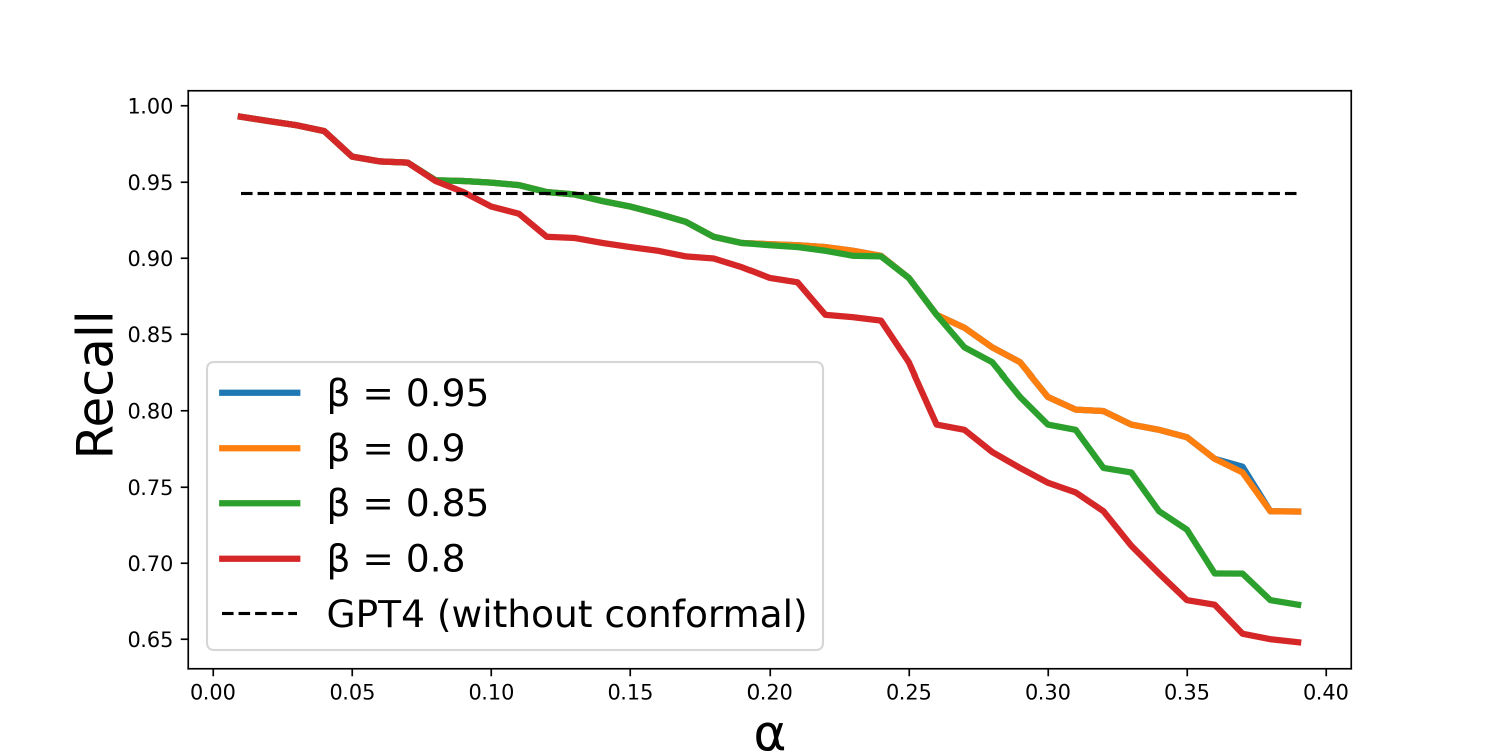}
        \caption{Cosine Similarity Centrality}
        \end{subfigure}
        \begin{subfigure}{0.48\textwidth}
        \includegraphics[width=0.95\linewidth, trim={30 0 60 40}, clip]{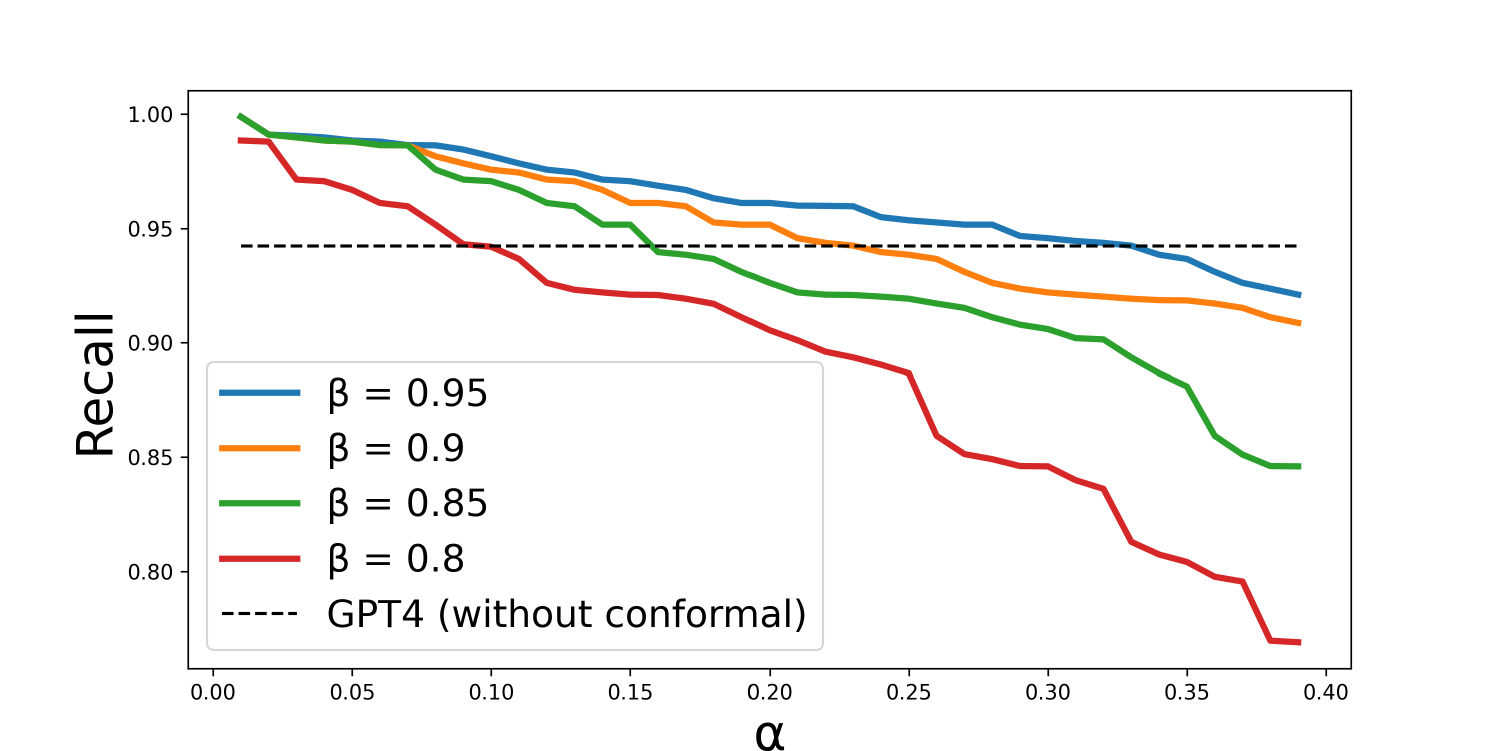}
        \caption{Sentence Centrality}
        \end{subfigure}
        \begin{subfigure}{0.48\textwidth}
        \includegraphics[width=0.95\linewidth, trim={30 0 60 40}, clip]{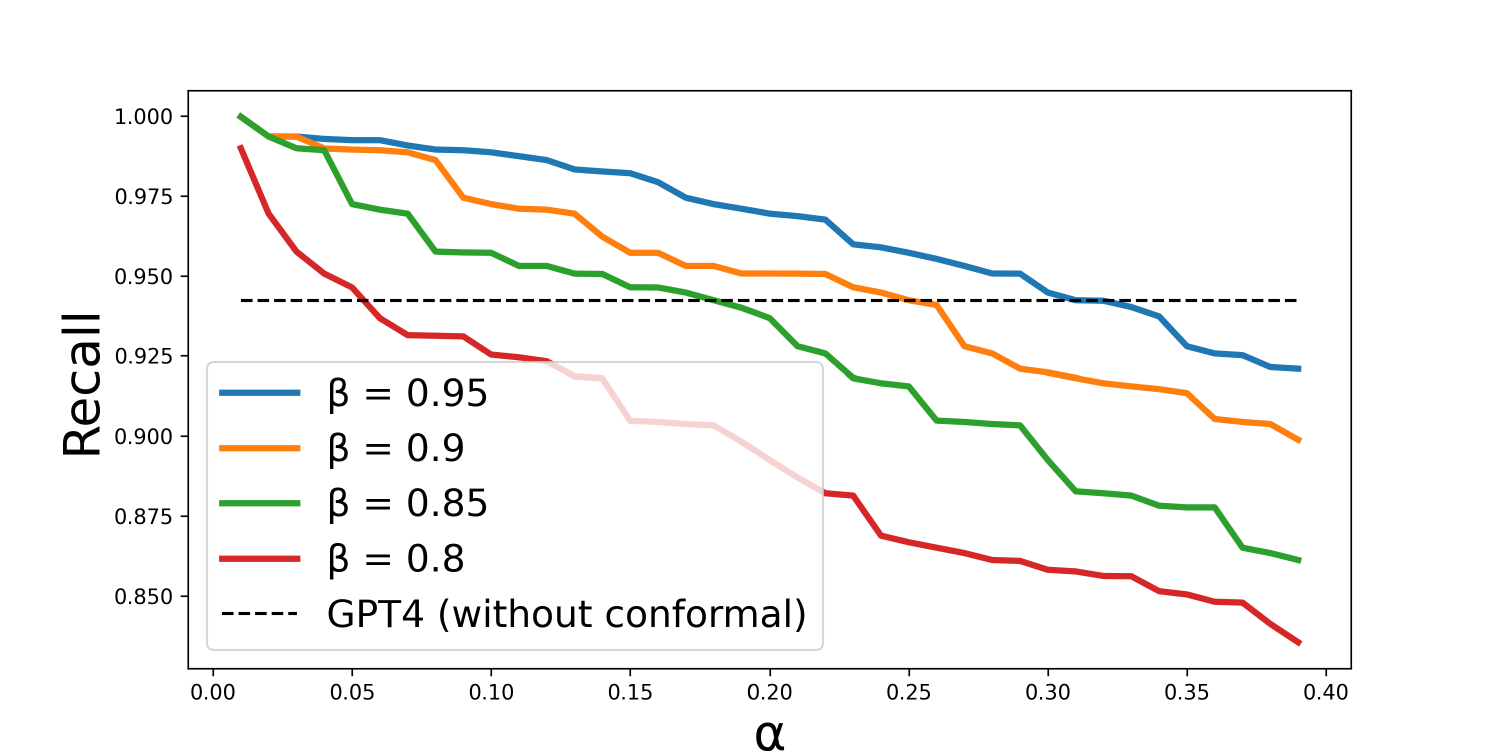}
        \caption{GUSUM}
        \end{subfigure}
        \begin{subfigure}{0.48\textwidth}
        \includegraphics[width=0.95\linewidth, trim={30 0 60 40}, clip]{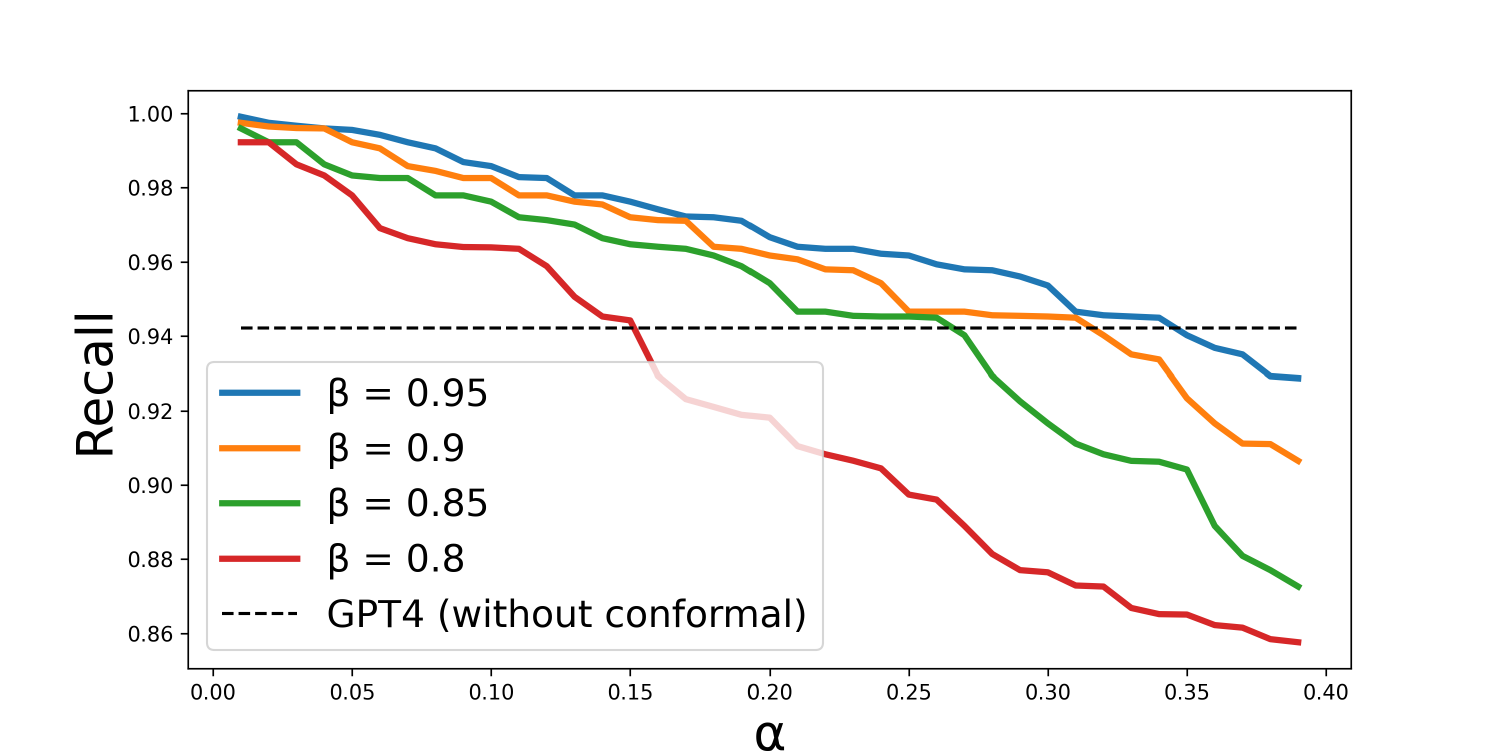}
        \caption{GPT-4o mini}
        \end{subfigure}
        \begin{subfigure}{0.48\textwidth}
        \includegraphics[width=0.95\linewidth, trim={30 0 60 40}, clip]{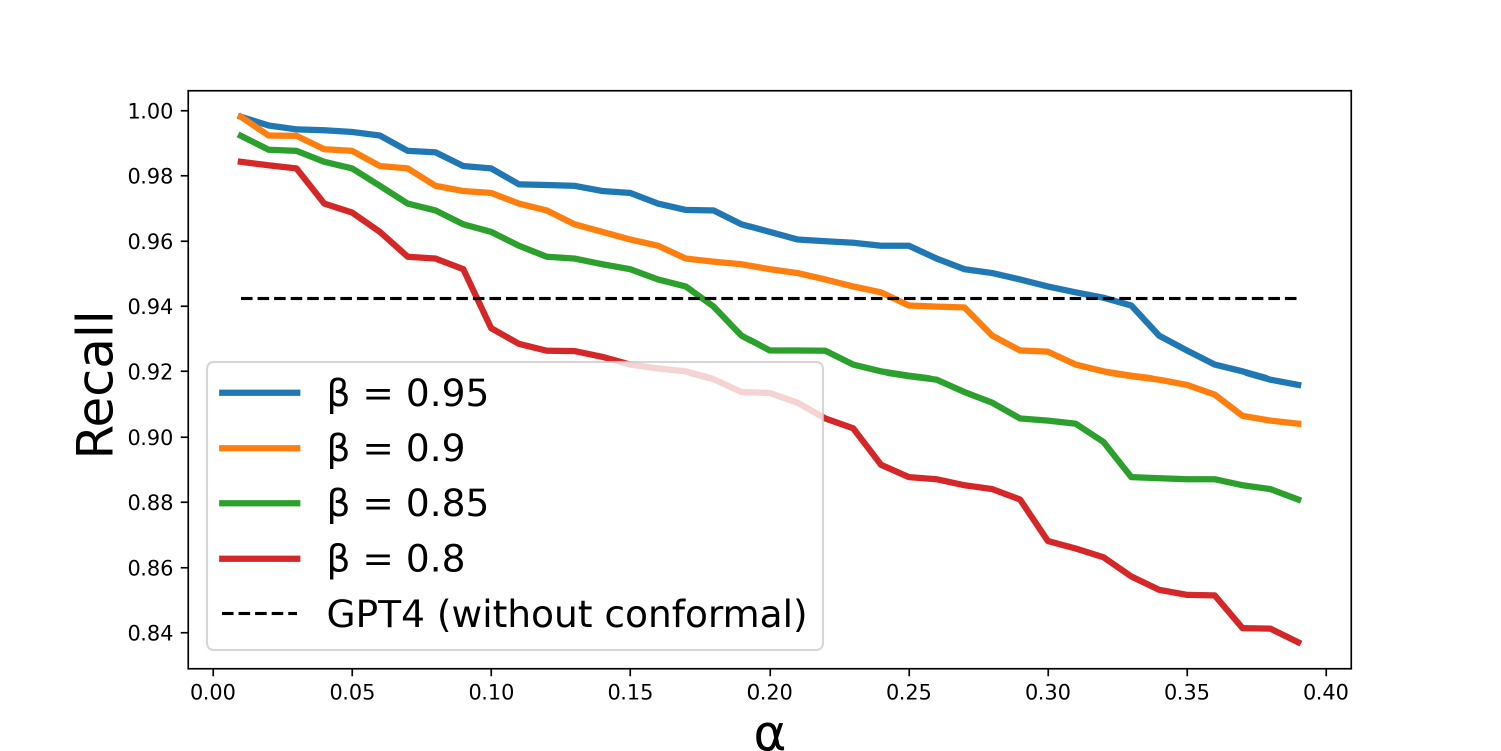}
        \caption{Llama 3}
        \end{subfigure}
        \begin{subfigure}{0.48\textwidth}
        \includegraphics[width=0.95\linewidth, trim={30 0 60 40}, clip]{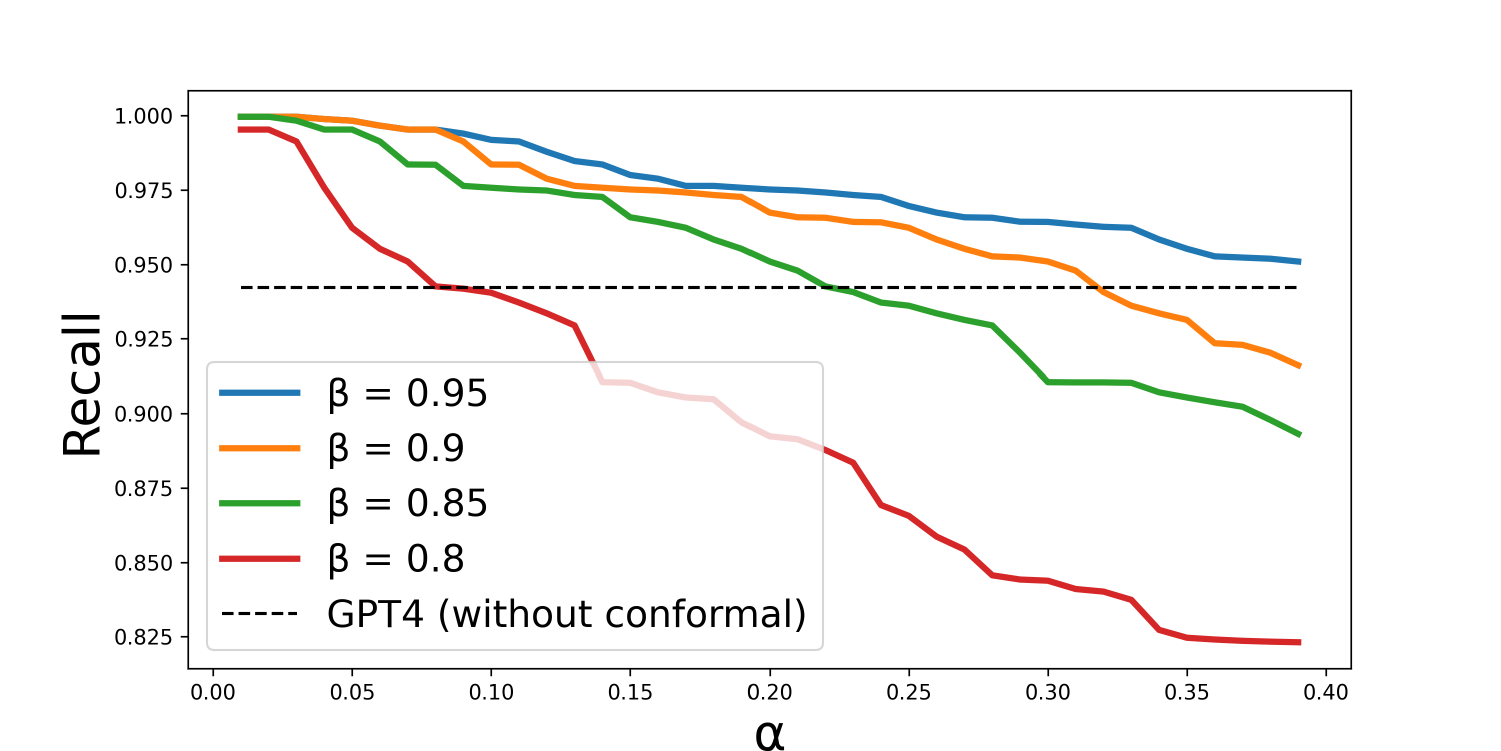}
        \caption{Qwen 3}
        \end{subfigure}
        \begin{subfigure}{0.48\textwidth}
        \includegraphics[width=0.95\linewidth, trim={30 0 60 40}, clip]{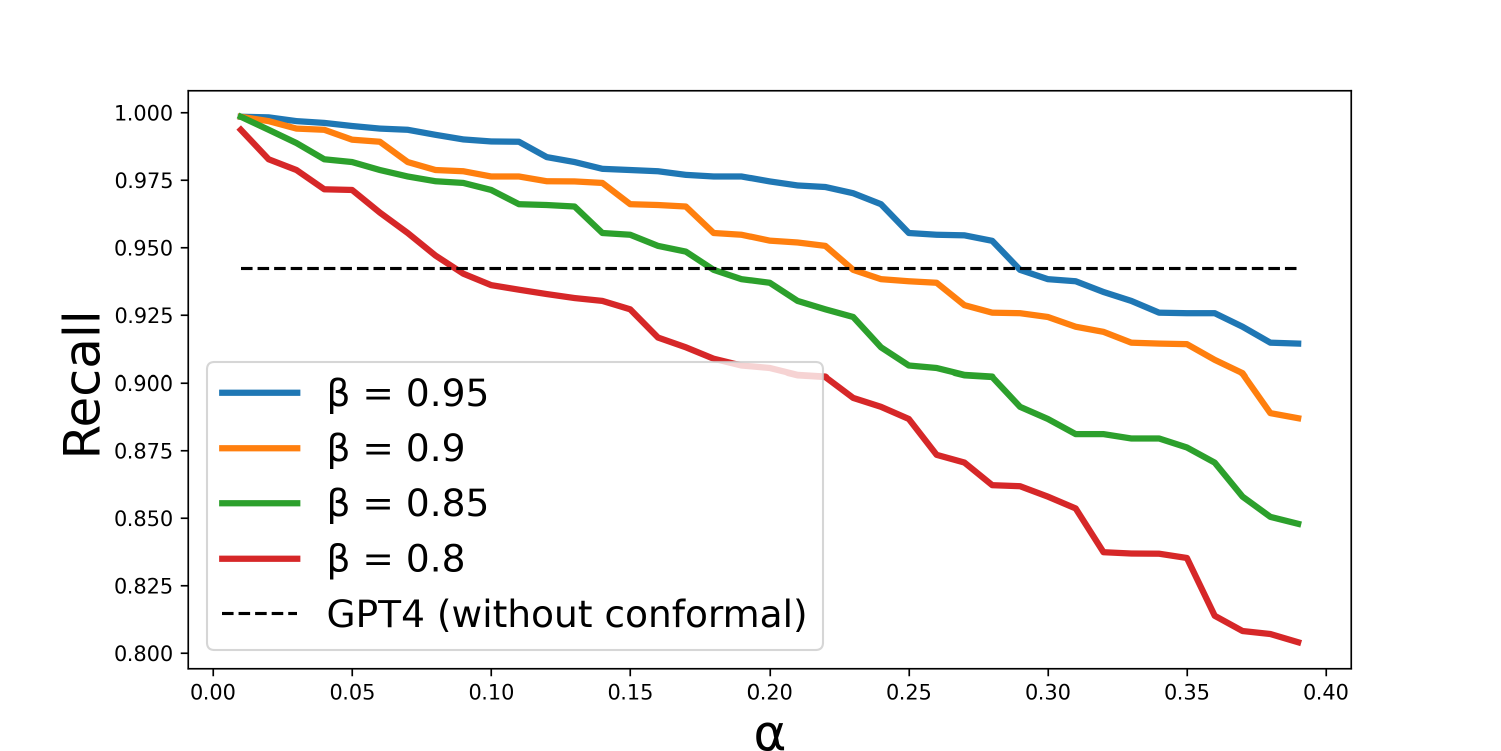}
        \caption{Gemini 2.0 Flash-Lite}
        \end{subfigure}
        \begin{subfigure}{0.48\textwidth}
        \includegraphics[width=0.95\linewidth, trim={30 0 60 40}, clip]{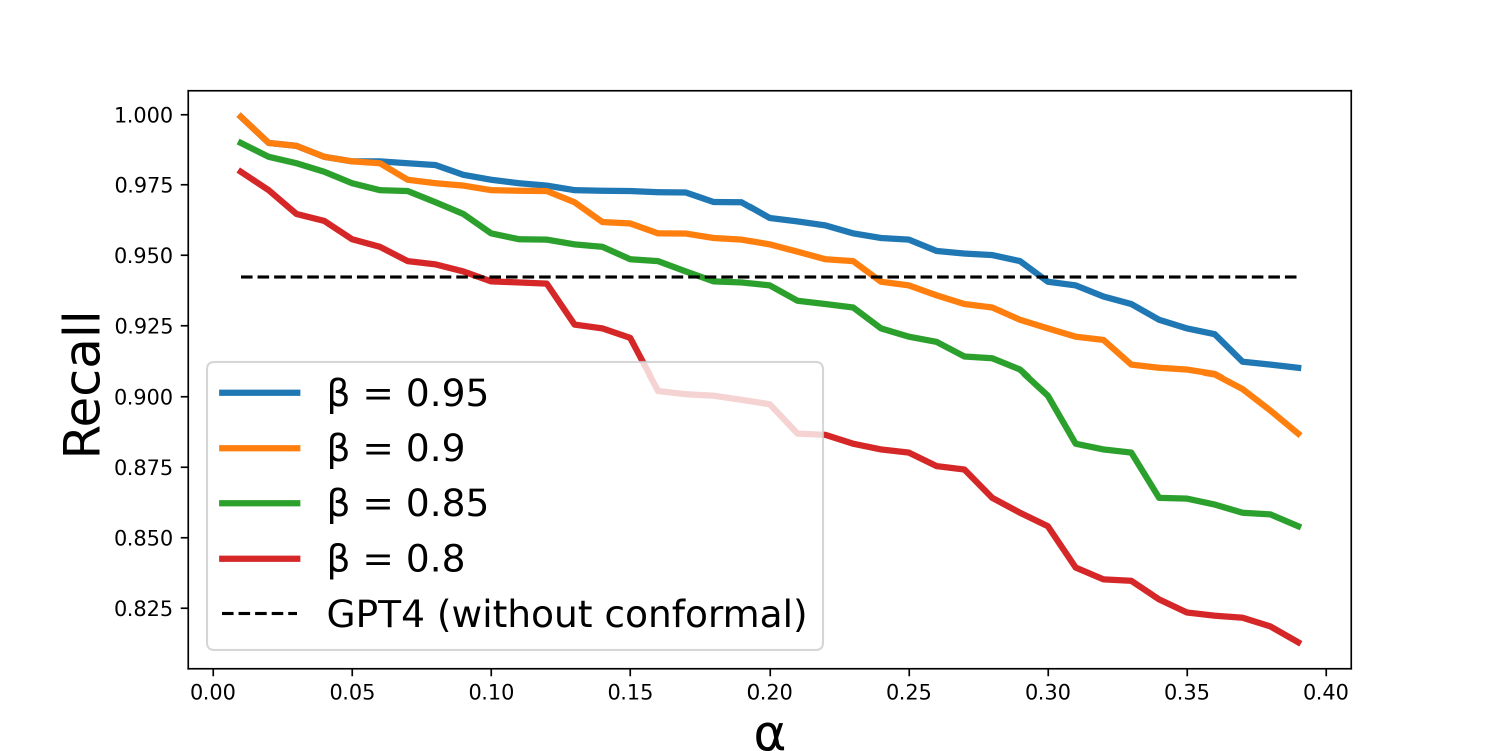}
        \caption{Gemini 2.5 Flash}
        \end{subfigure}
        \caption{Target error rate $\alpha$ versus empirical recall $B(y; y^*)$ of important sentences in summaries, averaged over the CSDS test set. The dashed line shows GPT-4o mini performance without using conformal prediction.}
    \label{fig:recall_versus_alpha_CSDS}
\end{figure}

\begin{figure}[t]
    \centering
        \begin{subfigure}{0.48\textwidth}
        \includegraphics[width=0.95\linewidth, trim={30 0 60 40}, clip]{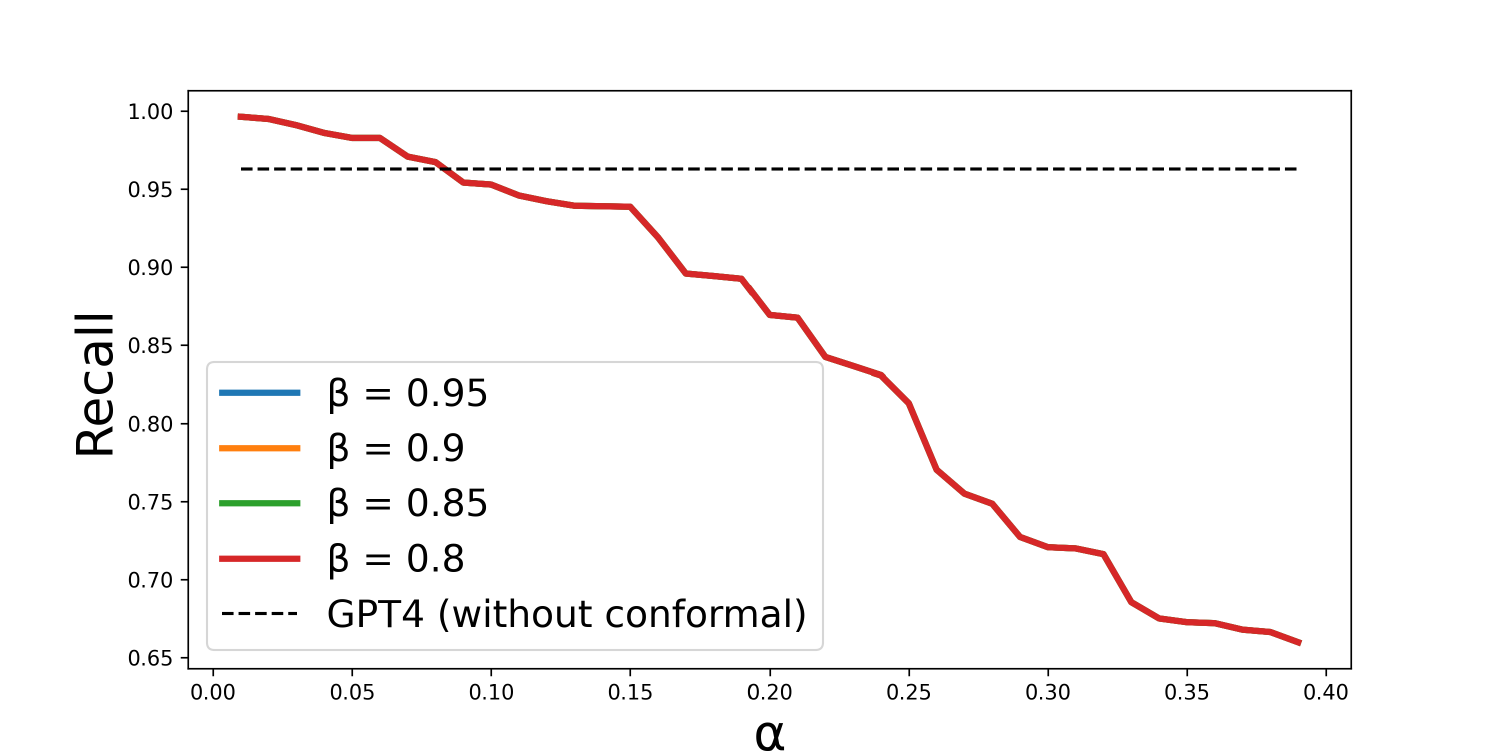}
        \caption{Cosine Similarity Centrality}
        \end{subfigure}
        \begin{subfigure}{0.48\textwidth}
        \includegraphics[width=0.95\linewidth, trim={30 0 60 40}, clip]{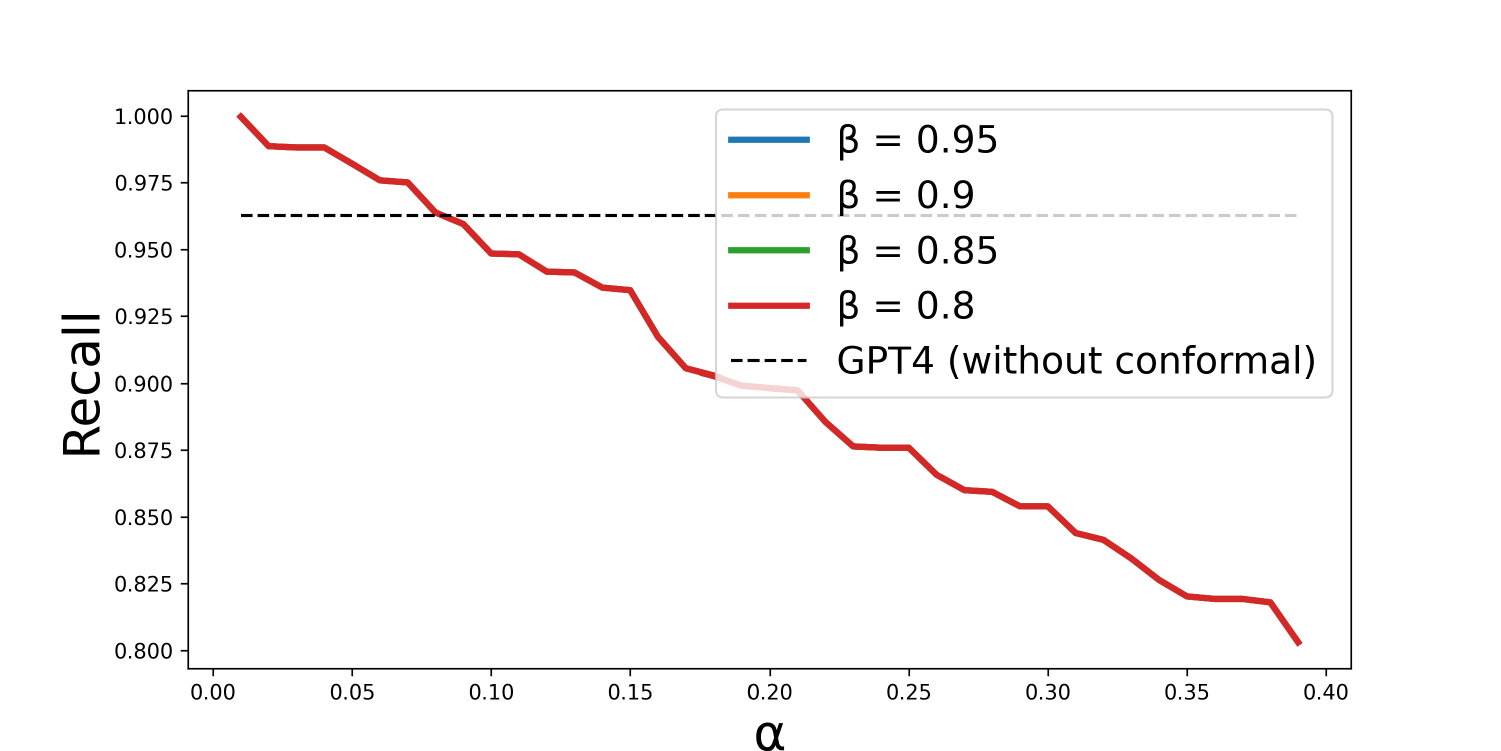}
        \caption{Sentence Centrality}
        \end{subfigure}
        \begin{subfigure}{0.48\textwidth}
        \includegraphics[width=0.95\linewidth, trim={30 0 60 40}, clip]{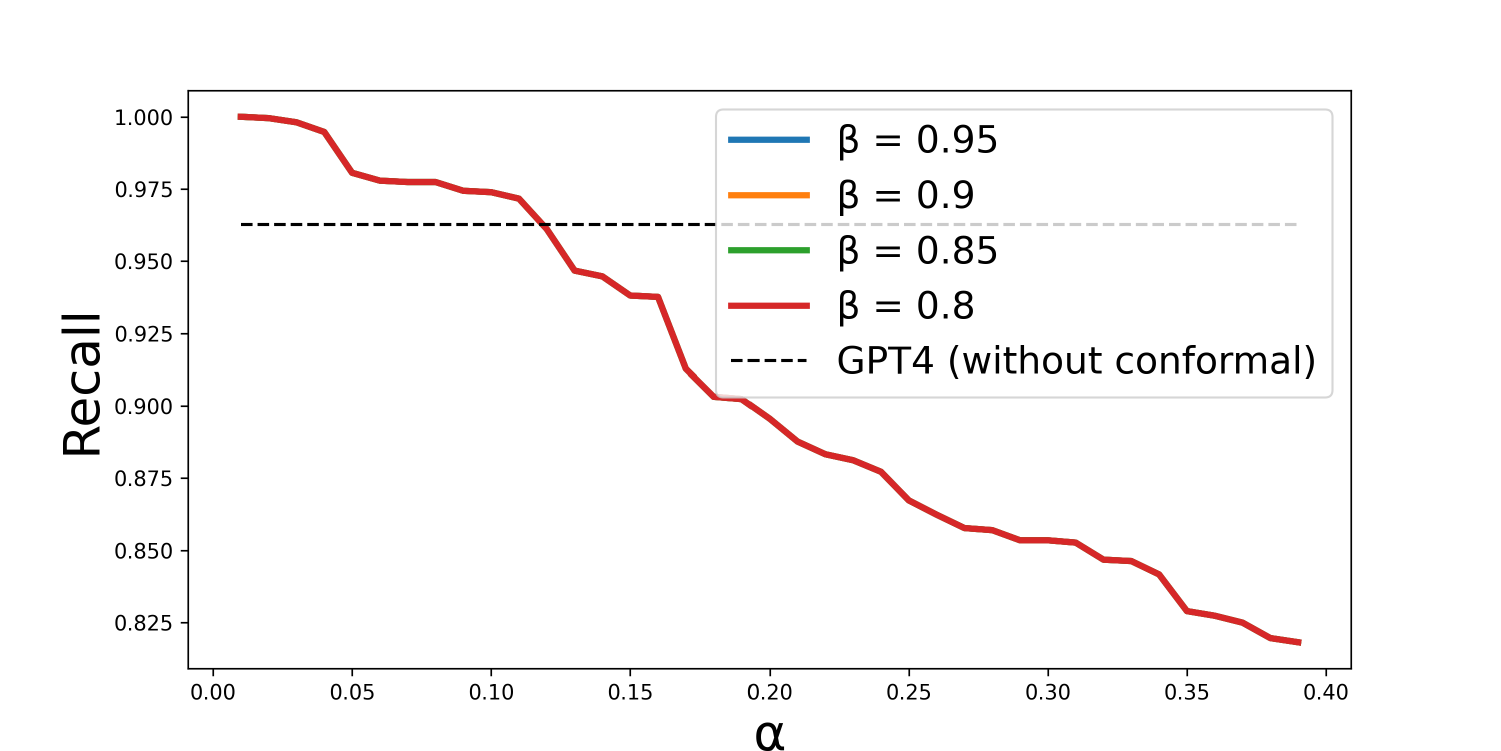}
        \caption{GUSUM}
        \end{subfigure}
        \begin{subfigure}{0.48\textwidth}
        \includegraphics[width=0.95\linewidth, trim={30 0 60 40}, clip]{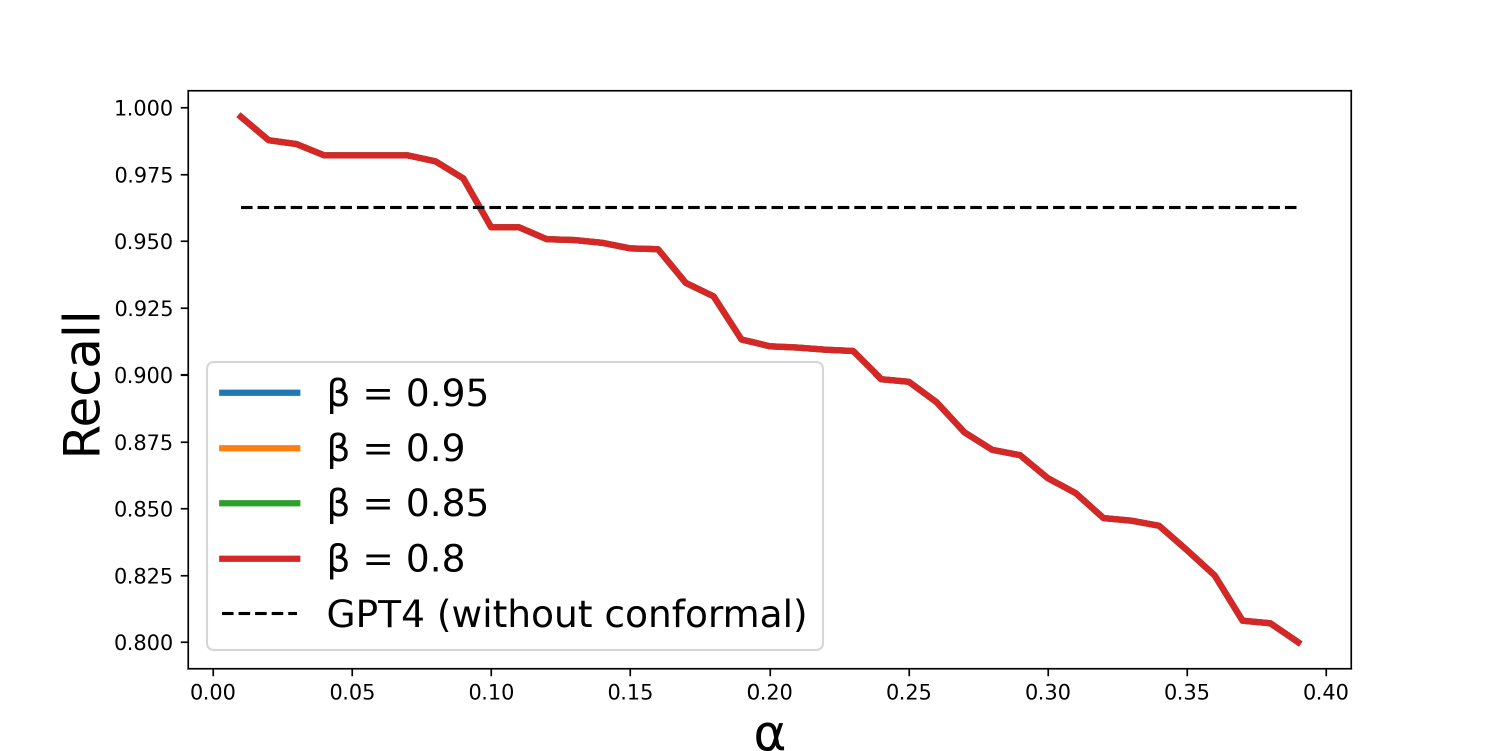}
        \caption{GPT-4o mini}
        \end{subfigure}
        \begin{subfigure}{0.48\textwidth}
        \includegraphics[width=0.95\linewidth, trim={30 0 60 40}, clip]{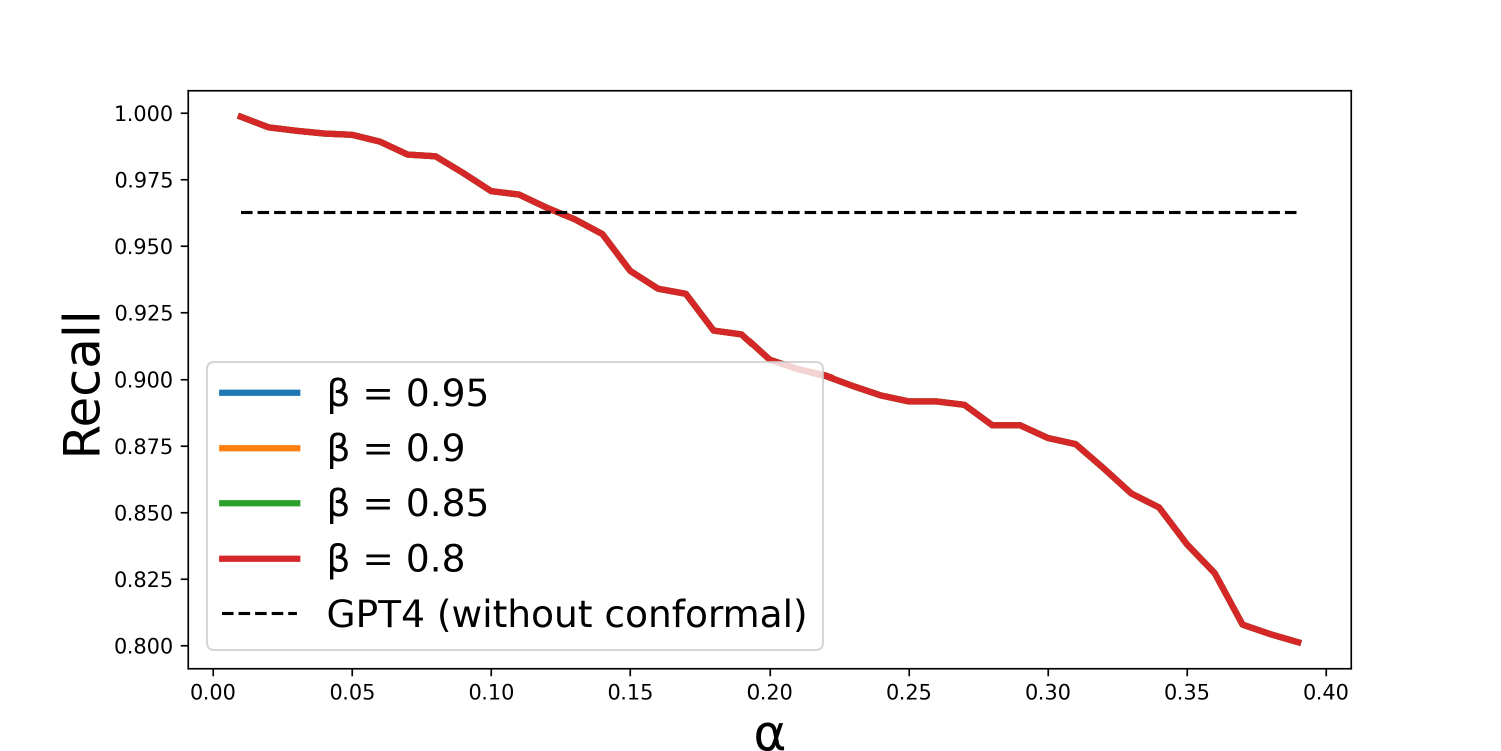}
        \caption{Llama 3}
        \end{subfigure}
        \begin{subfigure}{0.48\textwidth}
        \includegraphics[width=0.95\linewidth, trim={30 0 60 40}, clip]{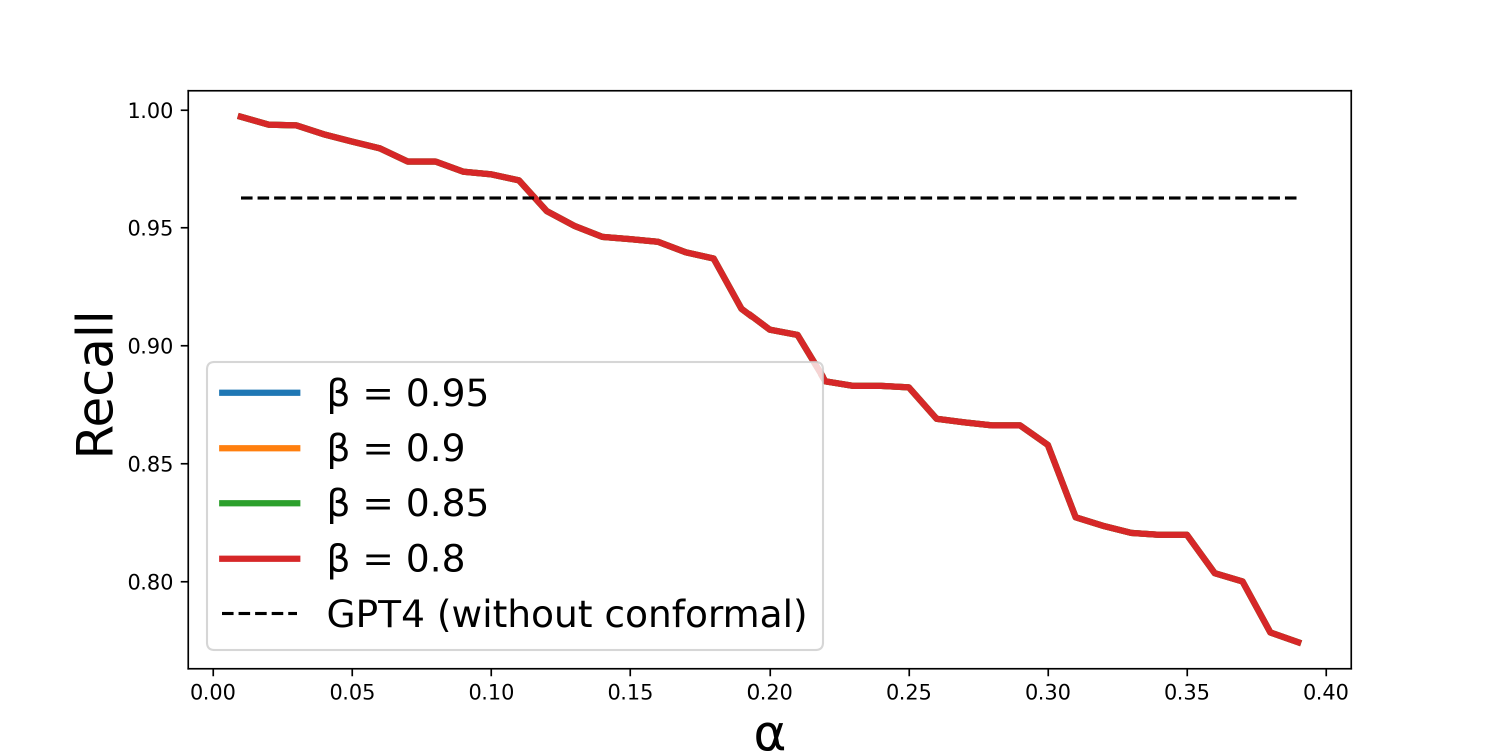}
        \caption{Qwen 3}
        \end{subfigure}
        \begin{subfigure}{0.48\textwidth}
        \includegraphics[width=0.95\linewidth, trim={30 0 60 40}, clip]{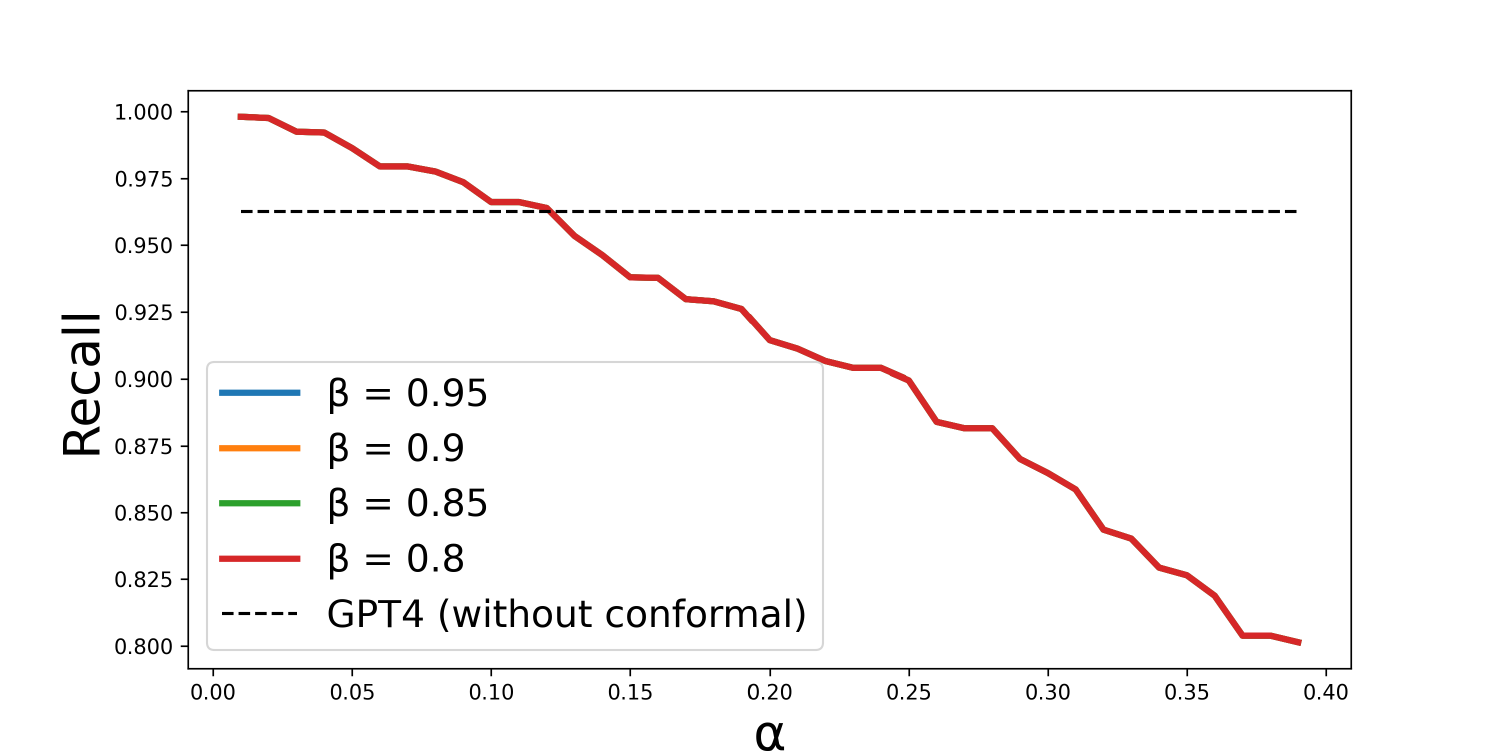}
        \caption{Gemini 2.0 Flash-Lite}
        \end{subfigure}
        \begin{subfigure}{0.48\textwidth}
        \includegraphics[width=0.95\linewidth, trim={30 0 60 40}, clip]{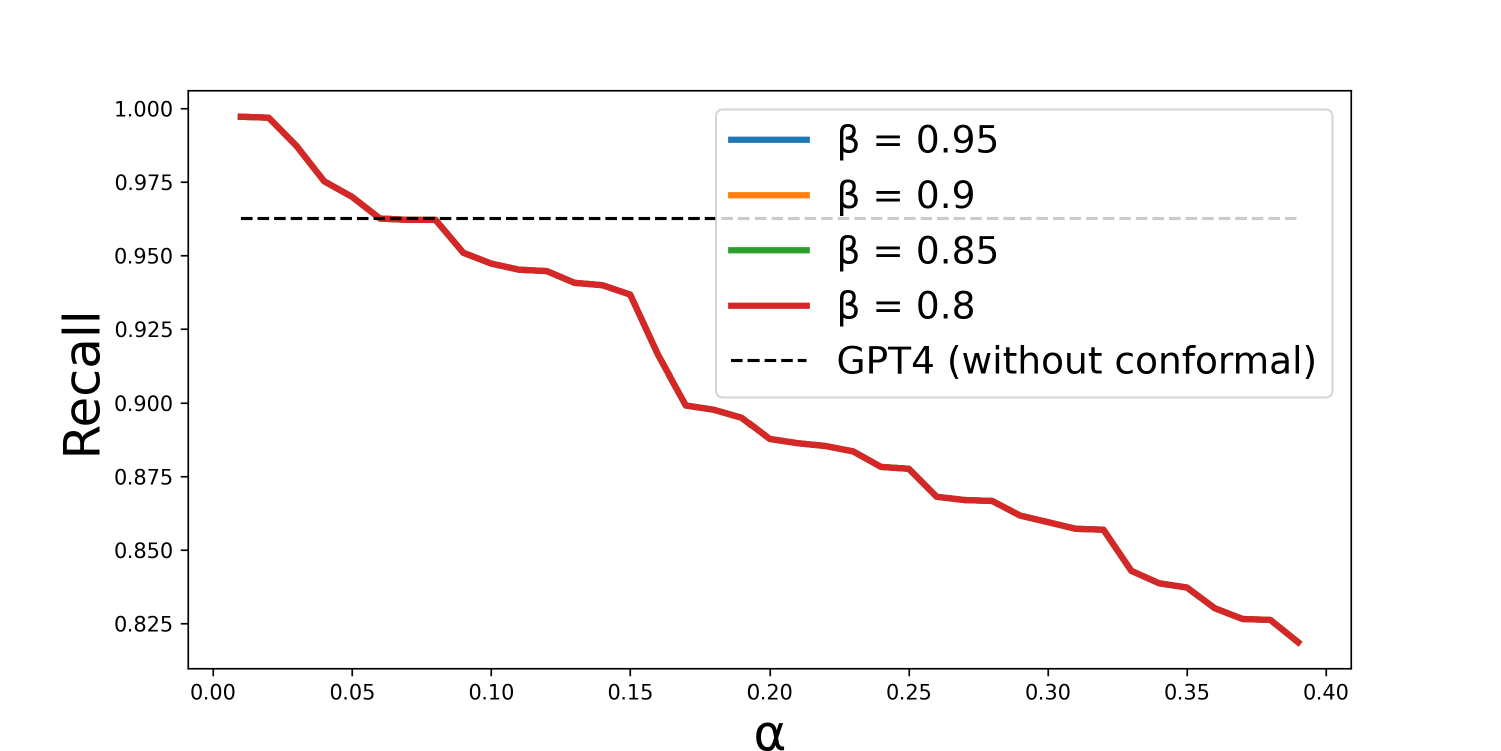}
        \caption{Gemini 2.5 Flash}
        \end{subfigure}
        \caption{Target error rate $\alpha$ versus empirical recall $B(y; y^*)$ of important sentences in summaries, averaged over the TLDR-AIC test set. The dashed line shows GPT-4o mini performance without using conformal prediction. Several curves overlap because all datapoints in TLDR-AIC contain a small number of ground-truth sentences, meaning there are only a few discrete levels of empirical recall possible, making some values of $\beta$ equivalent.}
    \label{fig:recall_versus_alpha_tldr}
\end{figure}

\begin{figure}[t]
    \centering
        \begin{subfigure}{0.48\textwidth}
        \includegraphics[width=0.95\linewidth, trim={30 0 60 40}, clip]{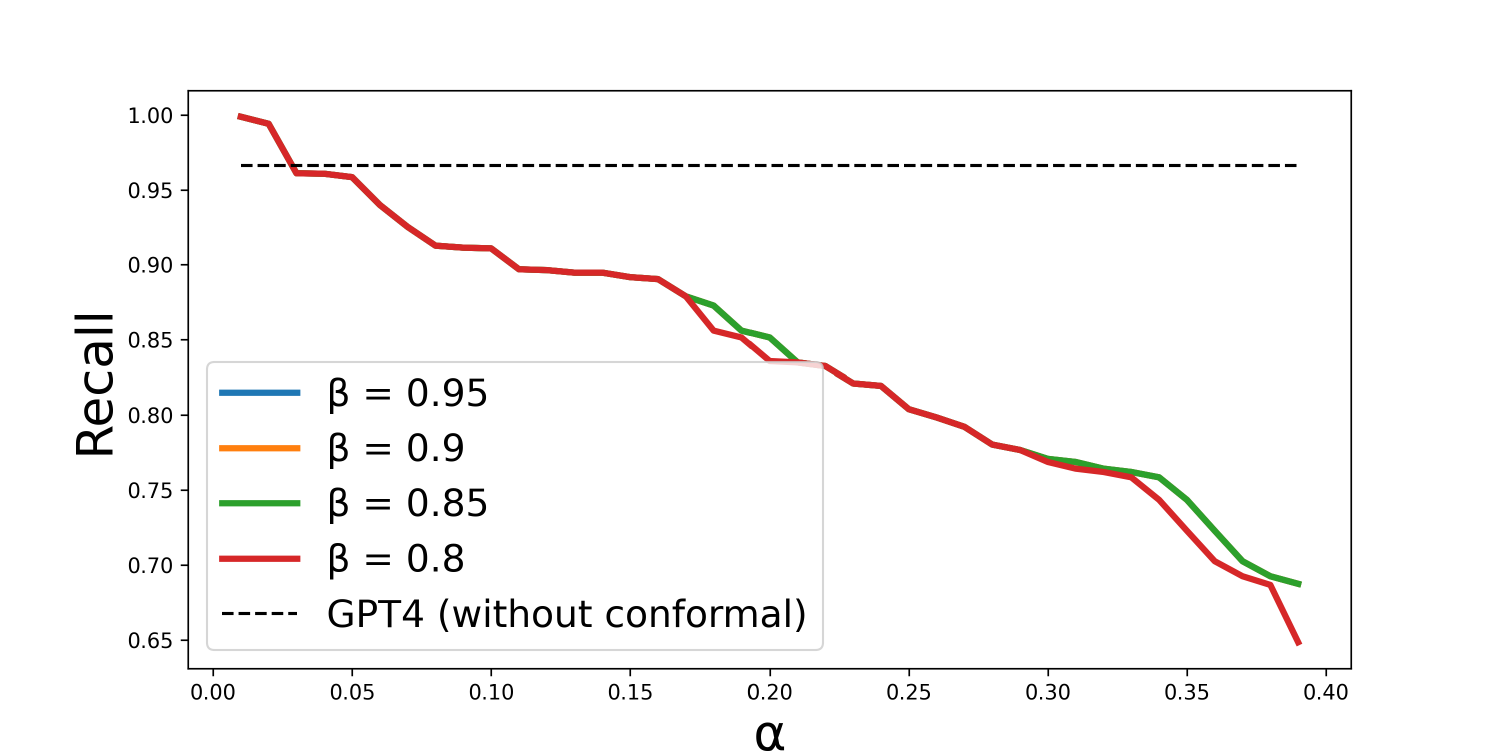}
        \caption{Cosine Similarity Centrality}
        \end{subfigure}
        \begin{subfigure}{0.48\textwidth}
        \includegraphics[width=0.95\linewidth, trim={30 0 60 40}, clip]{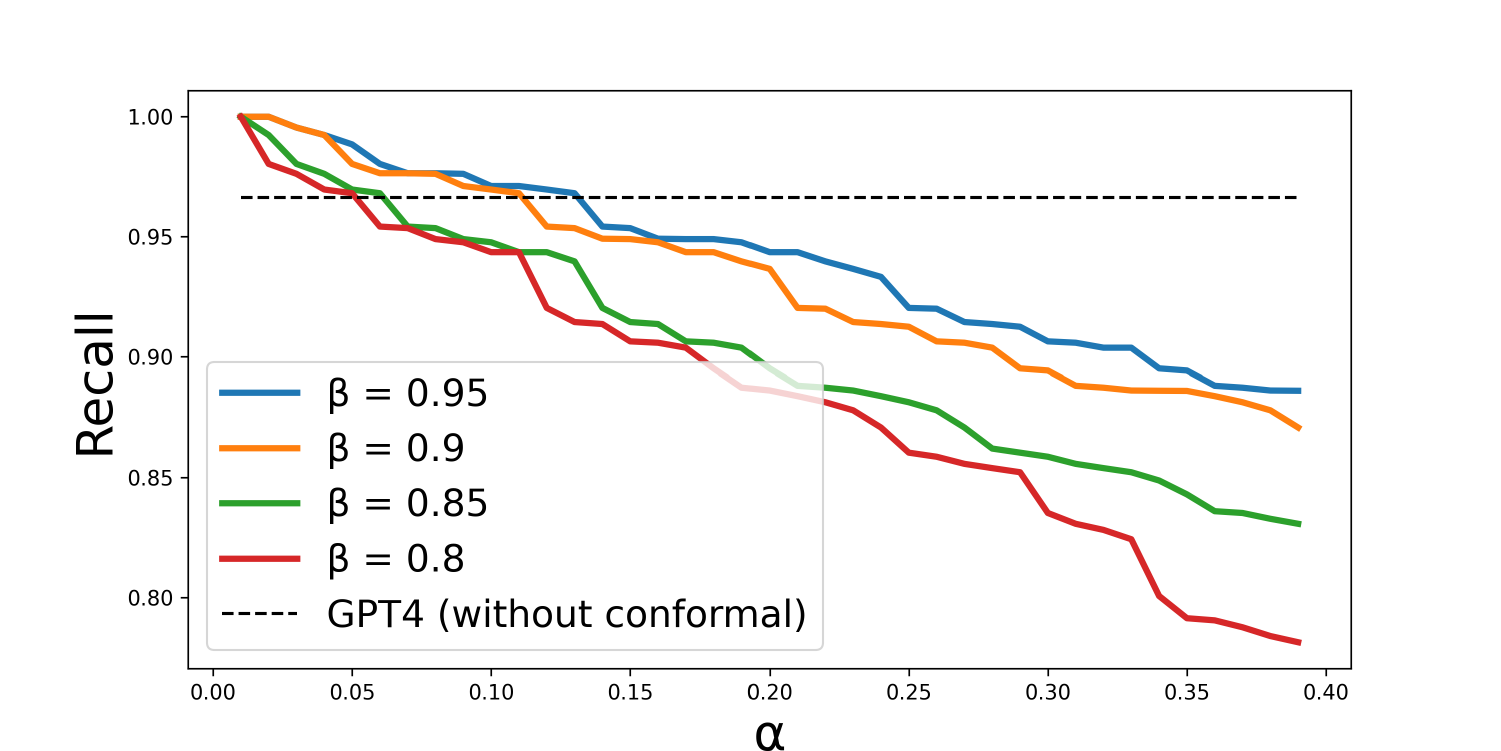}
        \caption{Sentence Centrality}
        \end{subfigure}
        \begin{subfigure}{0.48\textwidth}
        \includegraphics[width=0.95\linewidth, trim={30 0 60 40}, clip]{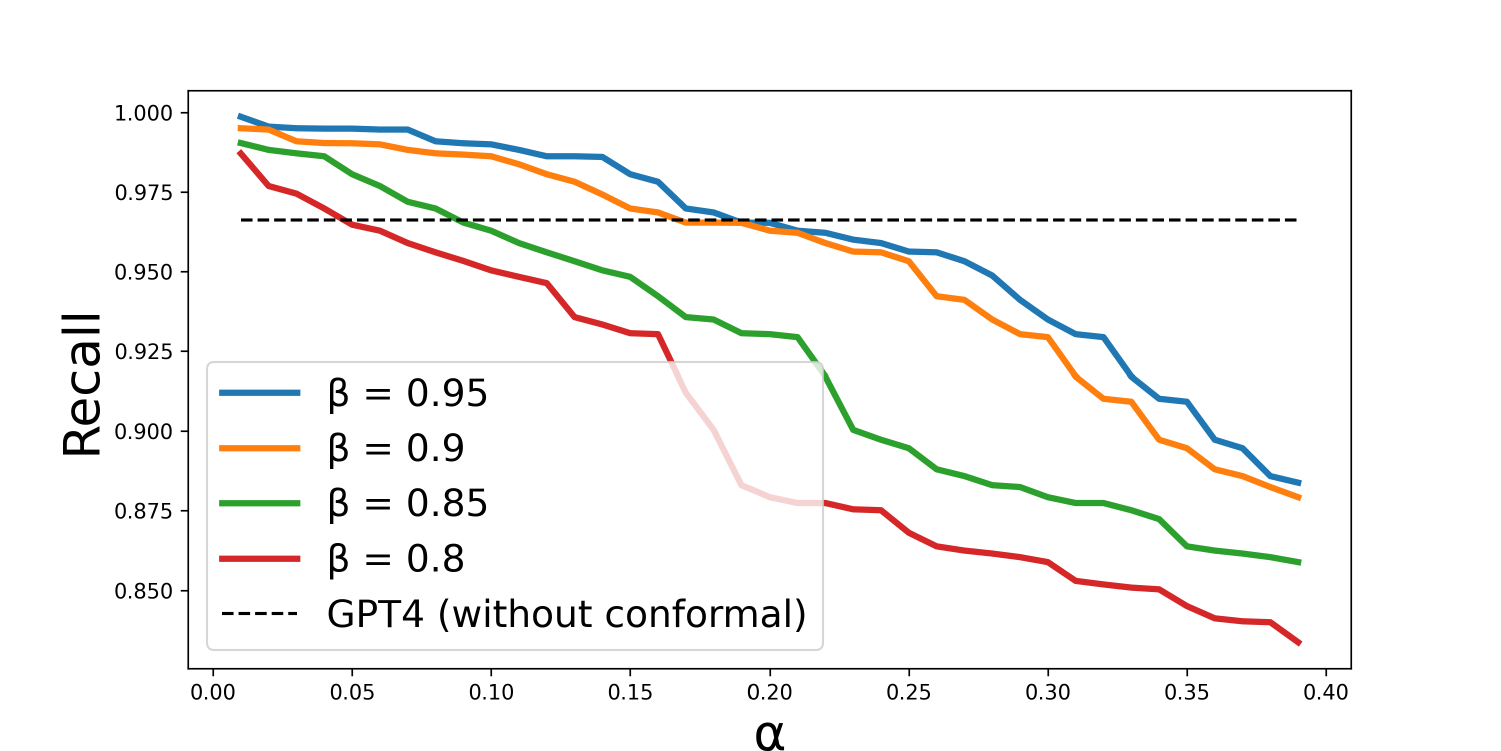}
        \caption{GUSUM}
        \end{subfigure}
        \begin{subfigure}{0.48\textwidth}
        \includegraphics[width=0.95\linewidth, trim={30 0 60 40}, clip]{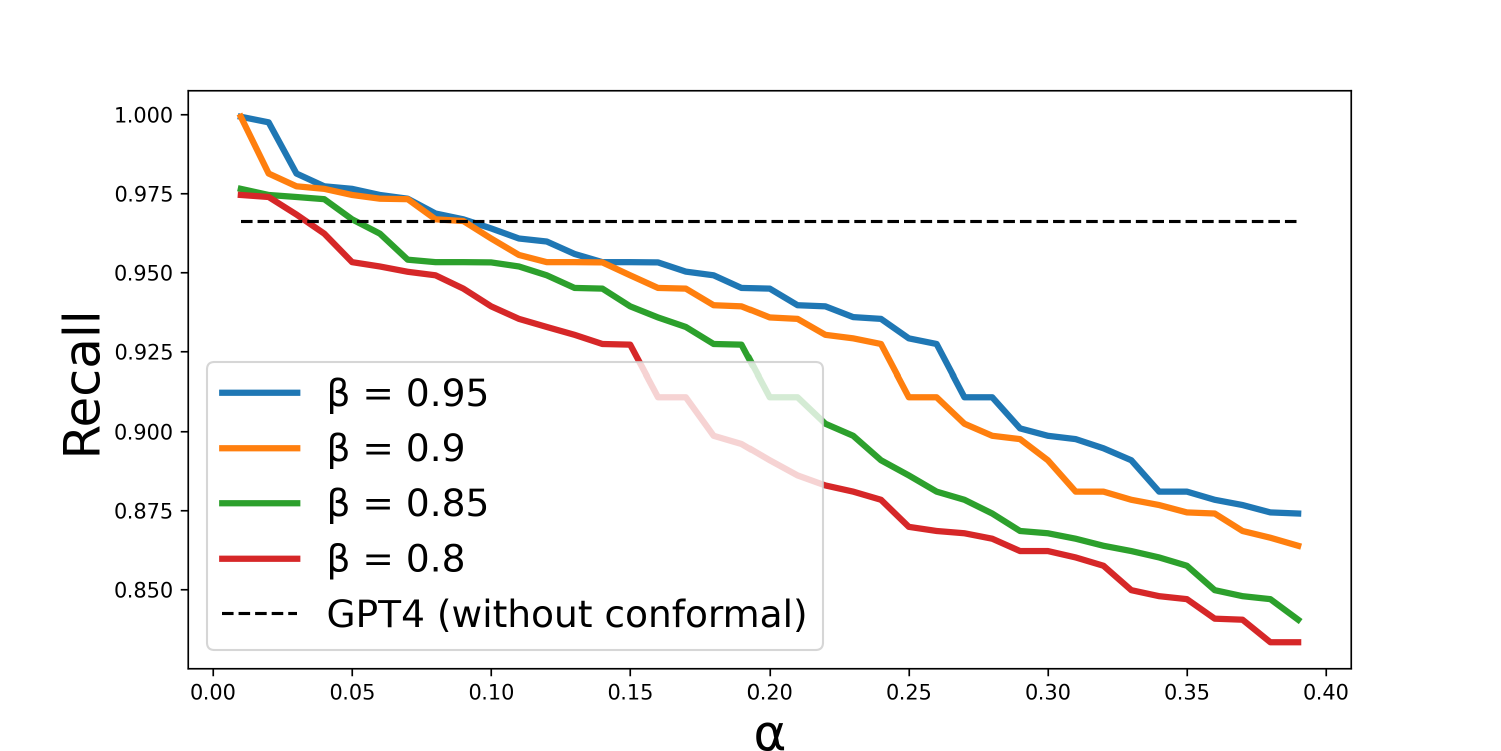}
        \caption{GPT-4o mini}
        \end{subfigure}
        \begin{subfigure}{0.48\textwidth}
        \includegraphics[width=0.95\linewidth, trim={30 0 60 40}, clip]{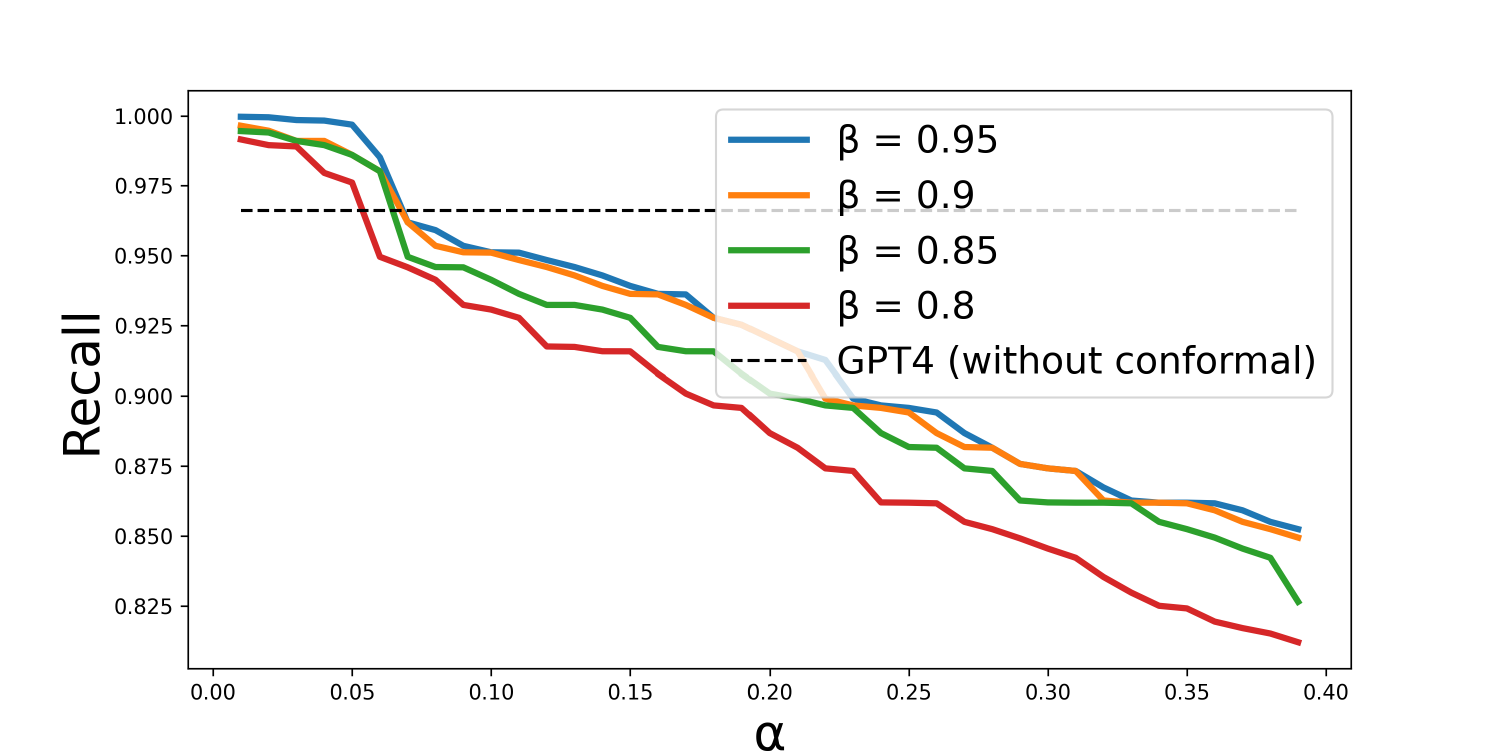}
        \caption{Llama 3}
        \end{subfigure}
        \begin{subfigure}{0.48\textwidth}
        \includegraphics[width=0.95\linewidth, trim={30 0 60 40}, clip]{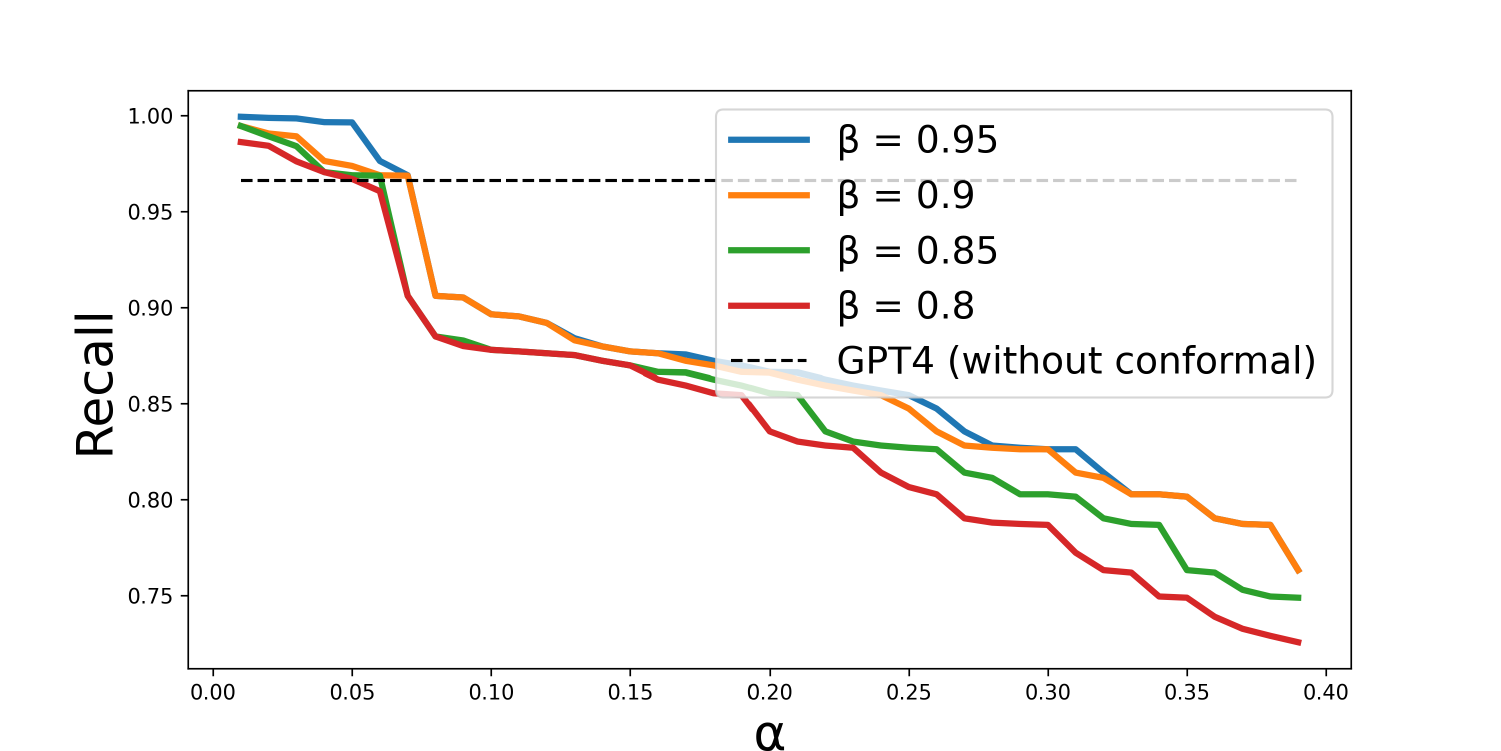}
        \caption{Qwen 3}
        \end{subfigure}
        \begin{subfigure}{0.48\textwidth}
        \includegraphics[width=0.95\linewidth, trim={30 0 60 40}, clip]{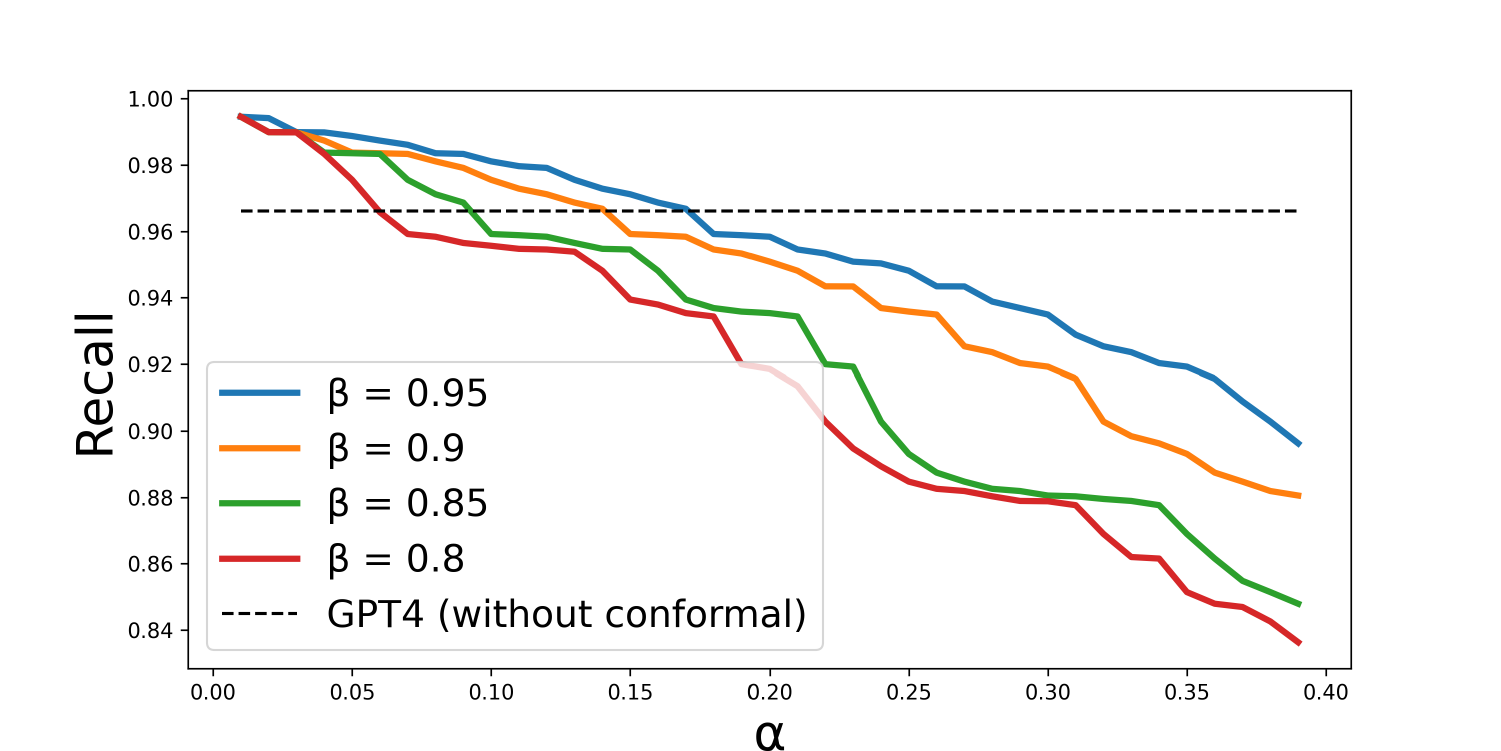}
        \caption{Gemini 2.0 Flash-Lite}
        \end{subfigure}
        \begin{subfigure}{0.48\textwidth}
        \includegraphics[width=0.95\linewidth, trim={30 0 60 40}, clip]{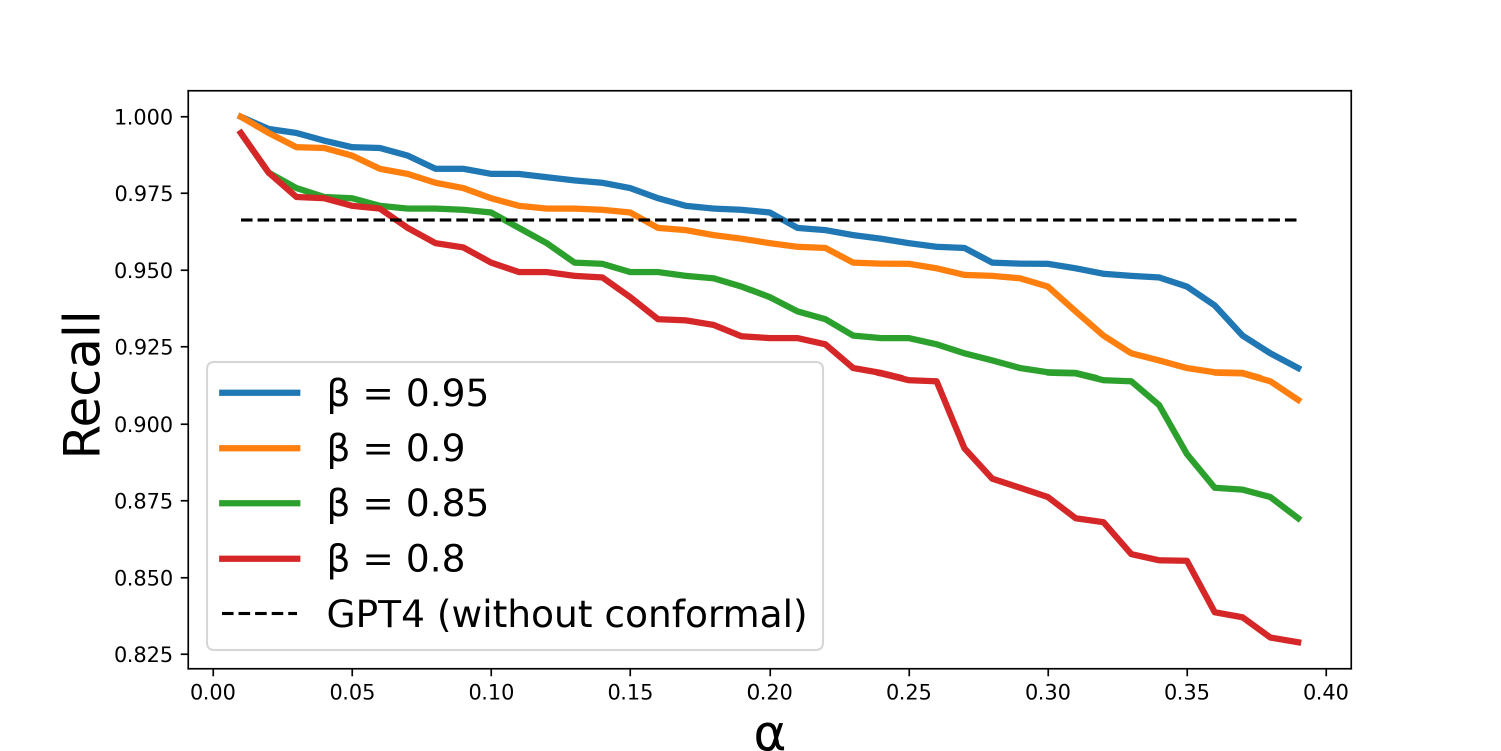}
        \caption{Gemini 2.5 Flash}
        \end{subfigure}
        \caption{Target error rate $\alpha$ versus empirical recall $B(y; y^*)$ of important sentences in summaries, averaged over the MTS test set. The dashed line shows GPT-4o mini performance without using conformal prediction. Several curves may overlap when there are only a few discrete levels of empirical recall possible, making some values of $\beta$ equivalent.}
    \label{fig:recall_versus_alpha_MTS}
\end{figure}

\begin{figure}[t]
    \centering
        \begin{subfigure}{0.48\textwidth}
        \includegraphics[width=0.95\linewidth, trim={30 0 60 40}, clip]{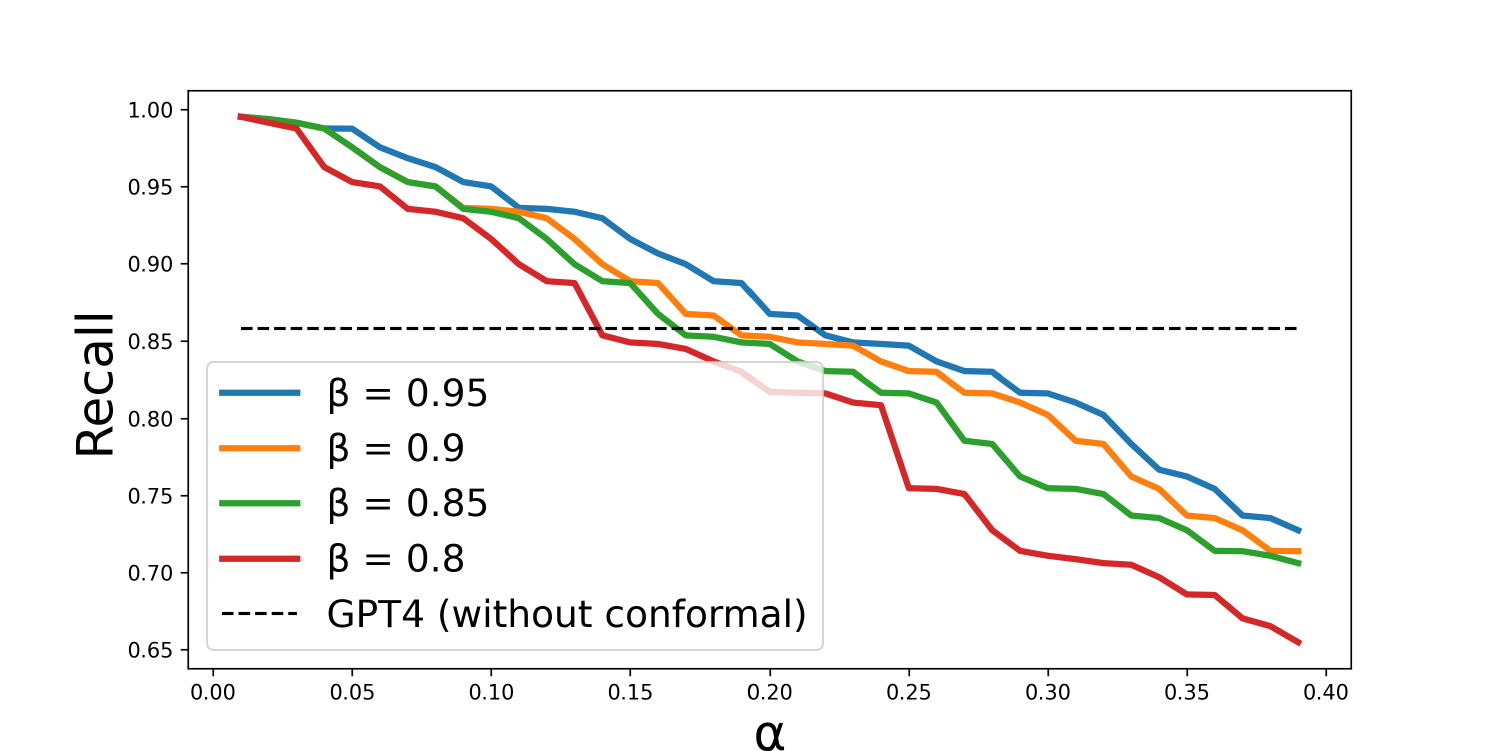}
        \caption{Cosine Similarity Centrality}
        \end{subfigure}
        \begin{subfigure}{0.48\textwidth}
        \includegraphics[width=0.95\linewidth, trim={30 0 60 40}, clip]{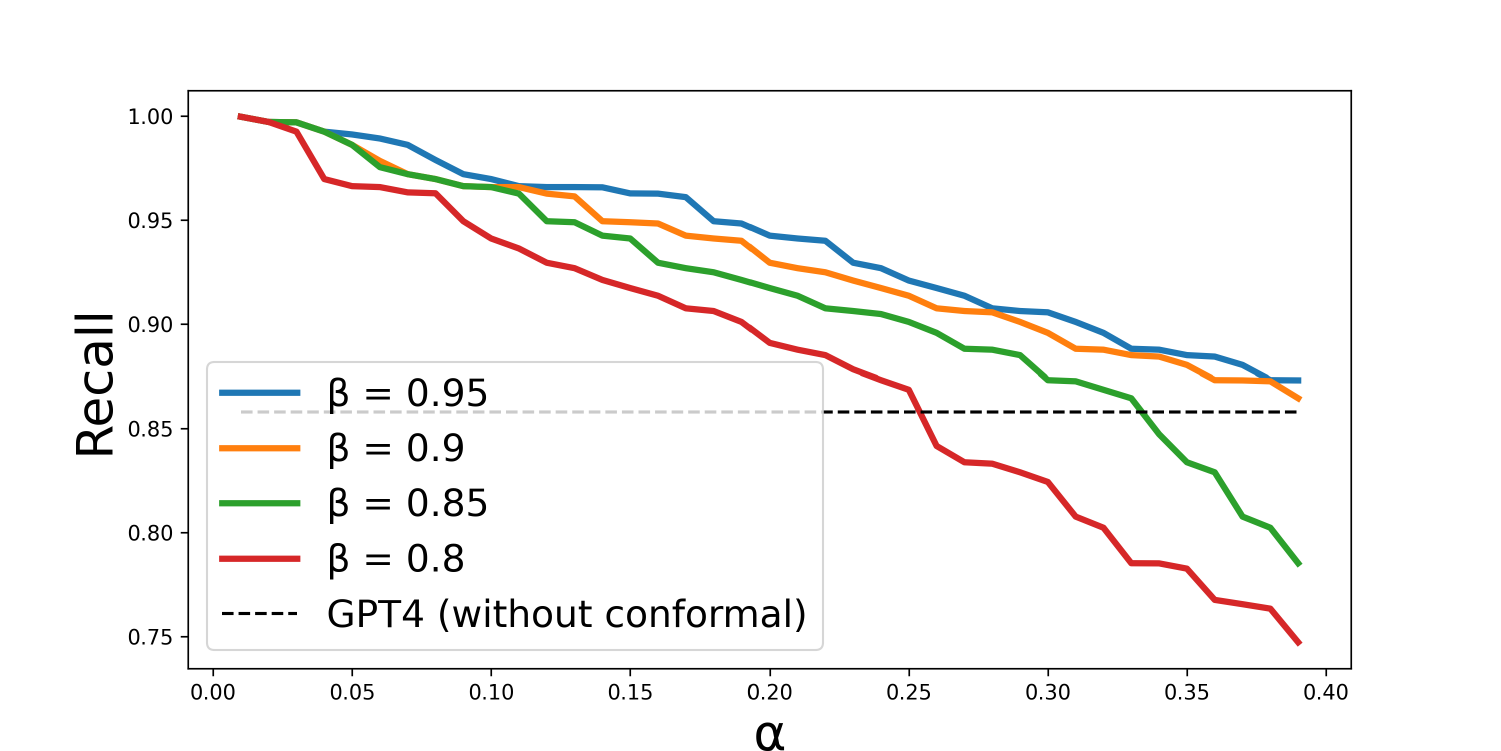}
        \caption{Sentence Centrality}
        \end{subfigure}
        \begin{subfigure}{0.48\textwidth}
        \includegraphics[width=0.95\linewidth, trim={30 0 60 40}, clip]{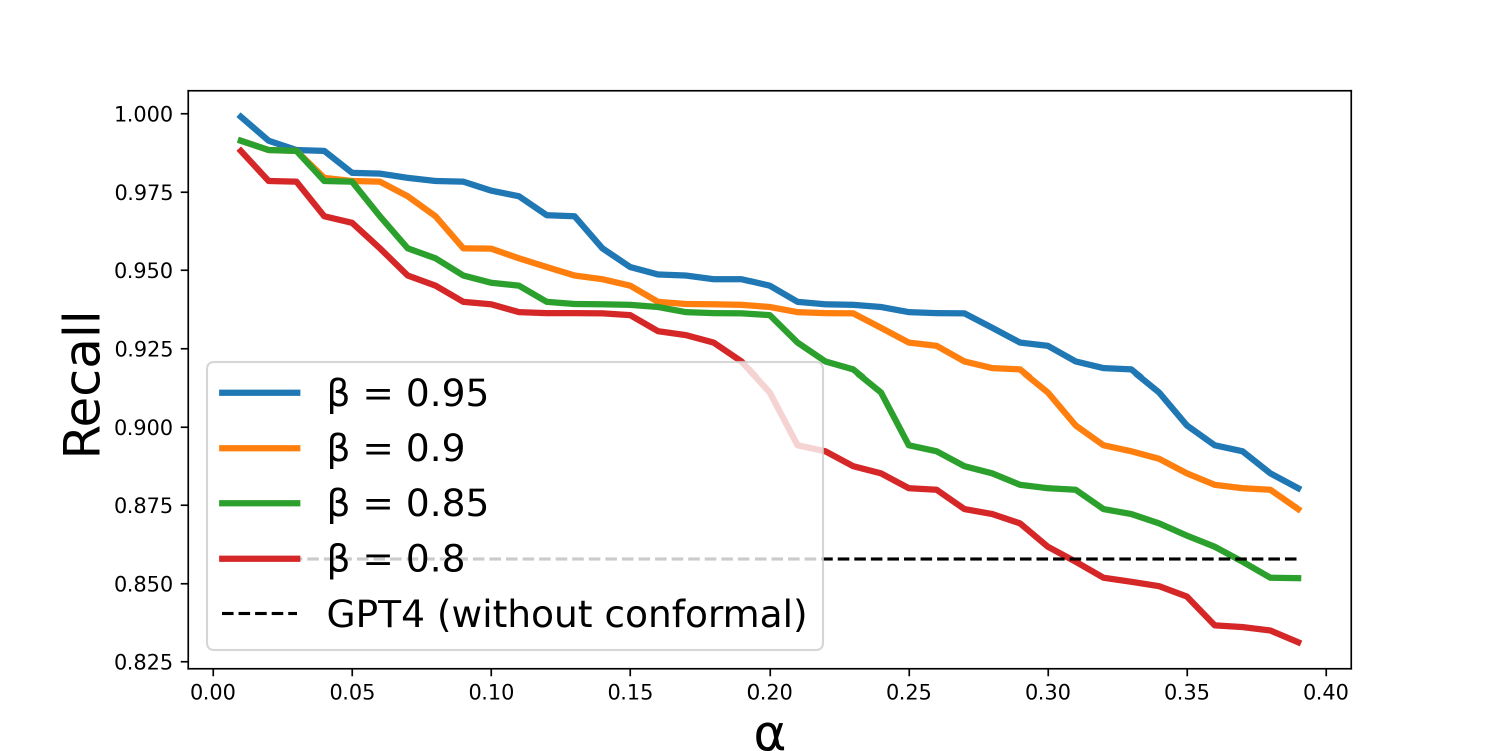}
        \caption{GUSUM}
        \end{subfigure}
        \begin{subfigure}{0.48\textwidth}
        \includegraphics[width=0.95\linewidth, trim={30 0 60 40}, clip]{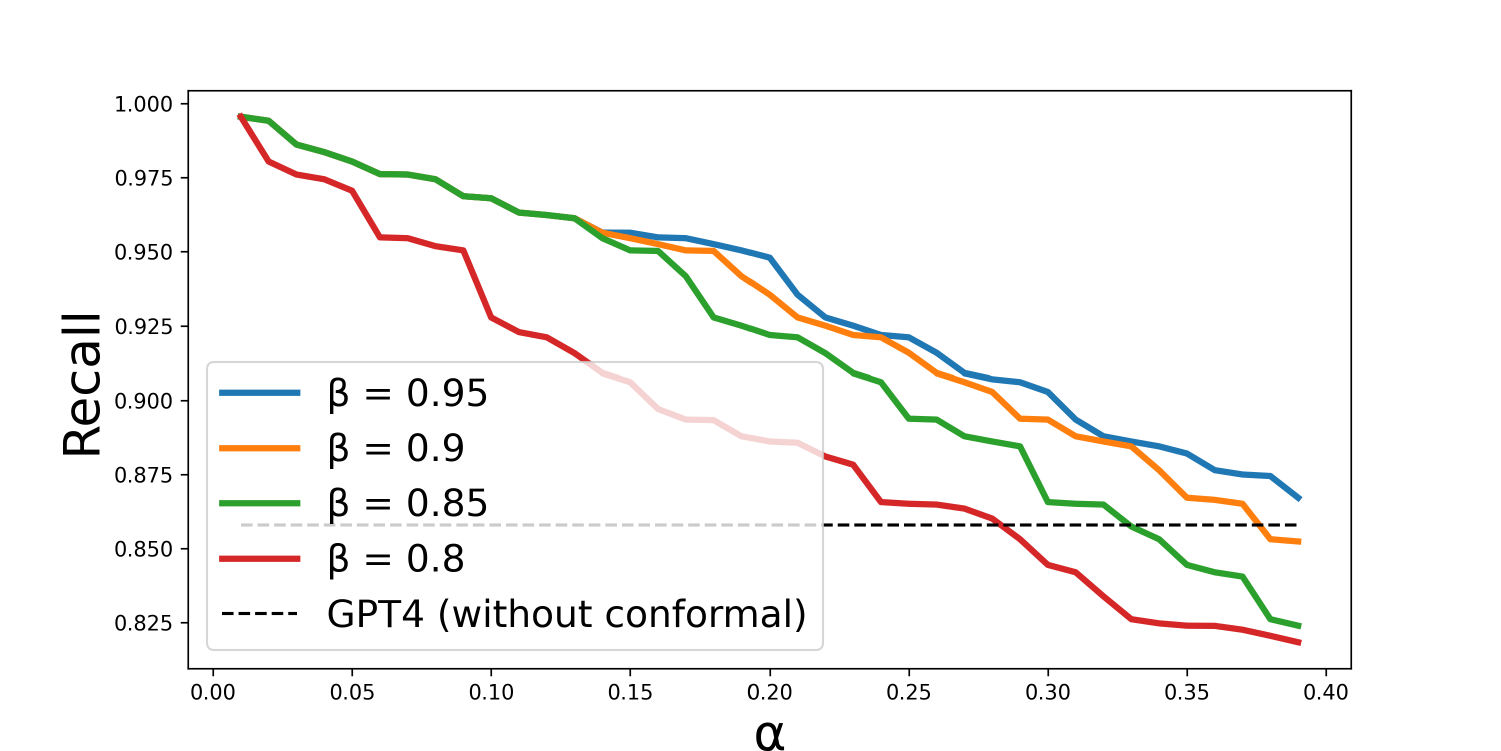}
        \caption{GPT-4o mini}
        \end{subfigure}
        \begin{subfigure}{0.48\textwidth}
        \includegraphics[width=0.95\linewidth, trim={30 0 60 40}, clip]{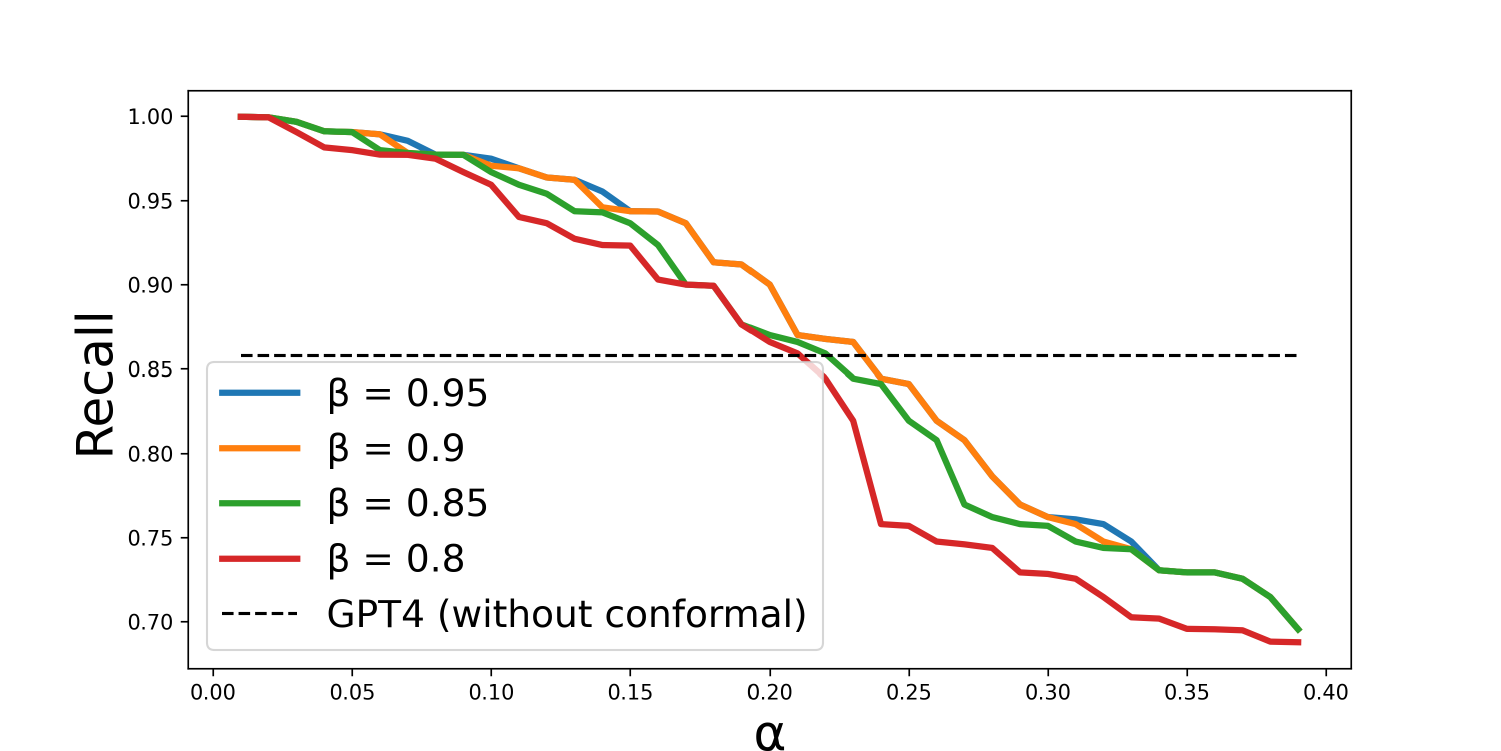}
        \caption{Llama 3}
        \end{subfigure}
        \begin{subfigure}{0.48\textwidth}
        \includegraphics[width=0.95\linewidth, trim={30 0 60 40}, clip]{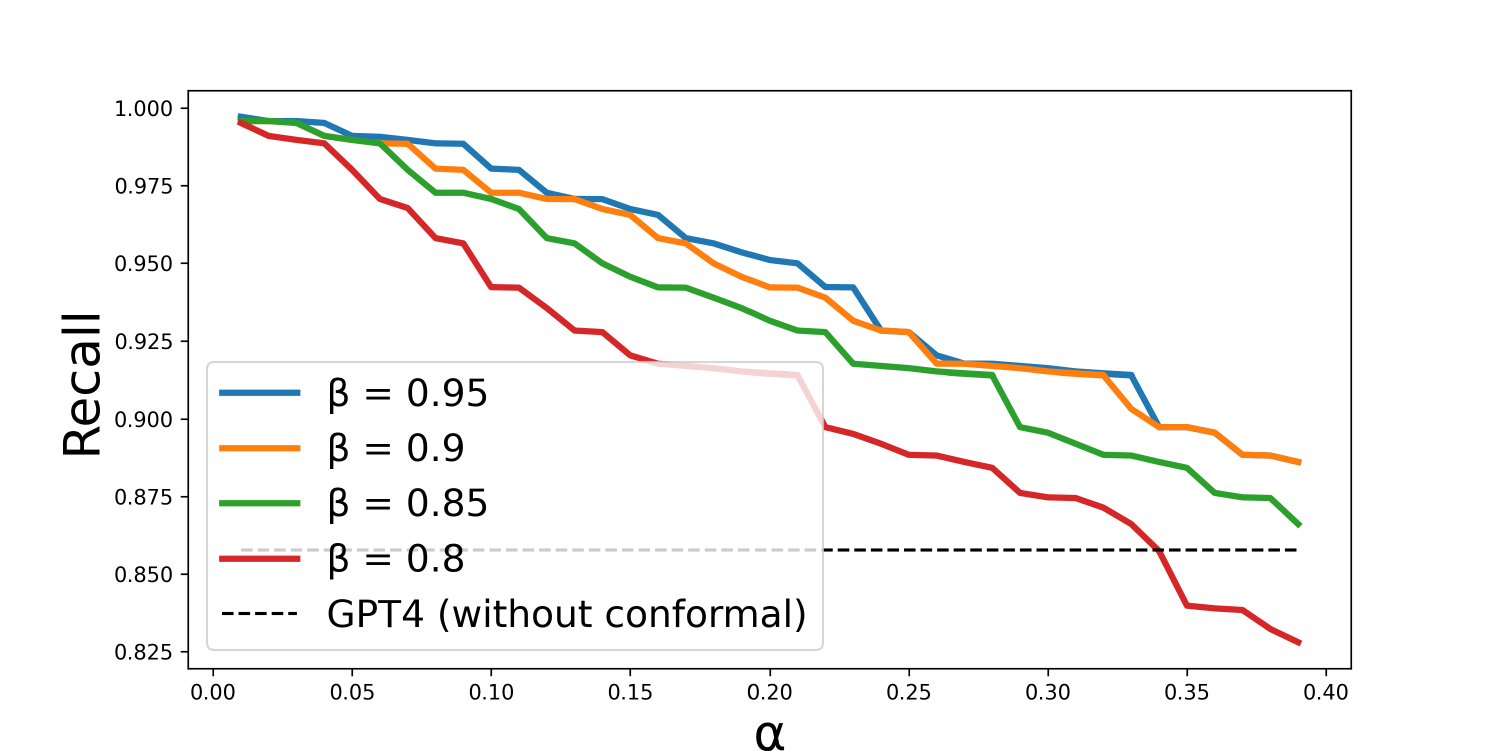}
        \caption{Qwen 3}
        \end{subfigure}
        \begin{subfigure}{0.48\textwidth}
        \includegraphics[width=0.95\linewidth, trim={30 0 60 40}, clip]{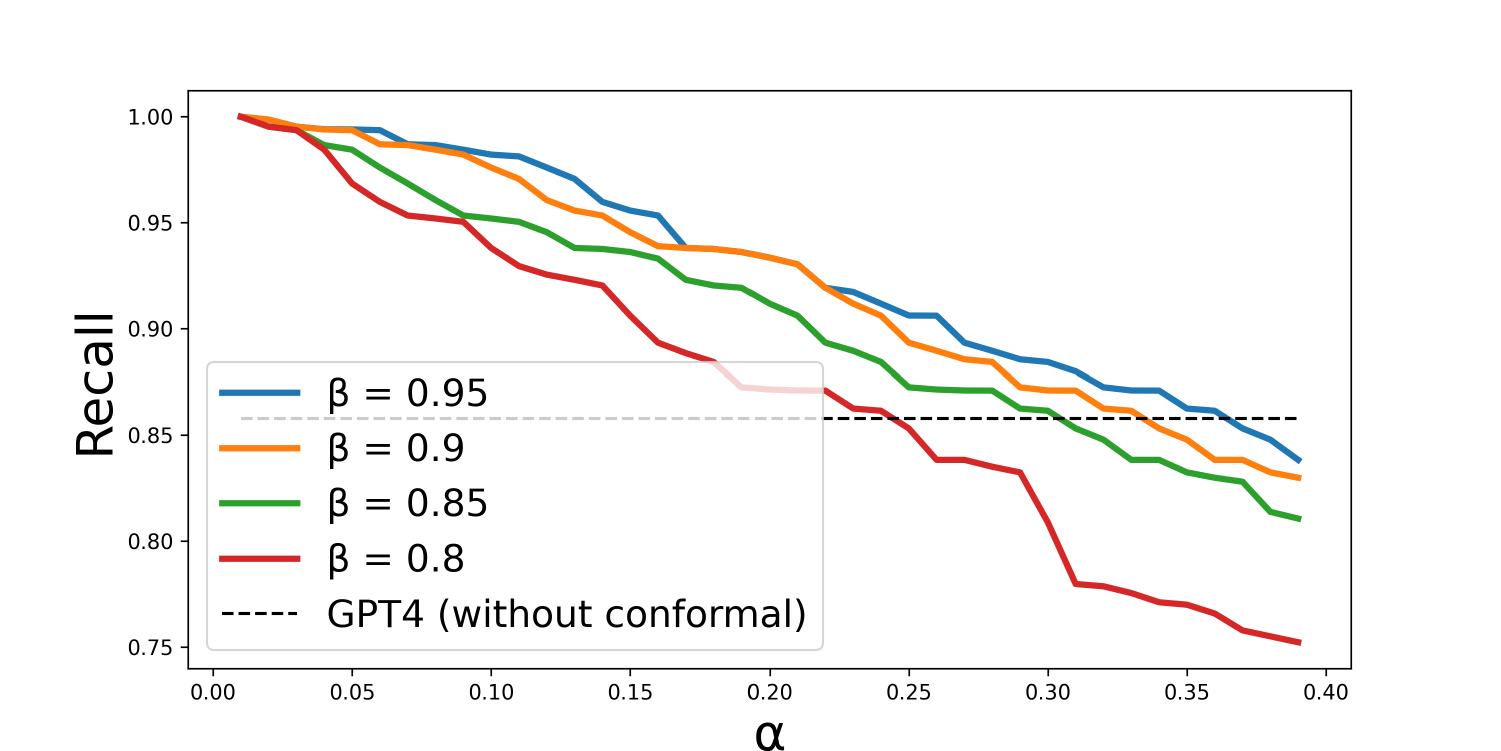}
        \caption{Gemini 2.0 Flash-Lite}
        \end{subfigure}
        \begin{subfigure}{0.48\textwidth}
        \includegraphics[width=0.95\linewidth, trim={30 0 60 40}, clip]{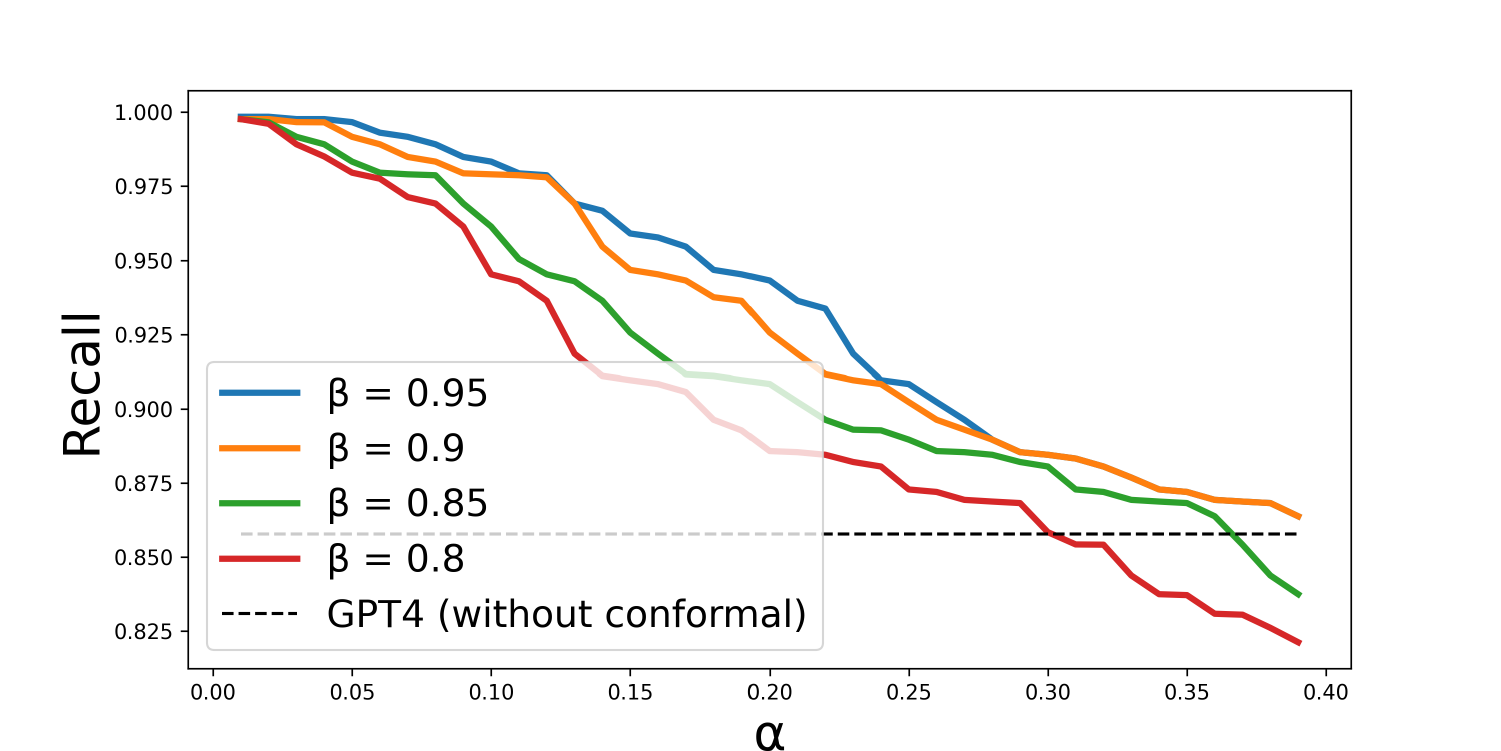}
        \caption{Gemini 2.5 Flash}
        \end{subfigure}
        \caption{Target error rate $\alpha$ versus empirical recall $B(y; y^*)$ of important sentences in summaries, averaged over the ECT test set. The dashed line shows GPT-4o mini performance without using conformal prediction.}
    \label{fig:recall_versus_alpha_ECT}
\end{figure}

Figures \ref{fig:recall_versus_alpha_CNNDM} - \ref{fig:recall_versus_alpha_ECT} show the empirical recall $B(y; y^*)$ based on the choice of error rate $\alpha$ for all datasets and methods\footnote{Due to computational constraints, we only compute this plot for LexRank on the CNN/DM dataset}, analogous to \Cref{fig:recall_vs_alpha} in \Cref{sec:empirical}. The trend is similar, with higher $\alpha$ leading to lower recall of important content. We notice that for TLDR-AIC and CNN/DM, many of the lines for different $\beta$ values overlap one another. This may be due to the low number of sentences in the long-texts for these datasets, making some $\beta$ values effectively equivalent for those $x$. 

\clearpage

\subsection{Target Recall vs. Conciseness Plots for all Datasets and Methods}\label{app:recall_conciseness}
\begin{figure}[t]
    \centering
        \begin{subfigure}{0.48\textwidth}
        \includegraphics[width=0.95\linewidth, trim={30 0 60 40}, clip]{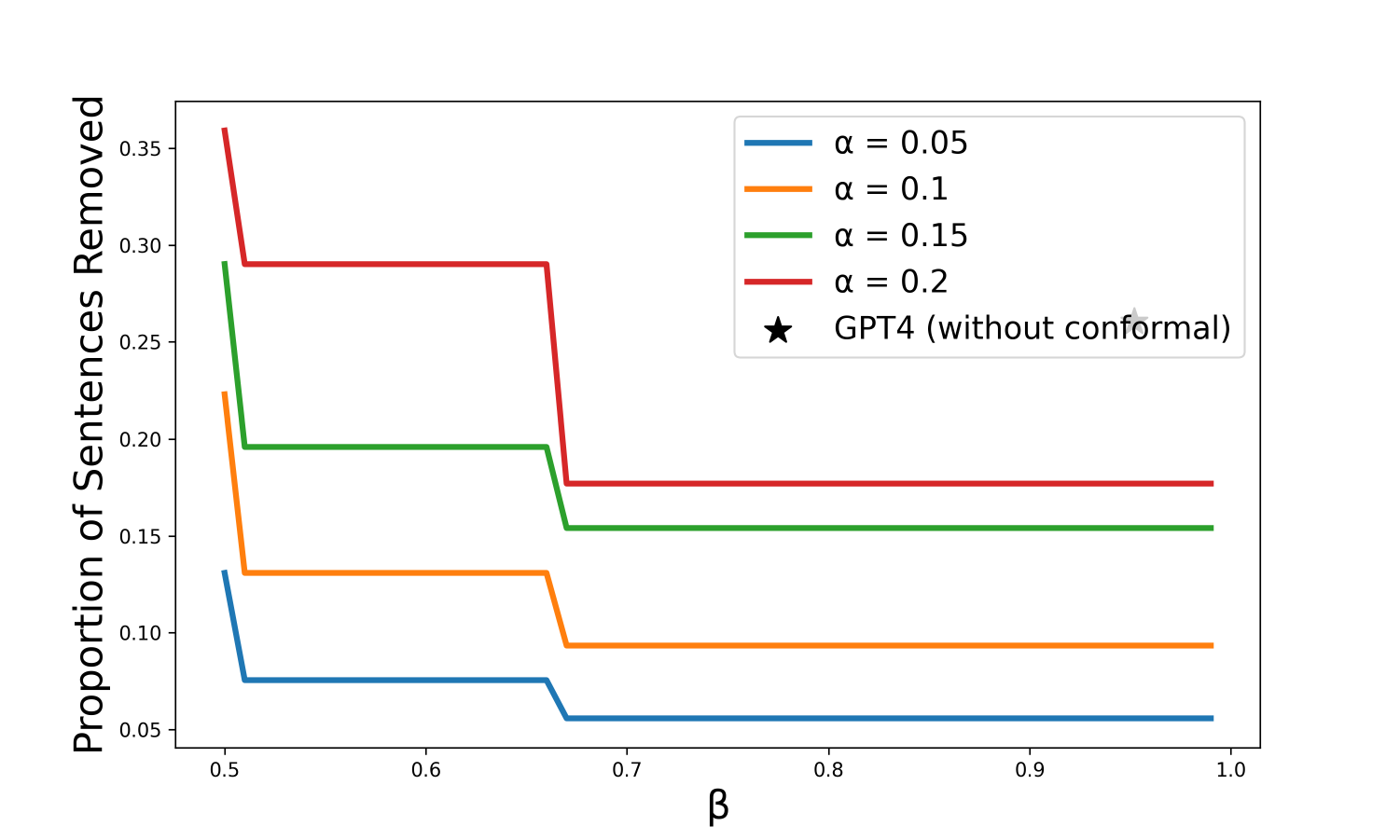}
        \caption{Cosine Similarity Centrality}
        \end{subfigure}
                \begin{subfigure}{0.48\textwidth}
        \includegraphics[width=0.95\linewidth, trim={30 0 60 40}, clip]{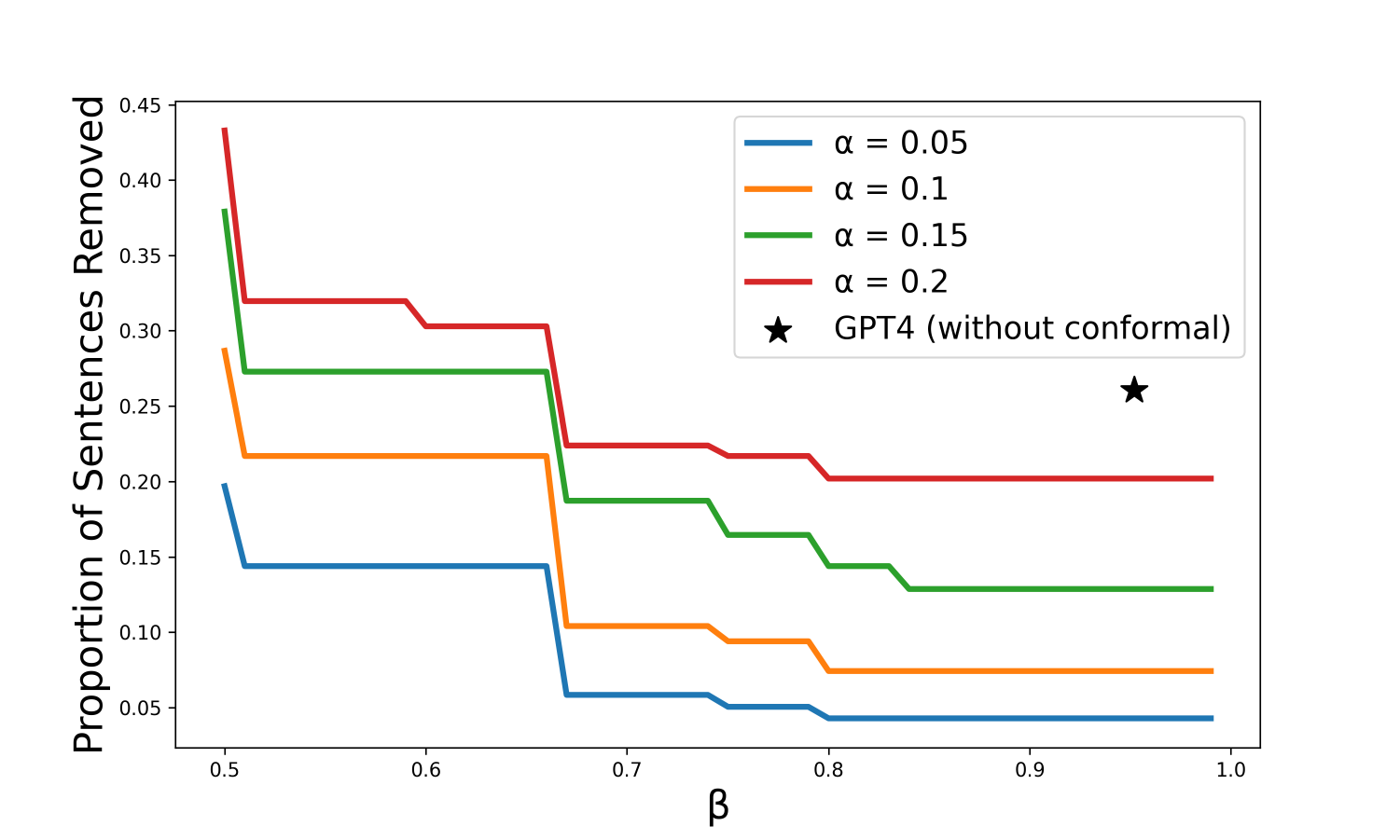}
        \caption{Sentence Centrality}
        \end{subfigure}
        \begin{subfigure}{0.48\textwidth}
        \includegraphics[width=0.95\linewidth, trim={30 0 60 40}, clip]{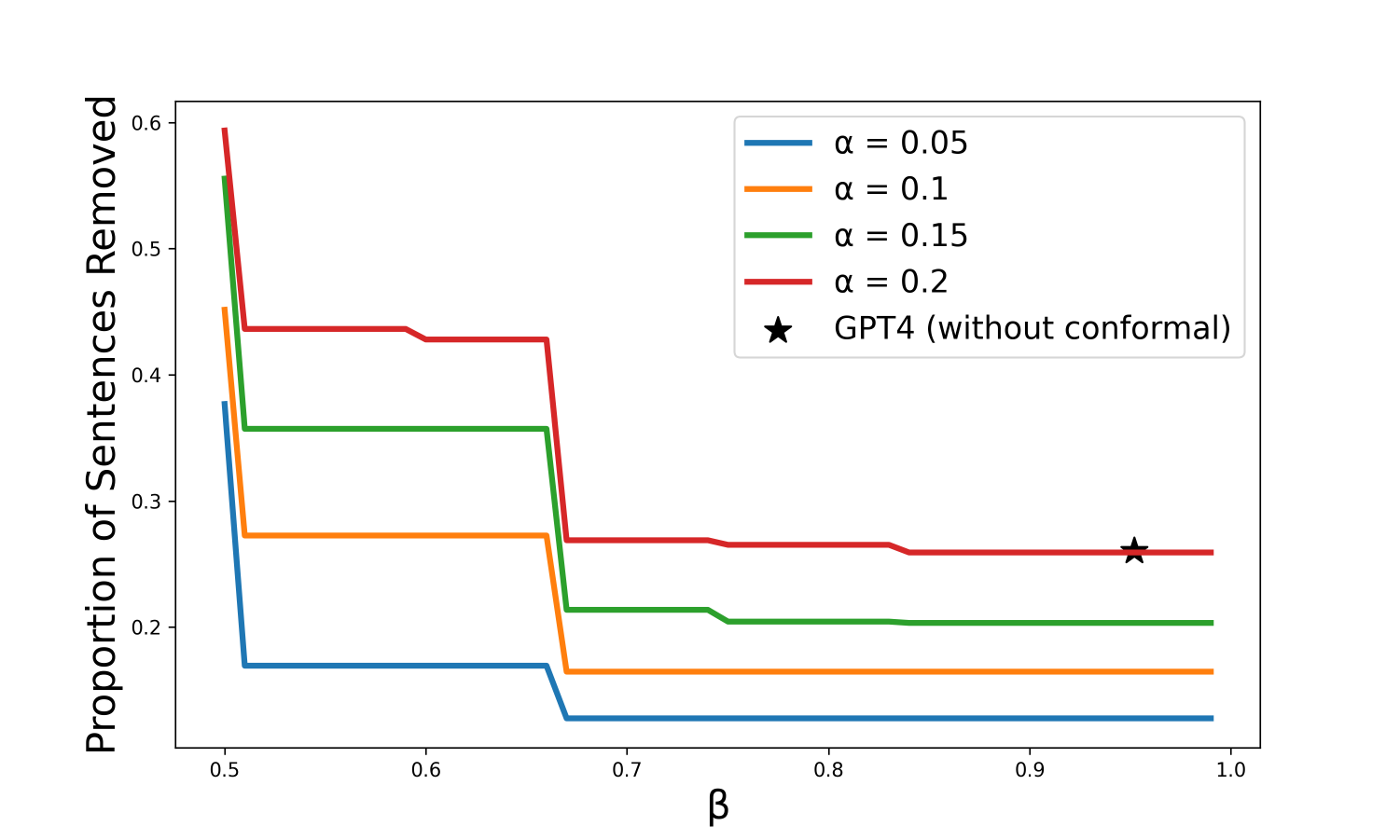}
        \caption{GUSUM}
        \end{subfigure}
        \begin{subfigure}{0.48\textwidth}
        \includegraphics[width=0.95\linewidth, trim={30 0 60 40}, clip]{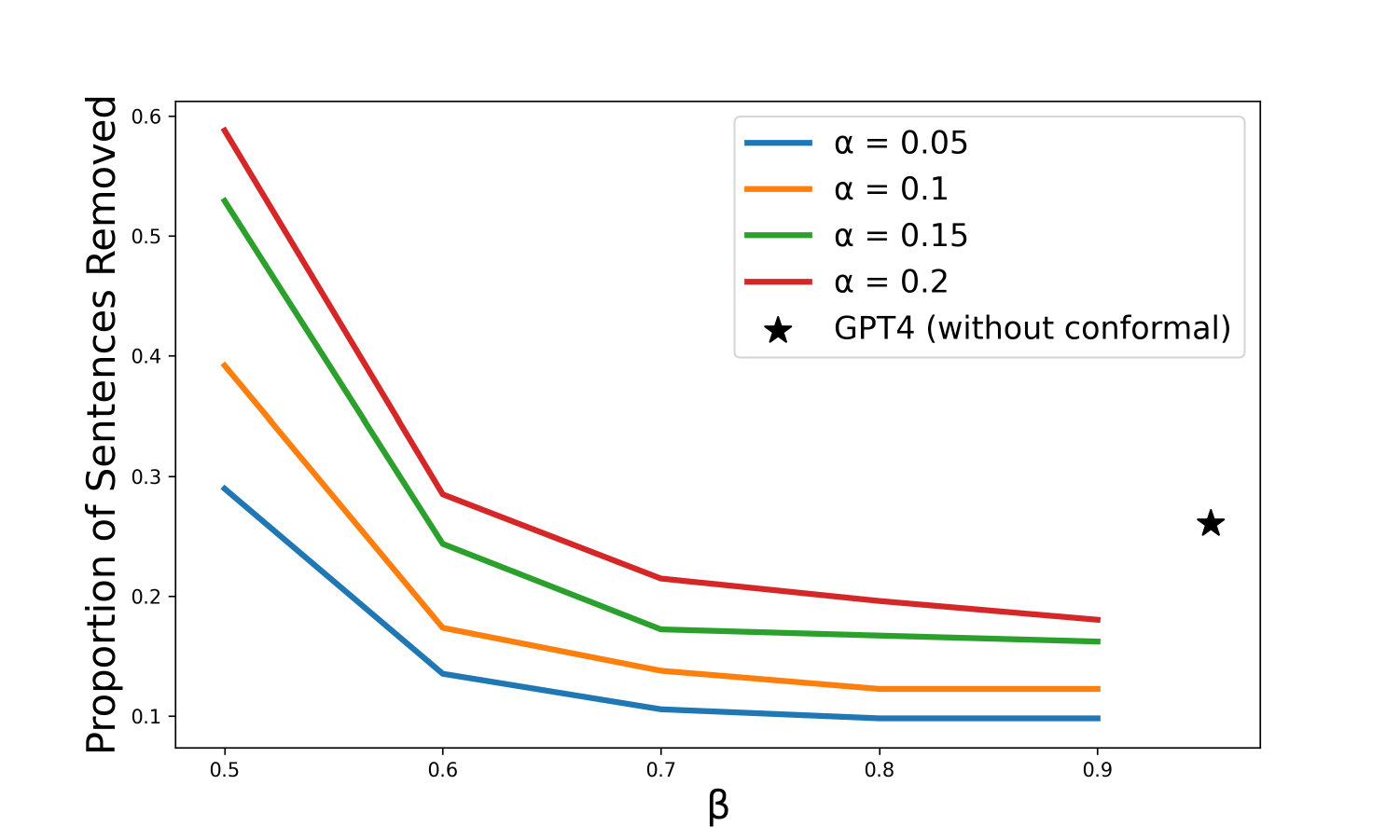}
        \caption{LexRank}
        \end{subfigure}
        \begin{subfigure}{0.48\textwidth}
        \includegraphics[width=0.95\linewidth, trim={30 0 60 40}, clip]{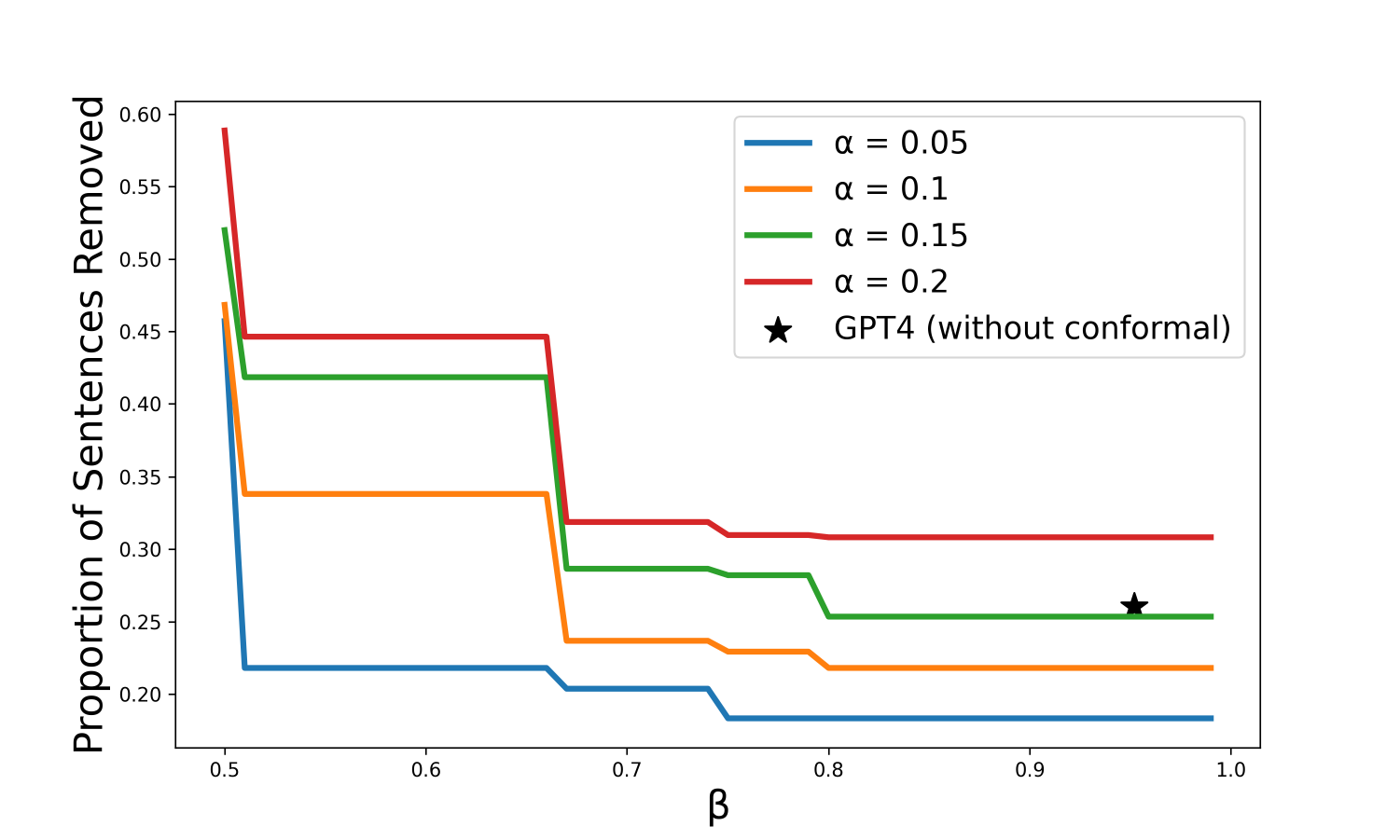}
        \caption{GPT-4o mini}
        \end{subfigure}
        \begin{subfigure}{0.48\textwidth}
        \includegraphics[width=0.95\linewidth, trim={30 0 60 40}, clip]{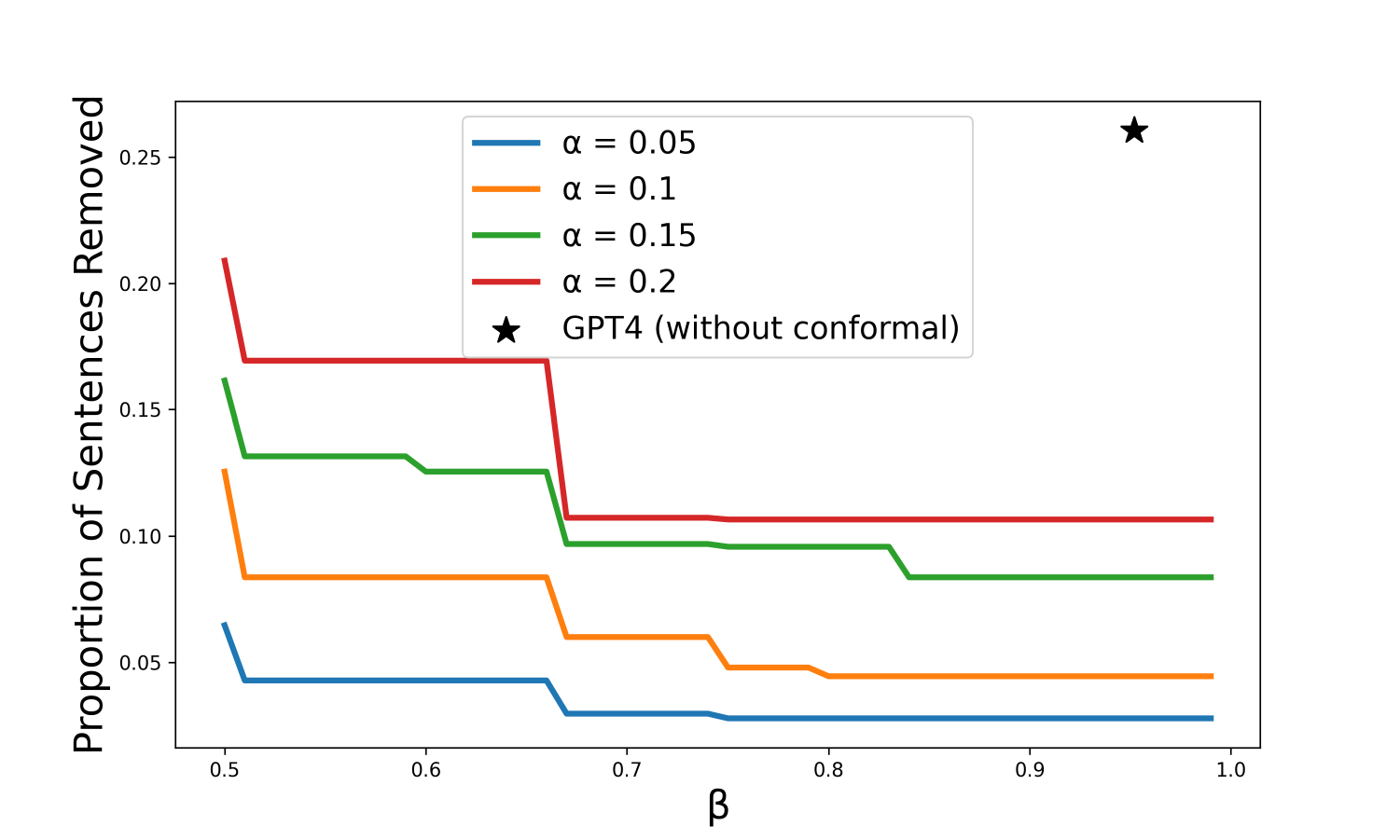}
        \caption{Llama 3}
        \end{subfigure}
        \begin{subfigure}{0.48\textwidth}
        \includegraphics[width=0.95\linewidth, trim={30 0 60 40}, clip]{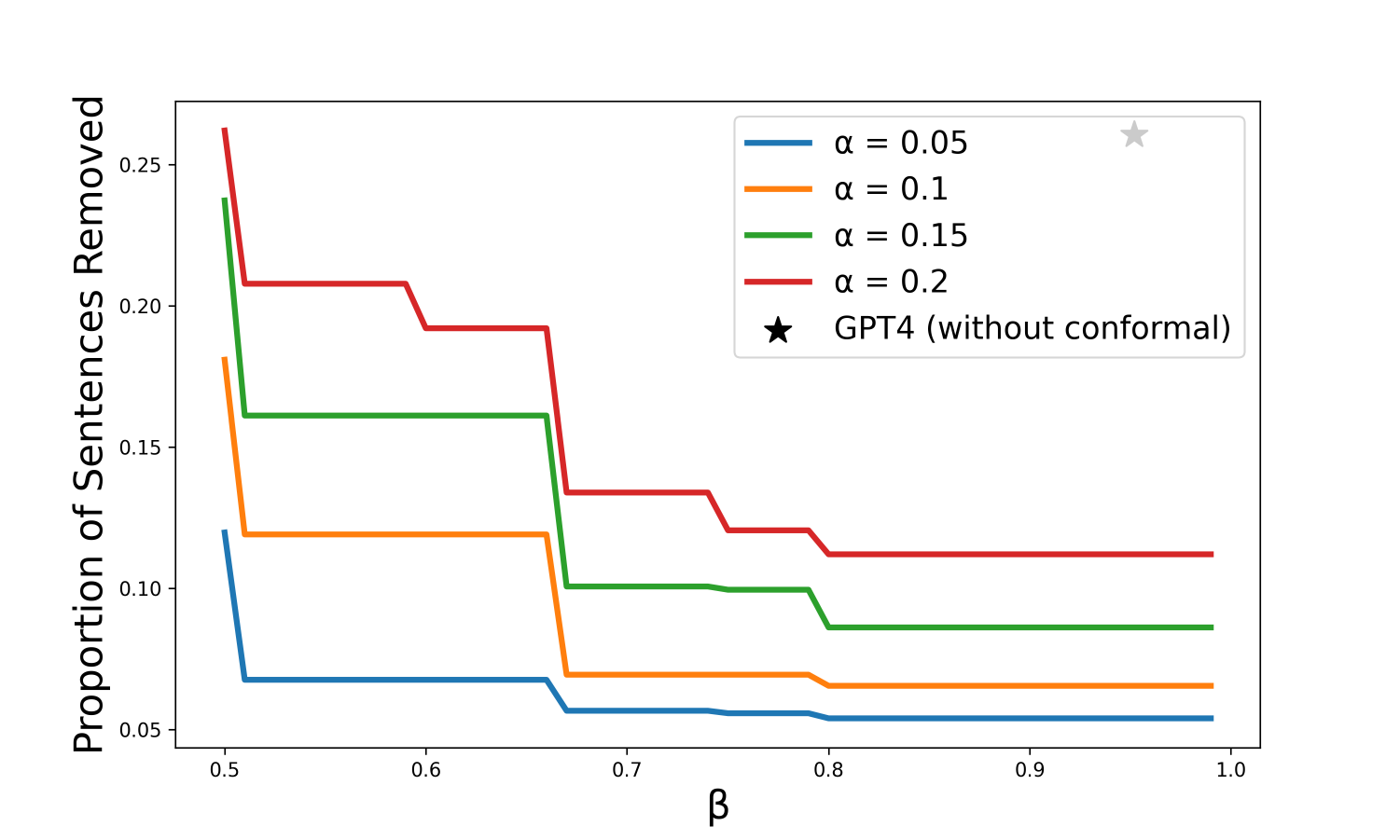}
        \caption{Qwen 3}
        \end{subfigure}
                \begin{subfigure}{0.48\textwidth}
        \includegraphics[width=0.95\linewidth, trim={30 0 60 40}, clip]{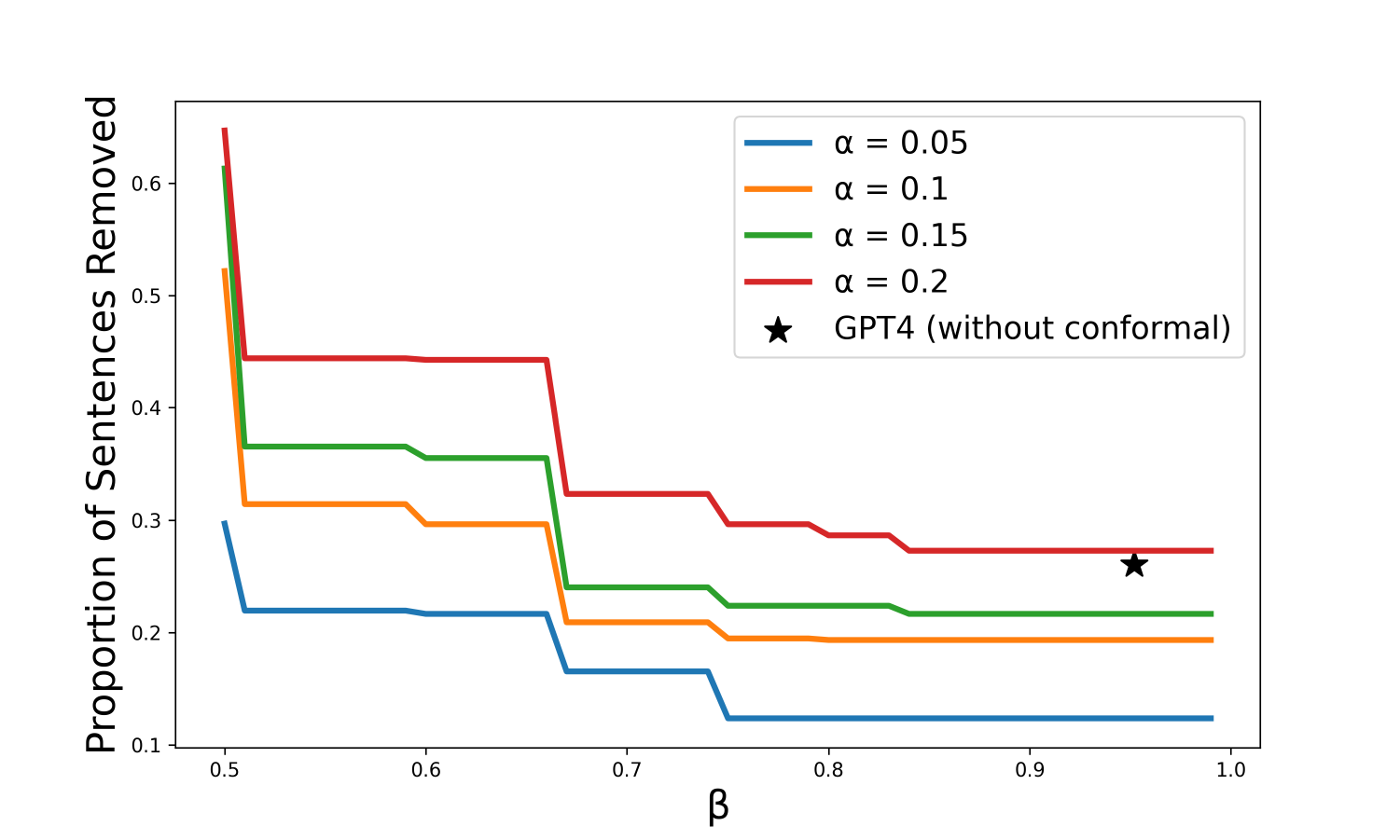}
        \caption{Gemini 2.0 Flash-Lite}
        \end{subfigure}
        \begin{subfigure}{0.48\textwidth}
        \includegraphics[width=0.95\linewidth, trim={30 0 60 40}, clip]{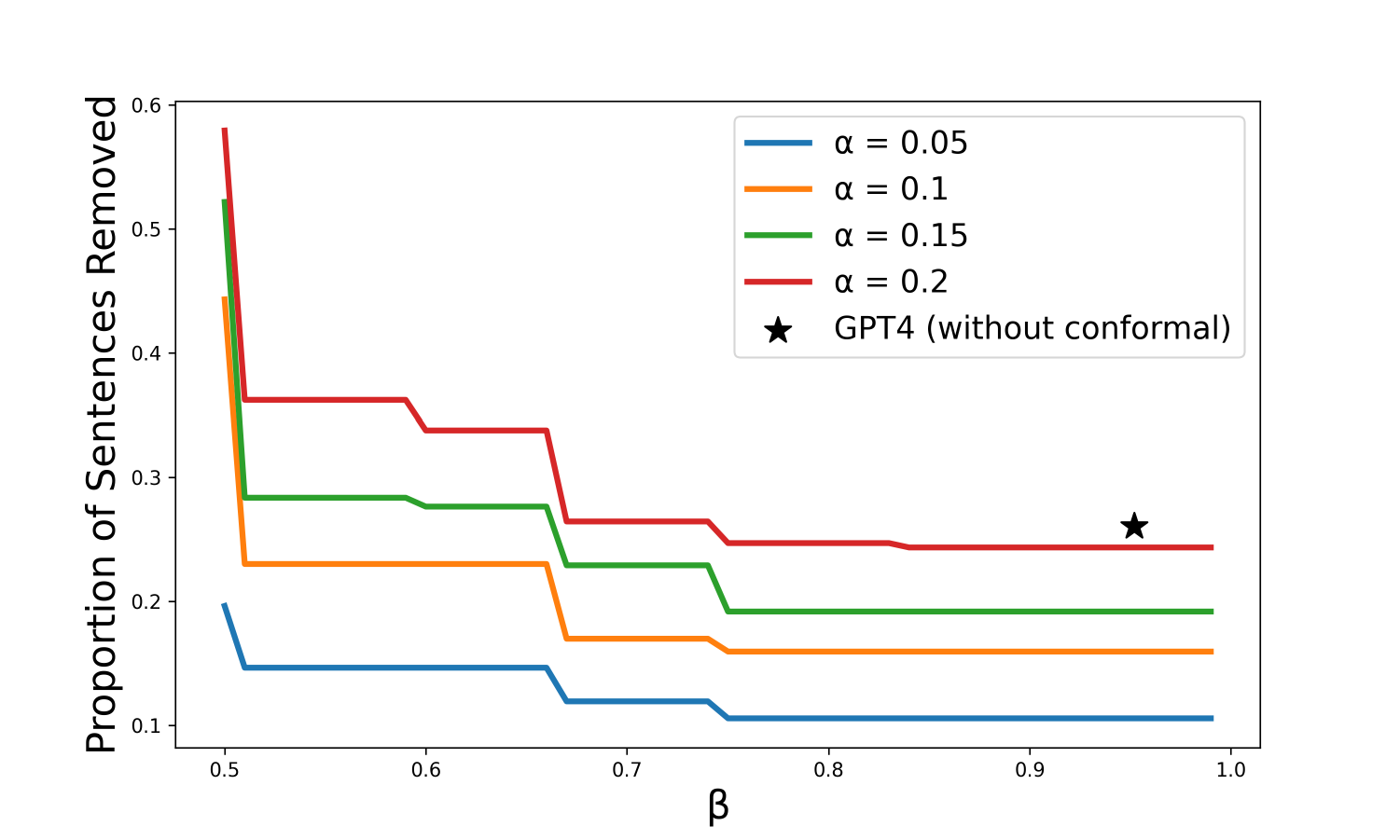}
        \caption{Gemini 2.5 Flash}
        \end{subfigure}
        \caption{Target recall $\beta$ vs. proportion of sentences removed (conciseness). Lines indicate different values for the target error rate $\alpha$ on CNN/DM.}
    \label{fig:reduction_versus_beta_CNNDM}
\end{figure}

\begin{figure}[t]
    \centering
        \begin{subfigure}{0.48\textwidth}
        \includegraphics[width=0.95\linewidth, trim={30 0 60 40}, clip]{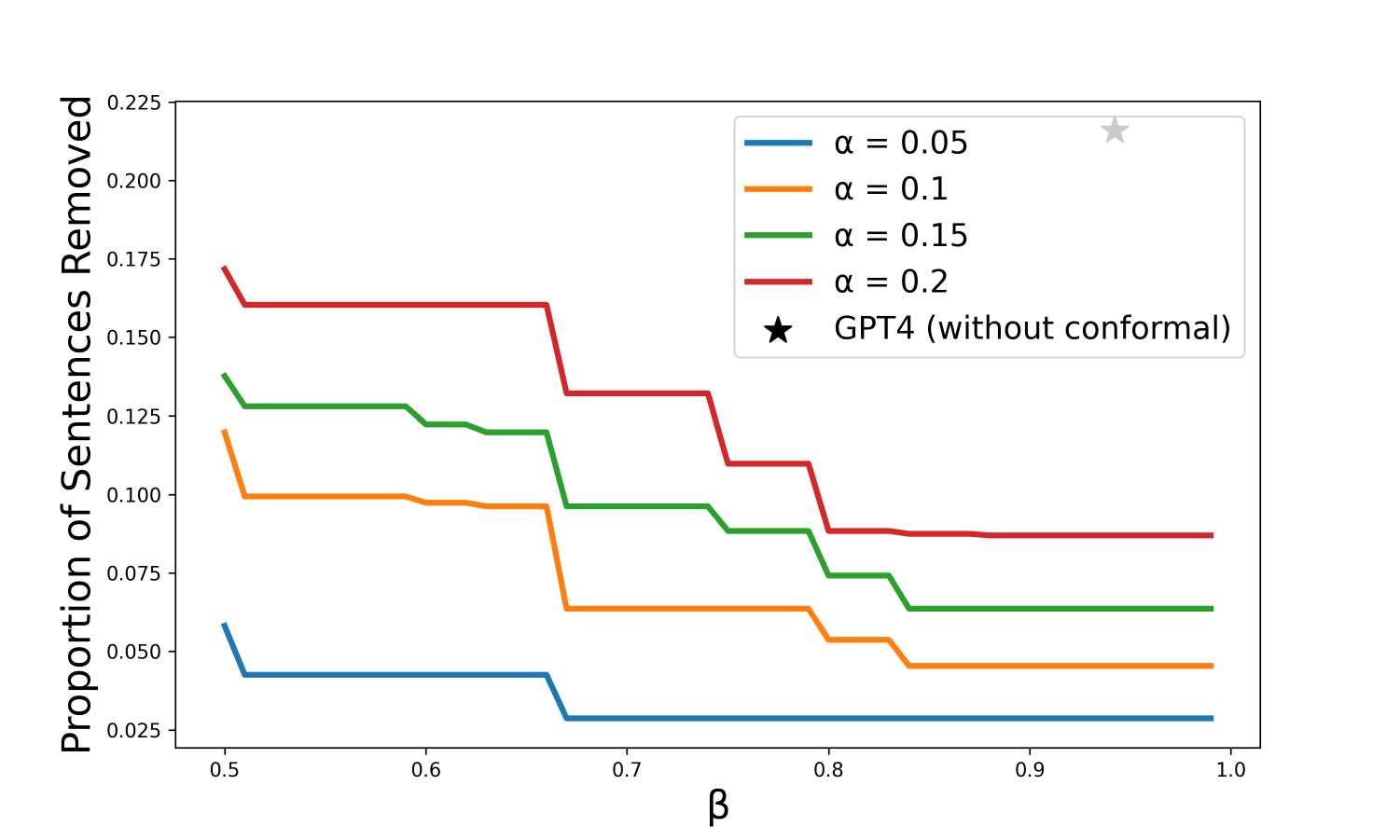}
        \caption{Cosine Similarity Centrality}
        \end{subfigure}
                \begin{subfigure}{0.48\textwidth}
        \includegraphics[width=0.95\linewidth, trim={30 0 60 40}, clip]{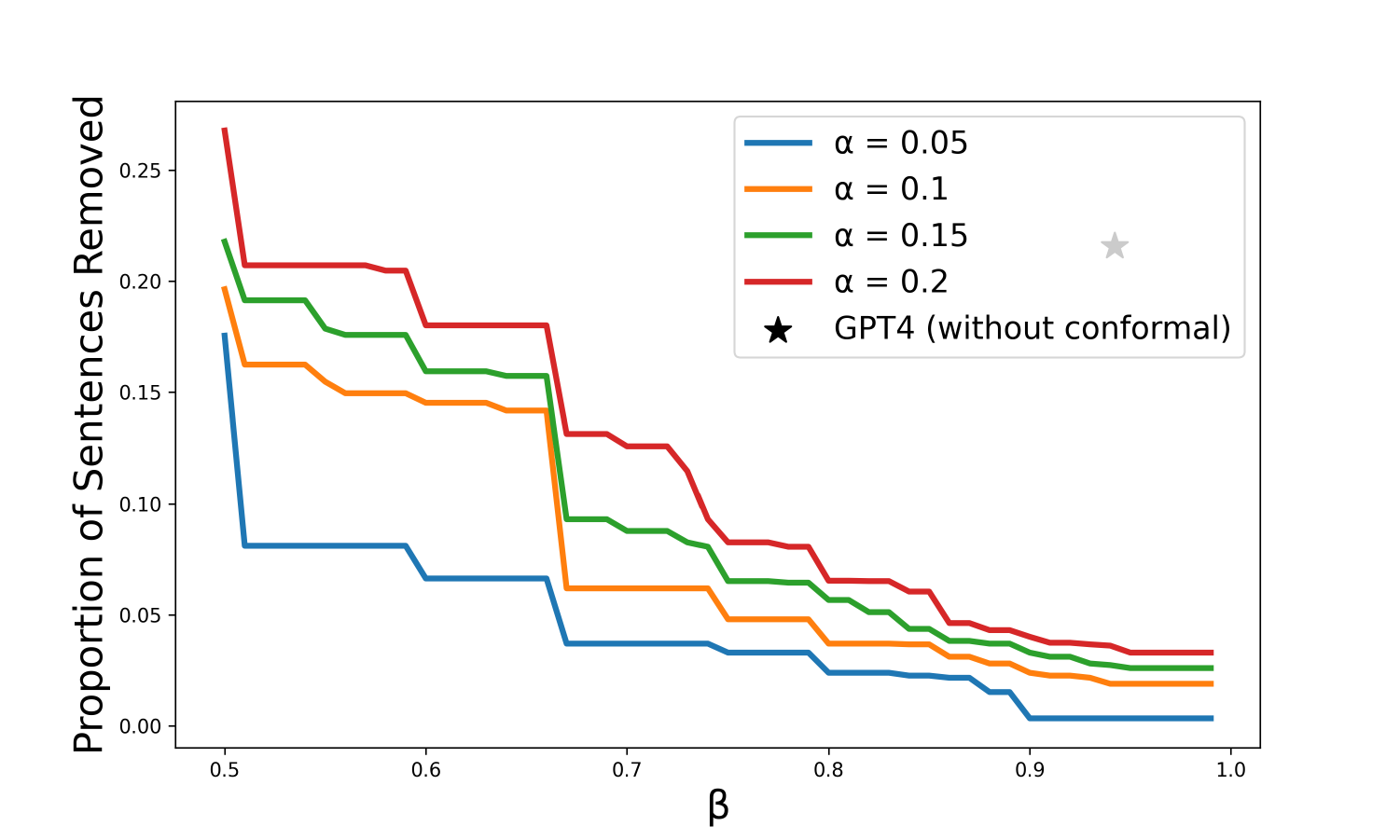}
        \caption{Sentence Centrality}
        \end{subfigure}
        \begin{subfigure}{0.48\textwidth}
        \includegraphics[width=0.95\linewidth, trim={30 0 60 40}, clip]{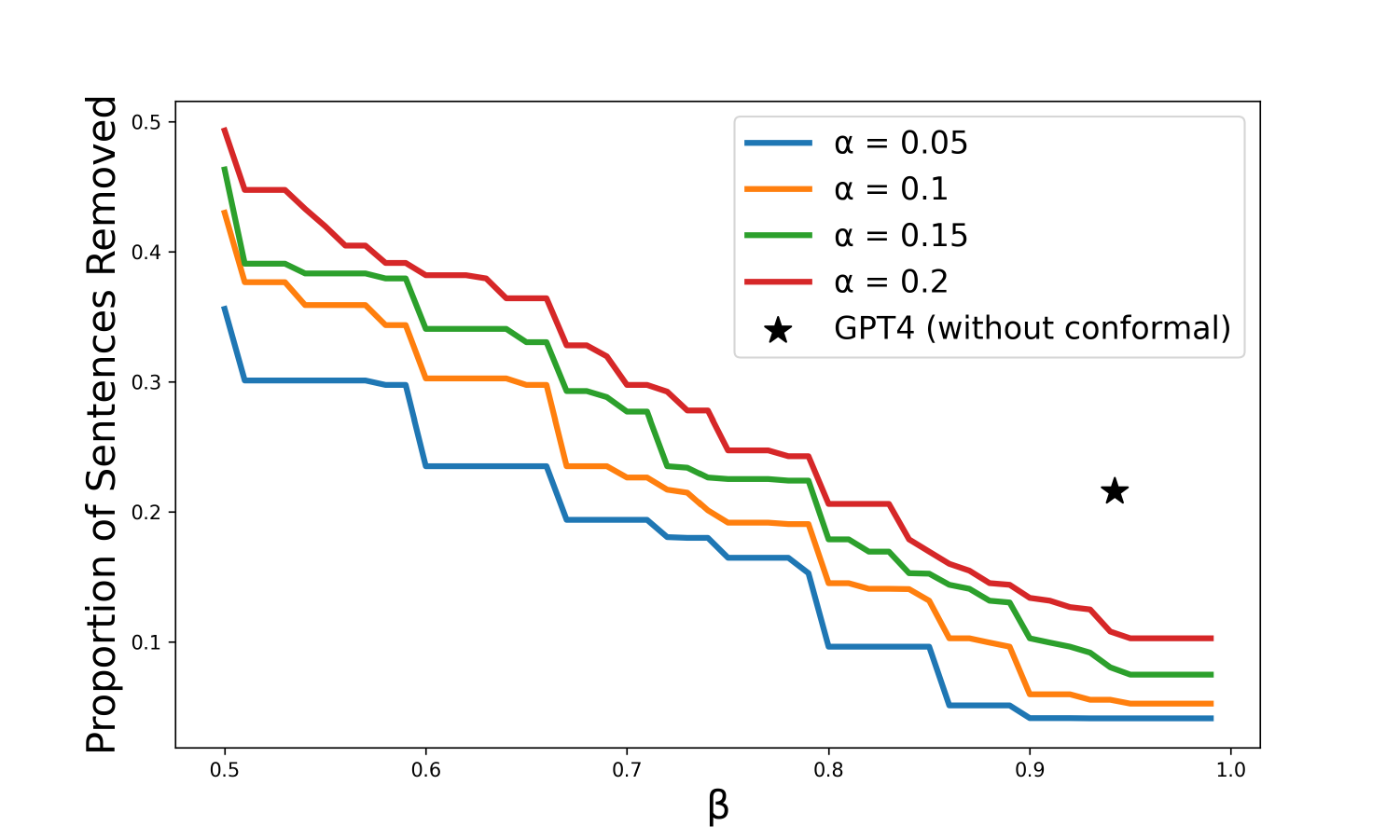}
        \caption{GUSUM}
        \end{subfigure}
        \begin{subfigure}{0.48\textwidth}
        \includegraphics[width=0.95\linewidth, trim={30 0 60 40}, clip]{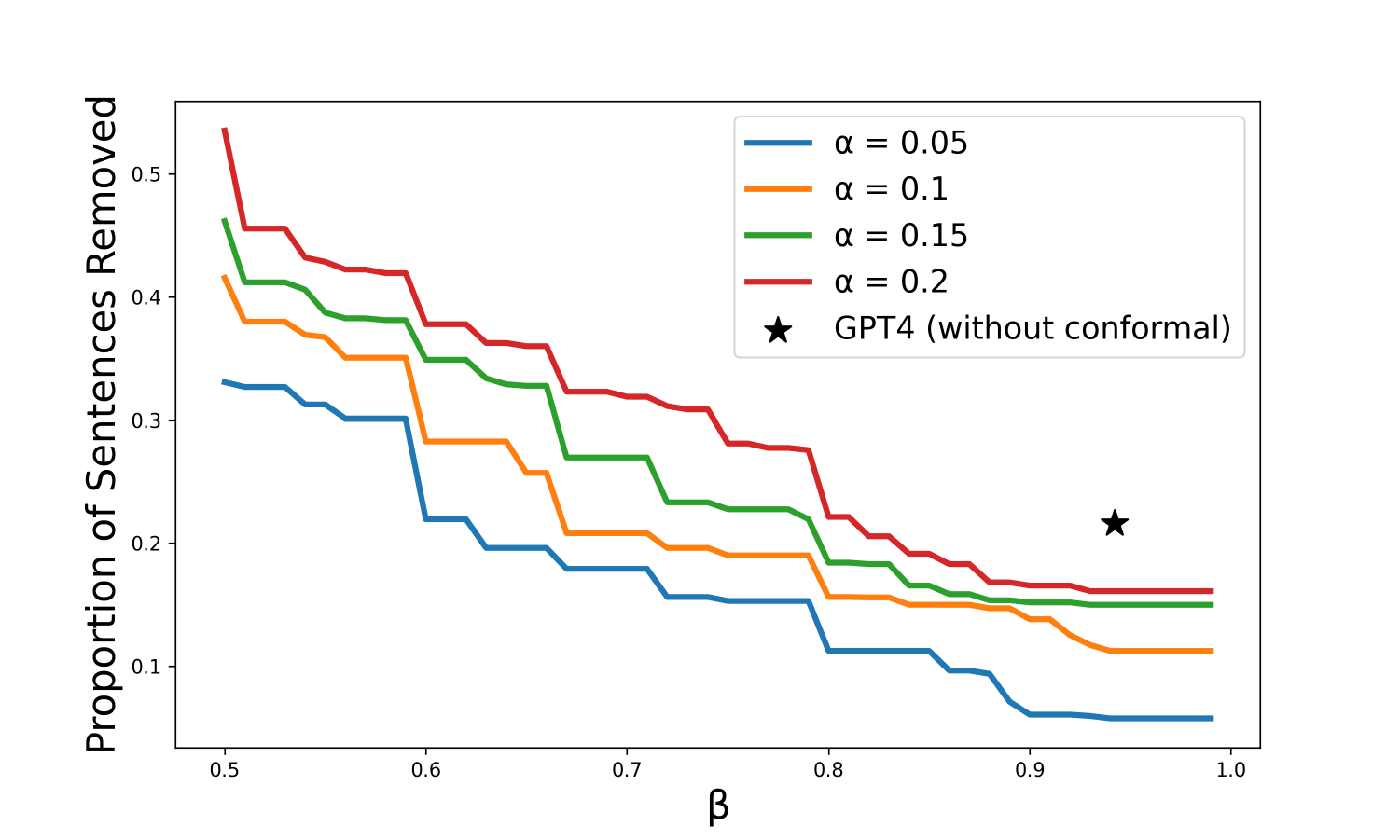}
        \caption{GPT-4o mini}
        \end{subfigure}
        \begin{subfigure}{0.48\textwidth}
        \includegraphics[width=0.95\linewidth, trim={30 0 60 40}, clip]{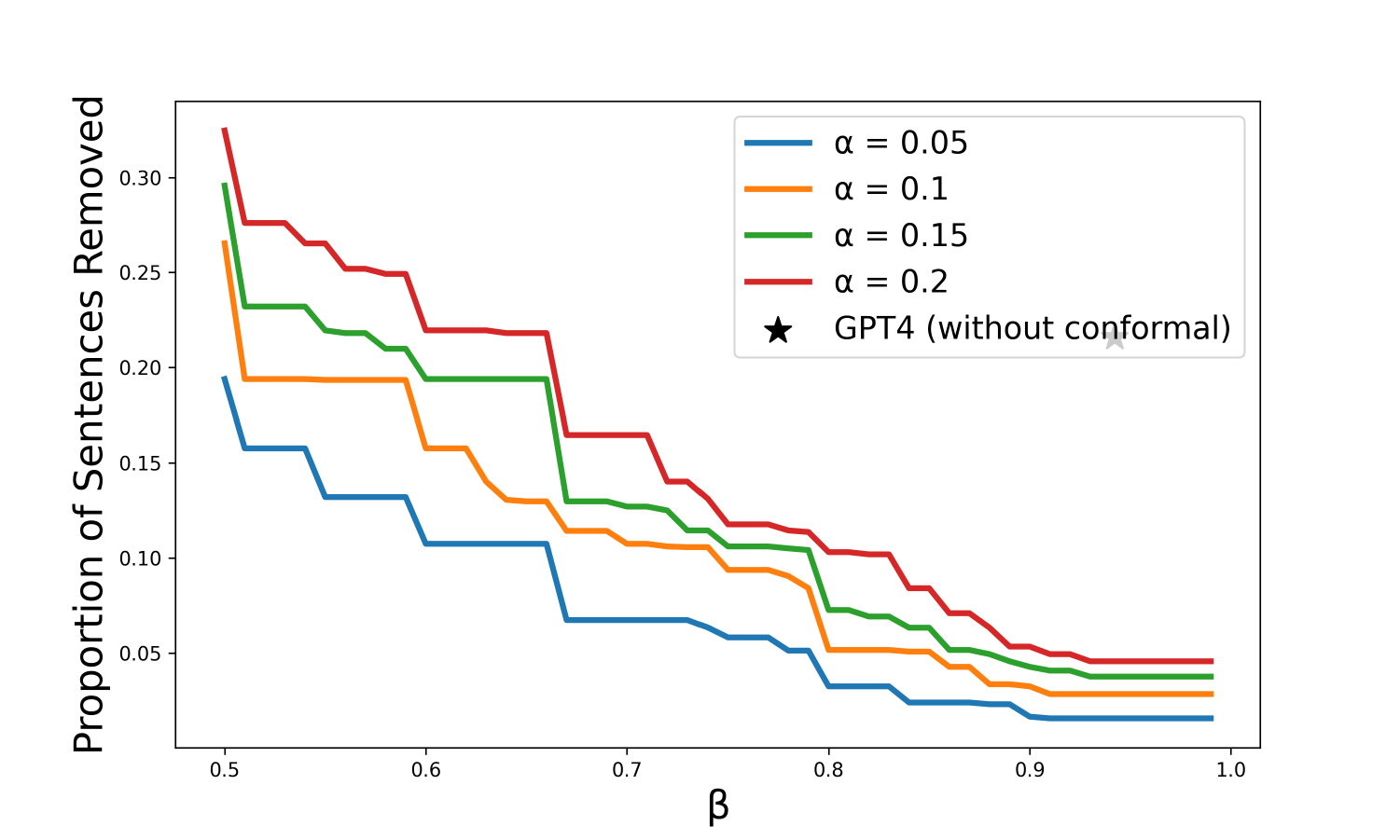}
        \caption{Llama 3}
        \end{subfigure}
        \begin{subfigure}{0.48\textwidth}
        \includegraphics[width=0.95\linewidth, trim={30 0 60 40}, clip]{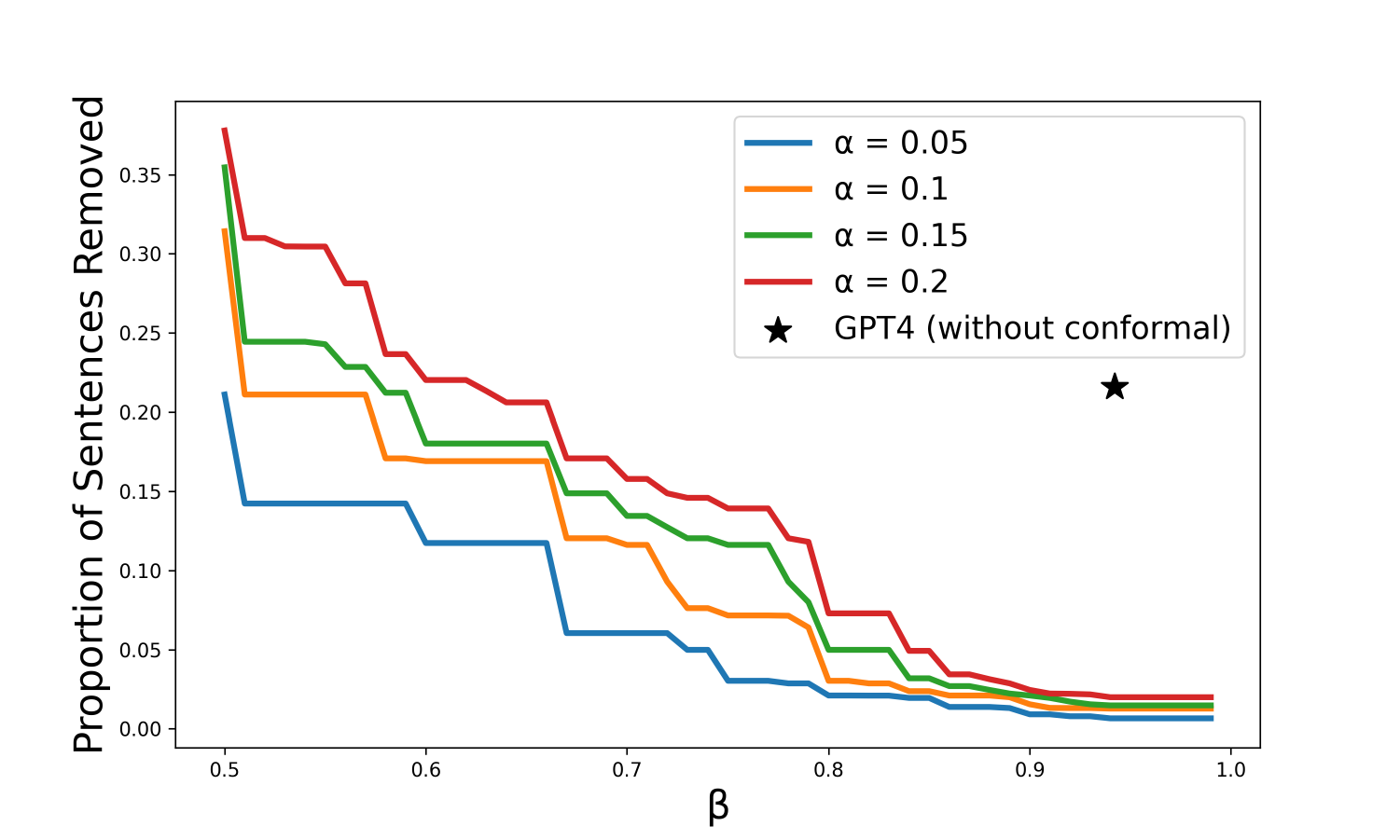}
        \caption{Qwen 3}
        \end{subfigure}
                \begin{subfigure}{0.48\textwidth}
        \includegraphics[width=0.95\linewidth, trim={30 0 60 40}, clip]{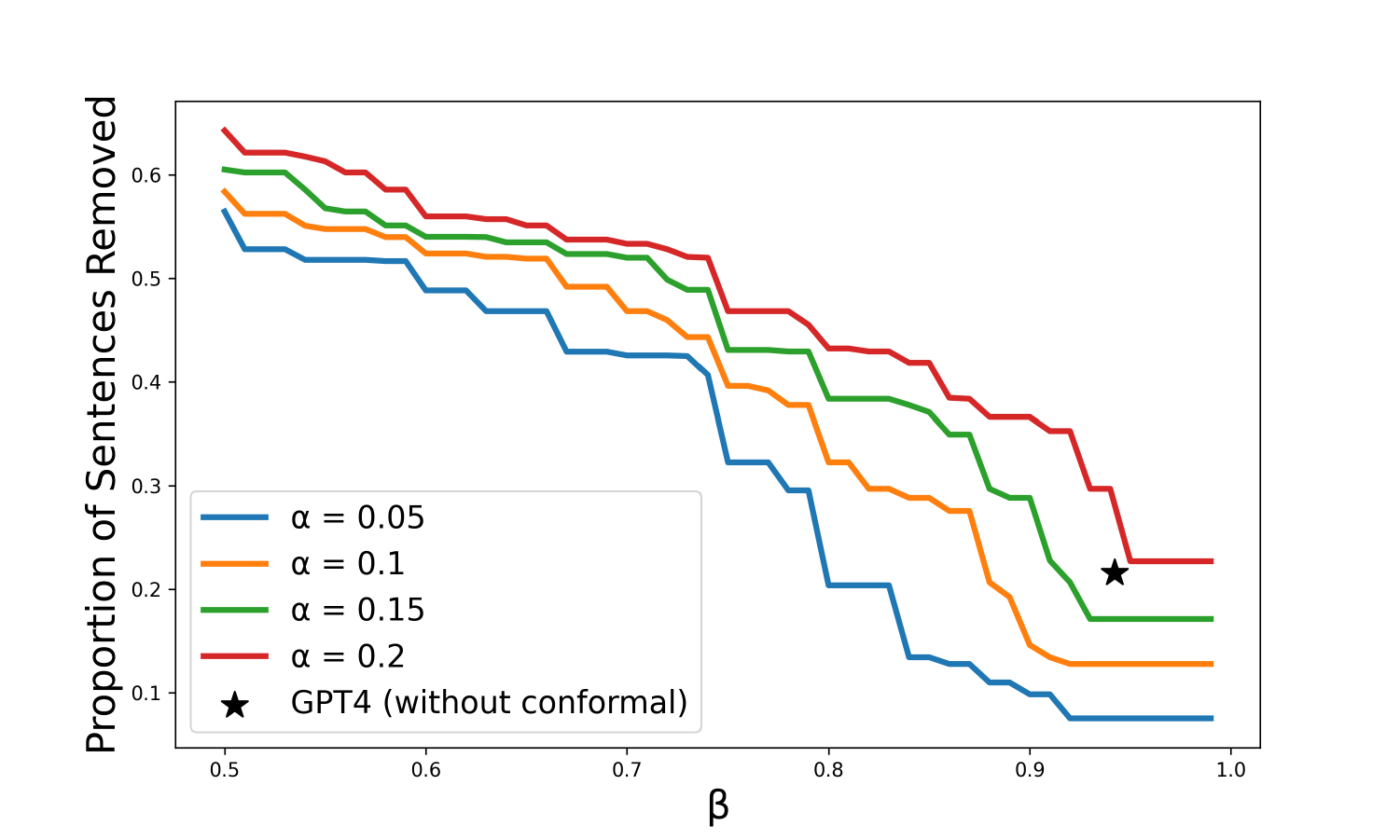}
        \caption{Gemini 2.0 Flash-Lite}
        \end{subfigure}
        \begin{subfigure}{0.48\textwidth}
        \includegraphics[width=0.95\linewidth, trim={30 0 60 40}, clip]{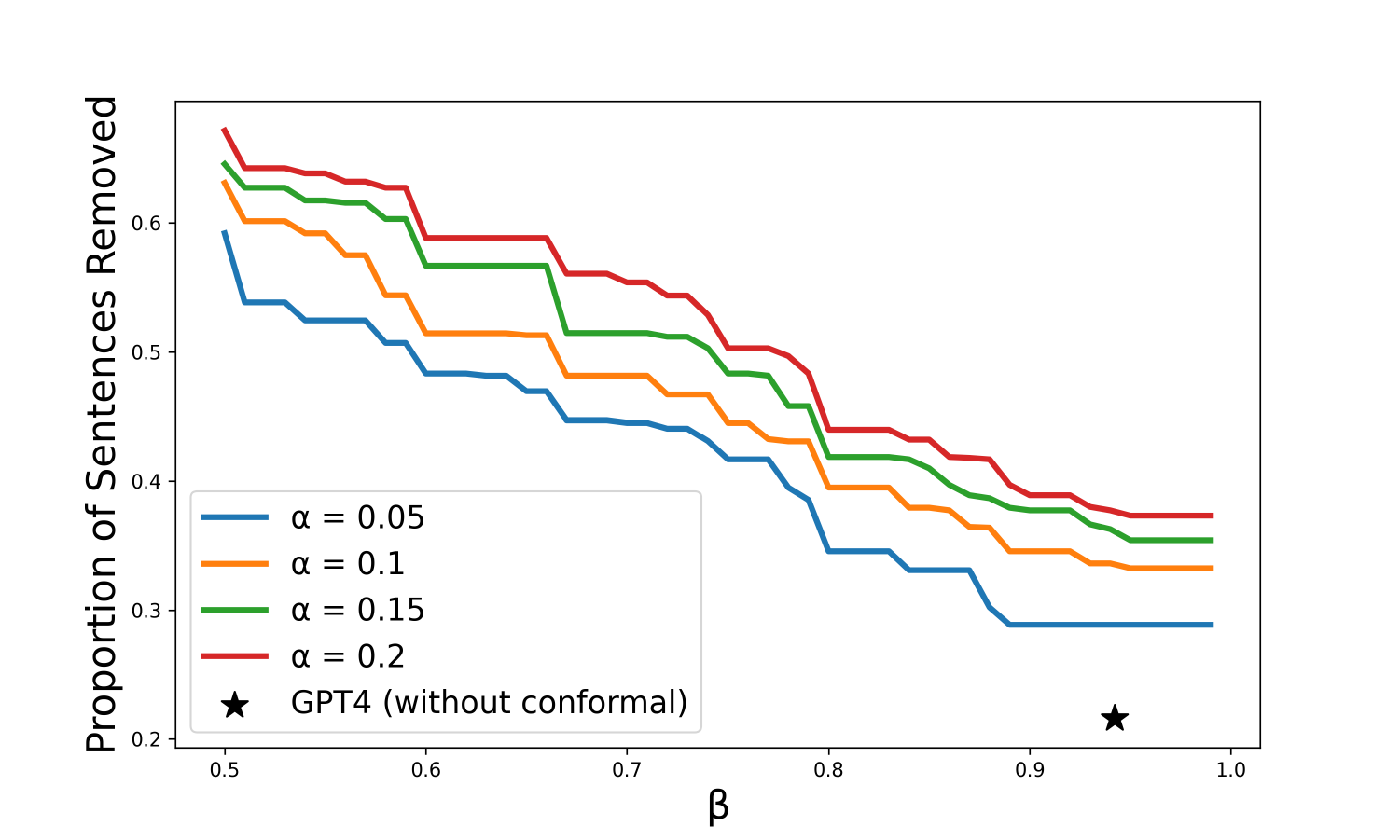}
        \caption{Gemini 2.5 Flash}
        \end{subfigure}
        \caption{Target recall $\beta$ vs. proportion of sentences removed (conciseness). Lines indicate different values for the target error rate $\alpha$ on CSDS.}
    \label{fig:reduction_versus_beta_CSDS}
\end{figure}

\begin{figure}[t]
    \centering
        \begin{subfigure}{0.48\textwidth}
        \includegraphics[width=0.95\linewidth, trim={30 0 60 40}, clip]{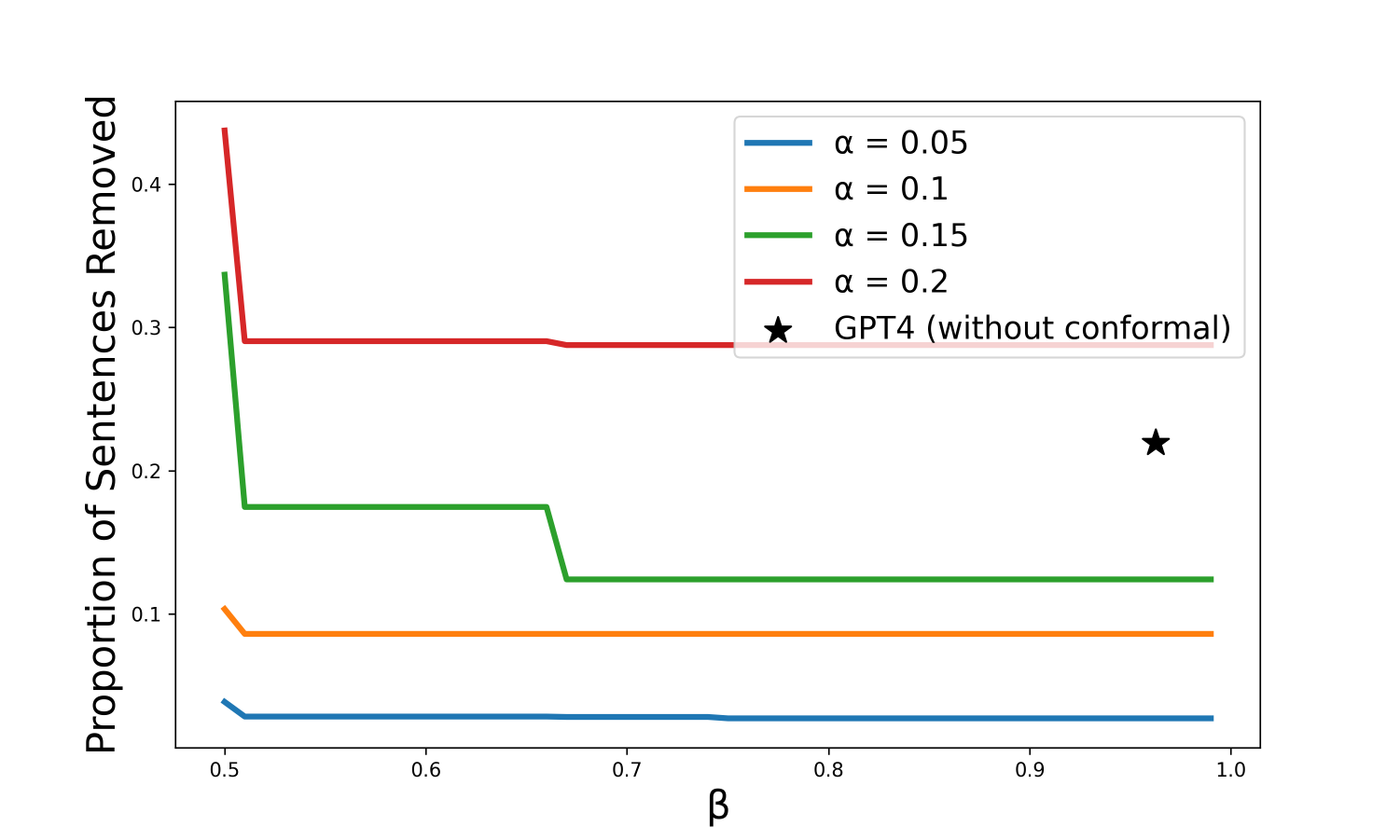}
        \caption{Cosine Similarity Centrality}
        \end{subfigure}
                \begin{subfigure}{0.48\textwidth}
        \includegraphics[width=0.95\linewidth, trim={30 0 60 40}, clip]{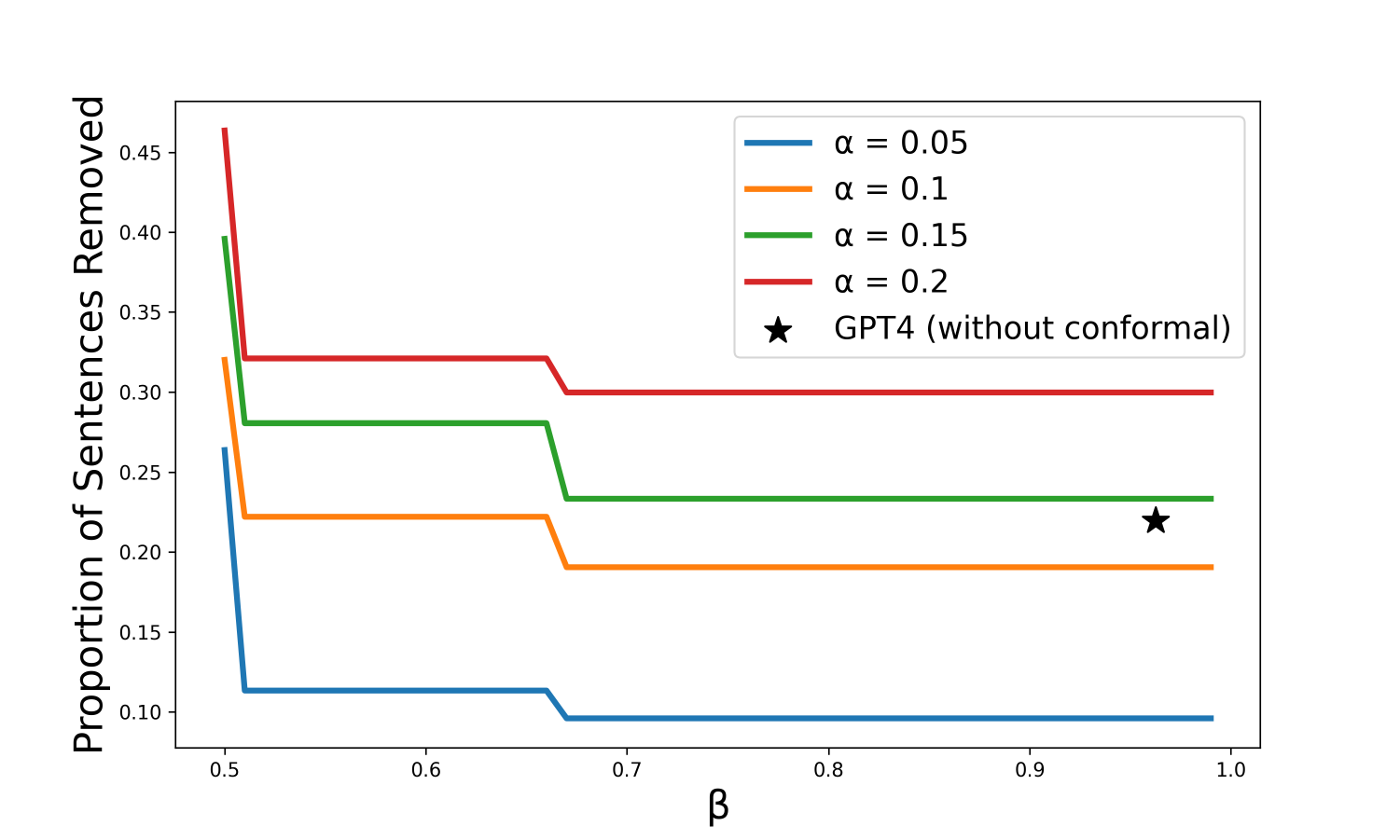}
        \caption{Sentence Centrality}
        \end{subfigure}
        \begin{subfigure}{0.48\textwidth}
        \includegraphics[width=0.95\linewidth, trim={30 0 60 40}, clip]{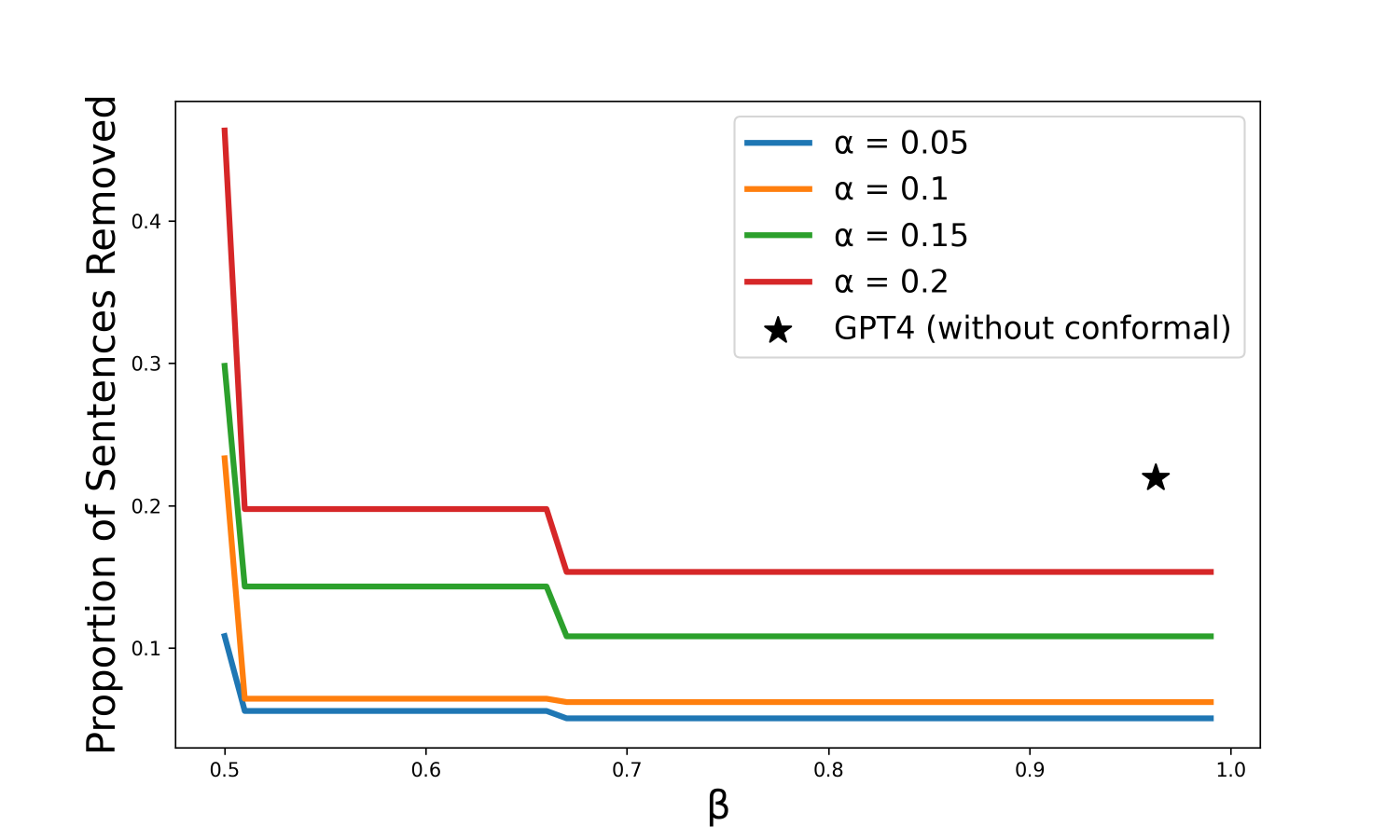}
        \caption{GUSUM}
        \end{subfigure}
        \begin{subfigure}{0.48\textwidth}
        \includegraphics[width=0.95\linewidth, trim={30 0 60 40}, clip]{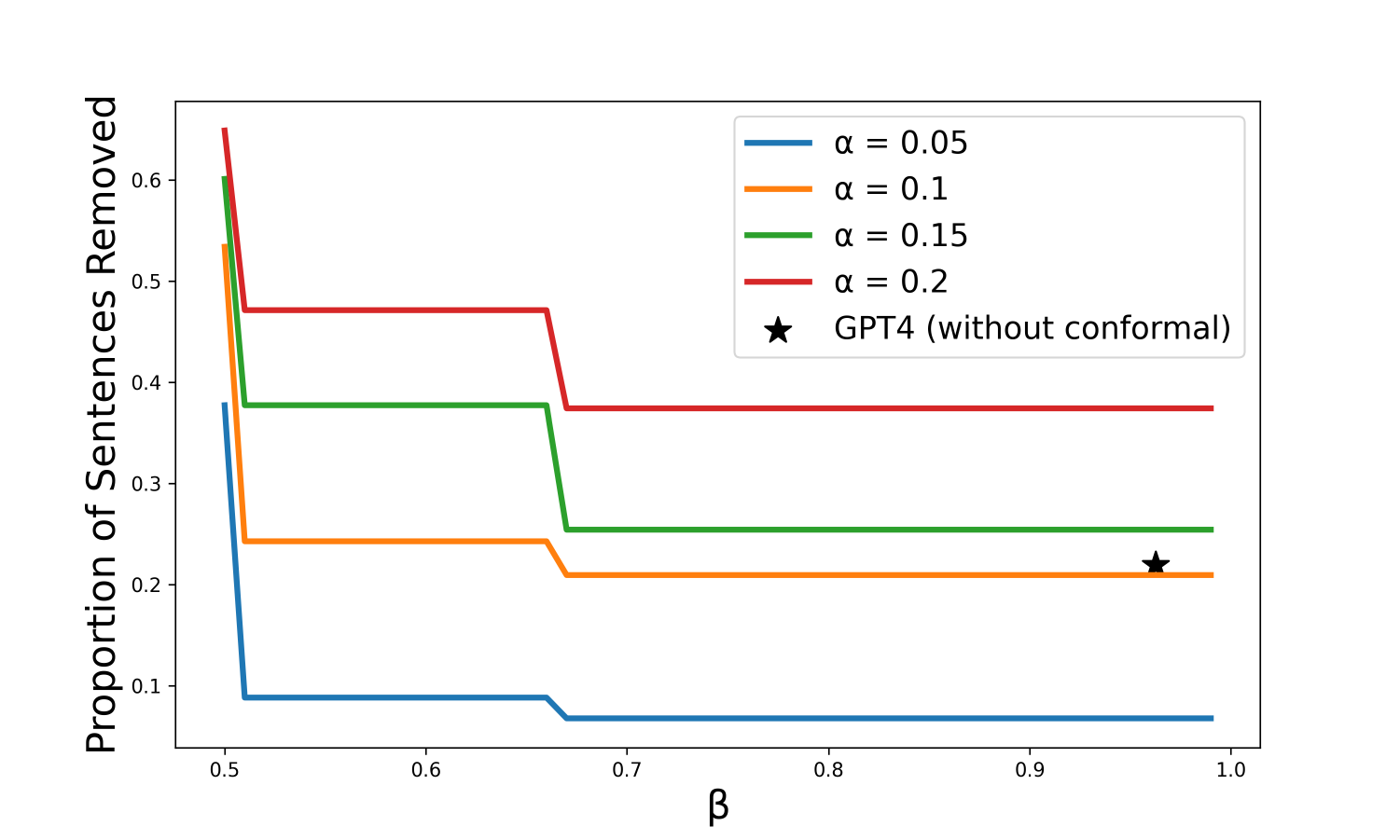}
        \caption{GPT-4o mini}
        \end{subfigure}
        \begin{subfigure}{0.48\textwidth}
        \includegraphics[width=0.95\linewidth, trim={30 0 60 40}, clip]{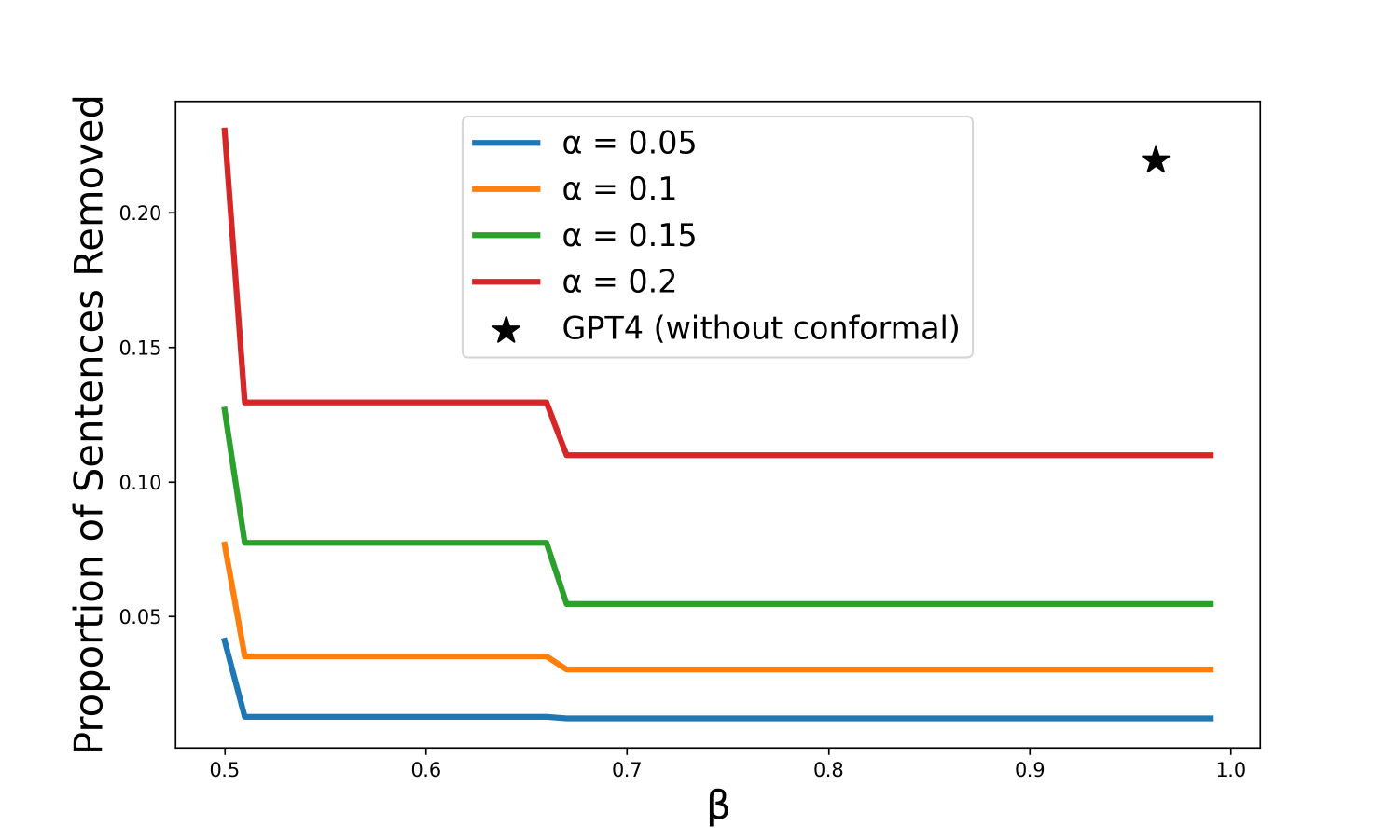}
        \caption{Llama 3}
        \end{subfigure}
        \begin{subfigure}{0.48\textwidth}
        \includegraphics[width=0.95\linewidth, trim={30 0 60 40}, clip]{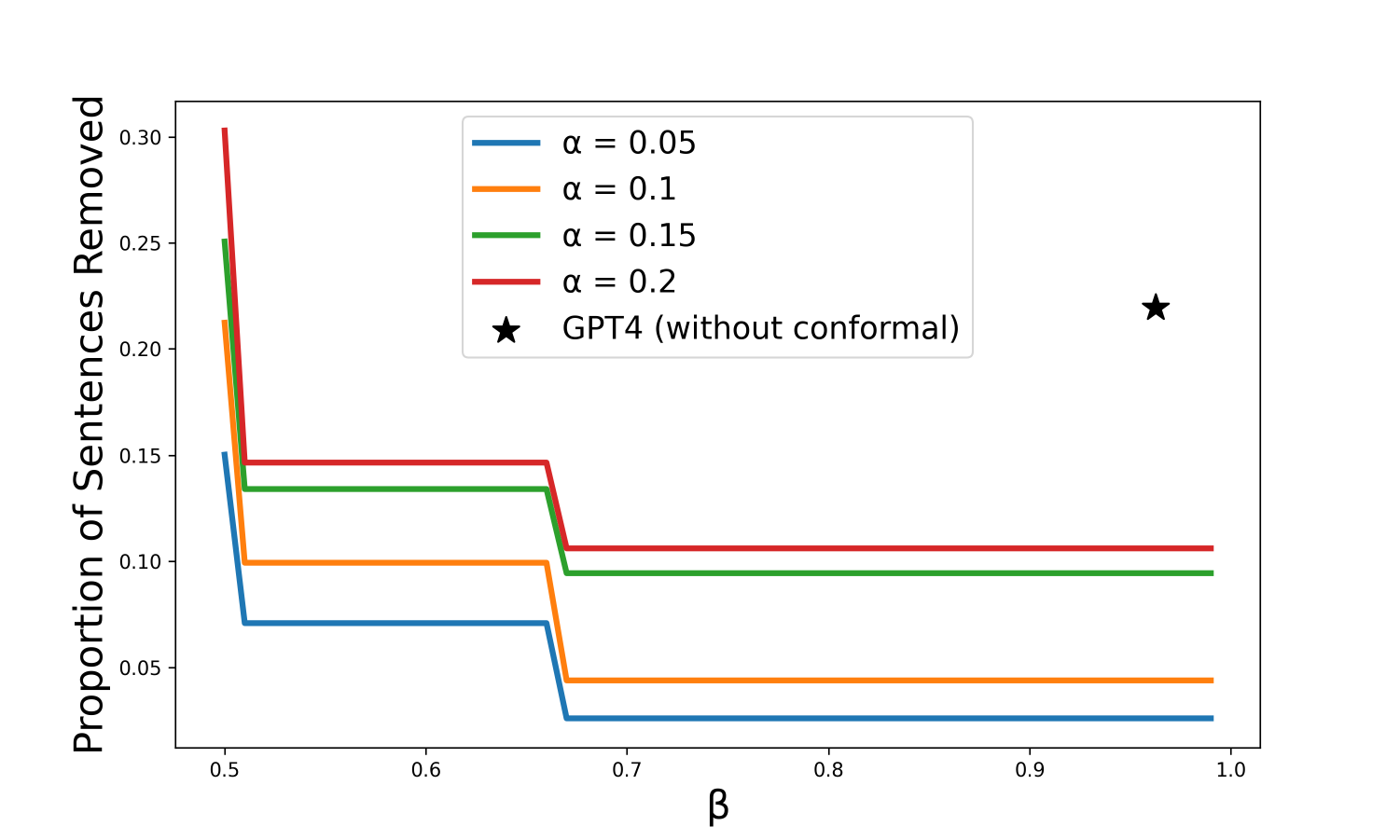}
        \caption{Qwen 3}
        \end{subfigure}
                \begin{subfigure}{0.48\textwidth}
        \includegraphics[width=0.95\linewidth, trim={30 0 60 40}, clip]{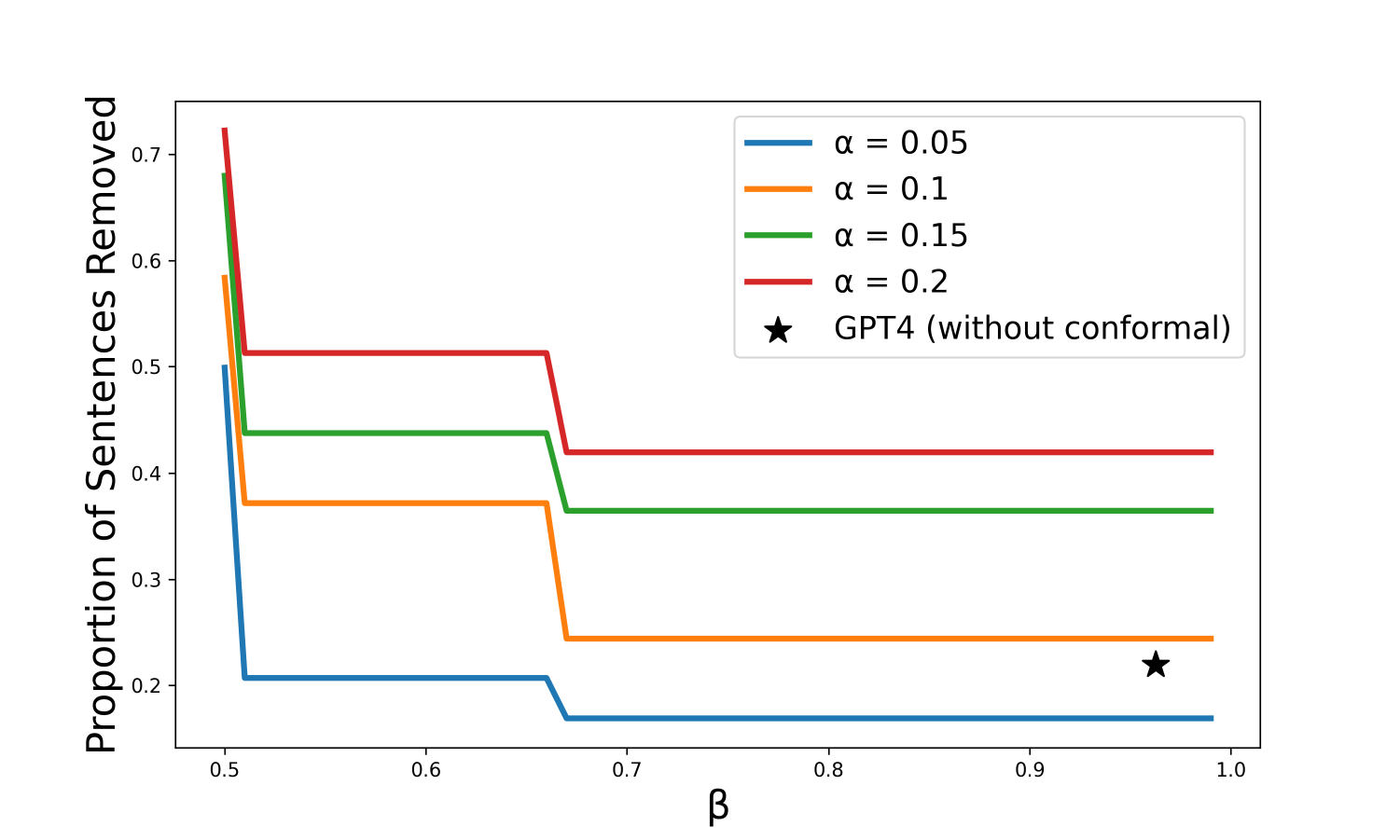}
        \caption{Gemini 2.0 Flash-Lite}
        \end{subfigure}
        \begin{subfigure}{0.48\textwidth}
        \includegraphics[width=0.95\linewidth, trim={30 0 60 40}, clip]{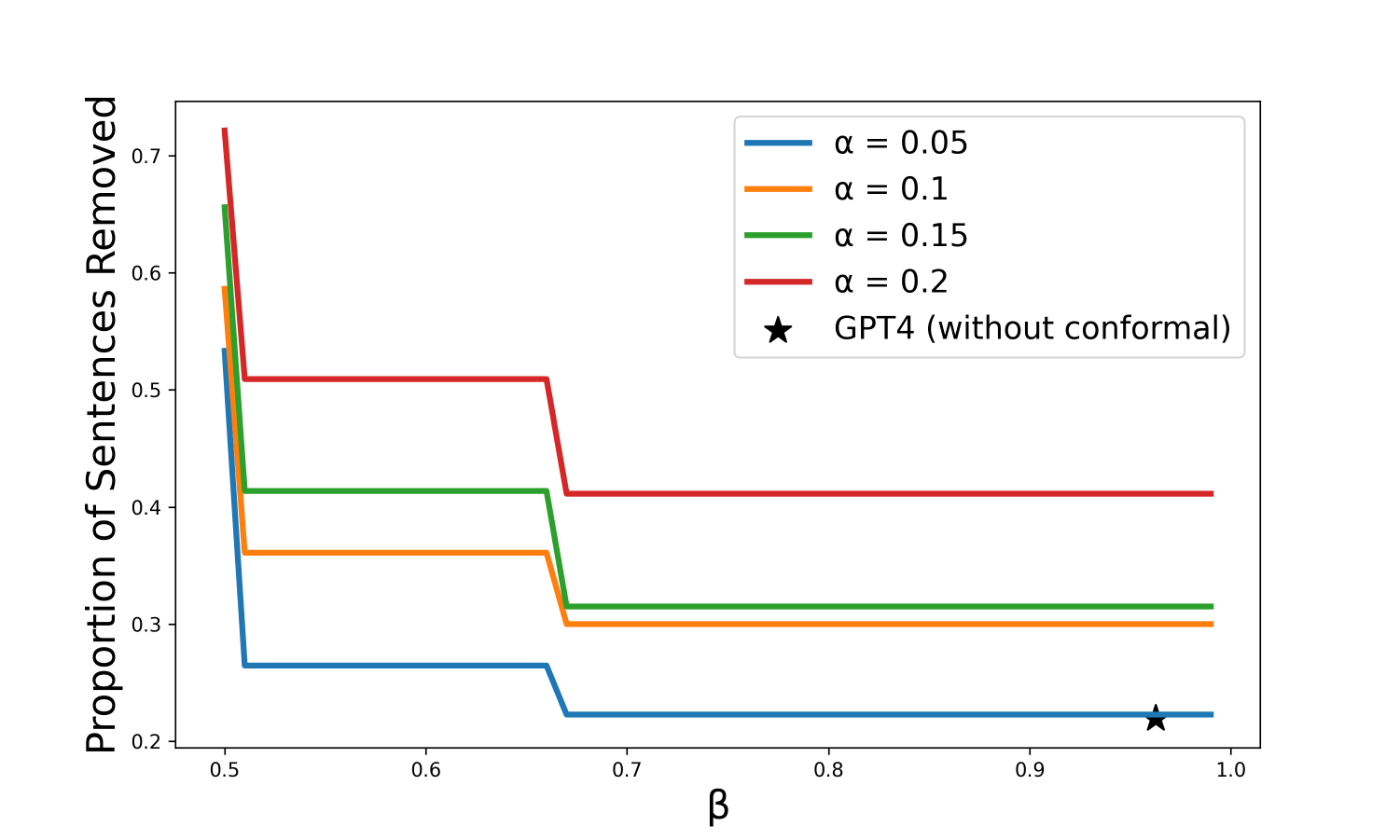}
        \caption{Gemini 2.5 Flash}
        \end{subfigure}
        \caption{Target recall $\beta$ vs. proportion of sentences removed (conciseness). Lines indicate different values for the target error rate $\alpha$ on TLDR-AIC.}
    \label{fig:reduction_versus_beta_tldr}
\end{figure}

\begin{figure}[t]
    \centering
        \begin{subfigure}{0.48\textwidth}
        \includegraphics[width=0.95\linewidth, trim={30 0 60 40}, clip]{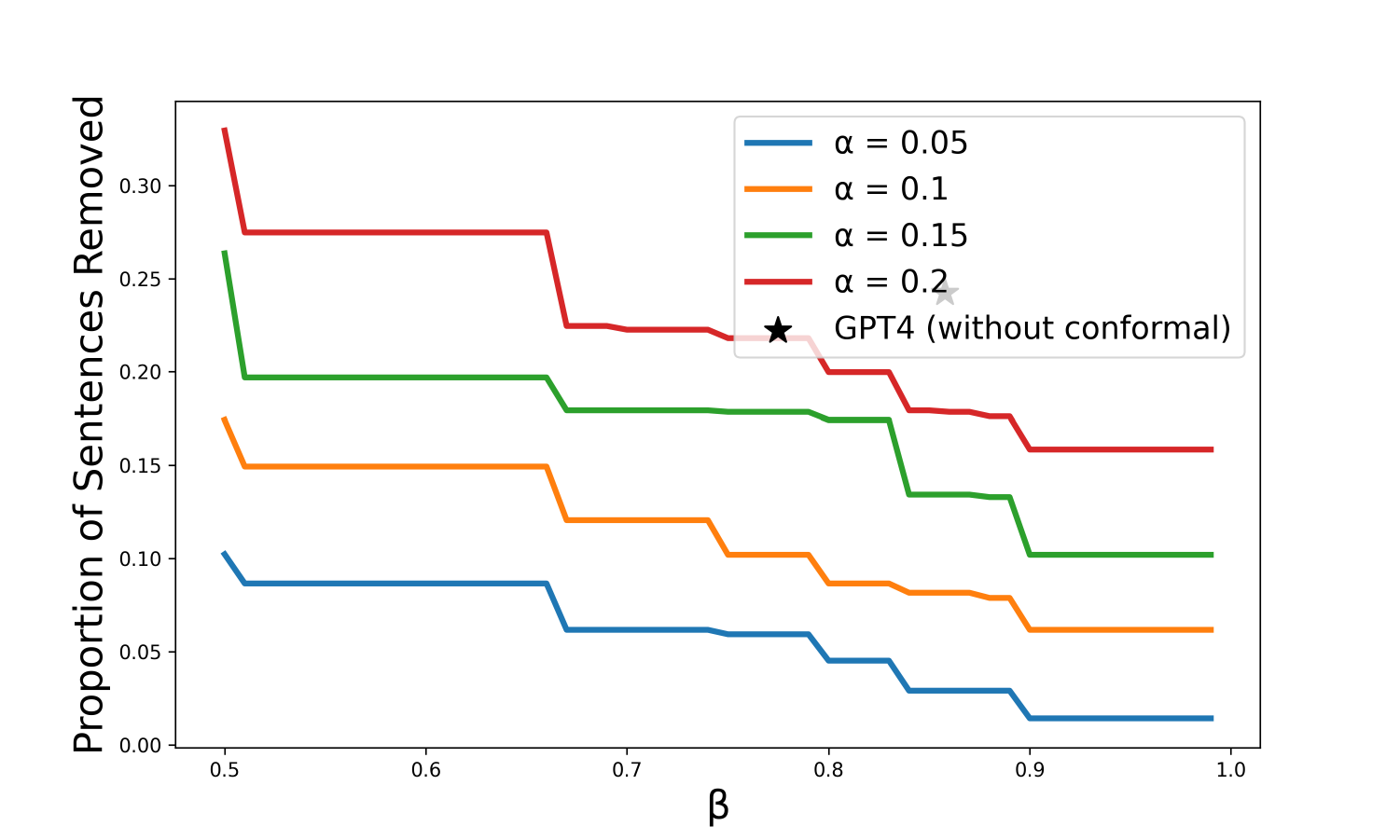}
        \caption{Cosine Similarity Centrality}
        \end{subfigure}
                \begin{subfigure}{0.48\textwidth}
        \includegraphics[width=0.95\linewidth, trim={30 0 60 40}, clip]{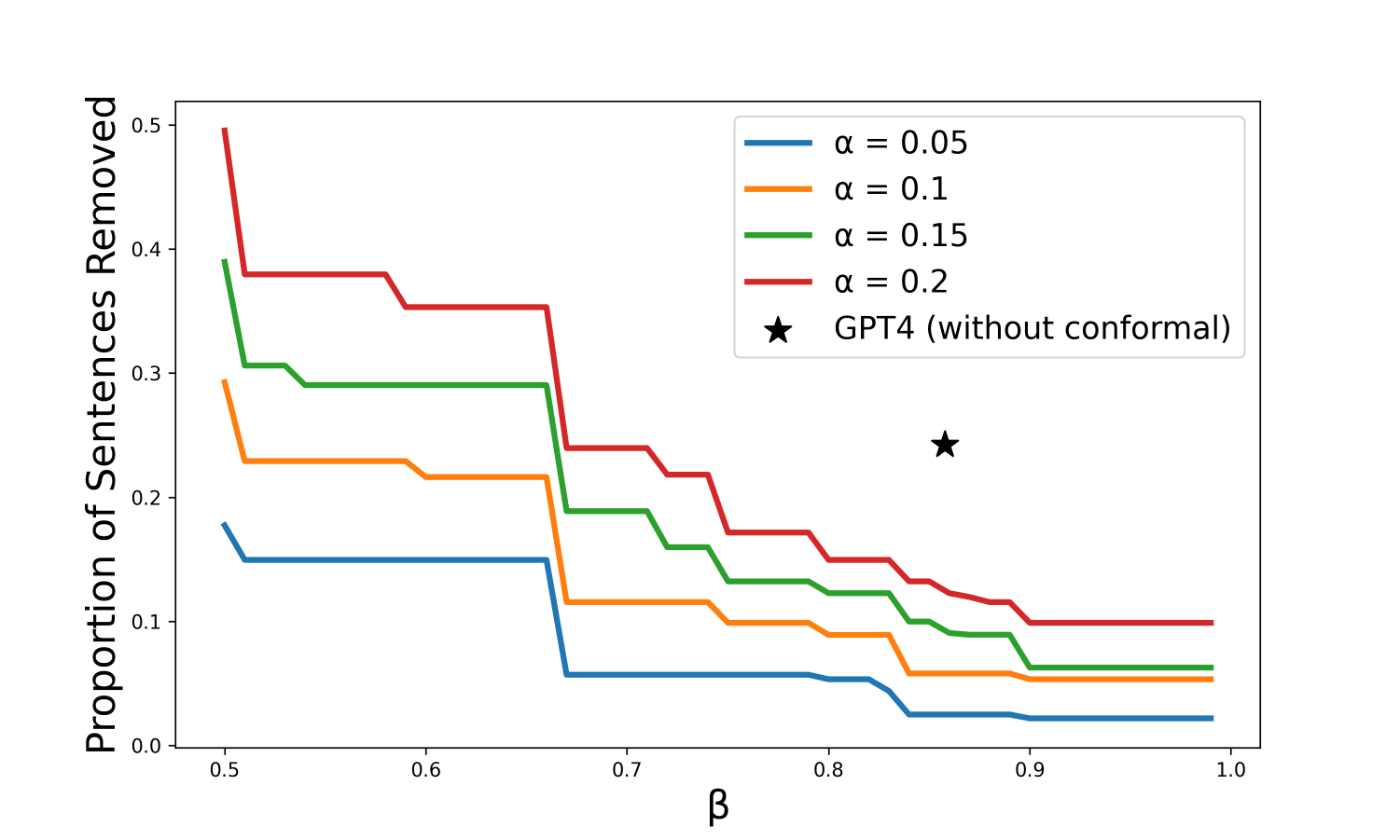}
        \caption{Sentence Centrality}
        \end{subfigure}
        \begin{subfigure}{0.48\textwidth}
        \includegraphics[width=0.95\linewidth, trim={30 0 60 40}, clip]{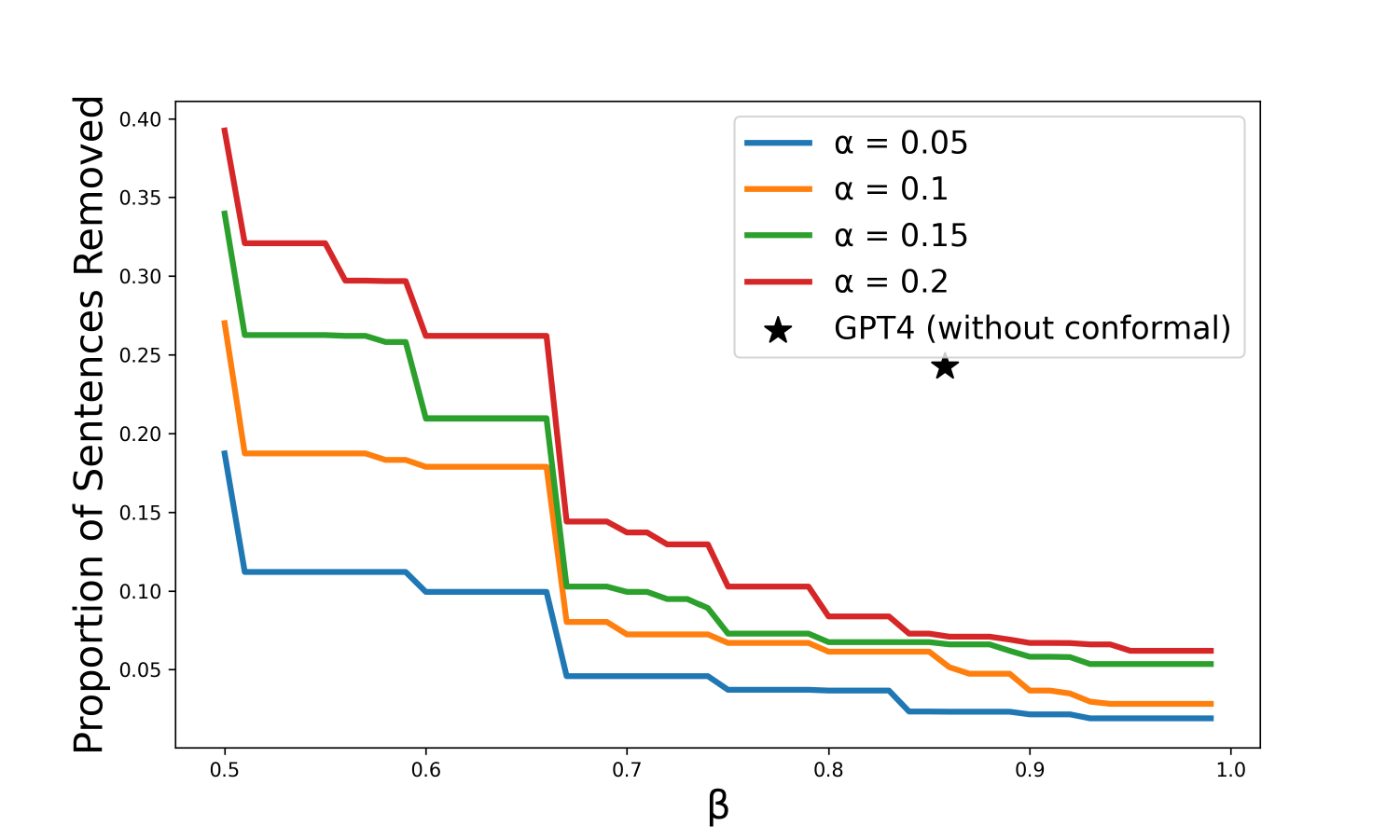}
        \caption{GUSUM}
        \end{subfigure}
        \begin{subfigure}{0.48\textwidth}
        \includegraphics[width=0.95\linewidth, trim={30 0 60 40}, clip]{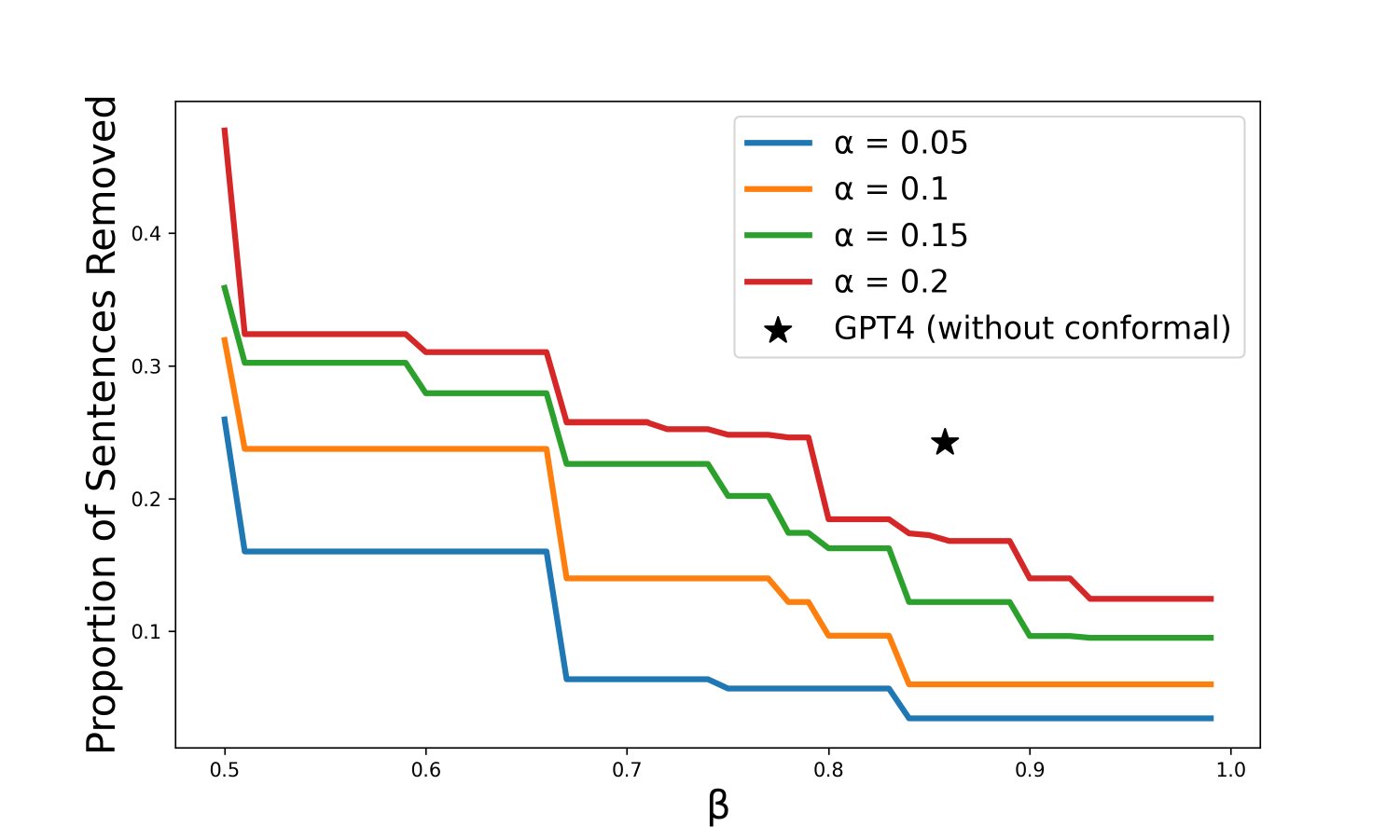}
        \caption{GPT-4o mini}
        \end{subfigure}
                \begin{subfigure}{0.48\textwidth}
        \includegraphics[width=0.95\linewidth, trim={30 0 60 40}, clip]{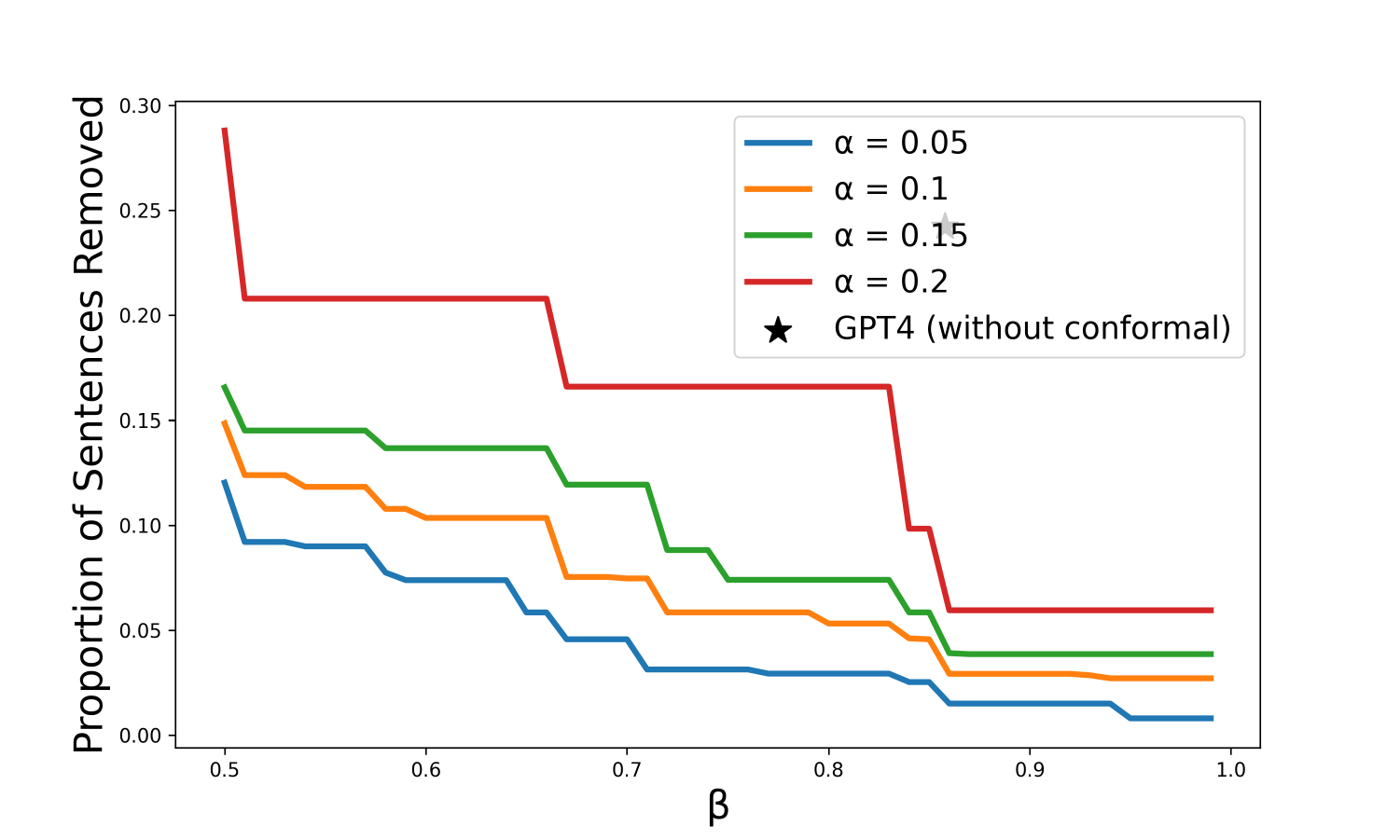}
        \caption{Llama 3}
        \end{subfigure}
        \begin{subfigure}{0.48\textwidth}
        \includegraphics[width=0.95\linewidth, trim={30 0 60 40}, clip]{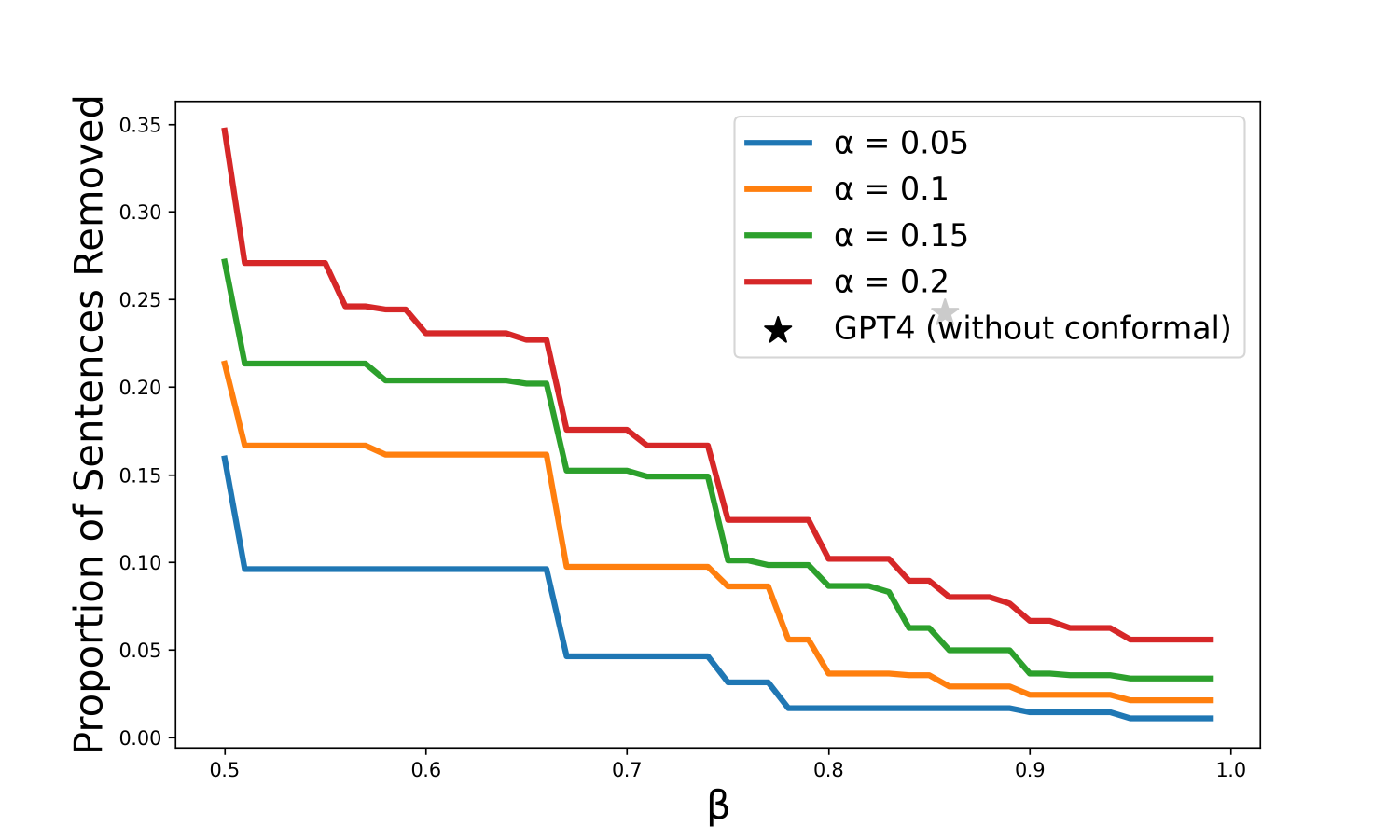}
        \caption{Qwen 3}
        \end{subfigure}
        \begin{subfigure}{0.48\textwidth}
        \includegraphics[width=0.95\linewidth, trim={30 0 60 40}, clip]{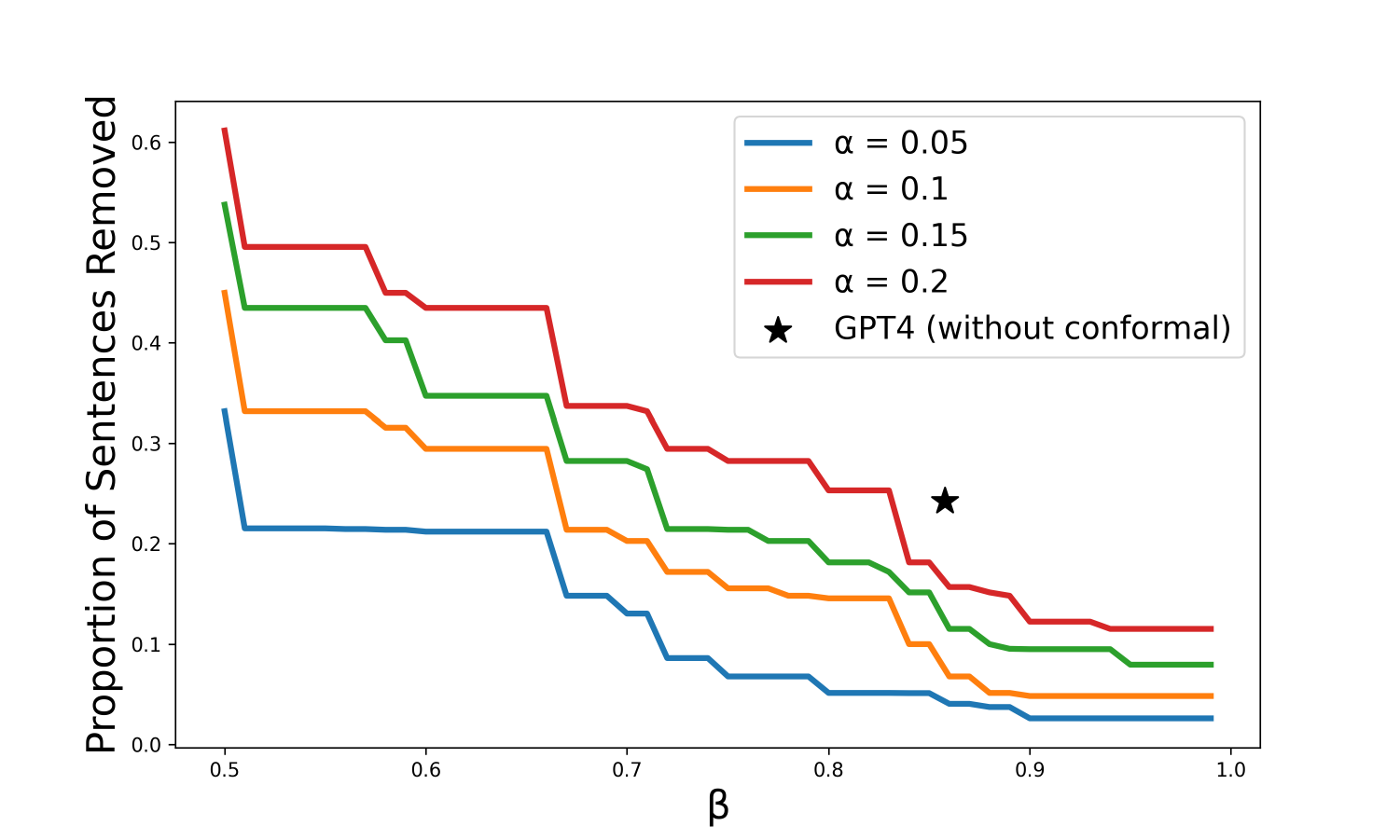}
        \caption{Gemini 2.0 Flash-Lite}
        \end{subfigure}
        \begin{subfigure}{0.48\textwidth}
        \includegraphics[width=0.95\linewidth, trim={30 0 60 40}, clip]{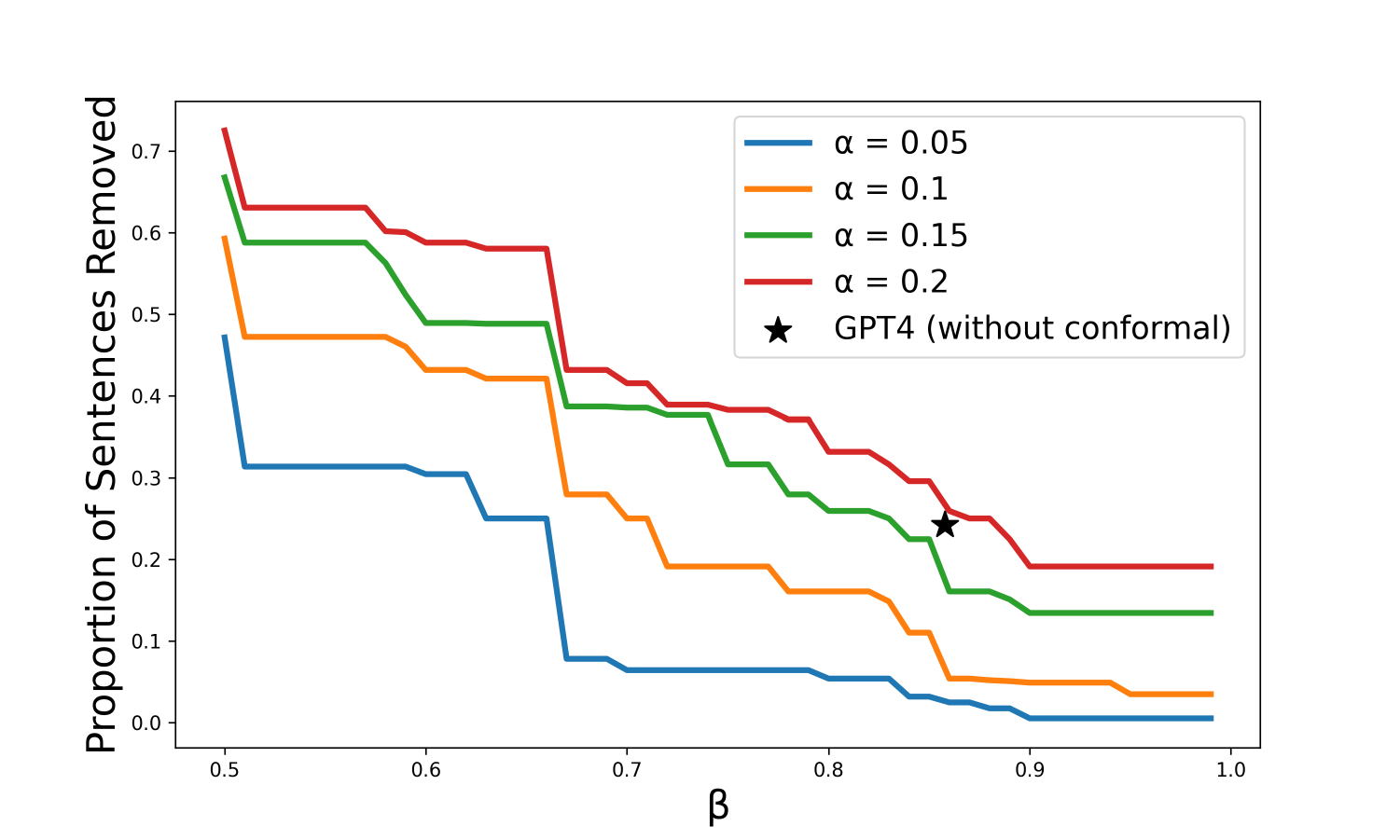}
        \caption{Gemini 2.5 Flash}
        \end{subfigure}
        \caption{Target recall $\beta$ vs. proportion of sentences removed (conciseness). Lines indicate different values for the target error rate $\alpha$ on ECT.}
    \label{fig:reduction_versus_beta_ECT}
\end{figure}

Figures \ref{fig:reduction_versus_beta_CNNDM} - \ref{fig:reduction_versus_beta_ECT} show the conciseness, the proportion of sentences removed, based on the choice of $\beta$ for all datasets and methods\footnote{Due to computational constraints, we only compute this plot for LexRank on the CNN/DM dataset}, analogous to \Cref{fig:reduction_versus_beta} in \Cref{sec:empirical}. Once again, the trend is highly similar across settings, with higher $\beta$ leading to a smaller reduction in length. 

\begin{figure}[t]
    \centering
        \begin{subfigure}{0.48\textwidth}
        \includegraphics[width=0.95\linewidth, trim={30 0 60 40}, clip]{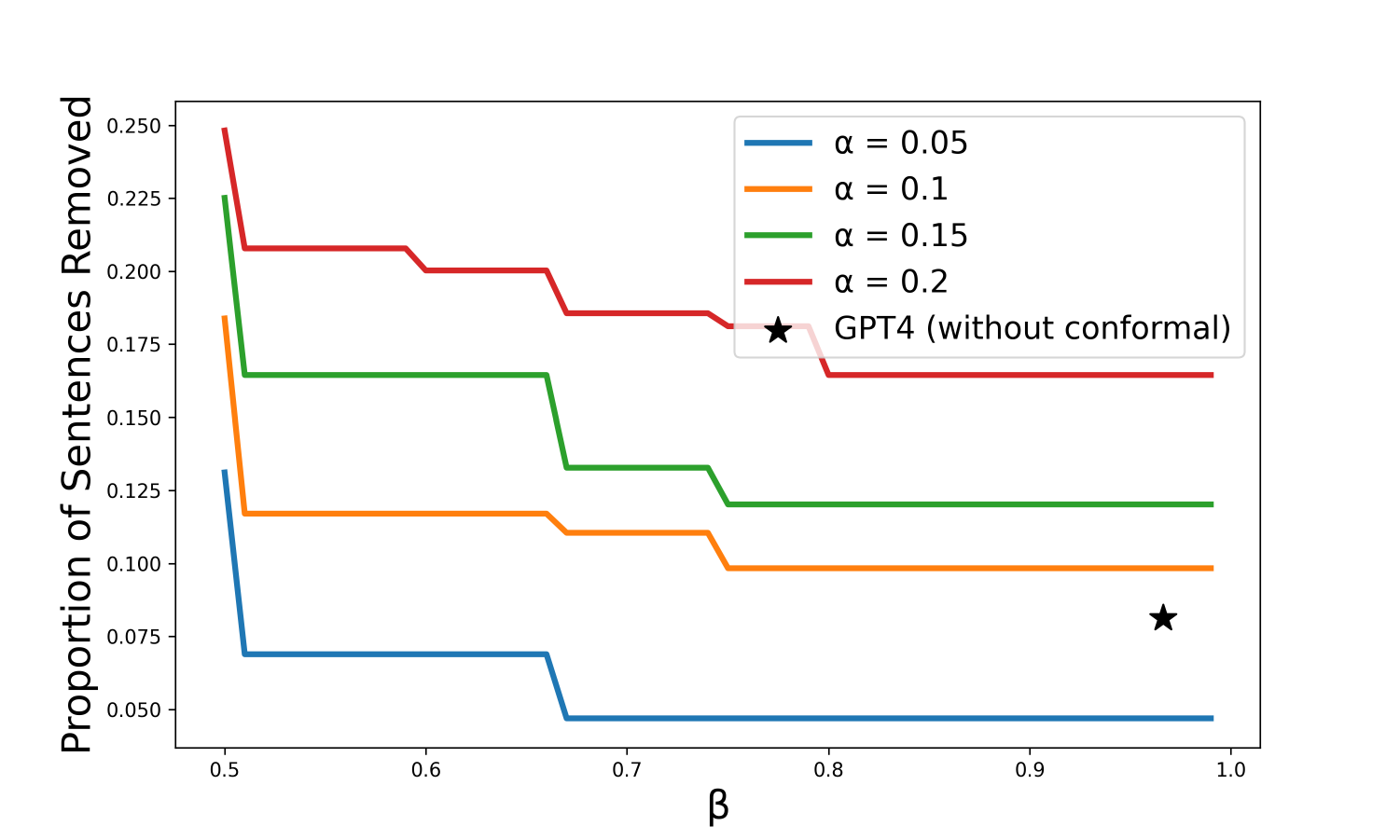}
        \caption{Cosine Similarity Centrality}
        \end{subfigure}
                \begin{subfigure}{0.48\textwidth}
        \includegraphics[width=0.95\linewidth, trim={30 0 60 40}, clip]{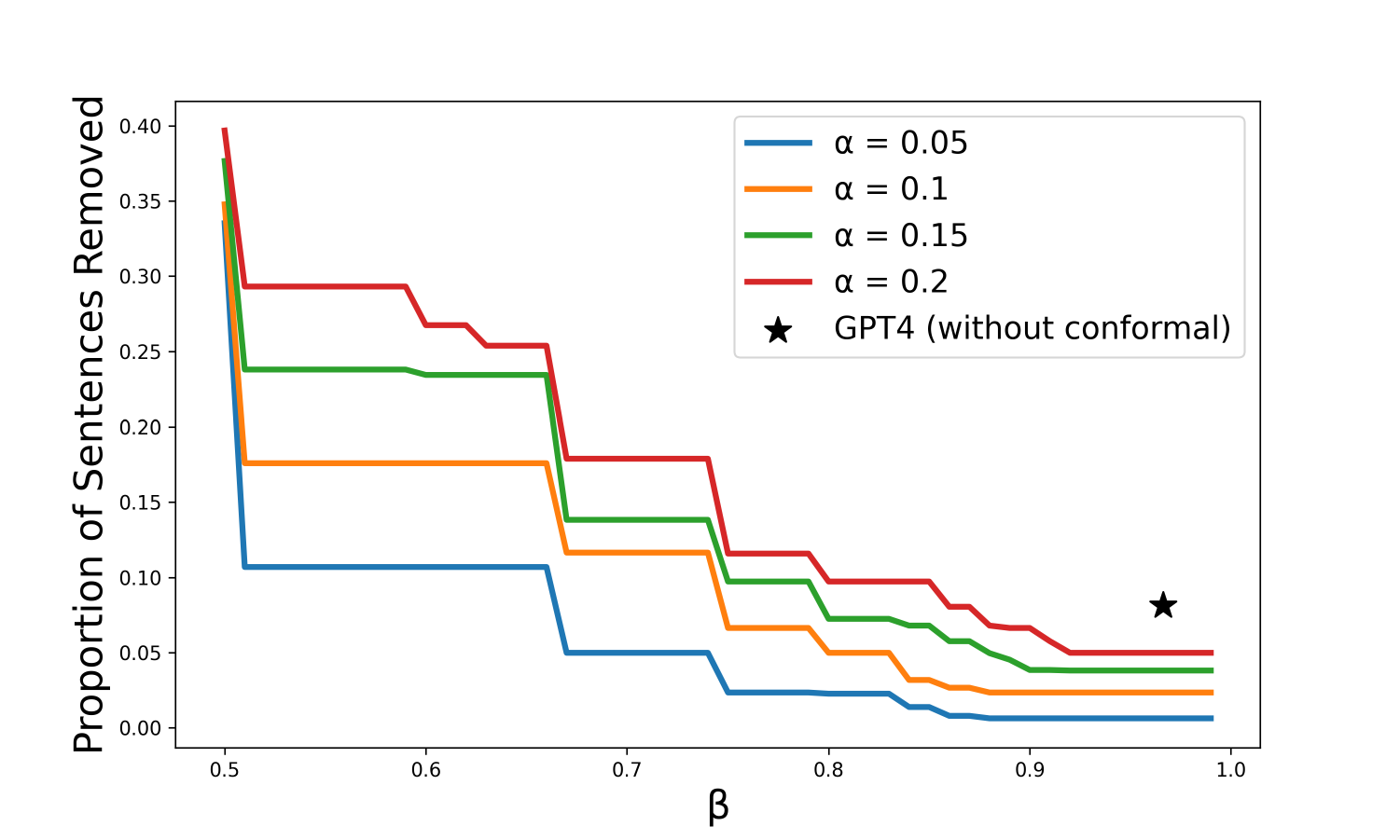}
        \caption{Sentence Centrality}
        \end{subfigure}
        \begin{subfigure}{0.48\textwidth}
        \includegraphics[width=0.95\linewidth, trim={30 0 60 40}, clip]{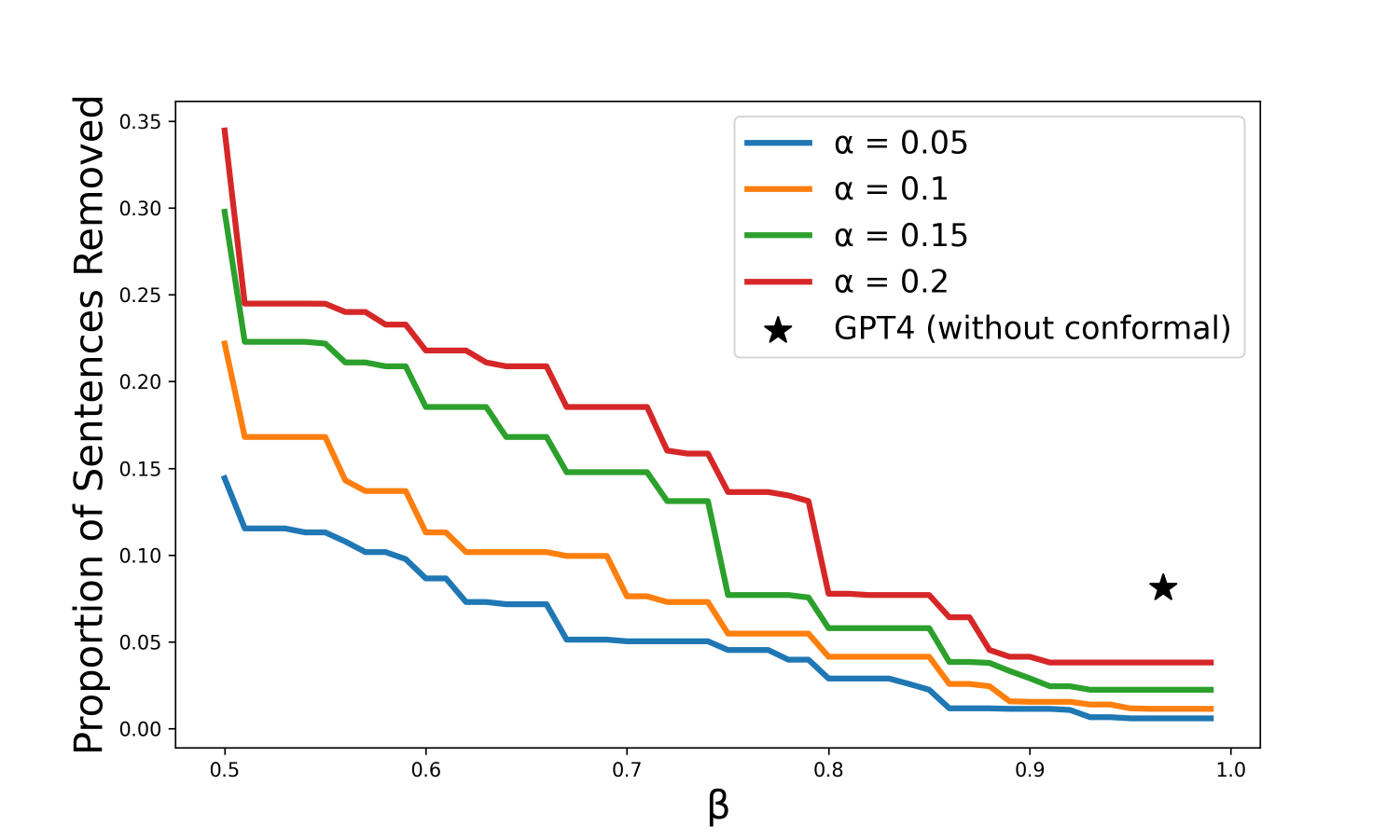}
        \caption{GUSUM}
        \end{subfigure}
        \begin{subfigure}{0.48\textwidth}
        \includegraphics[width=0.95\linewidth, trim={30 0 60 40}, clip]{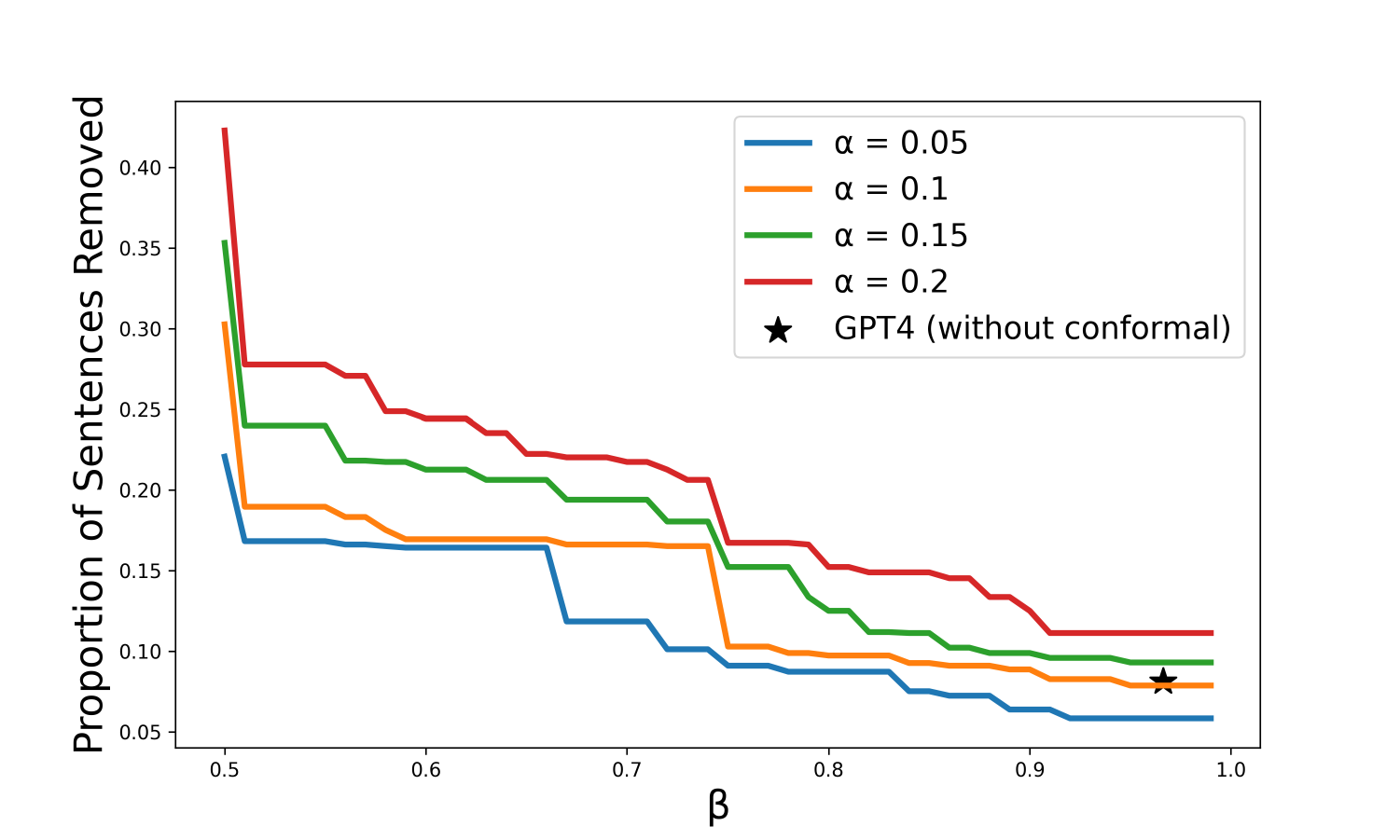}
        \caption{GPT-4o mini}
        \end{subfigure}
                \begin{subfigure}{0.48\textwidth}
        \includegraphics[width=0.95\linewidth, trim={30 0 60 40}, clip]{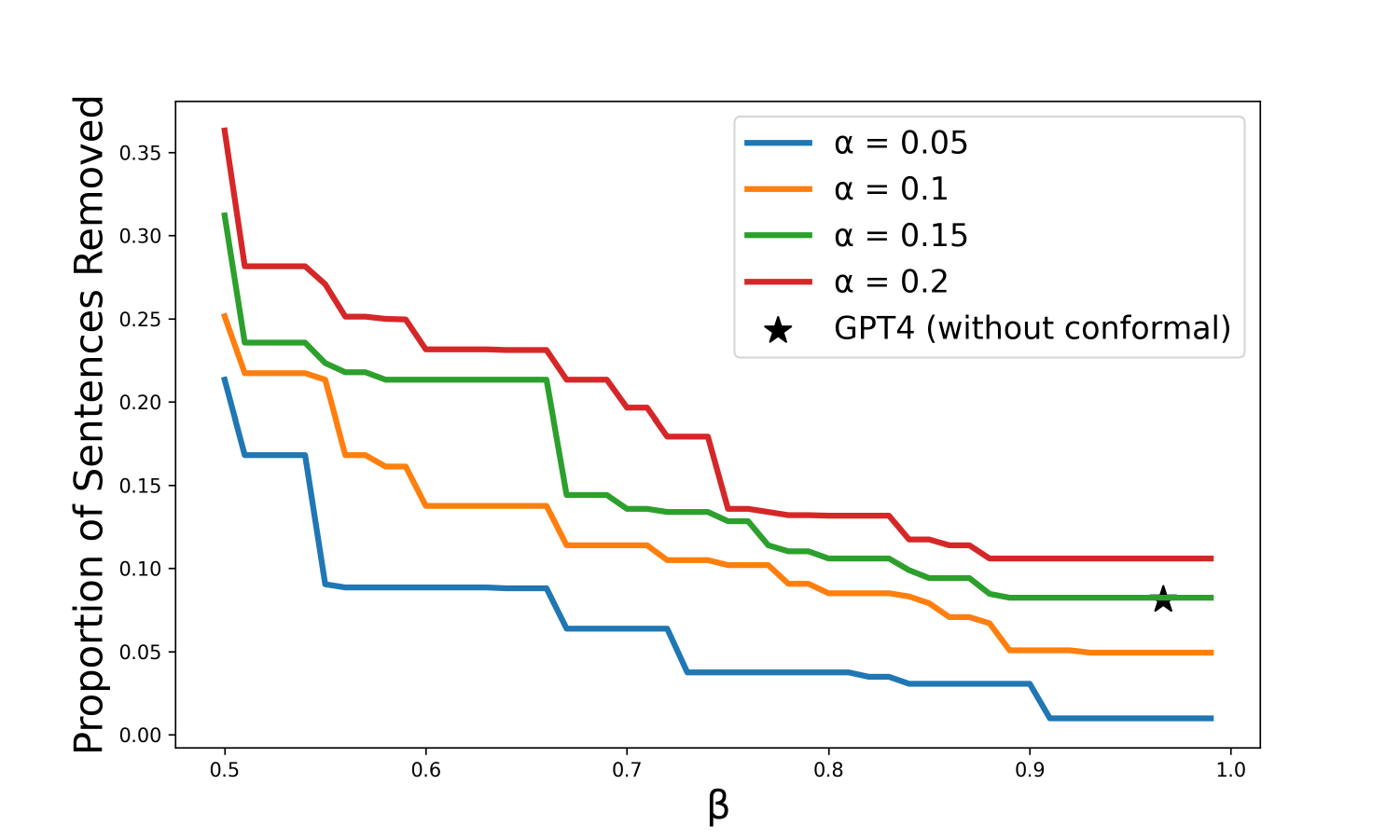}
        \caption{Llama 3}
        \end{subfigure}
        \begin{subfigure}{0.48\textwidth}
        \includegraphics[width=0.95\linewidth, trim={30 0 60 40}, clip]{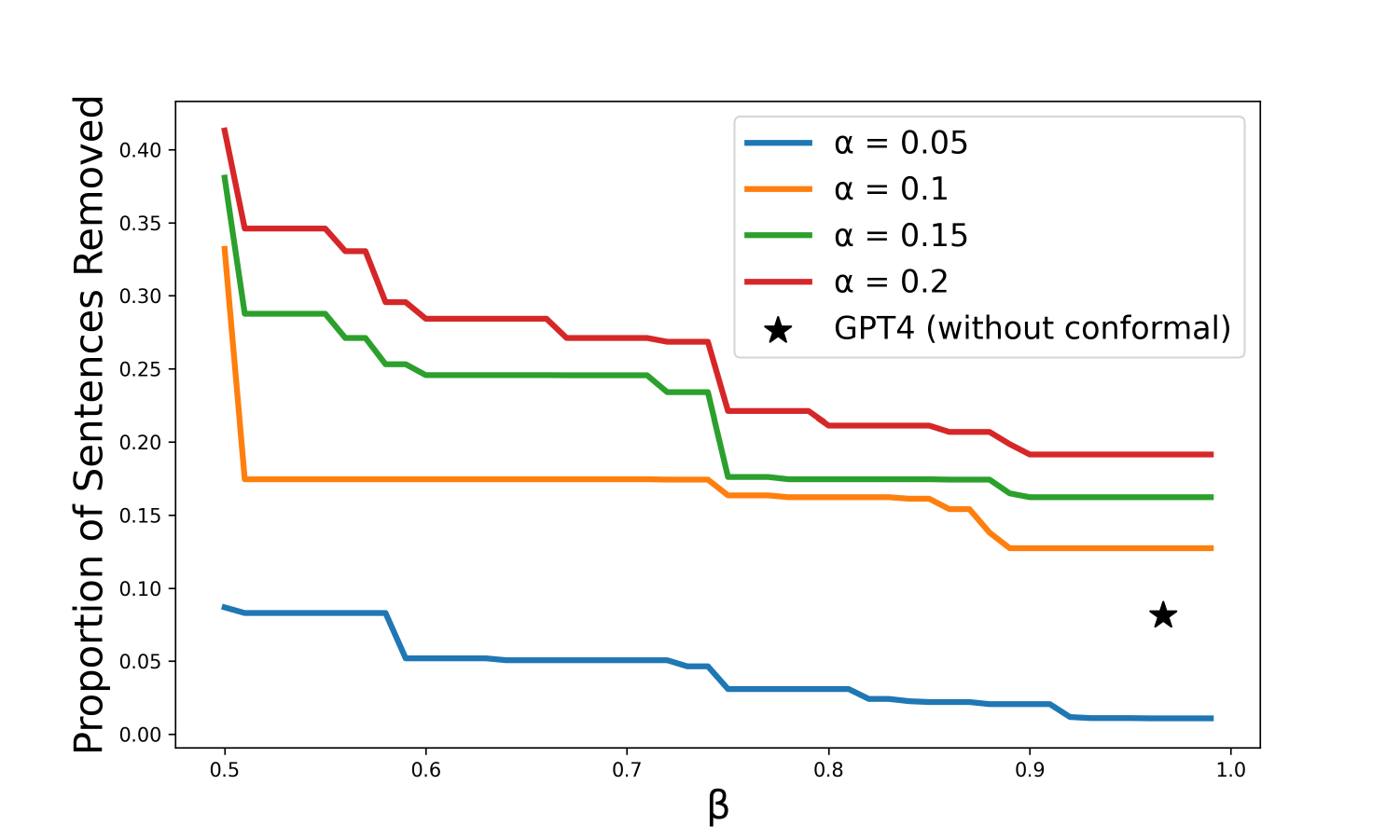}
        \caption{Qwen 3}
        \end{subfigure}
        \begin{subfigure}{0.48\textwidth}
        \includegraphics[width=0.95\linewidth, trim={30 0 60 40}, clip]{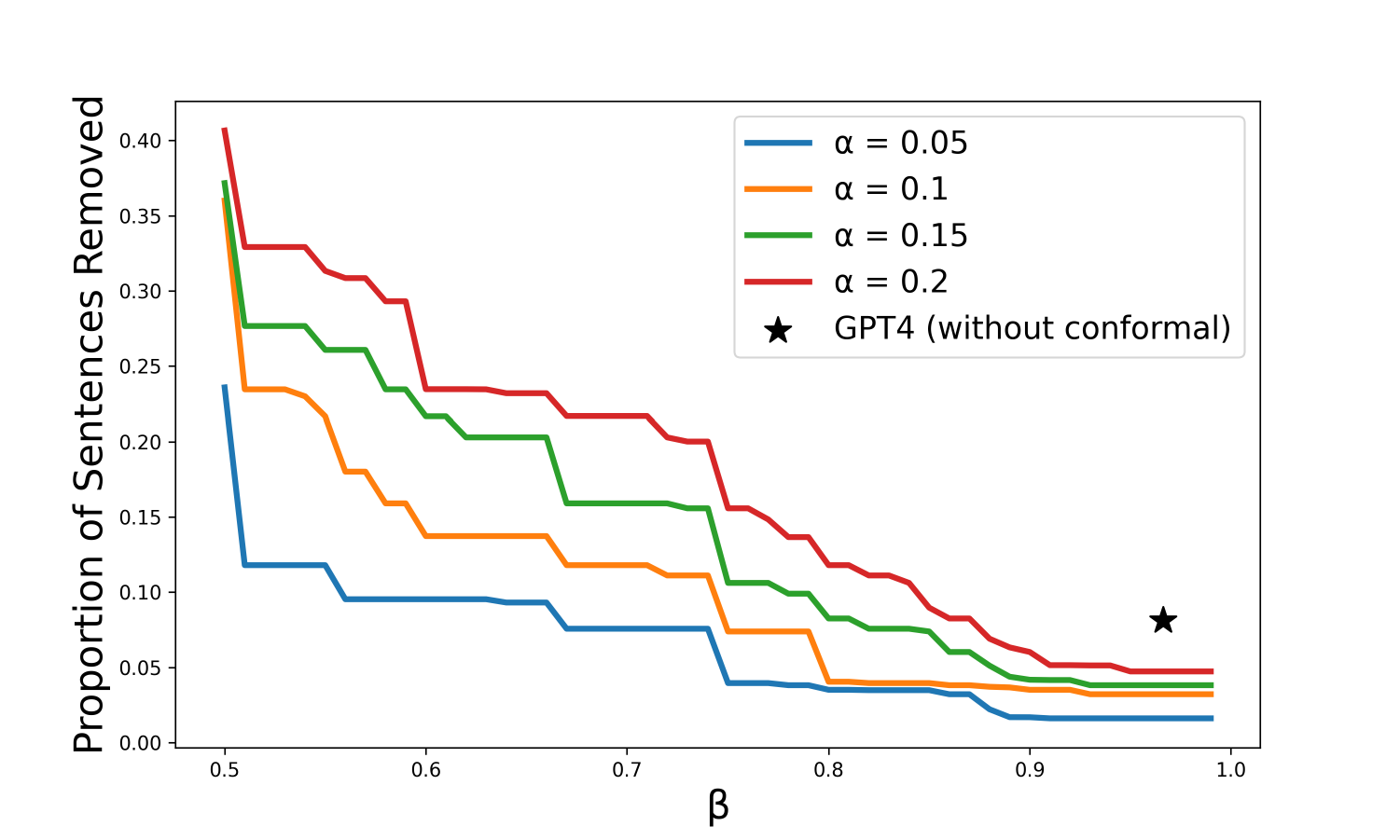}
        \caption{Gemini 2.0 Flash-Lite}
        \end{subfigure}
        \begin{subfigure}{0.48\textwidth}
        \includegraphics[width=0.95\linewidth, trim={30 0 60 40}, clip]{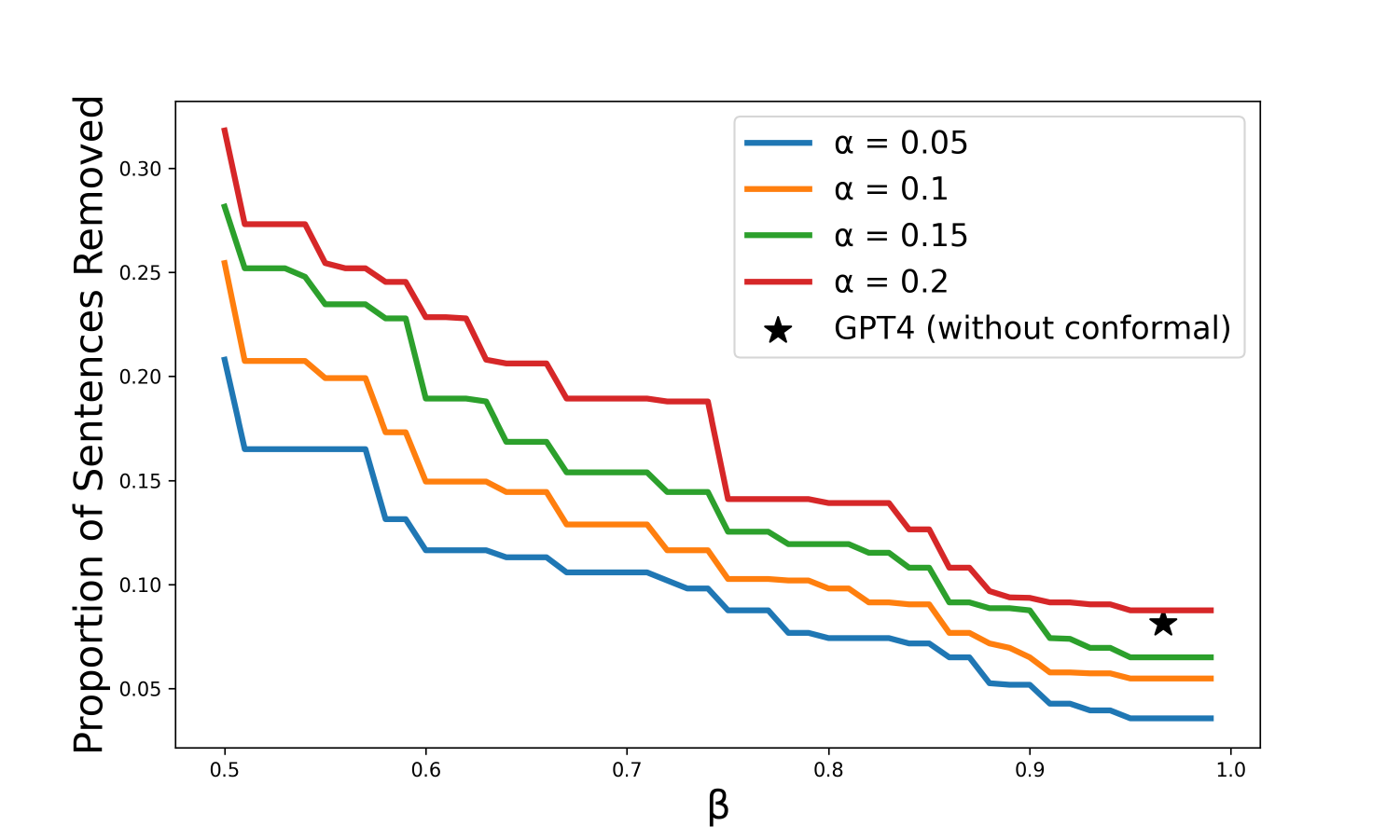}
        \caption{Gemini 2.5 Flash}
        \end{subfigure}
        \caption{Target recall $\beta$ vs. proportion of sentences removed (conciseness). Lines indicate different values for the target error rate $\alpha$ on MTS.}
    \label{fig:reduction_versus_beta_MTS}
\end{figure}

\clearpage

\subsection{ROUGE Score-based Ground Truth Performance}\label{app:rouge_score}

\Cref{tab:performance_precesion_rouge} and  \ref{tab:performance_reduction_rouge} respectively show the AUPRC and conciseness of summaries when we test our methods using labels generated from ROUGE scores, rather than cosine similarity, using \Cref{alg:greedy-extractive}. Since we only use this algorithm for CNN/DM and SciTLDR, we only display results for CNN/DM and TLDR-AIC. 

The results are similar to using cosine similarity: Gemini 2.0 Flash-Lite once again performs best in terms of AUPRC, and Gemini 2.5 Flash still performs very well in terms of sentence reduction length. The best numerical values for AUPRC and conciseness on each dataset are also comparable to the cosine similarity-based ground truth from \Cref{tab:performance_precesion_reduction}. 

\begin{table}[t]
    \centering
        \caption{Performance comparison of importance scoring methods, measured in AUPRC of claim rankings compared to ROUGE-1/2/L based ground truth labels. Higher is better.
}
    \label{tab:performance_precesion_rouge}
    \centering
        \setlength{\tabcolsep}{3pt} 
    \begin{tabular}{l|cc|cc|cc}
        \toprule
        Importance Score & \multicolumn{2}{c|}{ROUGE-1} & \multicolumn{2}{c|}{ROUGE-2} & \multicolumn{2}{c}{ROUGE-L} \\
        & CNN/DM & TLDR-AIC & CNN/DM & TLDR-AIC & CNN/DM & TLDR-AIC \\
        \midrule
        Dataset Positive Rate & 0.10 & 0.06 & 0.10 & 0.06 & 0.10 & 0.06 \\
        \midrule
        Cos. Sim. Centrality & 0.13& 0.30& 0.27 & 0.26& 0.31 & 0.29\\
        Sentence Centrality & 0.28& 0.27& 0.24 & 0.24& 0.27 & 0.27\\
        GUSUM & 0.31& 0.19& 0.30 & 0.18& 0.30 & 0.19\\
        LexRank & 0.32& 0.30& 0.28 & 0.26& 0.32 & 0.28\\
        \midrule
        GPT-4o mini (binary) & 0.13& 0.08& 0.13 & 0.08 & 0.13 & 0.08 \\
        GPT-4o mini & 0.38& 0.37& 0.34 & 0.32 & 0.37 & 0.33 \\
        Llama3-8B & 0.23& 0.18& 0.21 & 0.16 & 0.24 & 0.19 \\
        Qwen3-8B & 0.23& 0.20& 0.22 & 0.17 & 0.23 & 0.20 \\
        Gemini 2.0 Flash-Lite & \textbf{0.42}& \textbf{0.40}& \textbf{0.38} & \textbf{0.39}& \textbf{0.40} & \textbf{0.38} \\
        Gemini 2.5 Flash & 0.35& 0.36& 0.32 & 0.34& 0.34 & 0.36 \\
        \bottomrule

    \end{tabular}
\end{table}

\begin{table}[t]
    \centering
        \caption{Performance comparison of importance scoring methods, measured in conciseness of summaries (proportion of sentences removed) under Conformal Importance Summarization compared to ROUGE-1/2/L based ground truth labels. Higher is better.
}
    \label{tab:performance_reduction_rouge}
    \centering
        \setlength{\tabcolsep}{3pt} 
    \begin{tabular}{l|cc|cc|cc}
        \toprule
        Importance Score & \multicolumn{2}{c|}{ROUGE-1} & \multicolumn{2}{c|}{ROUGE-2} & \multicolumn{2}{c}{ROUGE-L} \\
        & CNN/DM & TLDR-AIC & CNN/DM & TLDR-AIC & CNN/DM & TLDR-AIC \\
        \midrule
        Original Article & 0.00 & 0.00 & 0.00 & 0.00 & 0.00 & 0.00 \\
        \midrule
        Cos. Sim. Centrality & 0.22 & 0.24& 0.23& 0.12& 0.21 & 0.22\\
        Sentence Centrality & 0.22 & 0.20& 0.25& 0.22& 0.19 & 0.22\\
        GUSUM & 0.21 & 0.07& 0.32& 0.10& 0.18 & 0.07\\
        LexRank & 0.17 & 0.26 & 0.17& 0.29& 0.16 & 0.25\\
        \midrule
        GPT-4o mini (binary) & 0.26 & 0.22 & 0.26 & 0.22 & 0.26 & 0.22 \\
        GPT-4o mini & \textbf{0.32} & 0.25 & \textbf{0.33} & 0.28 & \textbf{0.29} & 0.21 \\
        Llama3-8B& 0.25 & 0.27 & 0.31 & \textbf{0.37} & 0.27 & 0.27 \\
        Qwen3-8B & 0.07 & 0.06 & 0.15 & 0.08 & 0.11 & 0.08 \\
        Gemini 2.0 Flash-Lite& 0.11 & 0.09 & 0.09 & 0.10 & 0.09 & 0.11 \\
        Gemini 2.5 Flash & 0.22 & \textbf{0.28} & 0.30 & 0.31 & 0.22 & \textbf{0.29} \\
        \bottomrule
 
    \end{tabular}
\end{table}

\subsection{Ablation over Calibration Set Size}\label{app:cal_set_size}

\begin{table}[t]
    \centering
        \caption{Ablation of empirical coverage over calibration dataset size $n$.
}
    \label{tab:ablation_set_size}
    \small{
    \centering
    \resizebox{\textwidth}{!}{
        \setlength{\tabcolsep}{3pt} 
    \begin{tabular}{c|cccc|cccc}
        \toprule
        \multicolumn{1}{c|}{Target Coverage} & \multicolumn{4}{c|}{Mean}& \multicolumn{4}{c}{Standard Deviation}\\
        $1-\alpha$ & $n=25$ & $n=50$ & $n=75$ & $n=100$& $n=25$ & $n=50$ & $n=75$ & $n=100$  \\
        \midrule
        0.60 & 0.61 & 0.61 & 0.60 & 0.61 & 0.09 & 0.07 & 0.05 & 0.05\\
0.65 & 0.65 & 0.67 & 0.66 & 0.65 & 0.09 & 0.07 & 0.06 & 0.05\\
0.70 & 0.73 & 0.71 & 0.71 & 0.70 & 0.09 & 0.06 & 0.05 & 0.05\\
0.75 & 0.77 & 0.77 & 0.75 & 0.75 & 0.09 & 0.06 & 0.05 & 0.05\\
0.80 & 0.81 & 0.81 & 0.80 & 0.80 & 0.08 & 0.05 & 0.05 & 0.04\\
0.85 & 0.89 & 0.86 & 0.86 & 0.85 & 0.06 & 0.05 & 0.04 & 0.04\\
0.90 & 0.92 & 0.90 & 0.91 & 0.90 & 0.05 & 0.04 & 0.03 & 0.03\\
0.95 & 0.96 & 0.96 & 0.96 & 0.95 & 0.04 & 0.03 & 0.02 & 0.02\\
        \bottomrule
    \end{tabular}
    }
}
\end{table}

\begin{table}[t]
\centering
\caption{Ablation of summary conciseness (proportion of sentences removed) over calibration dataset size $n$. Results are taken over 20 random calibration/test splits.}\label{tab:ablation_set_size_performance}
\begin{tabular}{rcc}
\hline
$n$  & Mean & Std \\
\hline
25  & 0.28 & 0.09 \\
50  & 0.31 & 0.08 \\
75  & 0.32 & 0.08 \\
100 & 0.33 & 0.05 \\
\hline
\end{tabular}
\end{table}

Throughout the paper, we used a fixed calibration set size of $n=100$ samples to demonstrate that the method can operate with very little labeled data. However, in some regimes the availability of labeled data can be extremely limited, so in this section we test our method with even fewer calibration datapoints.

The effect of calibration dataset size in conformal prediction is well understood theoretically; the coverage guarantee is famously “valid in finite samples”, meaning that it holds statistically for any finite calibration dataset. In practice, $n$ controls the variance of the coverage viewed as a random variable over the calibration data. For a textbook-style explanation of these details, see Section 3.2 of \cite{angelopoulos2022gentle}.

We match the experimental setting of \Cref{fig:conformal_calibration} which uses $\beta=0.8$ and the Gemini 2.5 Flash scoring function to generate results on the ECTSum dataset, shown in \Cref{tab:ablation_set_size}. As guaranteed by \Cref{theorem}, the empirical coverage is no less than $1-\alpha$ for all values of $n$. Lower values of $n$ tend to overshoot the minimum coverage $1-\alpha$ by a bit more, because the upper bound of $1-\alpha + \frac{1}{n+1}$ becomes looser with $n$, but we still find all values within theoretical bounds (for example, $\frac{1}{25+1}\approx 0.04$, and $\frac{1}{50+1}\approx 0.02$). The primary reason to increase $n$ is to reduce the variance of the empirical coverage so that it is less likely any given instantiation has lower than expected coverage.

Given that we find the coverage guarantee to be satisfied, we can also check the main metric of our method's performance: the conciseness of summaries it produces (as the proportion of sentences removed). In \Cref{tab:ablation_set_size_performance} we find little difference in the performance when using $50$ to $100$ samples, although variance is increased on smaller datasets. Overall, this ablation demonstrates that our method is applicable in the very low labeled data regime.

\subsection{Direct Abstractive and Hybrid Extractive-Abstractive Comparison}\label{app:extractive_abstractive_comparison}

\begin{figure}[t]
    \centering
        \begin{subfigure}{0.32\textwidth}
        \includegraphics[width=1.0\linewidth, trim={8 10 10 10}, clip]{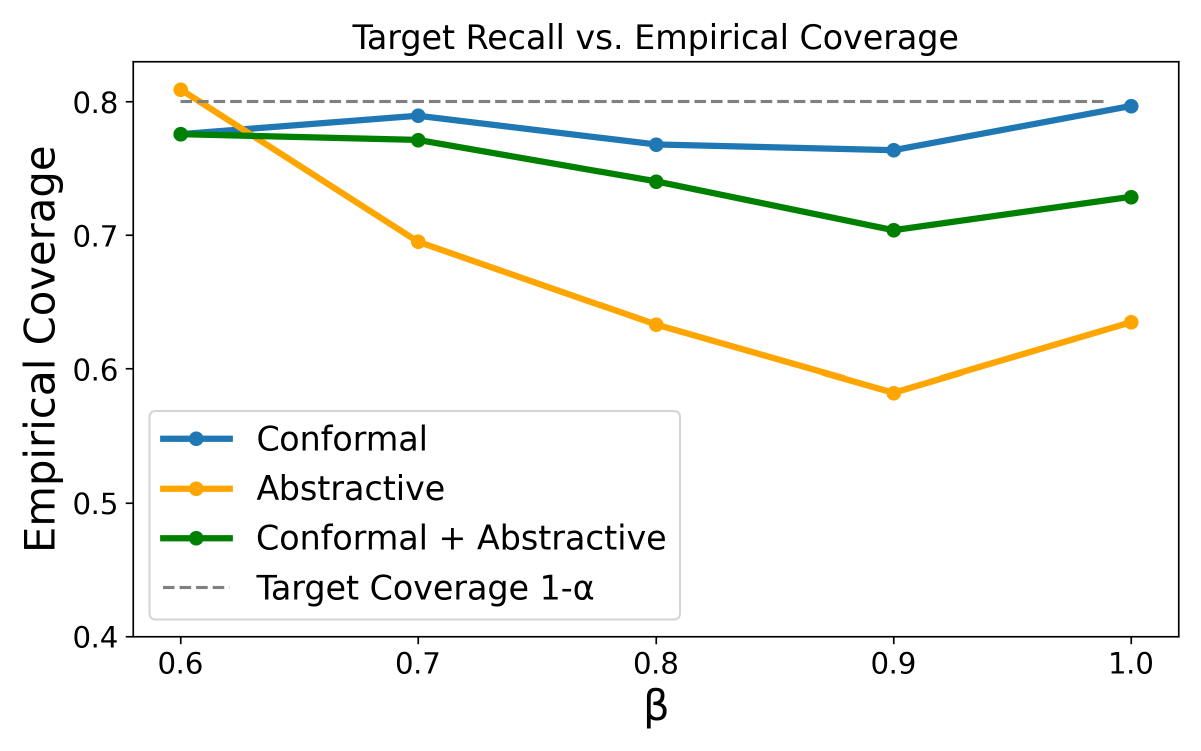}
        \end{subfigure} \
        \begin{subfigure}{0.32\textwidth}
            \includegraphics[width=1.0\linewidth, trim={8 10 10 10}, clip]{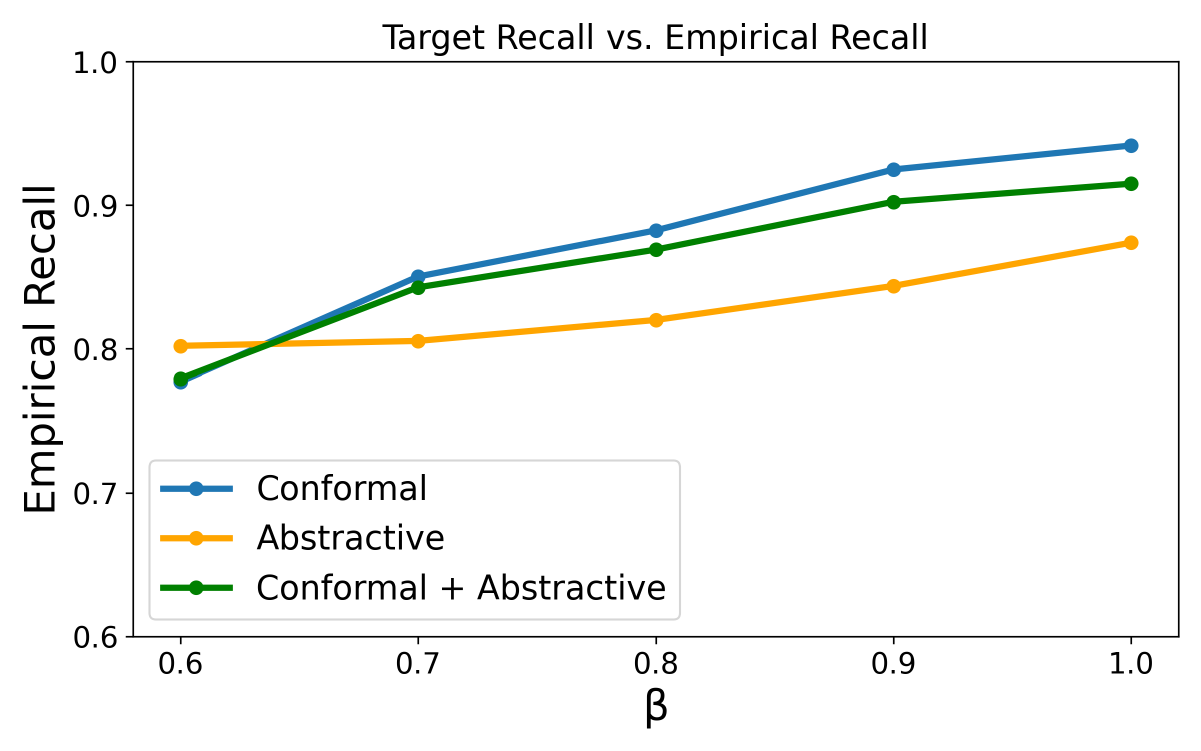}
        \end{subfigure} \
                \begin{subfigure}{0.32\textwidth}
            \includegraphics[width=1.0\linewidth, trim={8 10 10 10}, clip]{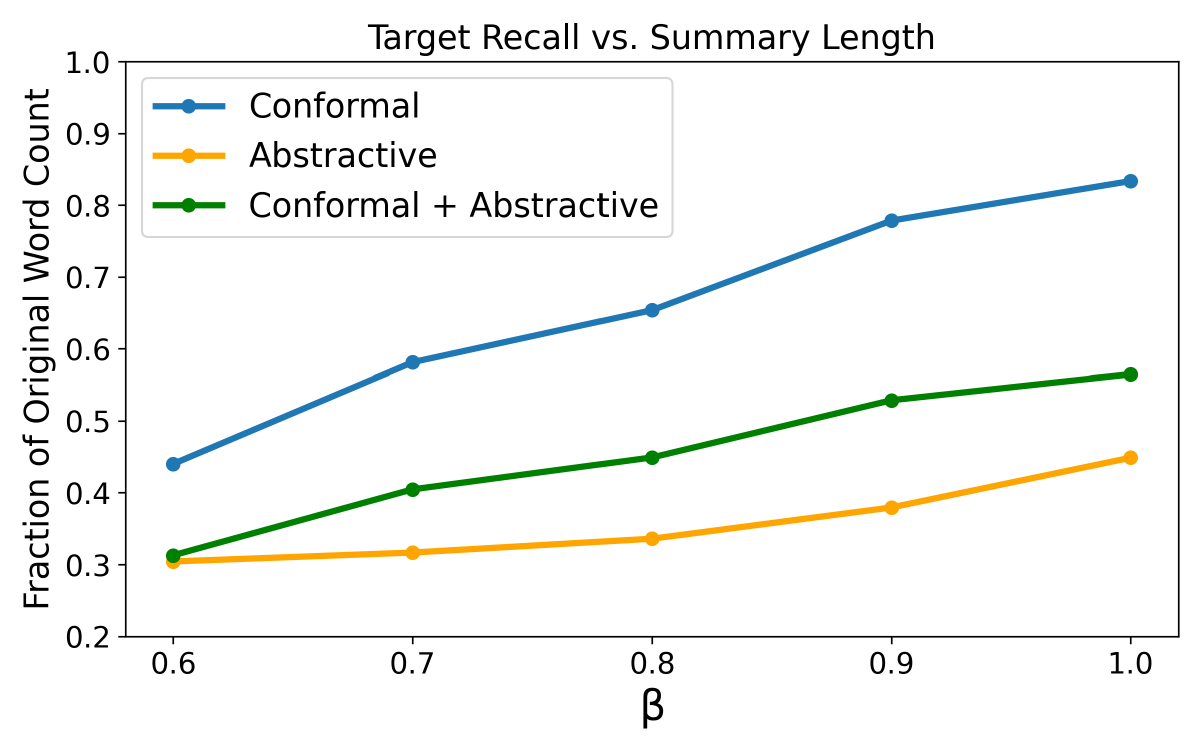}
        \end{subfigure}
        \caption{Comparison between extractive summarization with our method, abstractive summarization with an LLM, and our hybrid proposal on ECTSum. Here the target coverage is $1-\alpha=0.8$, the conformal approach uses Gemini 2.5 Flash scoring, and the abstractive model is also Gemini 2.5 Flash.}
\label{fig:extractive_abstractive_comparison_gemini}
\end{figure}

Here we provide additional plots using the same settings as in \Cref{subsec:extractive_abstractive_comparison}. \Cref{fig:extractive_abstractive_comparison_gemini} shows the comparison between extractive summarization with our conformal method, abstractive summarization with an LLM, and our hybrid proposal of applying abstractive summarization to our extractive summary, this time with Gemini 2.5 Flash used for both conformal scoring, and abstractive summarization. The results are highly comparable to \Cref{fig:extractive_abstractive_comparison_gpt}.

\end{document}